\documentclass[10pt,journal,compsoc]{IEEEtran}

\ifCLASSOPTIONcompsoc
  \usepackage[nocompress]{cite}
\else
  \usepackage{cite}
\fi

\usepackage{amsmath}
\usepackage{graphicx}
\usepackage{amssymb}
\usepackage{booktabs}
\usepackage{times}
\usepackage[x11names]{xcolor}
\usepackage{epsfig}
\usepackage{colortbl}
\usepackage{tabulary}
\usepackage{etoolbox}
\usepackage{multirow}
\usepackage[accsupp]{axessibility}
\usepackage[breaklinks]{hyperref}
\usepackage[capitalize]{cleveref}
\usepackage{soul}
\usepackage{placeins}
\usepackage{xspace}
\usepackage{tikz}
\usepackage[lmargin=13mm,rmargin=13mm,tmargin=15mm,bmargin=17mm]{geometry}

\makeatletter
\DeclareRobustCommand\onedot{\futurelet\@let@token\@onedot}
\def\@onedot{\ifx\@let@token.\else.\null\fi\xspace}

\def\eg{\emph{e.g}\onedot} 
\def\ie{\emph{i.e}\onedot}

\def\etal{\emph{et al}\onedot}
\makeatother

\renewcommand{\paragraph}[1]{\smallskip \noindent \textbf{#1}}
\definecolor{Gray}{gray}{0.85}
\newcolumntype{a}{>{\columncolor{Gray}}c}
\newcommand{\rulesep}{\unskip\ \vrule\ }

\newcommand{\beginsuppmat}{%
    \setcounter{section}{0}
    \renewcommand*{\thesection}{\Alph{section}}
 }

\makeatletter
\newcommand{\dashrule}[1][black]{%
  \color{#1}\rule[\dimexpr.5ex-.2pt]{4pt}{.4pt}\xleaders\hbox{\rule{4pt}{0pt}\rule[\dimexpr.5ex-.2pt]{4pt}{.4pt}}\hfill\kern0pt%
}
\newcommand{\rulecolor}[1]{%
  \def\CT@arc@{\color{#1}}%
}
\makeatother
\begin{document}
\newgeometry{left=13mm,right=13mm,top=15mm,bottom=24mm}
\title{FocalPose++: Focal Length and Object Pose Estimation via Render and Compare}

\author{Martin~Cífka$^*$,
        Georgy~Ponimatkin$^*$,
        Yann~Labbé,
        Bryan~Russell,
        Mathieu~Aubry,
        Vladimir~Petrik
        and~Josef~Sivic
    \IEEEcompsocitemizethanks{
    \IEEEcompsocthanksitem M. Cífka and G. Ponimatkin are with Faculty of Electrical Engineering, and Czech Institute of Informatics, Robotics and Cybernetics,
    Czech Technical University in Prague, 16000 Prague, Czech Republic. E-mail: \{\href{mailto:martin.cifka@cvut.cz}{martin.cifka}, \href{mailto:georgij.ponimatkin@cvut.cz}{georgij.ponimatkin}\}@cvut.cz.\protect
    \IEEEcompsocthanksitem V. Petrik and J. Sivic are with Czech Institute of Informatics, Robotics and Cybernetics, Czech Technical University in Prague, 16000 Prague, Czech Republic. E-mail: \{\href{mailto:vladimir.petrik@cvut.cz}{vladimir.petrik}, \href{mailto:josef.sivic@cvut.cz}{josef.sivic}\}@cvut.cz.\protect
    \IEEEcompsocthanksitem Y. Labbé is with Reality Labs, Meta Platforms, 8045 Zürich, Switzerland. E-mail: \href{mailto:ylabbe@meta.com}{ylabbe@meta.com}.\protect
    \IEEEcompsocthanksitem B. Russell is with Adobe Research, San Francisco, CA 94103 USA. E-mail: \href{mailto:brussell@adobe.com}{brussell@adobe.com}.\protect
    \IEEEcompsocthanksitem M. Aubry is with LIGM, École des Ponts, UGE, CNRS, 77455 Marne-la-Vallée, France. E-mail: \href{mailto:mathieu.aubry@imagine.enpc.fr}{mathieu.aubry@imagine.enpc.fr}.\protect
    }
    \thanks{$^*$Equal contribution}
}

\markboth{IEEE Transactions on Pattern Analysis and Machine Intelligence, 2024}%
{Cífka and Ponimatkin \MakeLowercase{\textit{et al.}}: FocalPose++: Focal Length and Object Pose Estimation via Render and Compare}

\IEEEtitleabstractindextext{%
\begin{abstract}
We introduce FocalPose++, a neural \textit{render-and-compare} method for jointly estimating the camera-object 6D pose and camera focal length given a single RGB input image depicting a known object.  
The contributions of this work are threefold.
First, we derive a focal length update rule that extends an existing state-of-the-art render-and-compare  6D pose estimator to address the joint estimation task.  
Second, we investigate several different loss functions for jointly estimating the object pose and focal length. We find that a combination of direct focal length regression with a reprojection loss disentangling the contribution of translation, rotation, and focal length leads to improved results. 
Third, we explore the effect of different synthetic training data on the performance of our method. Specifically, we investigate different distributions used for sampling object's 6D pose and camera's focal length when rendering the synthetic images, and show that parametric distribution fitted on real training data works the best.
We show results on three challenging benchmark datasets that depict known 3D models in uncontrolled settings. We demonstrate that our focal length and 6D pose estimates have lower error than the  existing state-of-the-art methods.

\end{abstract}

}

\maketitle
\IEEEdisplaynontitleabstractindextext

\begin{tikzpicture}[remember picture,overlay]
\node at (current page.south west) [anchor=south west]
{\footnotesize \hspace{12mm} \begin{minipage}{0.99\textwidth}\begin{center}
\noindent © 2024 IEEE. Personal use of this material is permitted. Permission from IEEE must be obtained for all other uses, in any current or future media, including reprinting/republishing this material for advertising or promotional purposes, creating new collective works, for resale or redistribution to servers or lists, or reuse of any copyrighted component of this work in other works.
\vspace{8mm}
\end{center}\end{minipage}};
\end{tikzpicture}
\IEEEraisesectionheading{\section{Introduction}\label{sec:introduction}}
\IEEEPARstart{T}{he} projection of a 3D object into an image depends not only on the object's relative pose to the camera but also on the camera's intrinsic parameters. 
While it is possible to capture objects in a controlled environment where the camera's intrinsic parameters are known (\eg, a calibrated camera on a robot), for many ``in-the-wild" images we do not have control over the capture process and these parameters are unknown, \eg, Internet pictures or archival photographs.

\begin{figure}[ht]
    \centering
    \includegraphics[width=\linewidth]{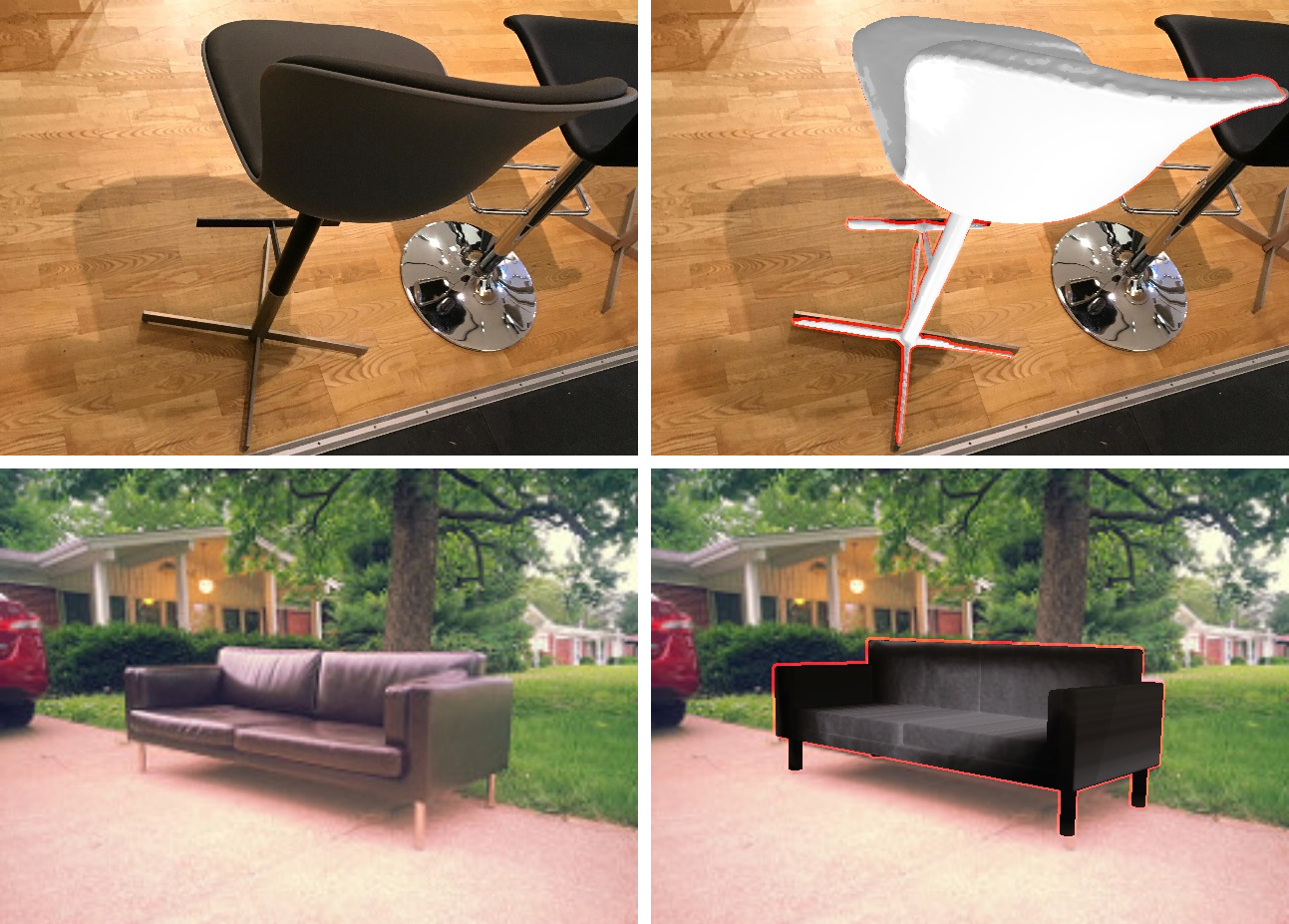}
    \vspace{-7mm}
    \caption{
        Given a single input photograph (\textbf{left}) and a known 3D model, our approach accurately estimates the 6D camera-object pose together with the focal length of the camera (\textbf{right}), here shown by overlaying the aligned 3D model over the input image. Our approach handles a large range of focal lengths and the resulting perspective effects. 
    }
    \label{fig:teaser}
\end{figure}

Given an input image, we seek to retrieve a 3D model of a depicted object from a database and estimate the relative camera-object 6D pose jointly with the camera's focal length (depicted in Fig.~\ref{fig:teaser}).
This problem has its origins in the early days of computer vision~\cite{Lowe1999-bf, Lowe1987-yf,Roberts1963-ck} and has important modern-day applications in augmented reality and computer graphics, such as applying in situ object overlays~\cite{kholgade20143d} or editing the position of an object via 3D compositing in uncontrolled
consumer-captured images.

The problem of 6D object pose estimation in an uncalibrated setting is, by its nature, challenging. First, it is difficult to distinguish subtle changes in the camera's focal length from changes in an object's depth. 
Second, including the camera's focal length increases the number of parameters that must be estimated and hence increases the optimization complexity. 
Finally, ``in-the-wild'' consumer-captured images may depict large appearance variation for a particular object instance in the model database. 
Variation may be due to differences in illumination and the depicted object having slightly different, non-identical shapes or surface appearance in different real-world instance captures. 
For example, consider different instances of the same car model that have a similar overall shape but may have different color, wear and tear, or customizable features.

Previous approaches for this task mainly rely on establishing local 2D-3D correspondences between an image and a 3D model using either hand-crafted~\cite{Aubry14,bay2006surf,Collet2010-zj,Collet2011-lj,Hinterstoisser2011-es,Lowe1999-bf} or CNN features~\cite{grabner2019gp2c,hu2019segmentation,Kehl2017-ek,Park2019-od,pavlakos20176,peng2019pvnet,Rad2017-de,song2020hybridpose,Tekin2017-hp,Tremblay2018-bd,Xiang2018-dv,zakharov2019dpod}, followed by robust camera pose estimation using PnP~\cite{lepetit2009epnp}. 
These approaches
\restoregeometry\noindent often fail in scenes with large texture-less areas where local correspondences cannot be reliably established. In contrast, the recent best-performing 6D object pose estimation methods are based on the render-and-compare strategy~\cite{labbe2020cosypose,li2018deepim,manhardt2018deep,oberweger2019generalized,zakharov2019dpod}, which performs a dense alignment over all pixels of rendered views of the 3D model to its depiction in the input image.
However, all prior render-and-compare methods fall short of handling the desired uncontrolled, uncalibrated setting, as they assume a controlled environment where the camera intrinsic parameters are fixed and known {\it a priori}. In addition, these previous methods typically operate on only a handful of known objects. 

To address these challenges, we build on the strengths of render and compare and extend it to handle our desired uncontrolled, uncalibrated setting. We introduce FocalPose++, a novel render-and-compare approach for jointly estimating an object's 6D pose and camera focal length from a monocular image. Our contributions are threefold. 
First, we extend one of the state-of-the-art~\cite{hodan2020bop} methods for 6D pose estimation (CosyPose~\cite{labbe2020cosypose}) by deriving and integrating focal length update rules in a differentiable manner, which allows our method to overcome the added complexity of including focal length. Second, we investigate several different loss functions for jointly estimating object pose and camera focal length. We find that a combination of direct focal length regression with a reprojection loss disentangling the contribution of translation, rotation, and focal length leads to the best performance and allows our method to distinguish subtle differences due to the focal length and the object's depth. Third, we explore the effect of different synthetic training data on the performance of our method. Specifically, we investigate different distributions used for sampling of object's 6D pose and camera's focal length when rendering the synthetic training images, and show that our parametric distribution fitted on real training data works the best.
We apply our method to three real-world consumer-captured image datasets with varying camera focal lengths and show that our focal length and 6D pose estimates have lower error compared to the state-of-the-art.
As an added benefit, our work is the first render-and-compare method applied to a large collection of 3D meshes (20-200 meshes for Pix3D~\cite{pix3d}, $\sim150$ for the car datasets~\cite{wang20183d}).

This paper is an extended version of FocalPose~\cite{ponimatkin2022focal}. 
We extend our previous work by investigating different parametric and non-parametric distributions for the synthetic training data and incorporating a new, more accurate, 3D model retrieval method. In addition, we derive and experimentally validate a new update rule for the translation component of the 6D pose update, which takes into account the focal length change between iterations. The full derivation of the update is provided in the supplementary material.
Overall, these contributions result in improvements of the measured metrics on all three datasets by almost 10\% on average compared to the original FocalPose~\cite{ponimatkin2022focal}, and outperforming other state-of-the-art methods with relative error reduction ranging from 10\% to 50\% on all three used datasets. Furthermore, we evaluate our method on the YCB-Video~\cite{Xiang2018-dv} dataset from the BOP challenge~\cite{hodan2024bop} and experimentally show the effect of focal length on the 6D pose estimation. Finally, we show two new applications of our method: 3D-aware image augmentation in computer graphics and imitation of manipulation skills in robotics. The code is publicly available through the project page at \href{https://cifkam.github.io/focalpose}{https://cifkam.github.io/focalpose}.

\section{Related Work}

\paragraph{6D pose estimation of rigid objects from RGB images.} This task is one of the oldest problems in computer vision~\cite{Roberts1963-ck,Lowe1987-yf,Lowe1999-bf} and has been successfully approached by estimating the pose from 2D-3D correspondences obtained via local invariant features~\cite{Lowe1999-bf,bay2006surf,Collet2010-zj,Collet2011-lj}, or by template-matching~\cite{Hinterstoisser2011-es}. Both of these strategies rely on shallow hand-designed image features and have been revisited with learnable deep convolutional neural networks (CNNs) \cite{Rad2017-de,Tremblay2018-bd,Kehl2017-ek,Tekin2017-hp,peng2019pvnet,pavlakos20176,hu2019segmentation,Xiang2018-dv,Park2019-od,song2020hybridpose,zakharov2019dpod}.
Recently, 3D-aware pose refiners~\cite{zhao20233d,liu2022gen6d,sinha2023sparsepose} have also been successfully used to refine the coarse 6D pose estimates. These methods backproject extracted 2D features either into a 3D volume or a collection of 3D points, and compare the extracted 2D features from the input image with features extracted from reference images while using the 3D information. For example, \cite{zhao20233d} predicts a relative pose between the object depicted in a query image and the object in a (single) reference image using a 3D-aware hypothesis-and-verification mechanism. Gen6D~\cite{liu2022gen6d} requires a set of reference images with known object poses, refining the pose of the object in the query image using features of the most similar reference images backprojected into 3D volumes. And SparsePose~\cite{sinha2023sparsepose} is inspired by 3D object reconstruction and predicts relative camera poses from a sparse set of input images depicting the same object. In contrast, we focus on a single-image input without having access to any reference image, but rely on the availability of the object 3D model.
One of the best-performing methods for 6D pose estimation from RGB images are based on variants of the deep {\em render-and-compare} strategy \cite{li2018deepim,manhardt2018deep,oberweger2019generalized,labbe2020cosypose,zakharov2019dpod}. However, these methods assume that the full perspective camera model is known so that the object can be rendered and compared with the input image. We build on the state-of-the-art render-and-compare approach of Labb\'{e} \etal~\cite{labbe2020cosypose} and extend it to the ``in-the-wild" uncontrolled set-up where the focal length of the camera is not known and has to be estimated together with the object's 6D pose directly from the input image. 

\begin{figure*}[th!]
  \centering
    \begin{minipage}{0.59\textwidth}
            \centering
            \includegraphics[width=\textwidth]{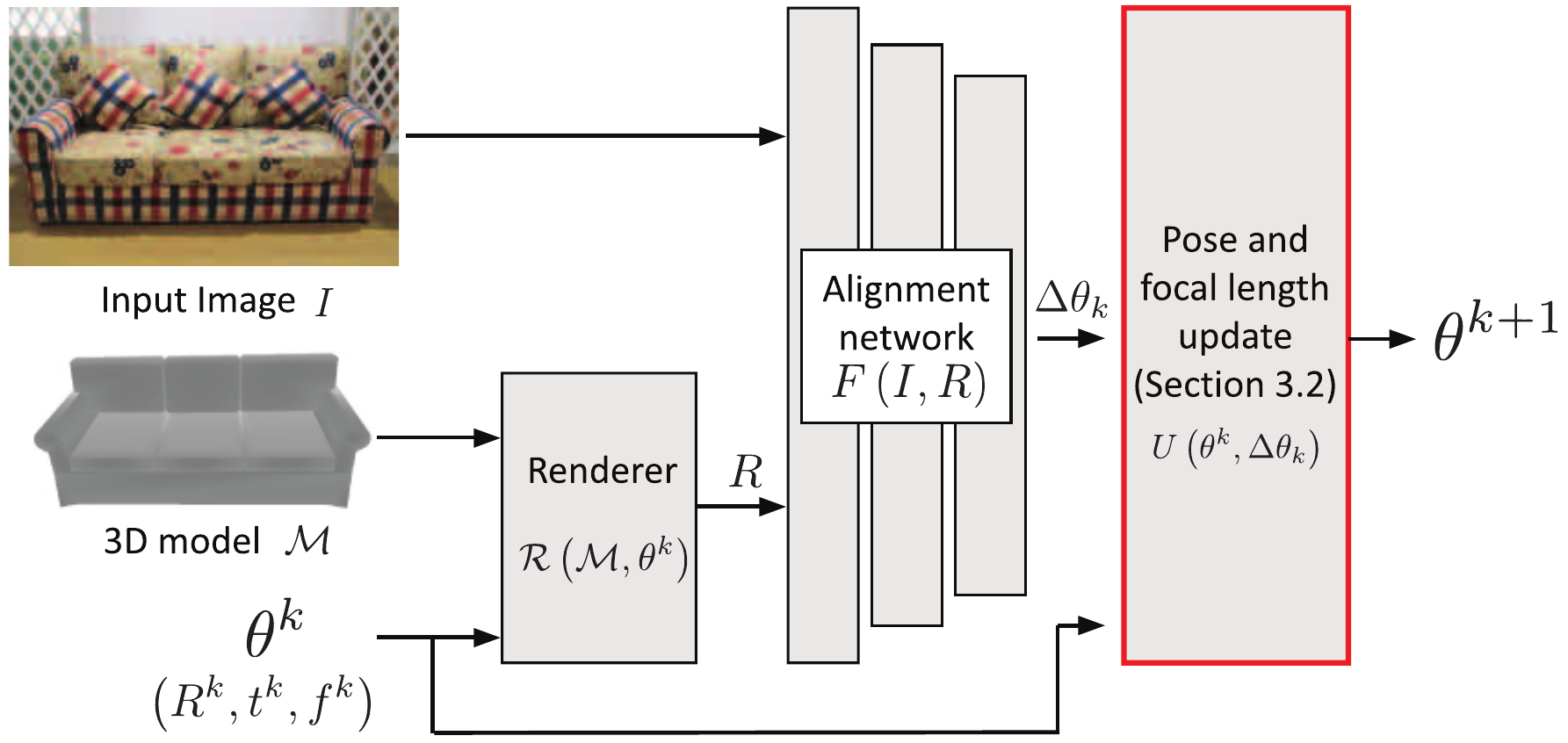}\\
    \small{(a)}
        \end{minipage}
        \rulesep
        \begin{minipage}{0.39\textwidth}
            \centering
            \includegraphics[width=\textwidth]{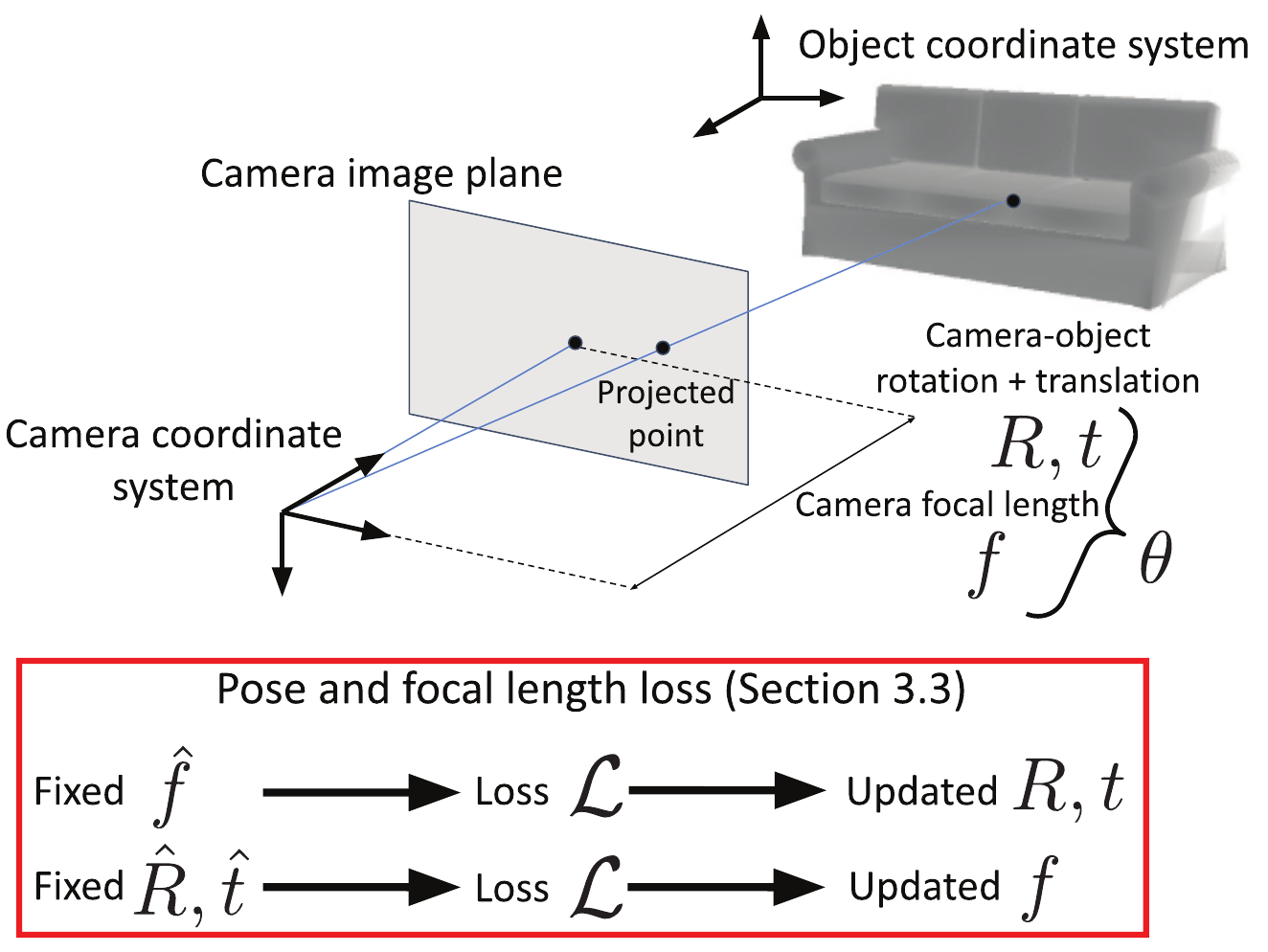}\\
    \small{(b)}
        \end{minipage} \\
  \vspace*{-1mm}
  \caption{ \textbf{ FocalPose overview.} \textbf{(a)}  Given a single  in-the-wild RGB input image $I$ of a known object 3D model $\mathcal{M}$, parameters $\theta^k$ composed of focal length $f^k$ and the object 6D pose (3D translation $t^k$ and 3D rotation $R^k$) are iteratively updated using our render-and-compare approach. The rendering $R$, together with the input image $I$, is given to a deep neural network $F$ that predicts the update $\Delta \theta_k$, which is then converted into the parameter update $\theta^{k+1}$ using a non-linear update rule $U$. \textbf{(b)} Illustration of the camera-object setup with parameters $\theta$ composed of 3D translation $t$, 3D rotation $R$ and focal length $f$. The alignment network is trained using a novel pose and focal length loss that disentangles the focal length and pose updates. The two main contributions of this work are highlighted by red boxes in the figure. 
}
    \vspace*{-4mm}
  \label{fig:method_overview}
\end{figure*}

\paragraph{Camera calibration.}
Camera calibration techniques ~\cite{faugeras1993three,andrew2001multiple,nakano2016versatile,penate2013exhaustive,zheng2014general,tsai1987versatile,dubska2014fully,szeliski2010computer} recover the camera model (intrinsic parameters) and its pose (extrinsic parameters) jointly. A limitation is that they require estimating 2D-3D correspondences in multiple images using structured object patterns~\cite{Forsyth12,Hartley2004,szeliski2010computer,tsai1987versatile}, identifying specific image elements such as lines or vanishing points~\cite{dubska2014fully,chen2004camera,szeliski2010computer} or structured features (\eg, human face landmarks~\cite{burgos2014distance}). These requirements limit their applicability to unconstrained images where these structures are not present.  Other works~\cite{workman2015deepfocal} have considered in-the-wild images, but only focus on recovering the focal length of the camera. In contrast, our approach recovers both components of the camera calibration (focal length and 6D camera pose) given a single image of a known object.

\paragraph{Joint 6D pose and focal length estimation from a single in-the-wild image.}
The prior work closest to our approach establishes point correspondences, followed by robust fitting of the camera model~\cite{wang20183d,grabner2019gp2c,han2020gcvnet}.
Wang \etal~\cite{wang20183d} uses Faster R-CNN with a scalar regression head and L1 loss to estimate the focal length, and the 6D pose is estimated by predicting 2D-3D correspondences followed by PnP.

GP2C~\cite{grabner2019gp2c} extends this approach via a two-step procedure that predicts initial 2D-3D correspondences and focal length with a similar direct regression, followed by applying a PnPf solver to jointly refine the 6D pose and the focal length. The model cannot be trained end-to-end, as it relies on a separate non-differentiable optimizer.
GCVNet~\cite{han2020gcvnet} uses an approximation of the PnPf solver for differentiability, but its results are limited by this approximation.
In contrast, our work builds on the success of recent render-and-compare methods~\cite{labbe2020cosypose,li2018deepim} for 6D rigid pose estimation. Our 6D pose and focal length updates are learned end-to-end using our novel focal length update parameterization coupled with a disentangled training loss. Our approach produces lower-error focal length and pose estimates compared to the two-step approach of GP2C~\cite{grabner2019gp2c} and the prior one-shot end-to-end approaches~\cite{wang20183d,han2020gcvnet}.

\section{Approach}  
Our goal is to estimate the 6D pose of objects in a photograph taken with unknown focal length.
To achieve this goal, we use a render-and-compare strategy, where we estimate jointly the camera focal length with the 6D pose.
We assume knowledge of a database of 3D models that may appear in the image, but our results show that the approach is effective even if the 3D models are only approximate.

\subsection{Approach Overview}
\label{approach-intro}

The first step of our approach identifies the object location in the input image and retrieves a 3D model from the database that matches the depicted object instance. We use an object detector~\cite{ridnik2021mldecoder} trained on real and synthetic images of these known objects. At test time, we run this detector on the test image to obtain a 2D bounding box of the object and its corresponding 3D model $\mathcal{M}$.

This bounding box and 3D model are used in a {\em render and compare} approach, illustrated in Fig.~\ref{fig:method_overview}, which iteratively estimates the focal length and 6D pose of the identified object. We denote the current estimate of focal length and 6D pose in iteration $k$ jointly as $\theta^k$.
The object model is first {\em rendered} using the current estimates $\theta^k$ into an image $\mathcal{R}(\mathcal{M},\theta^k)$ using a renderer $\mathcal{R}$. 
The rendering $\mathcal{R}(\mathcal{M},\theta^{k})$ and cropped input image $I$  are given to a deep neural network $F$ which predicts the pose and focal length update $\Delta \theta_k$:
\begin{equation}
    \Delta\theta_k=F(I,\mathcal{R}(\mathcal{M},\theta^k )).
    \label{eq:network-predictions}
\end{equation}
The intuition is that the neural network {\em compares} the cropped input image $I$ with the rendering $\mathcal{R}(\mathcal{M},\theta^k)$ and based on their (potentially subtle) differences predicts the update in the rendering parameters $\Delta\theta_k$. 
The pose and focal length updates $\Delta\theta_k$ are designed to be, as much as possible, free of nonlinearities and thus easy to predict by the neural network $F$. The pose and focal length at the next iteration $k+1$ is then computed by a non-linear update rule $U$:
\begin{equation}
    \theta^{k+1}=U(\theta^k,\Delta\theta_k),
    \label{eq:update-rule}
\end{equation}
where $\theta^{k}$ is the current estimate of the pose and focal length, $\Delta\theta_k$ is the prediction by the network $F$ given by eq.~\eqref{eq:network-predictions}, and $\theta^{k+1}$ are the updated pose and focal length.
Note that $U$ is not learned but derived from the 3D to 2D projection model and takes into account the non-linearities of the imaging process.
The neural network $F$ is trained in such a way that the updated pose and focal length $\theta^{k+1}$ are progressively closer to their ground truth. Our approach is summarized in Fig.~\ref{fig:method_overview}

\paragraph{Discussion.} Existing render-and-compare estimators~\cite{labbe2020cosypose,li2018deepim} require knowledge of the camera intrinsic parameters.
In our scenario, the problem is more challenging because the rendering also depends on the unknown focal length. 
We address this challenge by proposing an update rule for the focal length as well as a modification of the update rules for the 6D pose parameters that account for the unknown focal length (Sec~\ref{approach-inference}). We then introduce a novel loss function adapted for joint focal length and 6D pose estimation, which disentangles the effects of pose and focal length updates for better end-to-end training of the network (Sec.~\ref{approach-training}). Please, see Sec.~\ref{approach-details} for details of our implementation, $\theta^{0}$ parameter initialization, and our training data.

\subsection{Update rules with focal length estimation}
\label{approach-inference}
The standard render-and-compare approach to 6D pose estimation~\cite{li2018deepim,labbe2020cosypose} considers only translation $t^{k}$ and rotation $R^k$ as parameters $\theta^k$. We additionally estimate the focal length $f^k$ as unknown and thus need to build an appropriate rule $U$ (as defined in eq.~\eqref{eq:update-rule}) for updating jointly all parameters. In detail, we assume a pinhole camera model with focal length $f_{x}^{k}=f_{y}^{k}=f^k$ in which the optical center is set at the center of the image. We define the 6D pose of the object with respect to the camera by a 3D rotation $R^k$ and a 3D translation $t^k=[x^k,y^k,z^k]$. Next, we describe our updates for focal length and 6D pose. 

\paragraph{Focal length update.} To build an appropriate focal length update rule, we take into account the fact that it should remain strictly positive throughout the update iterations. We consider update rules that are multiplicative, \ie, they scale an initial guess $f^0$ by a sequence of multiplications. 
Let $f^{k}$ be the current estimate of the focal length at iteration $k$ and $v_{f}^{k}$ be the focal length update predicted by the network $F$ (see eq.~\eqref{eq:network-predictions}). We define the updated focal length $f^{k+1}$ as the multiplication,
\begin{equation}
\label{eq:f_update}
    f^{k+1} = e^{v_f^{k}}f^{k}.
\end{equation}
The sequence of multiplicative updates can be written as $f^{k+1} = e^{\sum_{i = 1}^{k} v_f^i}f^0$, where $f^0$ is the initial focal length and $v_{f}^i, \, i\in \{1,\ldots,k\}$ are the individual updates. An alternative to the above strategy would be to enforce the positivity of the focal length update via a sigmoid function instead of an exponential function. We found the exponential and sigmoid functions to behave similarly, but the sigmoid update requires setting an additional scale parameter. Hence, we opt for the simpler exponential updates as described in eq.~\eqref{eq:f_update}.

\paragraph{6D pose update.}
For the update of the 6D pose, we build on the update rule introduced in DeepIM~\cite{li2018deepim} that disentangles 3D rotation and 3D translation updates.
In more detail, the network $F$ is trained to predict a translation of the projected object center into the image $[v_{x}^{k},v_{y}^{k}]$ (measured in pixels), and a ratio $v_{z}^{k}$ of the camera-to-object depth between the observed and the rendered image. 
The 3D translation of the object is then updated from the quantities $[v_{x}^{k},v_{y}^{k}, v_{z}^{k}]$ predicted by the network $F$, taking into account the nonlinear projection equations derived from the camera model. In~\cite{li2018deepim} the focal length is known and fixed.
In our scenario, the focal length is not fixed and we use both the previous focal length estimate $f^k$ and the new predicted focal length $f^{k+1}$.
In detail, the updated 3D translation $[x^{k+1},y^{k+1},z^{k+1}]$ of the object with respect to the camera is obtained as:


\begin{align}
\label{eq:new-rules-x}
x^{k+1} &= \left(v_x^{k} + \frac{f^k x^k}{z^k}\right)\frac{z^{k+1}}{f^{k+1}}\\
\label{eq:new-rules-y}
y^{k+1} &= \left(v_y^{k} + \frac{f^k y^k}{z^k}\right)\frac{z^{k+1}}{f^{k+1}}\\
\label{eq:new-rules-z}
z^{k+1} &= v_z^{k} z^k,
\end{align}
where $[v_{x}^{k},v_{y}^{k},v_{z}^{k}]$ are the object translation updates predicted by network $F$ as part of $\Delta\theta$ (eq.~\ref{eq:network-predictions}), $[x^{k},y^{k},z^{k}]$ is the 3D translation vector of the relative camera-object pose at iteration $k$, $[x^{k+1},y^{k+1},z^{k+1}]$ is the new updated 3D translation vector, $f^{k}$ is the focal length from the previous iteration, and $f^{k+1}$ is the updated focal length of the camera given by eq.~\eqref{eq:f_update}. For detailed derivation of this update rule, please refer to the supplementary material.

To obtain the update of the rotation component of the object pose we use directly the prediction of the alignment network $F$ in a multiplicative update, which does not depend on the focal length. In particular, we parameterize the rotation update using two 3-vectors $v_{R,1}^{k}$, $v_{R,2}^{k}$ that define the rotation matrix $R(v_{R,1}^{k},v_{R,2}^{k})$ by Gram-Schmidt orthogonalization as described in~\cite{Zhou2018-eg}. This parameterization was found to work well for different prediction tasks~\cite{Zhou2018-eg} including 6D object pose estimation~\cite{labbe2020cosypose}. The resulting update rule is then written as 
\begin{equation}
\label{eq:R_update}
R^{k+1} = R(v_{R,1}^{k},v_{R,2}^{k}) R^{k},
\end{equation}
where $R^{k+1}$ is the new updated object rotation, $R^{k}$ is the current object rotation, and 
  $R(v_{R,1}^{k},v_{R,2}^{k})$ is the rotation matrix obtained by Gram-Schmidt orthogonalization from the two 3-vectors $v_{R,1}^{k}$, $v_{R,2}^{k}$ predicted by the alignment network $F$ as part of $\Delta\theta_k$.
Note that this rotation update is similar to that used in DeepIM~\cite{li2018deepim}.

\subsection{Pose and focal length training loss}
\label{approach-training}
We now present our network training loss, where we assume that the training data consist of image and aligned model pairs. Note that a training pair may be a real image with a manually aligned model or a rendered image of a model under a specified 6D pose and focal length. 
Given input parameters $\theta^{k}$, 
the output parameters $\theta^{k+1}$ are fully defined by the network outputs $\Delta\theta$ given by eq.~\eqref{eq:network-predictions} and the differentiable update rules described by eqs.~\eqref{eq:f_update}-\eqref{eq:R_update} in the previous section. In the following, we consider a single network iteration and denote $\theta=\{R, t, f\}$ as the estimated parameters.
For jointly learning to estimate the 6D pose and the focal length, we use the following loss that penalizes errors in the output 6D pose predictions $(R,t)$ and the estimated focal length $f$:
\begin{equation}
\begin{split}
\mathcal{L}(\theta, \hat{\theta}) &= \mathcal{L}_{\text{pose}}((R, t), (\hat{R}, \hat{t})) \\ &+ \alpha \mathcal{L}_{\text{focal}}((R, t, f), (\hat{R}, \hat{t}, \hat{f})),
\end{split}
\label{eq:full-loss}
\end{equation}
where $\theta=\{R,t,f\}$ are the estimated pose and focal length parameters, $\hat{\theta}=\{\hat{R},\hat{t},\hat{f}\}$ are the ground truth pose and focal length parameters, $\mathcal{L}_{{pose}}$ is a loss that penalizes errors in the 6D pose estimate, $\mathcal{L}_{focal}$ is our novel loss function that jointly takes into account the errors in the focal length and the 6D predicted pose, and $\alpha$ is a scalar hyper-parameter. 
This loss is written for a single instance, but our model is trained to minimize the average loss over all training images.
We now describe the individual losses $\mathcal{L}_{focal}$ and $\mathcal{L}_{pose}$.

\paragraph{Focal length loss.}
We use the following focal length loss:
\begin{equation}
    \mathcal{L}_{\text{focal}} = \beta \mathcal{L}_{H} (f, \hat{f}) + \mathcal{L}_{DR} ((R, t, f), (\hat{R}, \hat{t}, \hat{f})),
\label{eq:focal-length-loss}
\end{equation}
where  $\mathcal{L}_{H}$ is Huber regression loss, $\mathcal{L}_{DR}$ is disentangled reprojection loss and $\beta$ is a scalar hyper-parameter. The individual terms are explained next.
The Huber regression loss $\mathcal{L}_{H}$ measures the errors between the estimated and the ground truth focal length using a logarithmic parametrization of the focal length following the recommendations from Grabner \etal~\cite{grabner2019gp2c} for better training:
\begin{equation}
\begin{split}
    \mathcal{L}_{H}(f,\hat{f}) &= ||\log(f) - \log(\hat{f})||_{H},
\end{split}
\end{equation}
where again $\hat{f}$ is the ground truth focal length and $f$ is the focal length estimated by our model.

Although using only the loss $\mathcal{L}_{H}$ is possible to train our model, we found that better results are obtained by also considering the 2D errors of the projected 3D model in the image using the current estimates of the focal length and the object 6D pose. We first define the reprojection error:
\begin{equation}
  \begin{split}
  \mathcal{L}_{\text{proj}}((R, t, f), (\hat{R}, \hat{t}, \hat{f})) = \\
  \sum_{p \in \mathcal{M}}||\pi{\left(K(f), R, t, p\right)} - \pi{\left(K(\hat{f}), \hat{R}, \hat{t}, p\right)}||_1,
    \label{eq:proj}
\end{split}
\end{equation}
where $K(f)$ is the intrinsic camera matrix of our camera model with focal length $f$, $p\in\mathcal{M}$ are 3D points sampled on the object model, $\pi(K(f), R, t, p)$ is the projection of a 3D point $p$ using the current estimates of all the parameters, and $\pi(K(\hat{f}), \hat{R}, \hat{t}, p)$ is the projection of the same 3D point $p$ using ground truth parameters. This loss can be seen as the counterpart of the pose loss $\mathcal{L}_{pose}$ (defined below): instead of penalizing errors in 3D space, it penalizes reprojection errors in the image while also taking into account the estimated focal length $f$. However, this loss does not disentangle the effects of the pose and focal length predictions. We thus introduce our disentangled reprojection loss:
\begin{align}
\label{eq:DR_1}
  \mathcal{L}_{DR} &=  \frac{1}{2} \mathcal{L}_{\text{proj}}((R, t, \hat{f}), (\hat{R}, \hat{t}, \hat{f})) \\
  &+ \frac{1}{2} \mathcal{L}_{\text{proj}}((\hat{R}, \hat{t}, f), (\hat{R}, \hat{t}, \hat{f})),
   \label{eq:DR_2}
\end{align}
where each term separately measures the 2D reprojection errors resulting from errors in the 6D pose (the first term) and in the focal length (the second term). This disentanglement leads to faster convergence and better model accuracy, as we show in our ablation results. 

\paragraph{6D pose loss.} For $\mathcal{L}_{pose}$ (in eq.~\eqref{eq:full-loss}), we build on the loss used in CosyPose~\cite{labbe2020cosypose}. This loss is based on the point-matching loss~\cite{Xiang2018-dv,li2018deepim} that measures the error between the alignment of the points on the 3D model $\mathcal{M}$ transformed with the predicted pose $(R,t)$ and the ground truth pose $(\hat{R},\hat{t})$. CosyPose~\cite{labbe2020cosypose} extends this loss to take into account object symmetries and uses the disentanglement ideas of~\cite{Simonelli2019-da} to separate the influence of translation errors along the camera axis, image plane, and rotations.
In our approach, we do not consider object symmetries as they are nontrivial to obtain for 3D models in the wild considered in this work.
In detail, for the pose loss we utilize the following distance metric between two poses specified by $\{R_1, t_1\}$ and $\{R_2, t_2\}$:
\begin{equation}
    D(\{R_1, t_1\},\{R_2, t_2\}) = \frac{1}{|\mathcal{M}|} \sum_{p \in \mathcal{M}} ||(R_1p + t_1) - (R_2p + t_2)||_1,
    \label{eq:distance-pose}
\end{equation}
where $||\cdot||_1$ denotes $L_1$ norm, $R_i$ is a rotation matrix, $t_i$ is a translation vector and $p \in \mathcal{M}$ is a point sampled from the mesh $\mathcal{M}$. Following~\cite{labbe2020cosypose}, we disentangle the pose loss as
\begin{align}
    \begin{split}
        \mathcal{L}_{\text{pose}} &= D(U(\theta^k, \{v_x^k, v_y^k, \hat{v}_z^k, \hat{R}^k, \hat{v}_f^k\}), \hat{R}, \hat{t})    \\
            &+ D(U(\theta^k, \{\hat{v}_x^k, \hat{v}_y^k, v_z^k, \hat{R}^k, \hat{v}_f^k\}), \hat{R}, \hat{t}) \\
            &+ D(U(\theta^k, \{\hat{v}_x^k, \hat{v}_y^k, \hat{v}_z^k, R^k, \hat{v}_f^k\}), \hat{R}, \hat{t}),
    \end{split}
    \label{eq:disent-pose-loss}
\end{align}
where $\theta^k$ are the pose and focal length parameters at iteration $k$, $\hat{R}$ is a ground truth rotation, $\hat{t}$ is a ground truth translation, $D$ is a distance defined by eq.~\eqref{eq:distance-pose} and $U$ is an update function defined by \eqref{eq:update-rule}. The main idea of this loss is to separate the influence of translation errors in the $x-y$ plane, depth alignment errors along the $z$ axis, and rotation errors. In eq.~\eqref{eq:disent-pose-loss} the terms $\{v_x^k, v_y^k, \hat{v}_z^k, \hat{R}^k, v_f^k\}$, $\{\hat{v}_x^k, \hat{v}_y^k, v_z^k, \hat{R}^k, \hat{v}_f^k\}$ and $\{\hat{v}_x^k, \hat{v}_y^k, \hat{v}_z^k, R^k, \hat{v}_f^k\}$  represent the necessary updates that lead to such loss disentanglement. Here $[v_x^k, v_y^k, v_z^k]$ are translation updates at iteration $k$ as predicted by the network $F$, $R^k$ is a rotation update at iteration $k$ predicted by the network $F$ and $v_f^k$ is a focal length update at iteration $k$. The terms $\hat{v}_i^k$ and $\hat{R}^k$ then represent the updates needed to transform the current parameters into the ground truth values, which leads to the disentanglement along each of the dimensions. The first term in eq.~\eqref{eq:disent-pose-loss} leads to the disentanglement along the $x-y$ axis, since this term provides the gradients resulting from the $x-y$ alignment errors. Analogously, the second and third terms provide gradients that arise from depth and rotation alignment errors.

\subsection{Training data}
\label{sec:training_data}
The available datasets of images with objects annotated with 6D poses and focal lengths are quite small. This is especially true for the Pix3D dataset, which is split into individual object classes. Hence, training on such data is challenging. To address this issue, we train our neural network $F$ using the combination of real-world training data and synthetically generated data.
Synthetic data are generated by randomly sampling the object model, the 6D pose, and the focal length that are used to render the synthetic image with a random texture on the object. Usually, the common distribution of viewpoints is not uniform, \eg we usually do not see beds or cars upside down.
To respect this natural bias also in the synthetic dataset, the object pose and camera focal length are sampled from a parametric distribution fitted to the real training data. This is in contrast to~\cite{ponimatkin2022focal} which used a uniform distribution.

In detail, to model the distribution of 3D object rotations we use Bingham~distribution~\cite{bingham1974spherical}, an antipodally symmetric probability distribution on a surface of unit hyper-sphere. A 3D rotation can be represented as a unit quaternion, i.e. a point on unit 4D hyper-sphere. However, for an arbitrary unit quaternion~$q$, the quaternions $q$ and $-q$ represent the same rotation. The Bingham distribution, being antipodally symmetric, reflects this and can therefore be used to describe the rotation distribution on the SO(3) group. Before rendering, we fit the distribution parameters to the real training dataset, \ie orthogonal matrix $M \in \mathbb{R}^{4\times4}$ and diagonal matrix $Z=\textrm{diag}(z_1,z_2,z_3,0)\in\mathbb{R}^{4\times4}$. During the rendering, we sample from the distribution to acquire the unit quaternion representing the object rotation. We directly use the implementation from~\cite{Gilitschenski2020} for both fitting and sampling.

For the translations and focal lengths, we use two 2D normal distributions. We observe that the focal length $f$ and the $z$-component of translation are highly dependent as they both affect the object scale after projection. Also, we observe that after taking logarithm of both components, they follow the 2D normal distribution. We therefore fit one 2D normal distribution to focal lengths and $z$-translations after application of the logarithm function and separately fit one 2D normal distribution to $xy$-translations. During synthetic data generation, we sample from the fitted distributions and use the exponential function to get the correct values of $z$ and $f$. 

\begin{figure*}[tbp]
    \centering
    \hspace{-7mm}
    \includegraphics[height=56mm]{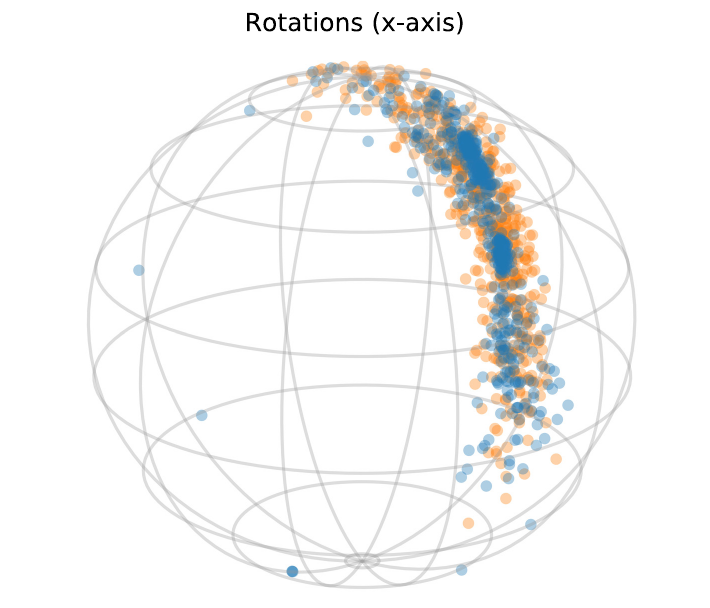}
    \hspace{-6mm}
    \includegraphics[height=56mm]{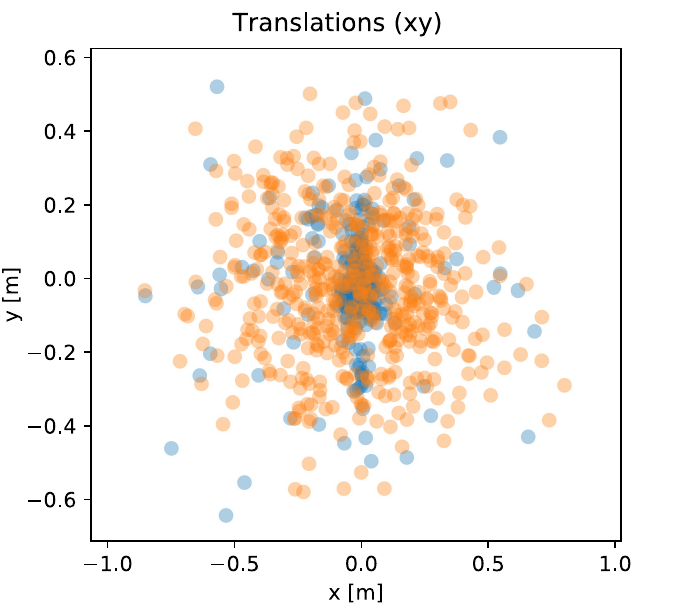}
    \hspace{-6mm}
    \includegraphics[height=56mm]{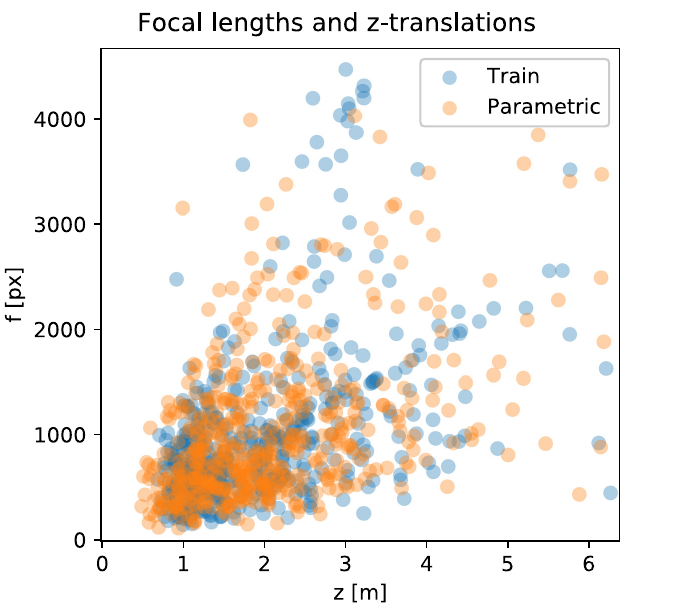}\\
    \vspace{-3mm}
    \caption{\textbf{ Parametric distribution of object poses and focal lengths in the training data.} We plot the poses and focal lengths of the real training dataset of Pix3D-sofa class (blue) together with poses and focal lengths sampled from the parametric distribution fitted to the data (orange). The number of samples from our distribution is the same as the number of data points in the real training dataset. We plot the rotations, xy-translations, and z-translations with focal lengths separately. To visualize the rotations, we plot the unit x-vector multiplied by the sampled rotations.}
    \label{fig:distr_param}
    \vspace{-3mm}
\end{figure*}

Fig.~\ref{fig:distr_param} visualizes samples from our parametric distribution together with 6D poses and focal lengths from the real Pix3D-sofa dataset. We describe the details of the parametric distributions and compare them with two other distributions (nonparametric and uniform) in Sec.~\ref{sec:ablation_synth_distr} and show that introducing a bias towards the real dataset increases the performance of our method.

\section{Experiments}
We evaluate our method for focal length and 6D pose estimation on three challenging benchmarks: the Pix3D~\cite{pix3d}, CompCars~\cite{wang20183d} and
StanfordCars~\cite{wang20183d} datasets.
In this section, we introduce the benchmark datasets and give details of the full pose estimation pipeline.  
Then, in Sec.~\ref{approach-details} we provide the implementation details of our approach. Sec.~\ref{sec:ablation_synth_distr} presents the ablation of different synthetic data distributions. In Sec.~\ref{synth-data-ablation} we provide an ablation of the ratio between real data and synthetic imagery. Sec.~\ref{sec:ablation_retrieval} ablates the 3D model retrieval method and Sec.~\ref{sec:update_rule_ablation} ablates the 6D pose update rule. In Sec.~\ref{sec:ablation_loss} we present the ablation of the main components of the proposed loss function. In Sec.~\ref{sec:sota} we compare our method with the state of the art~\cite{wang2018improving,grabner2019gp2c,han2020gcvnet} addressing the same task, and in Sec.~\ref{sec:6d_pose_benchmarks} we show a comparison on the YCB-Video~\cite{Xiang2018-dv} dataset using the metrics from the BOP~Challenge~\cite{hodan2024bop}, \ie a benchmark for 6D pose estimation with a calibrated camera. Sec.~\ref{sec:error_bars} reports the training stability between different runs, including rendering of new synthetic data. In Sec.~\ref{sec:limitations} we discuss the main limitations of our approach. Finally, in Sec.~\ref{sec:applications} we show applications of our method in computer graphics and robotics.

\paragraph{Datasets and evaluation criteria.} We consider three real-world in-the-wild datasets depicting objects with known 3D models annotated with ground-truth focal length and 6D pose of the object. Following Grabner~\etal~\cite{grabner2019gp2c}, we consider the {\em bed}, {\em chair}, {\em sofa}, and {\em table} classes in the Pix3D dataset~\cite{pix3d}. The images for each object class are considered as separate datasets. The Stanford cars and CompCars datasets~\cite{wang20183d} contain images of different car instances. Note that for the Pix3D chair images and both cars datasets, there are hundreds of different object instances in the dataset, which makes the task of recognizing the object instance challenging. 
We use the standard set of evaluation criteria used by prior work~\cite{wang2018improving,grabner2019gp2c,han2020gcvnet} that include detection accuracy and several 6D pose metrics.
The results are reported as median errors (smaller is better) between the prediction and ground truth (\eg, the $MedErr_R$ is the median rotation error) and accuracies (higher is better), which report the percentage of images with an error smaller than a certain threshold (\eg, $Acc_{R\frac{\pi}{6}}$ reports the percentage of test images with a rotation error smaller than $\frac{\pi}{6}$).    
See the supplementary material for a detailed description of all evaluation criteria.

\paragraph{The complete pose estimation pipeline.}
The first step of our pipeline returns bounding box coordinates for depicted model instances in the input image via a Mask R-CNN detector. One detector is trained for each object class. For each detected instance, we crop the input image given the bounding box and apply an instance classifier to obtain which 3D model instance to align. In our case we use the ML-Decoder~\cite{ridnik2021mldecoder}, an attention-based classification head with TResNet-L~\cite{ridnik2020tresnet} backbone, as the instance classifier. We align the 3D model instance corresponding to the top classifier score. Next, we estimate the coarse 6D pose and focal length using the full image, bounding box, and retrieved 3D model instance. Finally, the refiner FocalPose model iteratively refines the estimates for $N$ iterations given the coarse estimates.
Note, that we follow CosyPose~\cite{labbe2020cosypose} and
use two separate networks for coarse initialization and iterative refinement. Both networks follow the approach presented in Fig.~\ref{fig:method_overview}.

\subsection{Implementation details}
\label{approach-details}
We base our implementation on the render-and-compare approach of CosyPose~\cite{labbe2020cosypose} for 6D object pose estimation. We recall the main implementation details and explain the differences with~\cite{labbe2020cosypose}.

\paragraph{Network architecture.} The architecture of the network $F$ (eq.~\ref{eq:network-predictions}) relies on a ResNet-50~\cite{he2016deep} backbone, followed by average pooling and a linear layer for predicting the update $\Delta \theta$. The first input block in the backbone is inflated from 3 to 6 channels, to allow for the input of the merged RGB input image and the RGB rendered view.

 \paragraph{Data augmentations.} The (cropped) input image and rendering are resized to the input resolution: $640\times640$ for Pix3D dataset and $300\times200$ for StanfordCars and CompCars datasets. During training, one of the real data examples is sampled with probability $0.5\%$ while a synthetic data example is sampled with probability $99.5\%$. 
 Following~\cite{labbe2020cosypose}, we also use data augmentation to increase the variability of training images. Data augmentation includes adding blur, contrast, brightness, color, and sharpness image filters to the image, and replacing the background with an image from the Pascal VOC dataset~\cite{Everingham15}. We replace the background each time we encounter a synthetic image, while randomly replacing the background of real images with probability $10\%$. This is done by replacing pixels at positions where the object mask, provided with the training data, is equal to zero.

\paragraph{Initialization.} In all experiments, we set the initial focal length $f^{0} = 600$ pixels, which we found experimentally to be a good initial value for all datasets. This focal length could also be initialized using an EXIF file, or using a coarse estimate directly predicted by a different method.
The initialization of the 6D object pose $T^0$ follows~\cite{labbe2020cosypose} but relies on the initial focal length $f^{0}$ instead of using the ground truth focal length for computing an approximation of the object 3D translation. The initial depth of the object is set to $z = 1$ m, and the $x-y$ components of the 3D translation are analytically derived by computing the 3D position of the object center that reprojects to the center of the 2D detection, assuming the camera projection model defined by $f^{0}$. The initial object rotation is set to the identity: $R^0=I_{3}$.

\paragraph{Coarse estimate and refinement.} 
We follow CosyPose~\cite{labbe2020cosypose} and use two separate networks for coarse initialization and iterative refinement. The coarse network corrects the largest errors (between the observed state and the fixed initialization $\theta^{0}$) during the first iteration $k=1$. A separate refinement network iteratively refines the estimates by correcting smaller errors. The refinement network runs for multiple iterations, we run $K$ iterations of the refinement network at test time in our experiments, with $K=15$ on Pix3D and $K=55$ on the Stanford cars/CompCars datasets in our experiments.

\paragraph{Training input error distribution.} We use the same network architecture defined above for the coarse and refinement networks, but both are trained with different error distributions to simulate what each network will see at test time. During training, the initialization of the coarse network is the same as the one used at test time and described in the previous paragraph. Simulating the error distribution of the refinement network is more complicated as its input is not fixed and depends on the coarse estimate. To simulate the errors in focal length that the refinement network will see, we sample the focal length $f^{k}$ from a Gaussian distribution centered on the ground truth $f^{\text{gt}}$, with variance $0.15 f^{\text{gt}}$. The error of the input pose given to the refiner is sampled from a Gaussian with standard deviation of 1 cm around the $x-y$ components of translation, $5$ cm for the depth, and noise is added to the ground truth rotation matrix using three Euler angles sampled from Gaussian distributions with variance of $15^\circ$.

\paragraph{Training procedure.} The coarse and refinement networks are initialized using a classification network pretrained on ImageNet, and are trained using the same procedure as in~\cite{labbe2020cosypose}. Training is performed on 40 NVIDIA A100 GPUs using a global batch size of 1280. The average training time for one coarse/refiner model is around 5 hours. Each network is trained for $10$M iterations using the Adam optimizer~\cite{DBLP:journals/corr/KingmaB14} with a learning rate of $3\times 10^{-4}$. We use a linear warm-up of the learning rate during the first $700$K iterations and decrease it to $3\times 10^{-5}$ after $7$M iterations. During inference, the network can process 32 $640\times 640$ pixel resolution images in approximately 10 seconds. This time includes coarse estimation and 15 refiner iterations.

\paragraph{2D detection and instance-recognition.} We use Mask R-CNN~\cite{he_2017_iccv} to predict a 2D bounding box of the object of interest, and ML-Decoder\cite{ridnik2021mldecoder} as object instance classifier. 
The Mask R-CNN is based on a ResNet-50~\cite{he2016deep} feature pyramid (FPN) backbone~\cite{lin2017feature}. The network is initialized from a network trained on MS COCO, and the first ten convolutional layers remain fixed during training. This detector is trained using only the data provided by the Pix3D and Stanford/Comp cars datasets.
The ML-Decoder~\cite{ridnik2021mldecoder} is an attention-based classification head, we use the network version with TResNet-L~\cite{ridnik2020tresnet} as a backbone. We initialize it from a network pretrained on MS COCO and fine-tune it using the real datasets with 1000 added synthetic images to increase the classifier performance.

\paragraph{Cropping strategy.} The images from the datasets are center cropped to $640 \times 640$px for Pix3D and $300 \times 200$px for Stanford cars and CompCars. The input image is padded to preserve the input aspect ratio.
The second cropping happens before the input to the network itself. Let us call $(x_c, y_c)$ the 2D coordinates resulting from the projection of the 3D object center by the camera with the intrinsic parameter matrix $K$ and $[x_1, y_1, x_2, y_2]$ the coordinates of the bounding box provided by external means (for example, the Mask R-CNN detector), where $[x_1,y_1]$ are the upper-left corner coordinates and $[x_2,y_2]$ are the lower-right corner coordinates of the provided bounding box. Then we define 
\begin{equation}
    x_{\text{dist}} = \max(|x_1 - x_c|, |x_2 - x_c|),
\end{equation}
\vspace{-4mm}
\begin{equation}
    y_{\text{dist}} = \max(|y_1 - y_c|, |y_2 - y_c|).
\end{equation}
Then, the cropped image width and height are given by
\begin{equation}
    w = \max(x_{\text{dist}}, y_{\text{dist}}/r)\cdot 2\lambda,
\end{equation}
\vspace{-4mm}
\begin{equation}
    h = \max(x_{\text{dist}}/r, y_{\text{dist}})\cdot 2\lambda,
\end{equation}
where $r$ is the aspect ratio of the input image and $\lambda = 1.4$ is a parameter controlling the enlargement of the input image to capture the whole object. This value was chosen following~\cite{li2018deepim}.
To~ensure the same perspective effects as in the original input image, we adjust the intrinsic parameter matrix $K$ when resizing and cropping the image. In particular, we change the principal point $[c_x,c_y]$ to keep it in the same image position when resizing, cropping, and padding the image. Note that image cropping and padding does not affect the focal length. We change the focal length only when resizing the image. In particular, we change the focal length proportionally to the ratio of the sizes between the new and the old images.

\paragraph{Loss weights.} We utilize $\alpha = 10^{-2}$ and $\beta = 1$ as weights for the losses given by eq.~\ref{eq:full-loss} and eq.~\ref{eq:focal-length-loss}.

\subsection{Ablation of different synthetic data distributions}
\label{sec:ablation_synth_distr}
When rendering synthetic data for training, we need to sample object's 6D pose and camera's focal length. We investigate the effect of distribution choice on overall pipeline performance. The intuition is that introduction of bias towards the distribution of the real training data can increase the performance, as long as the test data samples are from the same distribution. 
We compare here the parametric distribution used in our approach, and described in detail in~Sec.\ref{sec:training_data}, with two other distributions, uniform and nonparametric, described next. We show that the parametric distribution works the best, as reported in~Table~\ref{table:distribution_ablation}.

\paragraph{Uniform Distribution.}
The simplest way to sample the training data is using a uniform distribution. In that case, we sample the rotation of the object uniformly in the SO(3) space and sample its 3D position within a box of size of 15 cm. The camera-to-object distance is sampled within the interval (0.8, 3.0) meters for the StanfordCars and CompCars datasets and (0.8, 2.4) meters for Pix3D. The focal length is sampled within (200, 1000) pixels, which covers the range of focal lengths from all datasets.

\paragraph{Nonparametric Distribution.}
When using the nonparametric distribution, we select the hyperparameters $\delta_R$, $\delta_x$, $\delta_y$, $\delta_z$ and $\delta_f$ as explained in the following paragraph. When sampling, we randomly select a data point from the corresponding real training dataset and perturb the rotation, translation, and focal length. We adjust the rotation by a random angle $\alpha \in [0,\delta_R]$ around a random unit vector. For translation and focal length, we use additive perturbations. However, we entangle the $z$-axis of translation together with focal length, and the $x$-axis together with the $y$-axis, since the focal length and distance from camera are highly dependent. In detail, for each of the pairs, we sample a random vector within an ellipse with axes lengths $\{\delta_z,\delta_f\}$ and $\{\delta_x,\delta_y\}$ respectively, and use the components of the sampled vectors as an additive perturbation.
The hyperparameters $\delta_x$, $\delta_y$, $\delta_z$ and $\delta_f$ are selected using the real dataset by finding the nearest neighbor of each data point, computing the Euclidean distance, and then taking the 95 percentile from all the distances (also computed separately for the $z$-axis with focal length and the $x$-axis with $y$-axis). The $\delta_R$ is computed in a similar way, except that we measure the angle between two rotations instead of using the Euclidean distance.

\begin{table}[t]
    \centering
    \caption{\textbf{Ablation of different synthetic training data distributions on Pix3D sofa.} When rendering synthetic images for training, we sample the object's 6D pose and camera's focal length randomly. We find that sampling from distribution that is close to the training data distribution (b. and c.) improves the performance compared to sampling from uniform distribution (a.). The parametric distribution performs the best.}
    \label{table:distribution_ablation}
    \small
    \setlength{\tabcolsep}{3.2pt}
    \begin{tabular}{l|c|c|c}
    \toprule
    Distribution & $MedErr_R$ & $MedErr_t \cdot 10$ & $MedErr_f \cdot 10$ \\
    \midrule
        a. Uniform & 2.95 & 1.27 &  1.34 \\
        b. Nonparametric & 2.82 & {\bf 1.14} & 1.24 \\
        c. Parametric & {\bf 2.77} & {\bf 1.14} & {\bf 1.18} \\
		\bottomrule
    \end{tabular}
\end{table}

\subsection{Ablation of real vs. synthetic training data}
\label{synth-data-ablation}
Manually annotating real in-the-wild images~\cite{wang20183d,wang2018improving} with the focal length and 6D pose is difficult because it requires significant effort and the ambiguities can be hard to resolve.
This setting results in relatively few training images available. Moreover, the annotations are often of poor quality (see Sec.~\ref{sec:sota} and in Fig.~\ref{fig:qual_examples}). Using synthetic data allows one to generate many images with accurate annotations. In Table~\ref{tab:data_ablation}, we report the results of our coarse model trained with only real data, only synthetic data, or a mix of synthetic and real data in each mini-batch (the fraction of real data in the mixed-data mini-batch is indicated in the table row). Using (exact) synthetic data in addition to a small number of (human-labeled) real images in each mini-batch yields the lowest median error. 
 Please note that this ablation was performed using the original FocalPose version of the model as presented in~\cite{ponimatkin2022focal}.
\begin{table}[t]
    \centering
    \caption{\textbf{Ablation for combining real and synthetic training data on Pix3D sofa dataset.} Mix of mostly synthetic data with a small number of real images in each mini-batch performs best.}
    \label{tab:data_ablation}
    \small
    \setlength{\tabcolsep}{3.2pt}
    \begin{tabular}{l|c|c|c}
    \toprule
    Dataset & $MedErr_R$ & $MedErr_t \cdot 10$ & $MedErr_f \cdot 10$ \\
    \midrule
         Synth only & 5.44 & 2.18 & 2.04 \\
		 Synth + Real 0.5\% & \textbf{2.98} & \textbf{1.29} & \textbf{1.36} \\
		 Synth + Real 5\% & 3.08 & 1.33 & 1.40 \\
		 Real only & 4.13 & 1.92 & 1.91 \\
		\bottomrule
    \end{tabular}
\end{table}

\subsection{Ablation of different model retrieval methods}
\label{sec:ablation_retrieval}
Even though our approach is effective even with only an approximate 3D model, the selection of the correct 3D model affects the whole pipeline. In this section, we evaluate two different methods for retrieving the 3D model. In the previous version of this work~\cite{ponimatkin2022focal} we fine-tuned the DINO architecture~\cite{caron2021emerging} on real images as the instance classifier. We compare this approach with another method: the ML-Decoder~\cite{ridnik2021mldecoder}, an attention-based classification head with TResNet-L~\cite{ridnik2021mldecoder} backbone, which offers a reasonable speed-accuracy trade-off. We report the classifier accuracy on all 3 datasets in Table~\ref{table:retrieval_accuracy}. Furthermore, we render additional 1000 synthetic images and add them to the training set to increase classifier performance (denoted as ``ML-Decoder+1k'' in Table~\ref{table:retrieval_accuracy}).  Although the ML-Decoder significantly increases retrieval accuracy, the performance of the entire 6D pose estimation pipeline improves mainly in rotation error and does not change for other metrics, as reported in Table~\ref{table:retrieval_ablation}.

\begin{table}[th]
    \vspace{2mm}
    \caption{
    \textbf{Model retrieval accuracy.} The use of ML-Decoder instead of DINO model as an instance classifier significantly improves the model retrieval accuracy. For Pix3D dataset, where number of real training images is very low, the classification accuracy can be further increased by adding synthetic images to the training data (``ML-Decoder+1k'').
    }
    \label{table:retrieval_accuracy}
    \small
    \setlength{\tabcolsep}{3.2pt}
    \begin{tabular}{l|c|c|c}
    \toprule
    Dataset & Pix3D Acc & CompCars Acc & Stanford Acc \\
    \midrule
        \cite{caron2021emerging} DINO       & 62.1\% & 79.0\% &  71.2\% \\
        \cite{ridnik2021mldecoder} ML-Decoder & 72.8\% & 93.3\% & \textbf{95.1\%}\\
        \cite{ridnik2021mldecoder} ML-Decoder+1k & {\bf 77.6\%} & \textbf{93.5\%} & 94.7\%\\
    \bottomrule
    \end{tabular}
\end{table}

\begin{table}[t]
    \centering
    \caption{\textbf{6D pose estimation with improved model retrieval on Pix3D sofa.} Although the ML-Decoder has significantly higher model retrieval accuracy compared to the DINO model (see Table~\ref{table:retrieval_accuracy}), it has only moderate effects when incorporated in the whole 6D pose estimation pipeline where it reduces mainly the rotation error. Other metrics do not change much.}
    \label{table:retrieval_ablation}
    \small
    \setlength{\tabcolsep}{3.2pt}
    \begin{tabular}{l|c|c|c}
    \toprule
    Retrieval Method & $MedErr_R$ & $MedErr_t \cdot 10$ & $MedErr_f \cdot 10$ \\
    \midrule
        \cite{caron2021emerging} DINO & 3.00 & {\bf 1.13} & 1.20 \\
        \cite{ridnik2021mldecoder} ML-Decoder+1k & {\bf 2.77} & 1.14 & {\bf 1.18} \\
		\bottomrule
    \end{tabular}
\end{table}

\subsection{Ablation of the update rule}
\label{sec:update_rule_ablation}
Building on the previous version of this work (FocalPose~\cite{ponimatkin2022focal}), we further refined the $x$ and $y$ components of the 6D pose update rule. We evaluate the new update rule here. As shown in Table~\ref{table:update_rule_ablation}, the new update rule achieves slightly better results. For detailed derivation of the update rule and discussion of the difference compared to~\cite{ponimatkin2022focal}, please refer to the supplementary material.

\begin{table}[t]
    \centering
    \caption{\textbf{Update rule ablation on Pix3D sofa.} Compared to the original update rule used in FocalPose~\cite{ponimatkin2022focal}, our new update rule improves the 6D pose estimation performance.}
    \label{table:update_rule_ablation}
    \small
    \setlength{\tabcolsep}{3.2pt}
    \begin{tabular}{l|c|c|c}
    \toprule
    Update Rule & $MedErr_R$ & $MedErr_t \cdot 10$ & $MedErr_f \cdot 10$ \\
    \midrule
        \cite{ponimatkin2022focal} FocalPose & 2.81 & 1.16 & 1.19 \\
        FocalPose++ (ours) & \textbf{2.77} & \textbf{1.14} & \textbf{1.18} \\
		\bottomrule
    \end{tabular}
\end{table}

\subsection{Loss ablation study}
\label{sec:ablation_loss}

In this section, we ablate the different components of our proposed loss function.
We train the coarse and refinement networks with the three different losses introduced in Sec.~\ref{approach-training}. We report the results in Table~\ref{tab:loss_ablation}. First, our solution (c.) combining the Huber regression loss with the 2D reprojection error taking into account the object 3D model and its 6D pose results in significantly lower errors than simply using the regression loss (a.)\ used in Grabner \etal~\cite{grabner2019gp2c}. 
Second, our new loss (c.), which disentangles the effects of focal length and pose, results in lower median errors compared to the standard reprojection loss that does not disentangle pose and focal length (b.). Please note that this ablation was performed using the original FocalPose as presented in~\cite{ponimatkin2022focal}.

\begin{table}[t]
    \caption{\textbf{Training loss ablation on Pix3D sofa.} The median alignment errors for refinement models trained using different loss functions. Our proposed combination of Huber regression loss with a disentangled reprojection loss (c.) performs best. 
    }
    \label{tab:loss_ablation}
    \small
    \setlength{\tabcolsep}{3.2pt}
    \begin{tabular}{l|c|c|c}
    \toprule
    Loss & $MedErr_R$ & $MedErr_t \cdot 10$ & $MedErr_f \cdot 10$ \\
    \midrule
        a. $\mathcal{L}_H$ & 6.61 & 1.51 &  4.17 \\
		b. $\mathcal{L}_H + \mathcal{L}_{\text{proj}}$& 3.28 & 1.42 & 1.45 \\
		c. $\mathcal{L}_H + \mathcal{L}_{DR}$ & {\bf 2.98} & {\bf 1.29} & {\bf 1.36} \\
		\bottomrule
    \end{tabular}
\end{table}

\subsection{Comparison to the state-of-the-art}
\label{sec:sota}
Below we report the results of our approach (FocalPose++) on the three different datasets and compare with other methods for 6D object pose and focal length estimation~\cite{wang2018improving,grabner2019gp2c,han2020gcvnet}. We also compare our method (FocalPose++) to its previous version (FocalPose) as presented in \cite{ponimatkin2022focal}. In particular, in~\cite{ponimatkin2022focal}, the synthetic training dataset was sampled using the uniform distribution, the DINO~\cite{caron2021emerging} model was used as an instance classifier instead of the ML-Decoder~\cite{ridnik2021mldecoder}, and only an approximate version of the 6D pose update rule was applied during pose estimation. The differences between the two update rules are described in detail in the supplementary material.

\paragraph{Pix3D dataset.}
We report the average for the four classes (bed, chair, sofa, table) in Table~\ref{table:pix3d} (top). On average, our method (FocalPose++) significantly outperforms the other methods~\cite{wang2018improving,grabner2019gp2c,han2020gcvnet} in 5 out of the 8 metrics. In particular, we see a clear improvement in the projection error (36\% relative error reduction), translation error (30\% reduction), focal length and pose errors (22\% reduction) and rotation error (14\% reduction). We also note clear improvements over the original FocalPose~\cite{ponimatkin2022focal}.
Please note that the 3D translation is related to the focal length because of the focal length/depth ambiguity.
These improvements are significant and validate the contribution of our method. 

\paragraph{CompCars and Stanford cars.} A similar pattern of results is shown in Table~\ref{table:pix3d} (middle, bottom) also  for the CompCars and Stanford cars datasets that contain hundreds of different car models. Our approach obtains again the best results in 5 of the 8 reported metrics. In particular, for StanfordCars our method significantly improves the translation, pose and focal length errors (almost 50\% relative error reduction), but also the projection error (36\% reduction) and the rotation error (21\% reduction). For CompCars our method significantly improves the projection error (30\% relative error reduction), translation, pose and focal length errors (about 20\% reduction), and the rotation error (10\% reduction). Again, these improvements are significant and validate the contribution of our method.

\paragraph{Per class results on the Pix3D dataset} 
\label{per-class-pix3d}
In Table~\ref{table:pix3d-perclass} we show the performance of our FocalPose++ approach on individual Pix3D classes. 
For \textbf{bed}, \textbf{sofa} and \textbf{table} our algorithm clearly outperforms the prior methods in 5 out of 8 reported metrics, with relative error reduction ranging from 12\% to 56\%, which validates the contribution of our work. For \textbf{chair} our approach clearly outperforms other methods in the projection median error with almost 50\% relative error reduction. In rotation, translation, pose, and focal length errors, our new approach is slightly worse than the original FocalPose~\cite{ponimatkin2022focal}, but it still outperforms the other methods by 3\%-10\%.
All tested methods, including ours, perform significantly worse on the \textbf{table} class in the rotation and projection metrics. We believe that this could be attributed to the fact that tables are often symmetric, which makes the 6D object pose estimation hard and often ambiguous. Object symmetries are one of the main failure models of our approach (see Sec. \ref{sec:limitations}).

\definecolor{lightgreen}{RGB}{200,240,217}
\definecolor{lightred}{RGB}{240,200,200}
\begin{table*}[t]
	\centering
	\caption{\textbf{Comparison with the state of the art for 6D pose and focal length prediction} on the Pix3D, CompCars and Stanford cars datasets. \textbf{Bold} denotes the best result among directly comparable methods.
	Our approach clearly outperforms other competing methods in 5 out the 8 reported metrics on all three datasets (with a relative reduction in the median error ranging from 10\% to 50\%), validating our approach and demonstrating that our method deals well with the focal length/depth ambiguity.
	}
	\small
	\setlength{\tabcolsep}{3.2pt}
	\begin{tabular}{lccc|cc|c|c|c|cc}
		\toprule
		\multicolumn{3}{c}{}&\multicolumn{1}{c}{\bf Detection}&\multicolumn{2}{c}{\bf Rotation}&\multicolumn{1}{c}{\bf Translation}&\multicolumn{1}{c}{\bf Pose}&\multicolumn{1}{c}{\bf Focal}&\multicolumn{2}{c}{\bf Projection}\\

		\cmidrule(lr){4-4}\cmidrule(lr){5-6}\cmidrule(lr){7-7}\cmidrule(lr){8-8}\cmidrule(lr){9-9}\cmidrule(lr){10-11}
		\multirow{2}{*}{Method}&\multicolumn{1}{c}{\multirow{2}{*}{Dataset}}&&\multicolumn{1}{c}{\multirow{2}{*}{$Acc_{D_{0.5}}$}}&\multicolumn{1}{c}{$MedErr_R$}&\multicolumn{1}{c}{\multirow{2}{*}{$Acc_{R\frac{\pi}{6}}$}}&\multicolumn{1}{c}{$MedErr_{t}$}&\multicolumn{1}{c}{$MedErr_{R,t}$}&\multicolumn{1}{c}{$MedErr_f$}&\multicolumn{1}{c}{$MedErr_{P}$}&\multicolumn{1}{c}{\multirow{2}{*}{$Acc_{P_{0.1}}$}}\\
		&&\multicolumn{1}{c}{}&\multicolumn{1}{c}{}&\multicolumn{1}{c}{$\cdot1$}&\multicolumn{1}{c}{}&\multicolumn{1}{c}{$\cdot10^{1}$}&\multicolumn{1}{c}{$\cdot10^{1}$}&\multicolumn{1}{c}{$\cdot10^{1}$}&\multicolumn{1}{c}{$\cdot10^{2}$}\\
		\midrule
		\midrule
		\cite{wang20183d}&\multirow{5}{*}{Pix3D}&&96.0\%&7.25&87.8\%&2.52&1.76&2.41&6.33&71.5\%\\
		\cite{grabner2019gp2c}-LF&&&96.2\%&6.92&88.4\%&1.85&1.30&1.72&3.85&85.5\%\\
		\cite{grabner2019gp2c}-BB&&&\bf97.7\%&6.89&\bf{90.8\%}&1.94&1.30&1.75&3.66&\bf{88.0\%}\\
            \cite{ponimatkin2022focal}~FocalPose &&& 95.5\% & 4.92 & 84.1\% & 1.49 & 1.09 & 1.53 & 2.97 & 79.2\%\\
		FocalPose++ (ours) &&& 95.5\% & \textbf{4.19} & 85.1\% & \textbf{1.31} & \textbf{0.99} & \textbf{1.34} & \textbf{2.34} & 81.5\%\\
	    \midrule
		\cite{wang20183d}&\multirow{6}{*}{CompCars}&&\bf98.9\%&5.24&97.6\%&3.30&2.35&3.23&7.85&73.7\%\\
		\cite{grabner2019gp2c}-LF&&&98.8\%&5.23&97.9\%&2.61&1.86&2.97&4.21&95.1\%\\
		\cite{grabner2019gp2c}-BB&&&\bf98.9\%&4.87&98.1\%&2.55&1.84&2.95&3.87&\bf{95.7\%}\\
		\cite{han2020gcvnet}-TwoStep&&&-&4.37&98.1\%&3.22&1.90&3.79&4.54&90.2\%\\
		\cite{han2020gcvnet}-GCVNet&&&-&3.99&98.4\%&3.18&1.89&3.76&4.31&90.5\%\\
            \cite{ponimatkin2022focal}~FocalPose &&& 98.2\% & 3.99 & 98.4\% & 2.35 & 1.67 & 2.65 & 2.95 & 93.0\%\\
		FocalPose++ (ours) &&& 98.2\% & \textbf{3.61}  & \textbf{98.5}\% & \textbf{1.96} & \textbf{1.44} & \textbf{2.37} &         \textbf{2.70}  & 94.2\%\\
		\midrule
		\cite{wang20183d}&\multirow{6}{*}{Stanford}&&99.6\%&5.43&98.0\%&2.33&1.80&2.34&7.46&76.4\%\\
		\cite{grabner2019gp2c}-LF&&&\bf99.6\%&5.38&\bf{98.3\%}&1.93&1.51&2.01&3.72&96.2\%\\
		\cite{grabner2019gp2c}-BB&&&\bf99.6\%&5.24&\bf{98.3\%}&1.92&1.47&2.07&3.25&\bf{96.5\%}\\
		\cite{han2020gcvnet}-TwoStep&&&-&5.09&97.5\%&2.29&1.52&2.52&3.78&93.6\%\\
		\cite{han2020gcvnet}-GCVNet&&&-&4.92&97.5\%&2.20&1.46&2.43&3.65&94.6\%\\
            \cite{ponimatkin2022focal}~FocalPose&&& 99.5\% & 4.44 & 95.1\% &         1.00  &         0.84  &         1.09  & 2.55 & 93.8\%\\
		FocalPose++ (ours) &&& 99.5\% & \textbf{3.87} & 96.2\% & \textbf{0.94} & \textbf{0.76} & \textbf{1.04} & \textbf{2.07}  & 95.1\%\\
		\bottomrule
	\end{tabular}
	\label{table:pix3d}
\end{table*}

\begin{table*}[ht]
	\centering
	\caption{\textbf{Comparison with the state of the art for 6D pose and focal length prediction} on the Pix3D dataset split by class. \textbf{Bold} denotes the best result among directly comparable methods. See Sec.~\ref{per-class-pix3d} for a more detailed analysis of the results.}
	\small
	\setlength{\tabcolsep}{3.2pt}
	\begin{tabular}{lccc|cc|c|c|c|cc}
		\toprule
		\multicolumn{3}{c}{}&\multicolumn{1}{c}{\bf Detection}&\multicolumn{2}{c}{\bf Rotation}&\multicolumn{1}{c}{\bf Translation}&\multicolumn{1}{c}{\bf Pose}&\multicolumn{1}{c}{\bf Focal}&\multicolumn{2}{c}{\bf Projection}\\

		\cmidrule(lr){4-4}\cmidrule(lr){5-6}\cmidrule(lr){7-7}\cmidrule(lr){8-8}\cmidrule(lr){9-9}\cmidrule(lr){10-11}
		\multirow{2}{*}{Method}&\multicolumn{1}{c}{\multirow{2}{*}{Dataset}}&\multicolumn{1}{c}{\multirow{2}{*}{Class}}&\multicolumn{1}{c}{\multirow{2}{*}{$Acc_{D_{0.5}}$}}&\multicolumn{1}{c}{$MedErr_R$}&\multicolumn{1}{c}{\multirow{2}{*}{$Acc_{R\frac{\pi}{6}}$}}&\multicolumn{1}{c}{$MedErr_{t}$}&\multicolumn{1}{c}{$MedErr_{R,t}$}&\multicolumn{1}{c}{$MedErr_f$}&\multicolumn{1}{c}{$MedErr_{P}$}&\multicolumn{1}{c}{\multirow{2}{*}{$Acc_{P_{0.1}}$}}\\
		&&\multicolumn{1}{c}{}&\multicolumn{1}{c}{}&\multicolumn{1}{c}{$\cdot1$}&\multicolumn{1}{c}{}&\multicolumn{1}{c}{$\cdot10^{1}$}&\multicolumn{1}{c}{$\cdot10^{1}$}&\multicolumn{1}{c}{$\cdot10^{1}$}&\multicolumn{1}{c}{$\cdot10^{2}$}\\
		\midrule
		\midrule
		\cite{wang20183d}&\multirow{5}{*}{Pix3D}&\multirow{5}{*}{bed} & 98.4\% & 5.82 & 95.3\% & 1.95 & 1.56 & 2.22 & 6.05 & 74.9\%\\
		\cite{grabner2019gp2c} LF &&& 99.0\% & 5.13 & 96.3\% & 1.41 & 1.04 & 1.43& 3.52&90.6\%\\
		\cite{grabner2019gp2c} BB &&& \bf99.5\% & 5.40 & \bf \textbf{97.9\%} & 1.66 & 1.17 & 1.59 & 3.55 &\bf{93.2\%}\\
            \cite{ponimatkin2022focal}~FocalPose &&& 98.4\% & 3.16 & 91.6\% & 1.28 & 0.93 & 1.28 & 1.91 & 88.9\%\\
            FocalPose++ (ours)  &&& 98.4\% & \textbf{2.74} & 93.2\% & \textbf{1.15} & \textbf{0.78} & \textbf{1.21} & \textbf{1.53} & 90.0\%\\

		\midrule
		\cite{wang20183d}&\multirow{5}{*}{Pix3D}&\multirow{5}{*}{chair}&94.9\%&7.52&88.0\%&2.69&1.58&1.98&6.04&75.3\%\\
		\cite{grabner2019gp2c}-LF&&&95.2\%&7.52&88.8\%&1.92&1.21&1.62&3.41&88.2\%\\
		\cite{grabner2019gp2c}-BB&&&\bf97.3\%&6.95&\bf{91.0\%}& 1.68&1.08&1.58&3.24&\bf{90.9\%}\\
            \cite{ponimatkin2022focal}~FocalPose &&& 91.8\% & 3.56 & 85.4\% & \textbf{1.49} & \textbf{0.94} & \textbf{1.36} & 1.73 & 79.3\%\\
            FocalPose++ (ours)  &&& 91.8\% & \textbf{3.49} & 87.5\% & 1.63 & 1.02 & 1.51 & \textbf{1.69} & 82.5\%\\
		
		\midrule
		\cite{wang20183d}&\multirow{5}{*}{Pix3D}&\multirow{5}{*}{sofa}&96.5\%&4.73&94.8\%&2.28&1.62&2.42&4.33&82.2\%\\
		\cite{grabner2019gp2c} LF&&&96.5\%&4.49&95.0\%&1.92&1.33&1.79&2.56&93.7\%\\
		\cite{grabner2019gp2c} BB&&&\bf98.3\%&4.40&\textbf{97.0\%}&1.63&1.16&1.73&2.13&\bf95.6\%\\
		\cite{ponimatkin2022focal}~FocalPose &&& 96.9\% & 2.98 & 97.6\% & 1.29 & 0.83 & 1.36 & 1.52 & 93.9\%\\
            FocalPose++ (ours)  &&& 96.9\% & \textbf{2.77} & 95.6\% & \textbf{1.14} & \textbf{0.75} & \textbf{1.18} & \textbf{1.19} & 95.4\%\\

		\midrule
		\cite{wang20183d}&\multirow{5}{*}{Pix3D}&\multirow{5}{*}{table}&94.0\%&10.94&72.9\%&3.16&2.28&3.03&8.90&53.6\%\\
		\cite{grabner2019gp2c} LF&&&94.0\%& 10.53&73.5\%& 2.16& 1.62& 2.05&5.92&69.5\%\\
		\cite{grabner2019gp2c} BB&&& \bf95.7\%&10.80& \bf{77.2\%}&2.81&1.78&2.10&5.74&\bf{72.4\%}\\
		\cite{ponimatkin2022focal}~FocalPose &&& 94.9\% & 9.98 & 61.8\% & 1.90 & 1.68 & 2.13 & 6.72 & 54.7\%\\
            FocalPose++ (ours)  &&& 94.9\% & \textbf{7.75} & 64.1\% & \textbf{1.33} & \textbf{1.42} & \textbf{1.45} & \textbf{4.95} & 58.1\%\\
		 
	\bottomrule
	\end{tabular}

	\label{table:pix3d-perclass}
        \vspace{-3mm}
\end{table*}

\begin{figure}[t]
    \centering
    \begin{minipage}{0.15\columnwidth}\begin{center}
    \small{Input image}
    \end{center}\end{minipage}
        \begin{minipage}{0.27\columnwidth}
            {\small a\vspace{1mm}}
            \centering
            \includegraphics[width=\textwidth]{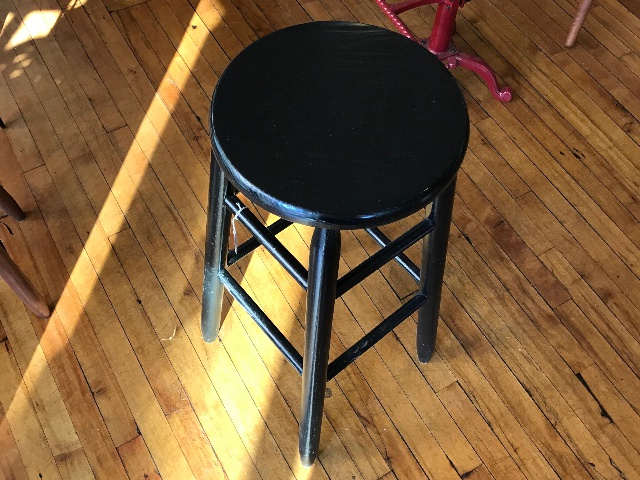}
        \end{minipage}
        \begin{minipage}{0.27\columnwidth}
            {\small b\vspace{1mm}}
            \centering
            \includegraphics[width=\textwidth]{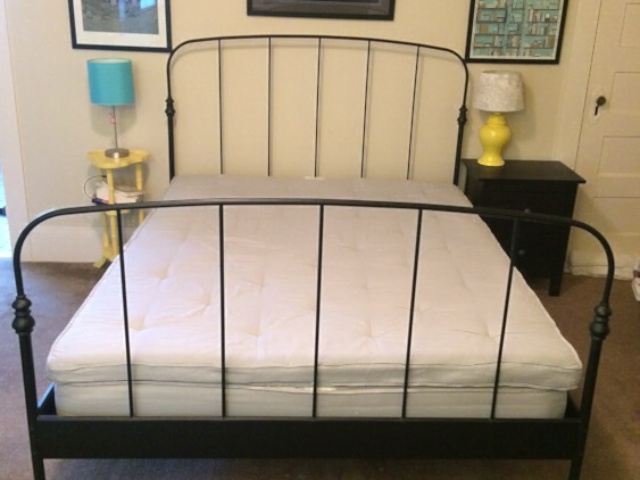}
        \end{minipage}
        \begin{minipage}{0.27\columnwidth}
        {\small c\vspace{1mm}}
            \centering
            \includegraphics[width=\textwidth]{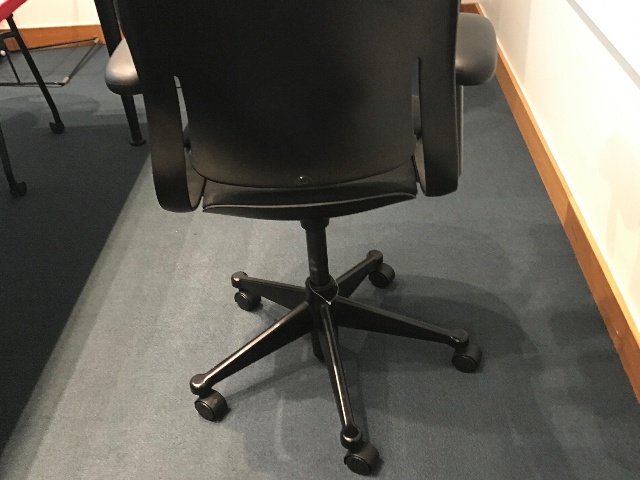}
        \end{minipage}\\[1mm]
    \begin{minipage}{0.15\columnwidth}\begin{center}
    \small{Ground truth}
    \end{center}\end{minipage}
        \begin{minipage}{0.27\columnwidth}
            \centering
            \includegraphics[width=\textwidth]{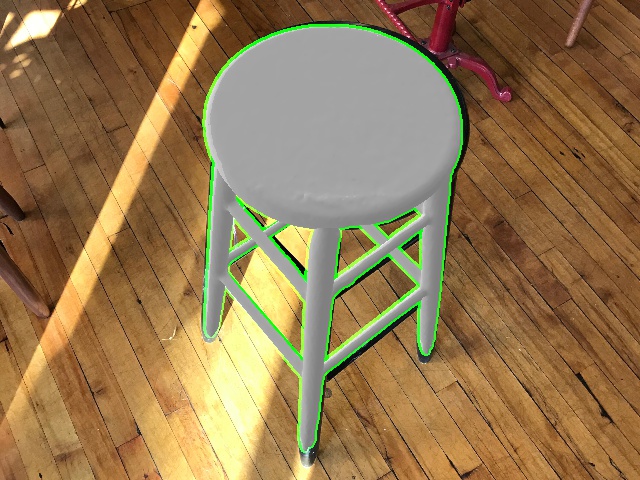}
        \end{minipage}
        \begin{minipage}{0.27\columnwidth}
            \centering
            \includegraphics[width=\textwidth]{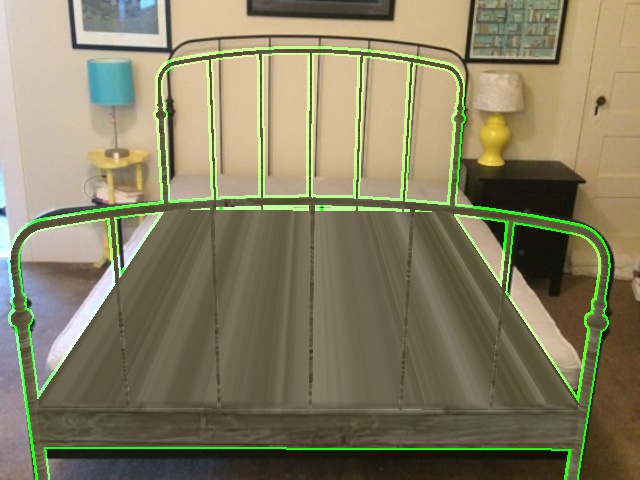}
        \end{minipage}
        \begin{minipage}{0.27\columnwidth}
            \centering
            \includegraphics[width=\textwidth]{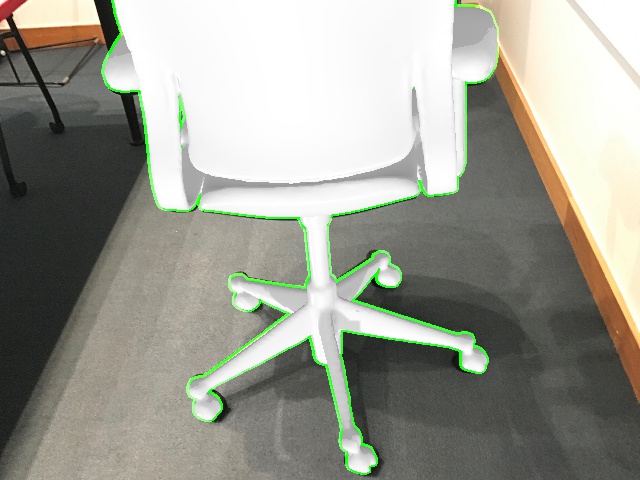}
        \end{minipage}\\[1mm]
    \begin{minipage}{0.15\columnwidth}\begin{center}
    \small{FocalPose \cite{ponimatkin2022focal}}
    \end{center}\end{minipage}
        \begin{minipage}{0.27\columnwidth}
            \centering
            \includegraphics[width=\textwidth]{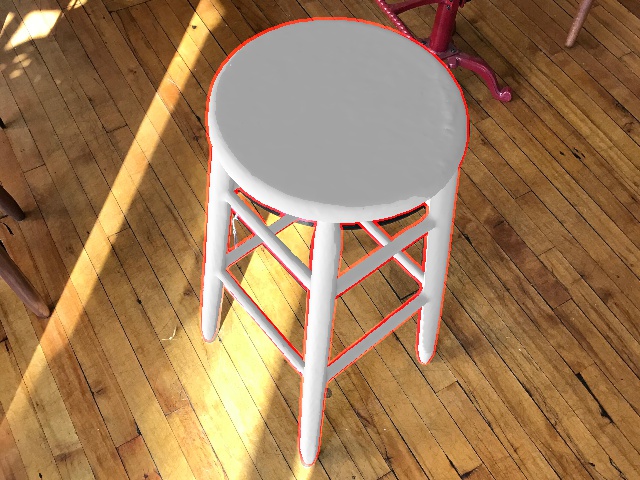}
        \end{minipage}
        \begin{minipage}{0.27\columnwidth}
            \centering
            \includegraphics[width=\textwidth]{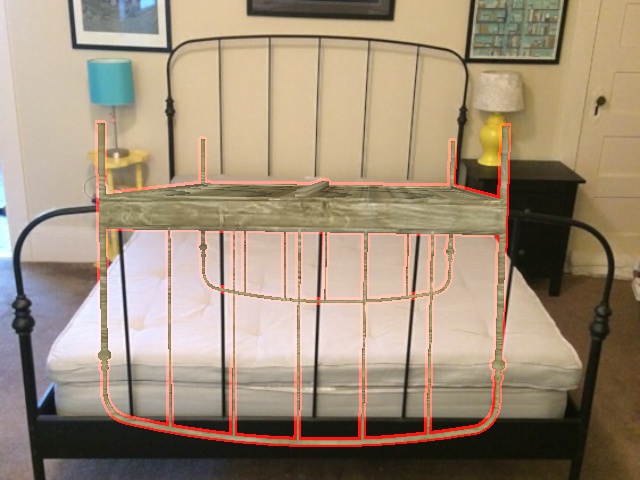}
        \end{minipage}
        \begin{minipage}{0.27\columnwidth}
            \centering
            \includegraphics[width=\textwidth]{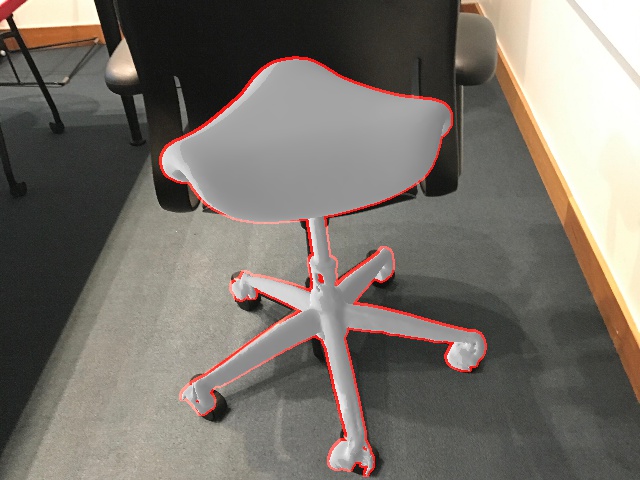}
        \end{minipage}\\[1mm]
    \begin{minipage}{0.15\columnwidth}\begin{center}
    \small{Our prediction}
    \end{center}\end{minipage}
        \begin{minipage}{0.27\columnwidth}
            \centering
            \includegraphics[width=\textwidth]{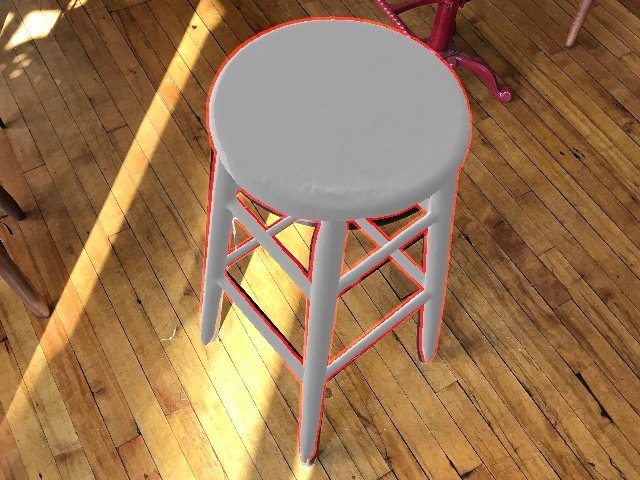}
        \end{minipage}
        \begin{minipage}{0.27\columnwidth}
            \centering
            \includegraphics[width=\textwidth]{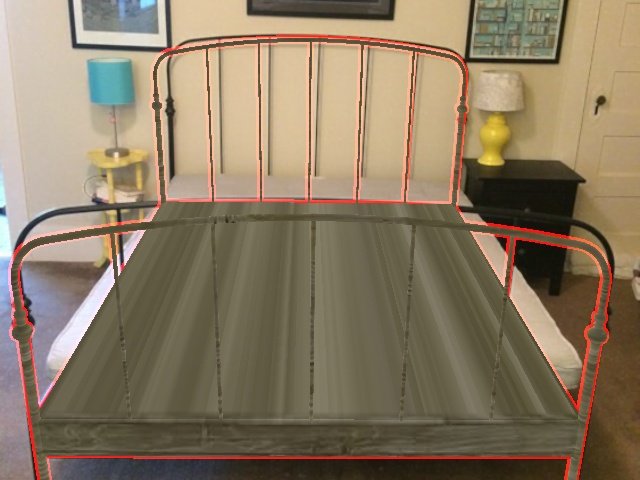}
        \end{minipage}
        \begin{minipage}{0.27\columnwidth}
            \centering
            \includegraphics[width=\textwidth]{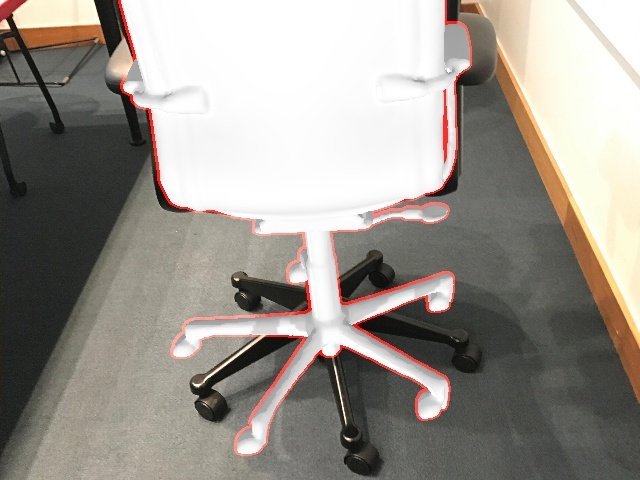}
        \end{minipage}\\[1mm]
        \vspace*{-1mm}
        \caption{\textbf{Main failure modes} are: (a) symmetric objects, (b) local minima, and (c) incorrect 3D models identified by the object detector.}
    \label{fig:failure_modes}
    \vspace{-4mm}
\end{figure}

\vspace{-3mm}
\subsection{Evaluation on 6D pose estimation benchmarks}
\label{sec:6d_pose_benchmarks}
In this section we compare our approach on a standard benchmark dataset for 6D pose estimation.
The standard 6D pose estimation benchmarks~\cite{hodan2024bop,Xiang2018-dv,rennie2016dataset,hodan2017tless,tyree20226dof} focus on the scenario with a known focal length. To compare with methods for 6D pose estimation with a calibrated camera, we need to modify not only the training of our approach but also the evaluation. Next, we focus on the BOP~Challenge~\cite{hodan2024bop} benchmark, describe our training and evaluation setup, and compare to the CosyPose~\cite{labbe2020cosypose} approach.

\paragraph{Training data.}
To train our model, we cannot use the official BOP Challenge real and synthetic training datasets as the model would overfit to the ground-truth focal length, which is constant for all training and testing images. For simplicity, we focus only on objects from the YCB-Video~\cite{Xiang2018-dv} dataset with a ground-truth focal length of 1067 px. Instead of using the official training data, we generate a new synthetic training dataset with a varying focal length randomly selected between 400 and 1600 pixels for each image. Similarly to the official synthetic BOP datasets, we use BlenderProc~\cite{denninger2019blenderproc} to render photorealistic images of scenes with objects dropped on a plane using a physics simulator.

\paragraph{Evaluation.}
To directly compare with CosyPose, we use the same object detections provided by PoseCNN~\cite{Xiang2018-dv}. We report pose metrics from the BOP Challenge, \ie  $\text{AR}_\text{VSD}$ (Visible Surface Discrepancy), $\text{AR}_\text{MSSD}$ (Maximum Symmetry-Aware Surface Distance) and $\text{AR}_\text{MSPD}$ (Maximum Symmetry-Aware Projection Distance). Their average is reported as a final AR score (see~\cite{hodan2024bop} for the definition of individual metrics). These metrics can also be used to evaluate the 6D pose without camera calibration, although $\text{AR}_\text{MSPD}$ and $\text{AR}_\text{VSD}$ need to be adapted to report correct scores by taking into account also the predicted focal length, as they rely on camera projection. Please note that our method, which jointly estimates the 6D pose and the camera focal length, solves a task that is by its nature more difficult, as small focal length and z-translation errors can be hard to distinguish.

\begin{figure*}[tb]
    \centering

    \small{1}
        \begin{minipage}{0.245\linewidth}
            {\small Input image\vspace{1mm}}
            \centering
            \includegraphics[width=\textwidth]{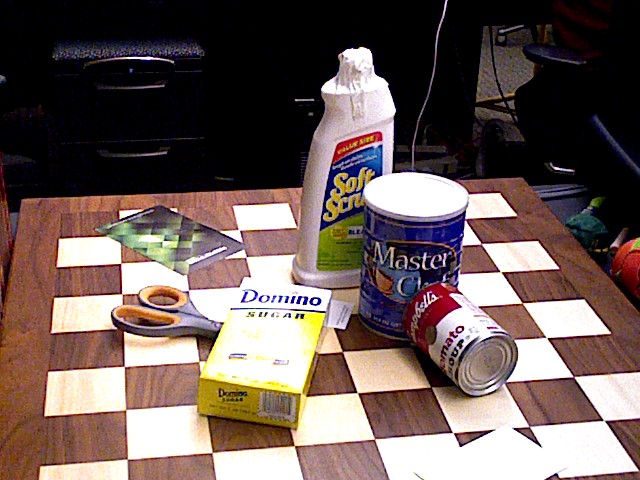}
        \end{minipage}
        \begin{minipage}{0.245\linewidth}
            {\small Ground truth\vspace{1mm}}
            \centering
            \includegraphics[width=\textwidth]{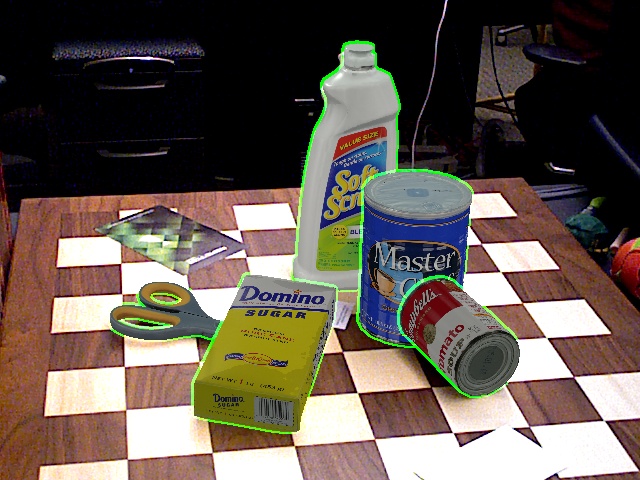}
        \end{minipage}
        \begin{minipage}{0.245\linewidth}
            {\small Our prediction\vspace{1mm}}
            \centering
            \includegraphics[width=\textwidth]{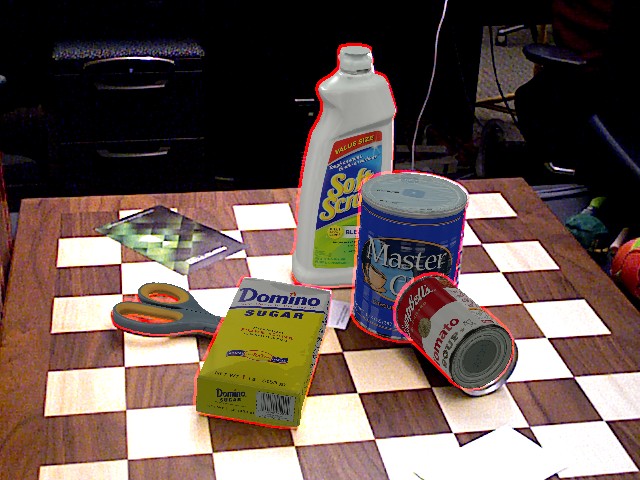}
        \end{minipage}\\[0.5mm]
    \small{2}
        \begin{minipage}{0.245\linewidth}
            \centering
            \includegraphics[width=\textwidth]{figures/ycbv/000052_000582_input.jpg}
        \end{minipage}
        \begin{minipage}{0.245\linewidth}
            \centering
            \includegraphics[width=\textwidth]{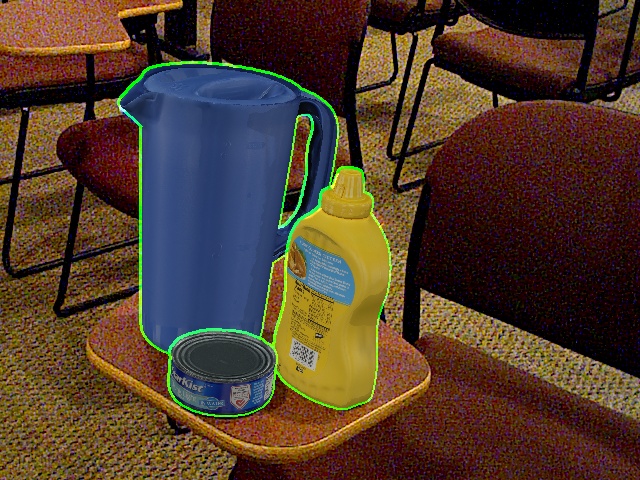}
        \end{minipage}
        \begin{minipage}{0.245\linewidth}
            \centering
            \includegraphics[width=\textwidth]{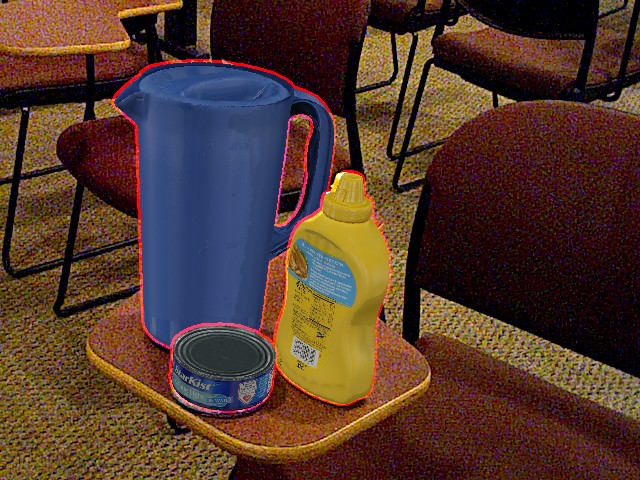}
        \end{minipage}\\[0.5mm]
    \small{3}
        \begin{minipage}{0.245\linewidth}
            \centering
            \includegraphics[width=\textwidth]{figures/ycbv/000054_001080_input.jpg}
        \end{minipage}
        \begin{minipage}{0.245\linewidth}
            \centering
            \includegraphics[width=\textwidth]{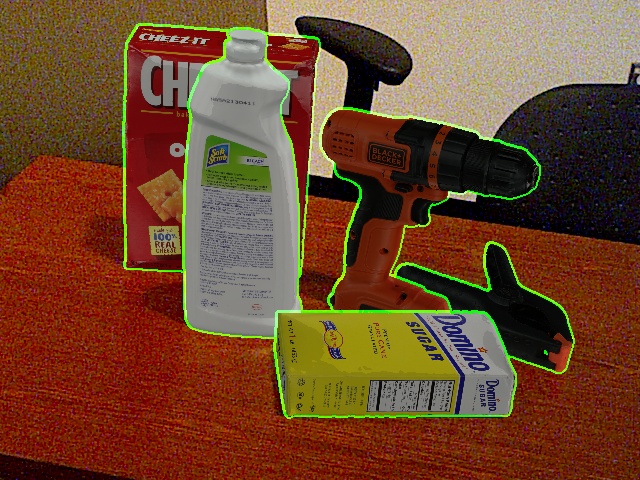}
        \end{minipage}
        \begin{minipage}{0.245\linewidth}
            \centering
            \includegraphics[width=\textwidth]{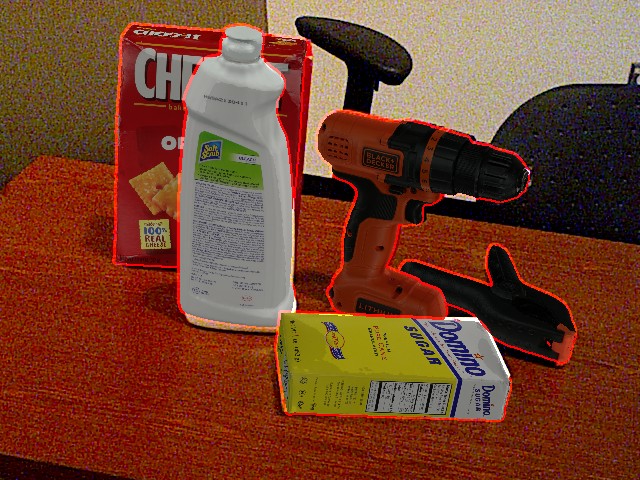}
        \end{minipage}\\[0.5mm]
    \small{4}
        \begin{minipage}{0.245\linewidth}
            \centering
            \includegraphics[width=\textwidth]{figures/ycbv/000059_000648_input.jpg}
        \end{minipage}
        \begin{minipage}{0.245\linewidth}
            \centering
            \includegraphics[width=\textwidth]{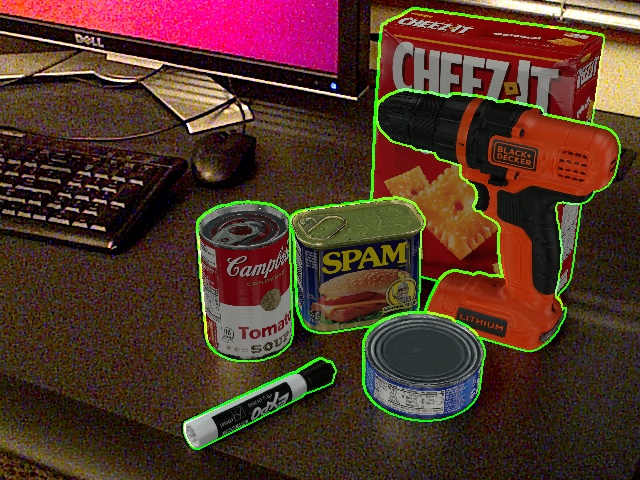}
        \end{minipage}
        \begin{minipage}{0.245\linewidth}
            \centering
            \includegraphics[width=\textwidth]{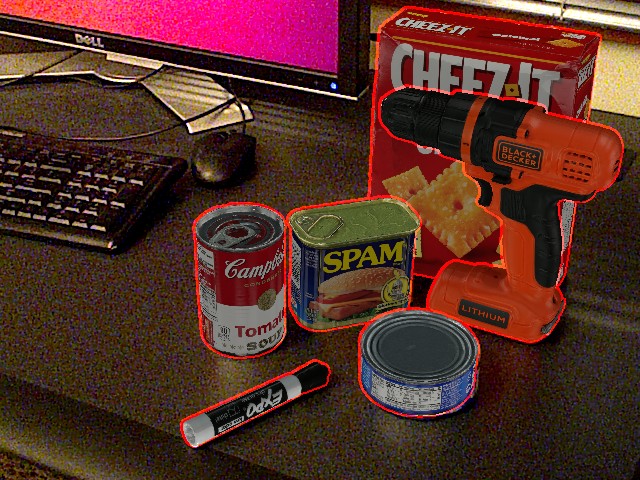}
        \end{minipage}\\[0.5mm]

    \caption{\textbf{Qualitative results for YCB-Video~\cite{Xiang2018-dv} dataset.} Our method (right) produces precise alignments that are often hard to distinguish from the ground-truth annotations (middle) when looking from the camera's point of view. We visualize the ground-truth annotations and our predictions by overlaying the input image with the objects rendered using the corresponding camera focal length and object 6D poses. We show contours of each object with green color when rendered in the ground-truth pose, and with red color when rendered in the estimated poses.}
    \label{fig:ycbv}
    \vspace{-3mm}
\end{figure*}

\paragraph{Results.}
In Table~\ref{table:ycbv_eval} we report the BOP metrics for our method using 55 refiner iterations with predicted focal length (FocalPose++ Pred) and also with focal length fixed to the ground-truth value (FocalPose++ GT). The latter is obtained using a modification of our method, where we multiply the focal length and the z-translation by a corrective factor obtained as the ratio between the ground truth and the predicted focal length after each iteration. We compare our method with CosyPose~\cite{labbe2020cosypose}, specifically with the model \textit{ECCV20-PBR-1VIEW} trained only on synthetic data. 
Our results with ground-truth focal length show an AR score similar to that of CosyPose, although the individual metrics differ. We hypothesize that this is mainly due to the difference in the training data and because our model was not trained with a fixed focal length. 
With the prediction of focal length, our model increases $\text{AR}_\text{MSPD}$, \ie we fit the model better from the perceptual point of view, as we have one more degree of freedom, but the other two metrics are lower compared to CosyPose GT that uses the ground truth focal length. Please note that this decrease (caused by errors in the predicted z-translation and the predicted focal length) is expected, since the task we are solving is significantly more difficult as small focal length errors can be compensated by adjusting the z-translation without almost any reduction in $\text{AR}_\text{MSPD}$. In detail, our method (FocalPose++ Pred) achieves a median focal length error $MedErr_f$ of 0.116 (or 11.6\%), which can be hardly noticeable from the camera's point of view.
To illustrate the effect of different focal lengths, we evaluate CosyPose with focal length multiplied by 0.95, 0.90, 0.85 and 0.80, which corresponds to 5\%, 10\%, 15\% 20\% focal length error, respectively, and report the result also in Table~\ref{table:ycbv_eval} (lines denoted as $\text{GT}\cdot0.95$, $\text{GT}\cdot0.90$, etc.). We observe that with 10\% focal length error, the results of CosyPose drop to performance similar to our FocalPose approach, which uses focal length prediction. Additionally, we report CosyPose results with the focal length set to the commonly used prior equal to the size of the image diagonal ($\sqrt{w^2+h^2}$). The results show that $\text{AR}_\text{MSPD}$ still maintains reasonable performance, but the $\text{AR}_\text{VSD}$ and $\text{AR}_\text{MSSD}$ metrics drop to zero. This is because this focal length prior corresponds to approximately 25\% relative focal length error on the YCB-video dataset, which causes errors in the estimated z-translation. These errors are above the thresholds used in the latter metrics for a pose to be considered correct. In Fig.~\ref{fig:ycbv}, we show the qualitative result of our method with the predicted focal length. The results show that even with some focal length errors, our method produces results that are often perceptually indistinguishable from the ground-truth annotations from the camera's point of view; however, the z-translation might be incorrect due to the incorrect focal length estimate.

Please note that the state-of-the-art single-view RGB-only
methods for 6D pose estimation with a {\em calibrated camera} report even higher AR scores in the BOP Challenge~\cite{hodan2024bop}; however, these results are not directly comparable to ours as those methods use both real and synthetic training data, whereas our method uses only synthetic images. Also, we do not aim to directly compete with those methods, as we solve a task that is more difficult, and we aim at different use cases, as discussed in Section~\ref{sec:applications}.

\begin{table}[ht!]
    \caption{
    \textbf{Comparison of our method with CosyPose\cite{labbe2020cosypose} on the YCB-Video~\cite{Xiang2018-dv} dataset.} We report BOP Challenge~\cite{hodan2024bop} 6D pose estimation metrics for our method trained on synthetic data with randomized focal length, showing results with focal length prediction (FocalPose++ Pred) and with the focal length fixed to its ground-truth value (FocalPose++ GT). We compare our method to CosyPose GT) trained on synthetic data with a fixed ground truth focal length (CosyPose GT) and show how its performance changes when the focal length is set to an incorrect value (CosyPose $\text{GT}\cdot0.95$, etc.). Additionally, we report CosyPose results with the focal length set to the commonly used prior equal to the size of the image diagonal, \ie $\sqrt{w^2+h^2}$), which corresponds to about 25\% relative focal length error for the YCB-Video dataset. The methods are ranked by the $\text{AR}$ metric (top is the best). The top two methods separated by the dashed line use the correct ground truth focal length.
    }
    \label{table:ycbv_eval}
    \small
    \setlength{\tabcolsep}{3.2pt}
    \begin{tabular}{ll|r|rrr}
    \toprule
    Method & Focal len. & $\text{AR}$ & $\text{AR}_\text{VSD}$ & $\text{AR}_\text{MSSD}$ & $\text{AR}_\text{MSPD}$ \\
    \midrule
        \textbf{FocalPose++ (Ours)} & \textbf{GT} & 57.9 & 43.3 & 63.7 & 66.5 \\
        CosyPose & $\text{GT}$ & 57.4 & 51.6 & 55.4 & 65.3
        \\[-\jot]
        \multicolumn{6}{@{}c@{}}{\makebox[\linewidth]{\dashrule[black!40]}} \\[-\jot]
        
        CosyPose & $\text{GT}\cdot0.95$ & 54.0 & 46.3 & 50.6 & 65.0 \\
        \textbf{FocalPose++ (Ours)} & \textbf{Pred} & 36.7 & 17.4 & 20.3 & 72.4 \\
        CosyPose & $\text{GT}\cdot0.90$ & 32.9 & 16.0 & 18.0 & 64.7\\
        CosyPose & $\text{GT}\cdot0.85$ & 23.3 &  2.7 &  2.6 & 64.6\\
        CosyPose & $\text{GT}\cdot0.80$ & 21.4 &  0.2 &  0.1 & 64.0\\
        CosyPose &     $\sqrt{w^2+h^2}$ & 21.2 &  0.0 & 0.0 & 63.5\\
    \bottomrule
    \end{tabular}
    \vspace{-3mm}
\end{table}

\subsection{Training stability}
\label{sec:error_bars}
We evaluate the stability of the training between different runs, including the rendering of a new synthetic training dataset. For the Pix3D sofa category, we render 3 synthetic datasets and train 3 coarse and refiner models on the real training dataset with one of the synthetic datasets added every time. Then, we run the evaluation of the models. We observe that the values of the reported metrics differ (relative to the mean of the 3 runs) at most by $\pm 3.9\%$ for the rotation error, $\pm 2.2\%$ for the translation and focal length errors, $\pm 1.2\%$ for the pose error, $\pm 1.1\%$ for the projection error, $\pm 0.5\%$ for the projection accuracy, and finally by $\pm 0.4\%$ for the rotation accuracy. These estimated errors are much lower than the relative reduction in median errors by our method compared to other state-of-the-art methods.
\vspace{-3mm}

\subsection{Applications}
\label{sec:applications}
In this section, we show two examples of applications of our method. First, we show an application in computer graphics where our method can be used for image editing/augmentation. Second, we show how our method can be used in robotics for imitation of object manipulation skills from Internet videos. Both applications target in-the-wild Internet data, where focal length is often unknown. In both applications, we assume that (at least approximate) 3D model of the depicted object is available, which could be reasonable for many common man-made objects such as vehicles, furniture, or kitchen utensils that are common in large-scale repositories of 3D data~\cite{deitke2022objaverse,deitke2023objaversexl} or can even be generated given the input image or video~\cite{wu2024reconstructing,long2023wonder3d,chen2024v3d}.

\begin{figure*}[tb]
    \centering

    \begin{minipage}{0.2464\linewidth}
        \centering Input image
    \end{minipage}
    \begin{minipage}{0.2464\linewidth}
        \centering Estimated pose
    \end{minipage}
    \begin{minipage}{0.4928\linewidth}
        \centering Augmented images
    \end{minipage}\\[1mm]

    \includegraphics[width=0.98\textwidth]{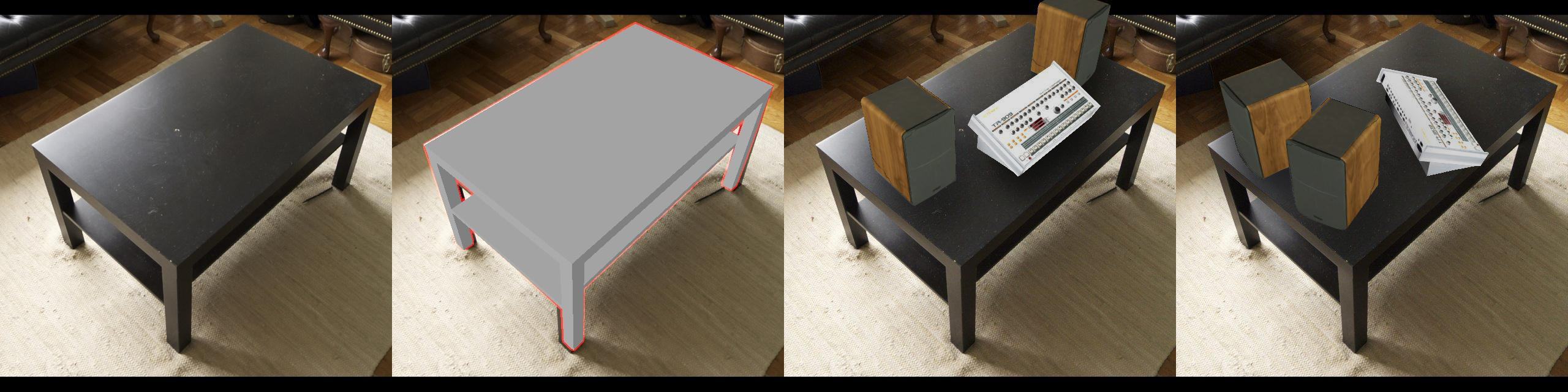}\\
    \vspace{1mm}
    \includegraphics[width=0.98\textwidth]{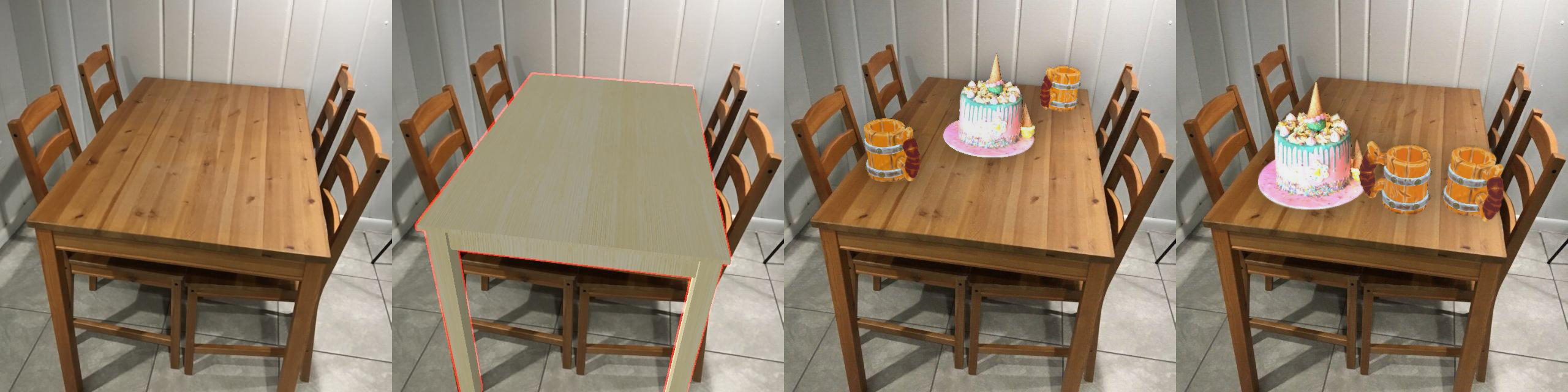}\\
    \vspace{1mm}
    \includegraphics[width=0.98\textwidth]{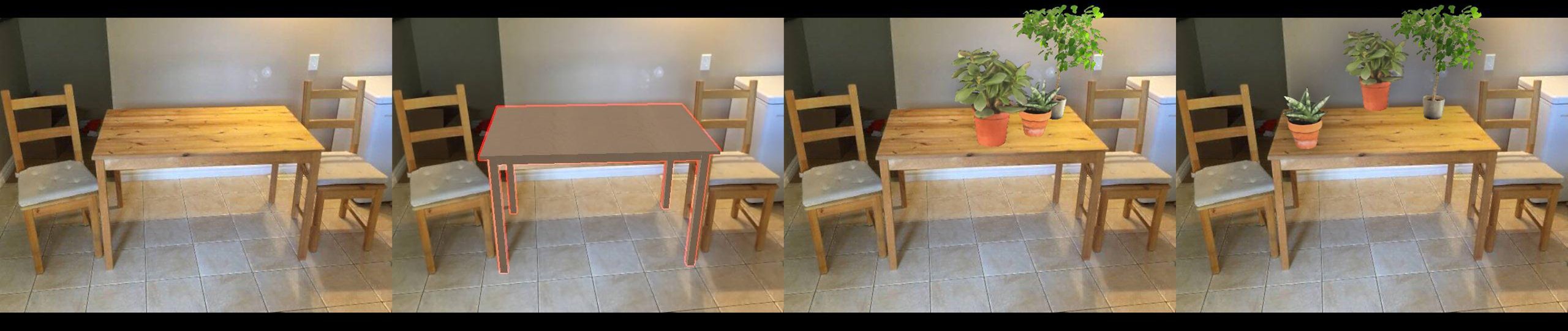}\\
    \caption{\textbf{Application I: 3D-aware image augmentation.} Given an input image (first column), we estimate the camera focal length and 6D pose of the table using our FocalPose++ approach (second column). The estimated geometry of the table allows us to randomly place three new 3D objects on the table and render them in the original image (third and fourth column). Note how the new objects are inserted into the scene respecting its geometry and perspective effects.}
    \label{fig:application_cg}
    \vspace{-3mm}
\end{figure*}

\paragraph{Application I: Image augmentation for in-the-wild images.}
Image editing is an important area of computer graphics that can take many forms. We focus on a specific type of image editing, where the objective is to augment the image by adding new objects to the scene; \ie we need to place the objects into the image in a way that is aware of the scene geometry~\cite{kholgade20143d}. We show such image augmentation on the Pix3D-table dataset: First, we estimate the 6D pose of the table and the camera focal length for several images from the dataset. Second, we take three objects and place them randomly on top of the table, using its estimated coordinate frame to properly align the newly inserted object with the perspective of the scene. Finally, we render the objects using the estimated focal length and camera pose and compose the rendered object with the original image. Fig.~\ref{fig:application_cg} shows examples of such image augmentation, depicting the original image, the estimated pose of the table, and two different outputs of our approach for each image. Note that a precise focal length estimation results in better pose estimation of the table; however, estimating the focal length is also required to match the perspective effects in the input image.

\begin{figure*}[tbp]
    \centering
    \rotatebox{90}{\hspace{-6mm} \small{Input video}}
    \begin{minipage}{0.97\linewidth}
        \centering\includegraphics[width=\textwidth]{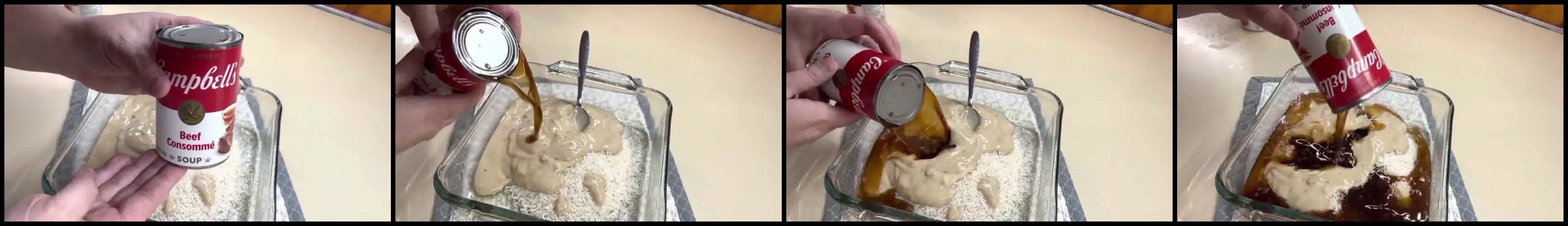}\\
    \end{minipage}
    \vspace{1mm}

    \rotatebox{90}{\hspace{-11mm} \small{Object trajectory}}
    \begin{minipage}{0.97\linewidth}
        \centering\includegraphics[width=\textwidth]{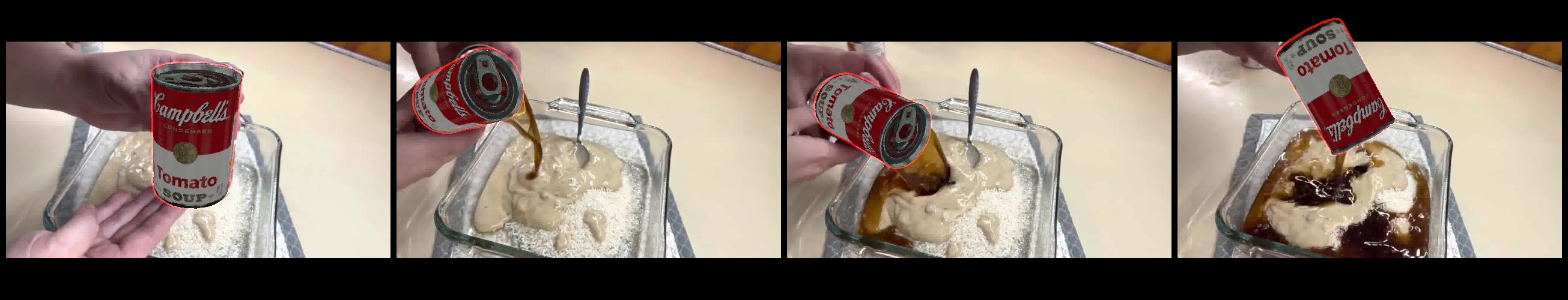}\\
    \end{minipage}
    \vspace{1mm}

    \rotatebox{90}{\hspace{-11mm} \small{Robot simulation}}\hspace{0.8mm}
    \begin{minipage}{0.97\linewidth}
        \centering\includegraphics[width=\textwidth]{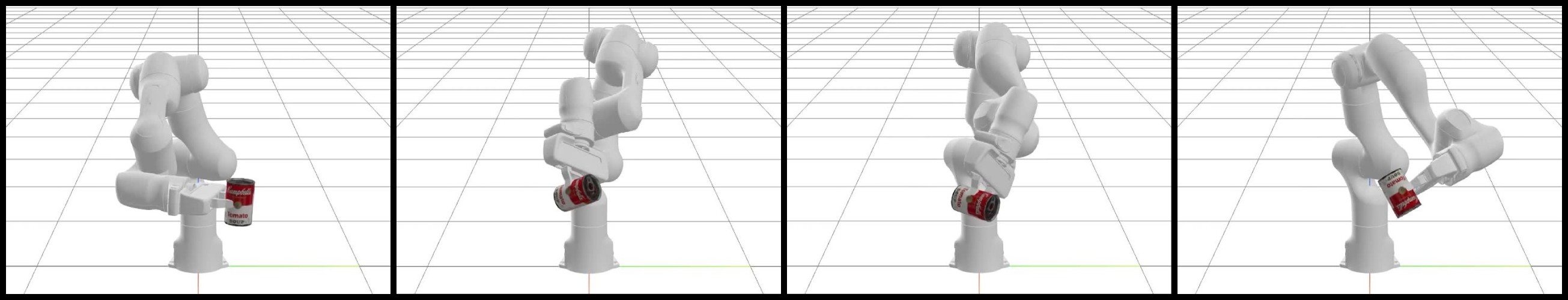}\\
    \end{minipage}
    \vspace{1mm}

    \rotatebox{90}{\hspace{-6mm} \small{Real robot}}
    \begin{minipage}{0.97\linewidth}
        \centering\includegraphics[width=\textwidth]{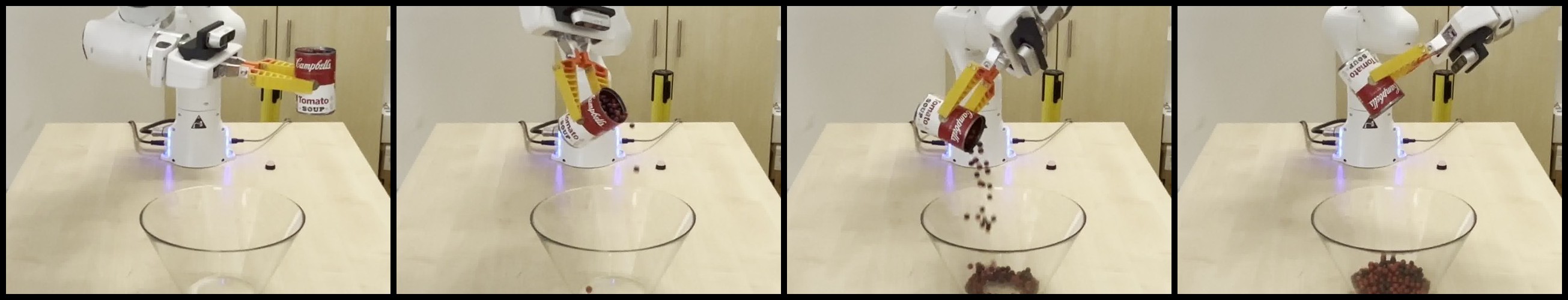}\\
    \end{minipage}
    
    \caption{\textbf{Application II: object manipulation from Internet video.} Given an input video (first row) with a known object, we estimate its 6D pose in each video frame. In the first frame, we estimate object's 6D pose and camera focal length using our method (coarse and refiner model). In the following frames, we reuse the 6D pose and focal length from the previous frame as initialization and apply only the refiner to track the object. To obtain the final trajectory, we use the estimated median focal length, recompute z-translation accordingly, and apply trajectory smoothing (second row). Finally, we compute inverse kinematics and imitate the object manipulation with a Franka Emika Panda in simulation (third row) and on the real robot (fourth row). Please see the supplementary video.}
    \label{fig:application_robo}
\end{figure*}

\begin{figure*}[tbp]
    \centering
    \includegraphics[width=\textwidth]{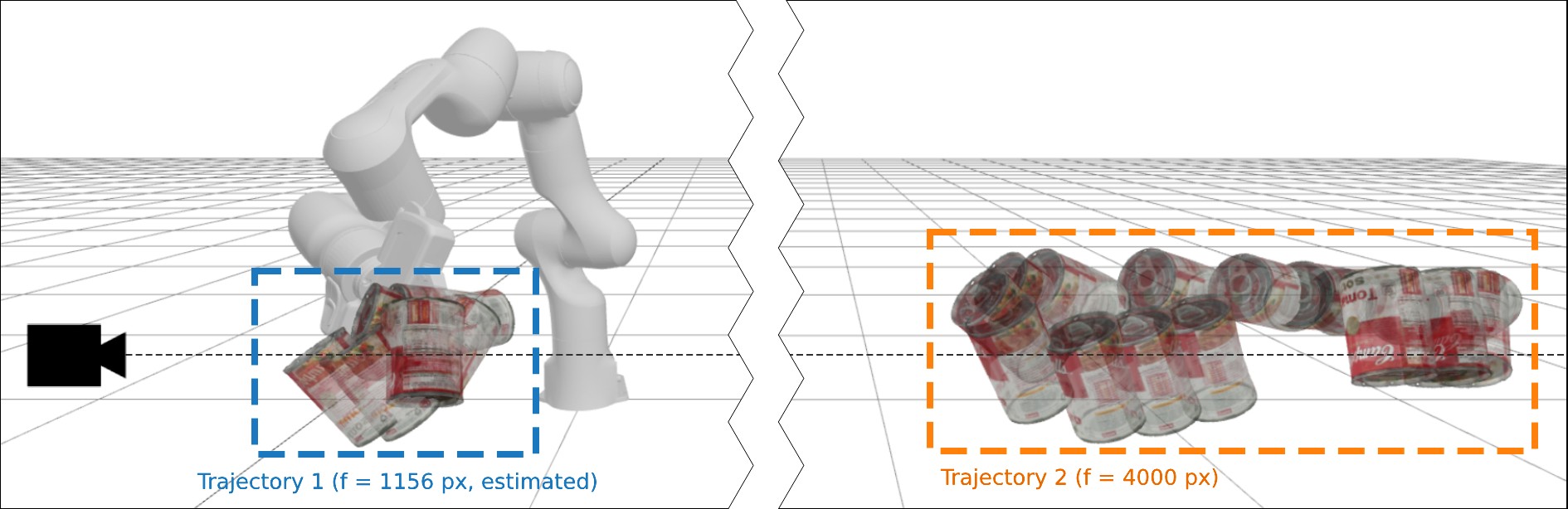}\\
    \caption{\textbf{Visualization of two object trajectories computed using different focal lengths.} We show a trajectory obtained using the estimated focal length (left) and a trajectory computed using an (incorrect) focal length of 4000 px (right), with the camera looking from left towards right. A larger focal length results in larger movement in the z-direction, which may be hard or impossible for the robot to follow in extreme cases, showing  the importance of focal length estimation.}
    \label{fig:application_robo_focal}
\end{figure*}

\paragraph{Application II: Robotic imitation of object manipulation skills from Internet videos.}
6D pose estimation has important application in robotics for object manipulation. In an uncalibrated setup, it can be used to imitate object manipulation skills from Internet videos where the camera intrinsic parameters are unknown. We use our approach to extract the pose of the Campbell's soup can from YouTube video and reproduce the trajectory of the object with the Franka Emika Panda robot. In detail, we first detect the object in the first frame of the video and run both the coarse and the refiner models to estimate the initial object 6D pose and camera focal length. The rest of the frames are treated differently: Since the object pose changes only slightly between the frames, we use the parameters from the previous video frame as initialization and run only the refiner model, resulting in a more stable trajectory. Our method thus predicts a new focal length for every frame, possibly introducing some noise in the pose estimation. However, this approach allows for more flexibility in object tracking and yields more focal length information. After processing all video frames, we post-process the outputs. First, we unify the focal length by recomputing the outputs to the median focal length from all frames. Second, we smooth the object trajectory by fitting a spline curve with smoothness constraint to the object translations, and by applying moving average to the rotations. We then use the estimated object 6D poses to compute the inverse kinematics, and imitate the object manipulation on the Panda robot. Note that although we manually align the robot's environment with the observed real-world scene, this can be done fully automatically as in~\cite{zorina2021learning}. In Fig.~\ref{fig:application_robo} we show several frames of the input video, visualization of the estimated 6D poses, and the robot that moves the object along the estimated trajectory. Please note that estimating the focal length is important in this task. In Fig.~\ref{fig:application_robo_focal} we show two object trajectories, one with the estimated focal length (left) and one with an (incorrect) focal length of 4000px (right), viewing from the side (\ie the camera is looking from left to right). Note how the larger focal length results in a larger movement in the z-direction. This can result in an object trajectory that is hard or impossible for the robot to follow.

\subsection{Limitations}
\label{sec:limitations}
There are three main failure modes of our approach, illustrated in Fig.~\ref{fig:failure_modes}.
First, we observe high rotation errors for \textit{symmetric objects} such as tables or stools, where the correct orientation is ambiguous. Please note that none of the evaluation criteria used takes into account the symmetries of objects. Second, our iterative alignment procedure can get stuck in a \textit{local minima} where the predicted object model in the predicted configuration is reasonably aligned but the errors are still high, \eg, because the object is flipped upside down. This failure mode is hard to completely eliminate; however, we succeeded in reducing the number of cases by using only synthetic training images with common viewpoints, \ie~introducing our parametric synthetic data distribution (Sec.~\ref{sec:training_data}). Also, this could be mitigated by running our approach from multiple initializations or running our refinement network on better coarse estimates. Finally, we observe that, in some situations, the 3D model retrieved by our pipeline is incorrect. These failure modes lead to large errors, which explains the lower accuracies measured by the $Acc_{R\frac{\pi}{6}}$ and $Acc_{P_{0.1}}$ metrics. 
Nevertheless, our approach achieves significantly lower median errors (5 out of the 8 reported metrics) compared to the current state-of-the-art methods, demonstrating the high precision of our approach outside of these failure modes. 

\noindent \textbf{Broader impact.} Our work has the potential to positively impact practical applications in augmented reality and robotics, among them overlaying artistic effects on viewed objects or for a robotic assistant that can manipulate real-world objects. However, our work could also potentially be used as a component for 3D-assisted manipulation of an image or video via object compositing to create misinformation.

\begin{figure}[tbp]
    \centering
    \small{1}
        \begin{minipage}{0.31\columnwidth}
            {\small Input image\vspace{1mm}}
            \centering
            \includegraphics[width=\textwidth]{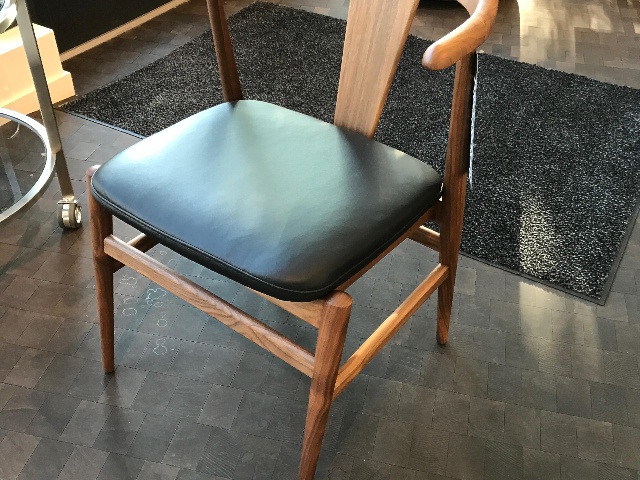}
        \end{minipage}
        \begin{minipage}{0.31\columnwidth}
            {\small Ground truth\vspace{1mm}}
            \centering
            \includegraphics[width=\textwidth]{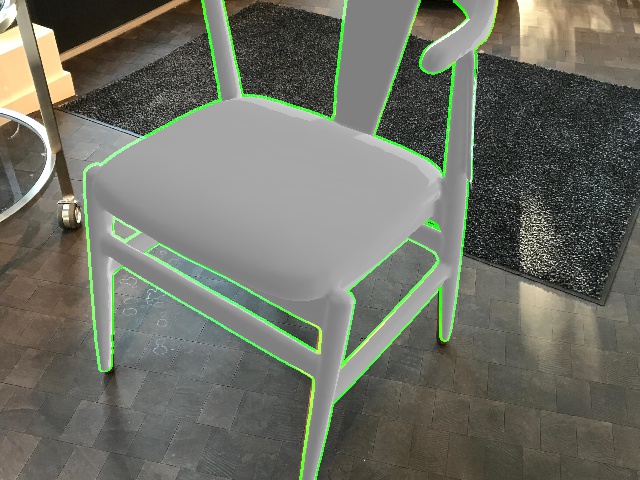}
        \end{minipage}
        \begin{minipage}{0.31\columnwidth}
            {\small Our prediction\vspace{1mm}}
            \centering
            \includegraphics[width=\textwidth]{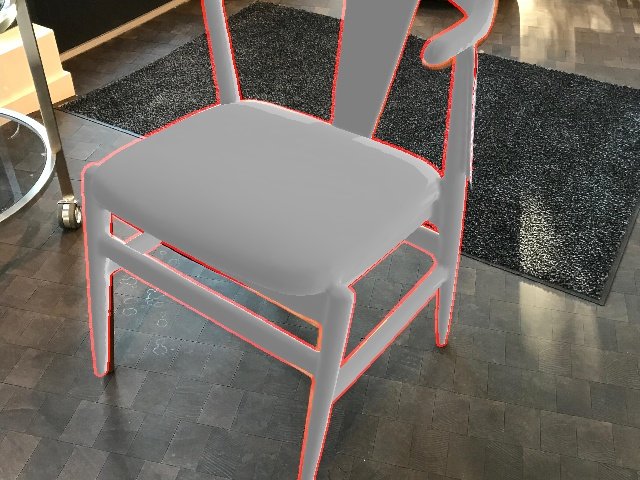}
        \end{minipage}\\[0.5mm]
    \small{2}
        \begin{minipage}{0.31\columnwidth}
            \centering
            \includegraphics[width=\textwidth]{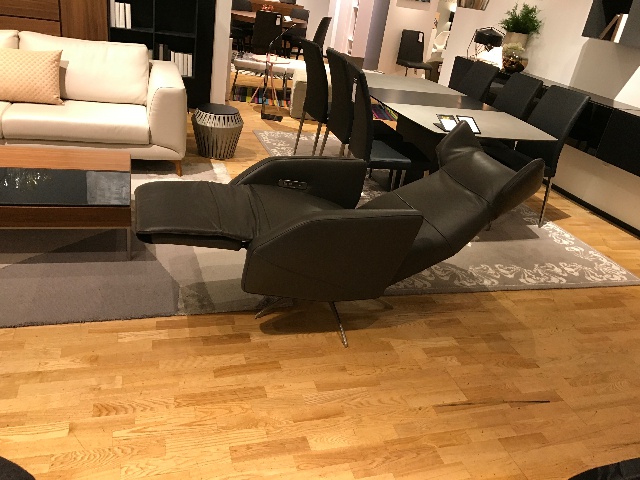}
        \end{minipage}
        \begin{minipage}{0.31\columnwidth}
            \centering
            \includegraphics[width=\textwidth]{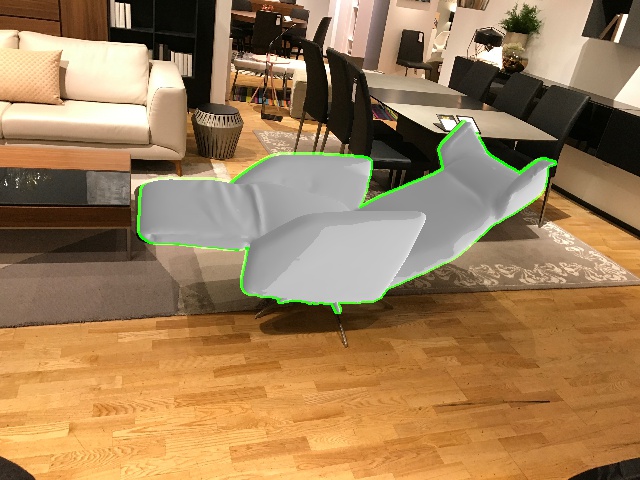}
        \end{minipage}
        \begin{minipage}{0.31\columnwidth}
            \centering
            \includegraphics[width=\textwidth]{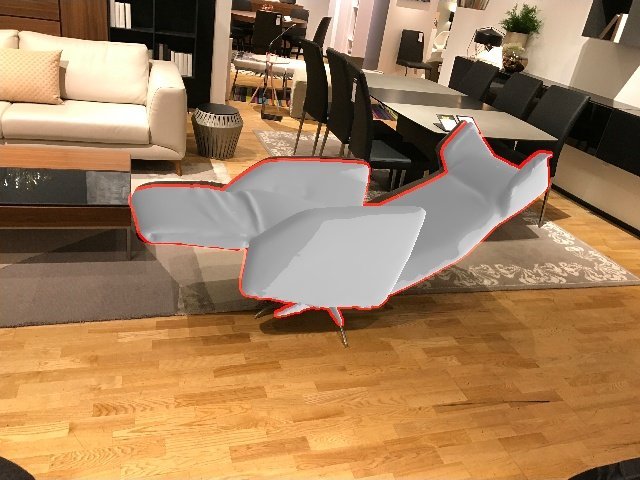}
        \end{minipage}\\[0.5mm]
    \small{3}
        \begin{minipage}{0.31\columnwidth}
            \centering
            \includegraphics[width=\textwidth]{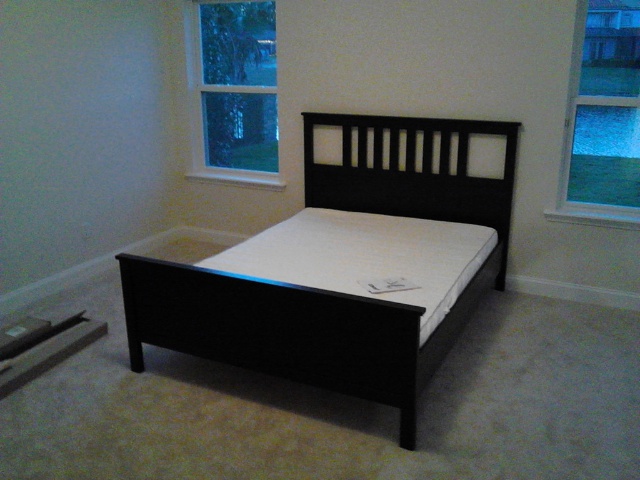}
        \end{minipage}
        \begin{minipage}{0.31\columnwidth}
            \centering
            \includegraphics[width=\textwidth]{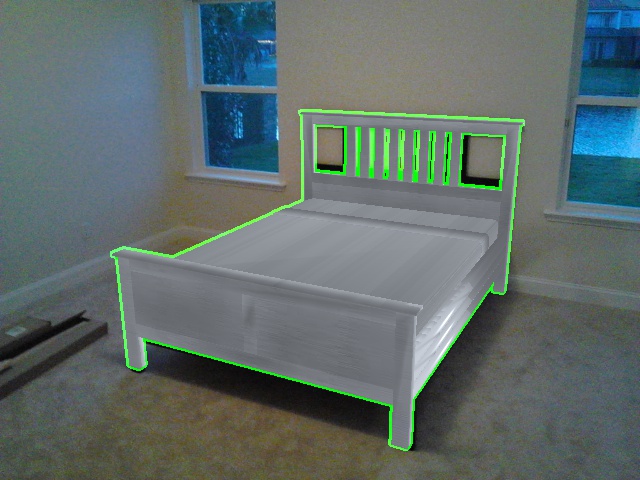}
        \end{minipage}
        \begin{minipage}{0.31\columnwidth}
            \centering
            \includegraphics[width=\textwidth]{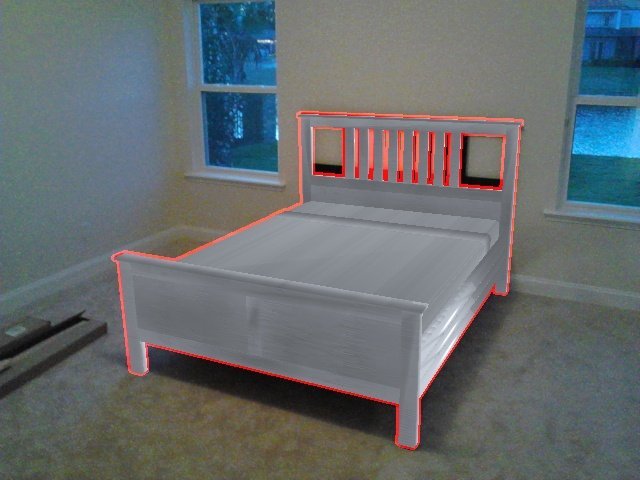}
        \end{minipage}\\[0.5mm]
    \small{4}
        \begin{minipage}{0.31\columnwidth}
            \centering
            \includegraphics[width=\textwidth]{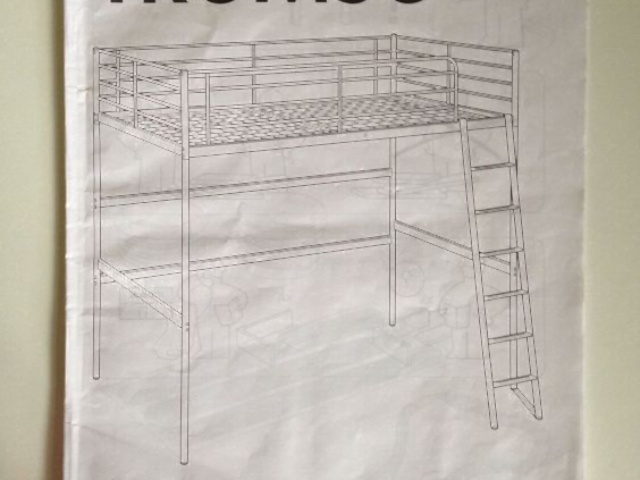}
        \end{minipage}
        \begin{minipage}{0.31\columnwidth}
            \centering
            \includegraphics[width=\textwidth]{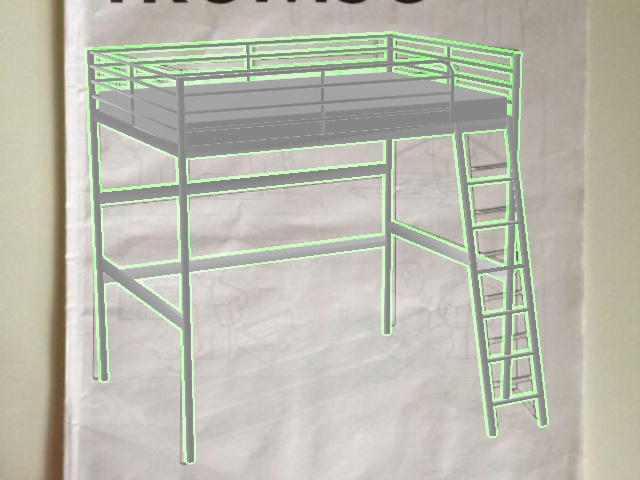}
        \end{minipage}
        \begin{minipage}{0.31\columnwidth}
            \centering
            \includegraphics[width=\textwidth]{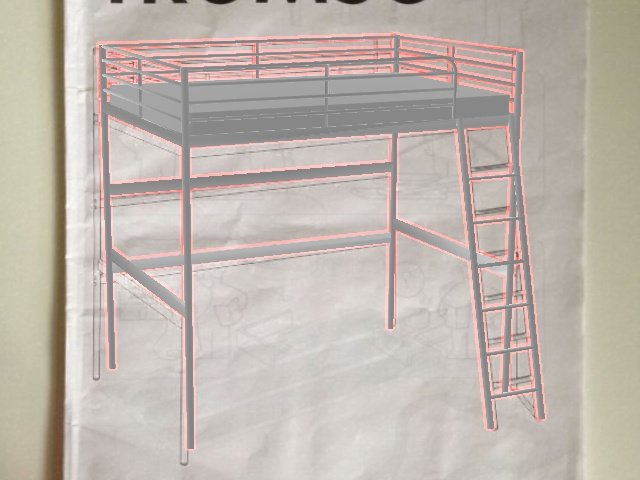}
        \end{minipage}\\[0.5mm]

    \small{5}
        \begin{minipage}{0.31\columnwidth}
            \centering
            \includegraphics[width=\textwidth]{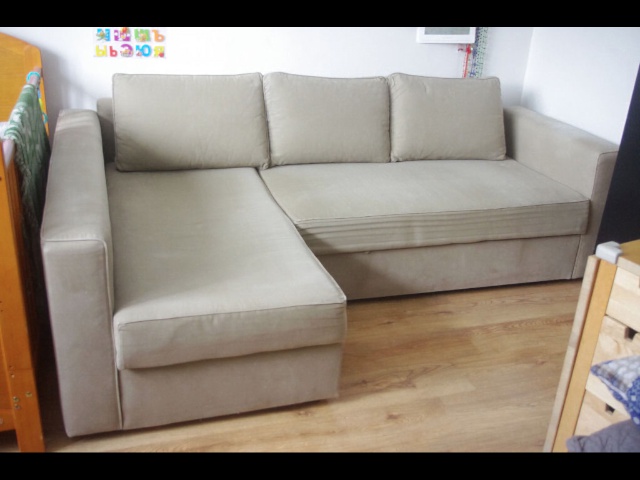}
        \end{minipage}
        \begin{minipage}{0.31\columnwidth}
            \centering
            \includegraphics[width=\textwidth]{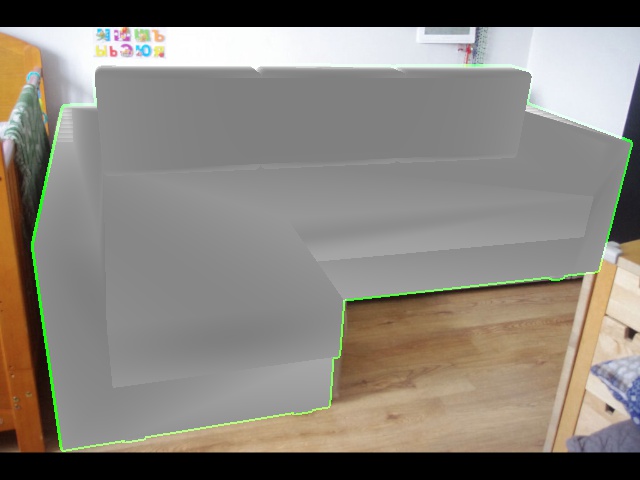}
        \end{minipage}
        \begin{minipage}{0.31\columnwidth}
            \centering
            \includegraphics[width=\textwidth]{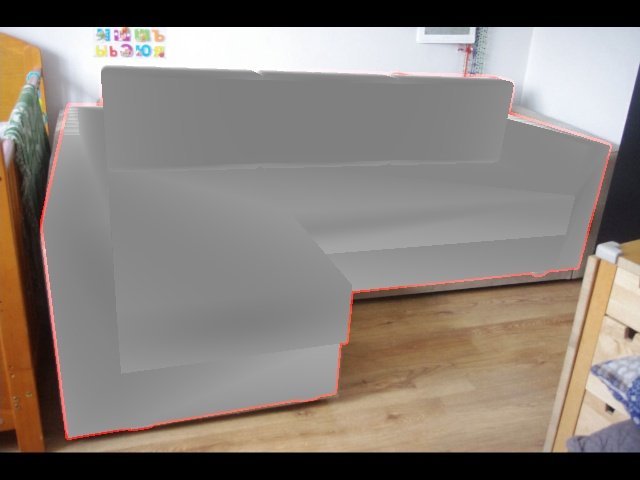}
        \end{minipage}\\[0.5mm]

    \small{6}
        \begin{minipage}{0.31\columnwidth}
            \centering
            \includegraphics[width=\textwidth]{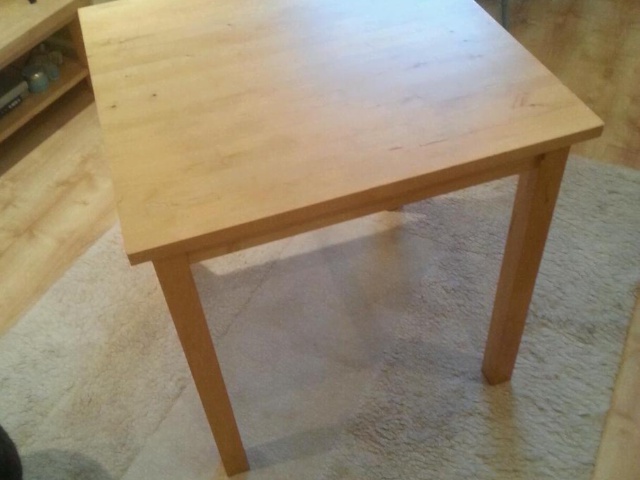}
        \end{minipage}
        \begin{minipage}{0.31\columnwidth}
            \centering
            \includegraphics[width=\textwidth]{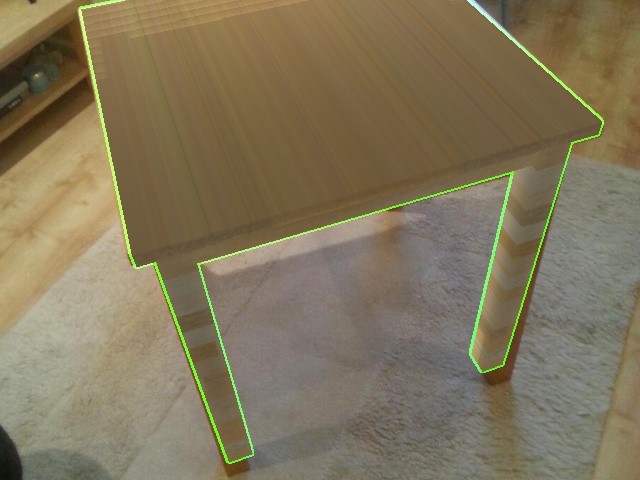}
        \end{minipage}
        \begin{minipage}{0.31\columnwidth}
            \centering
            \includegraphics[width=\textwidth]{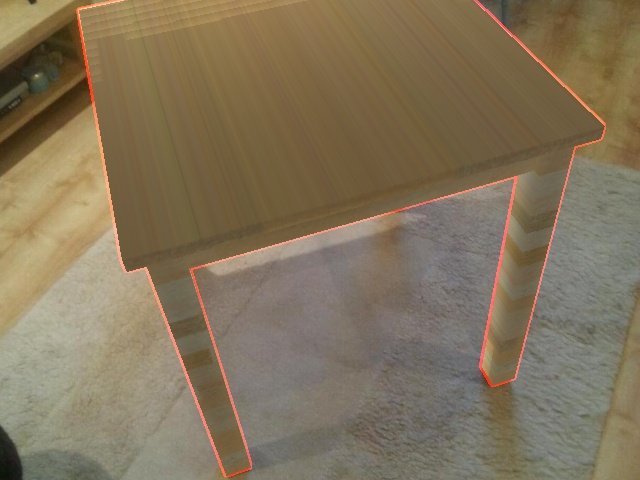}
        \end{minipage}\\[0.5mm]

    \small{7}
        \begin{minipage}{0.31\columnwidth}
            \centering
            \includegraphics[width=\textwidth]{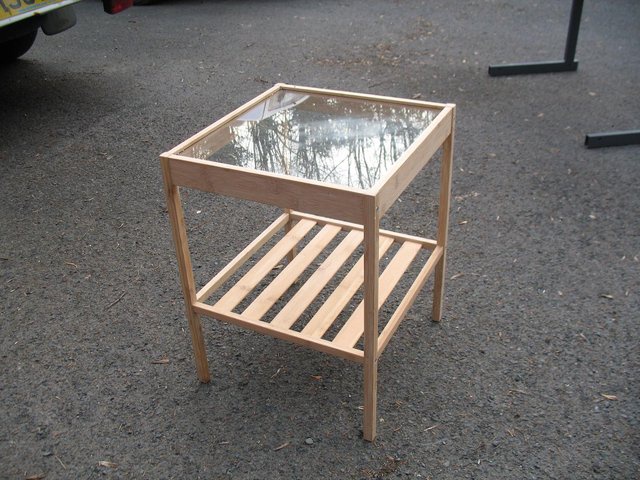}
        \end{minipage}
        \begin{minipage}{0.31\columnwidth}
            \centering
            \includegraphics[width=\textwidth]{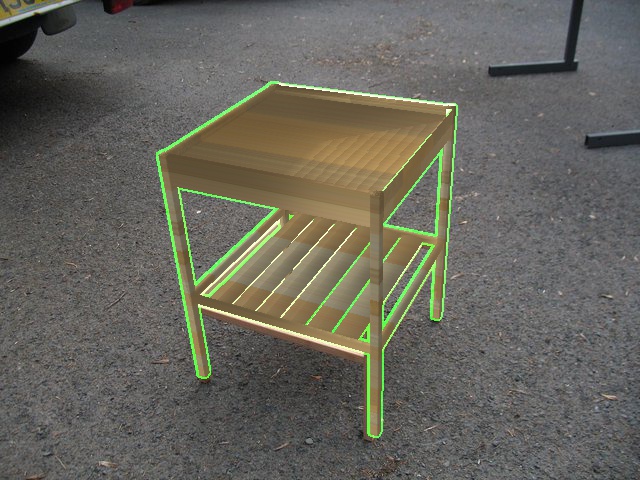}
        \end{minipage}
        \begin{minipage}{0.31\columnwidth}
            \centering
            \includegraphics[width=\textwidth]{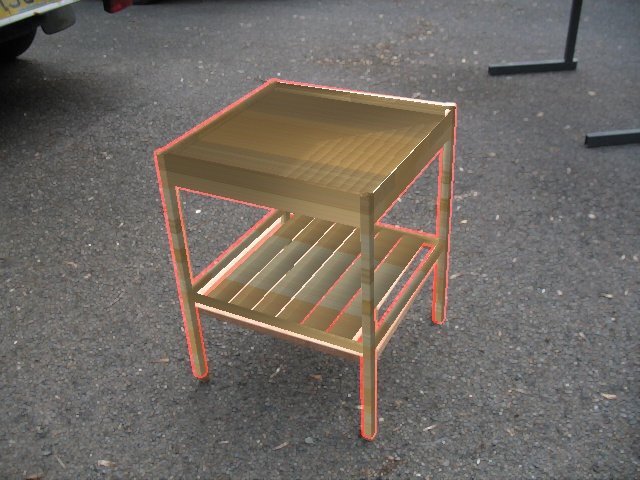}
        \end{minipage}\\[0.5mm]
        
    \small{8}
    \begin{minipage}{0.31\columnwidth}
            \centering
            \includegraphics[width=\textwidth]{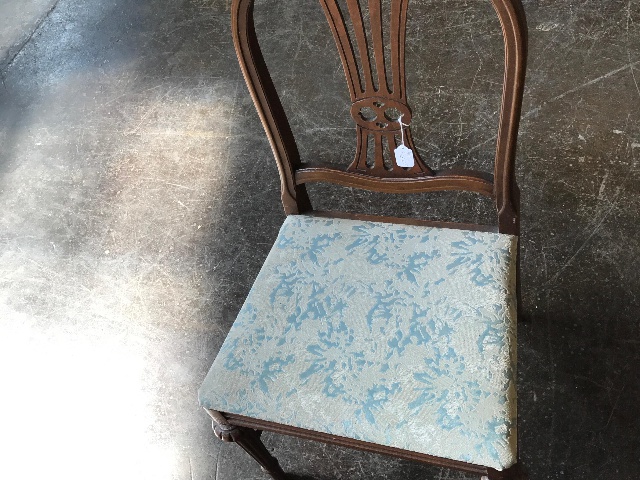}
        \end{minipage}
        \begin{minipage}{0.31\columnwidth}
            \centering
            \includegraphics[width=\textwidth]{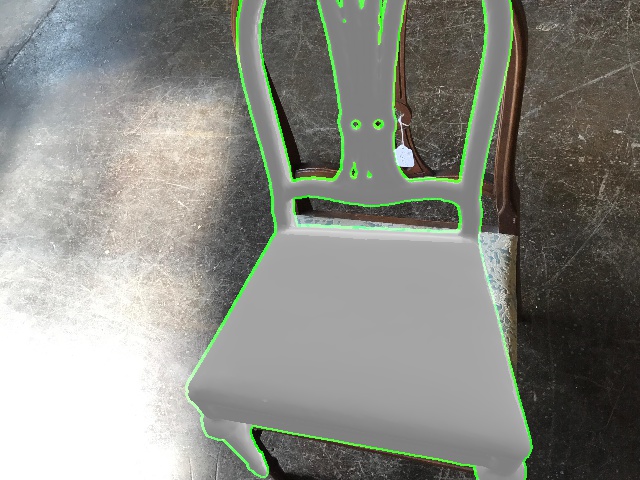}
        \end{minipage}
        \begin{minipage}{0.31\columnwidth}
            \centering
            \includegraphics[width=\textwidth]{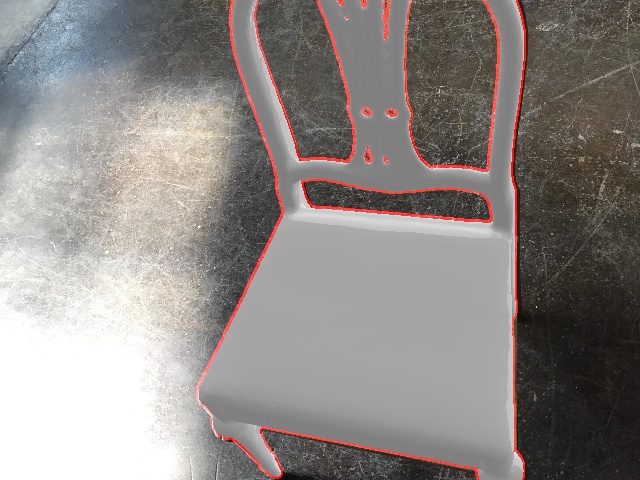}
        \end{minipage}\\[0.5mm]
    
    \small{9}
    \begin{minipage}{0.31\columnwidth}
            \centering
            \includegraphics[width=\textwidth]{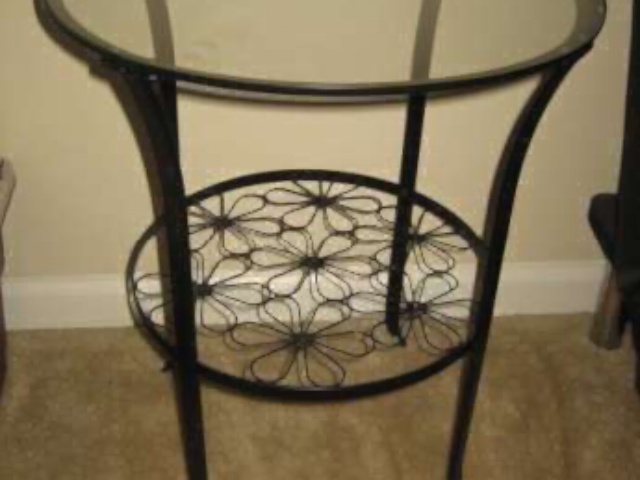}
        \end{minipage}
        \begin{minipage}{0.31\columnwidth}
            \centering
            \includegraphics[width=\textwidth]{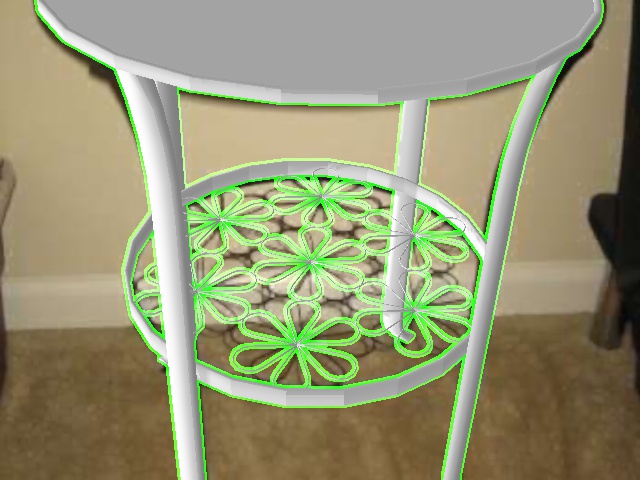}
        \end{minipage}
        \begin{minipage}{0.31\columnwidth}
            \centering
            \includegraphics[width=\textwidth]{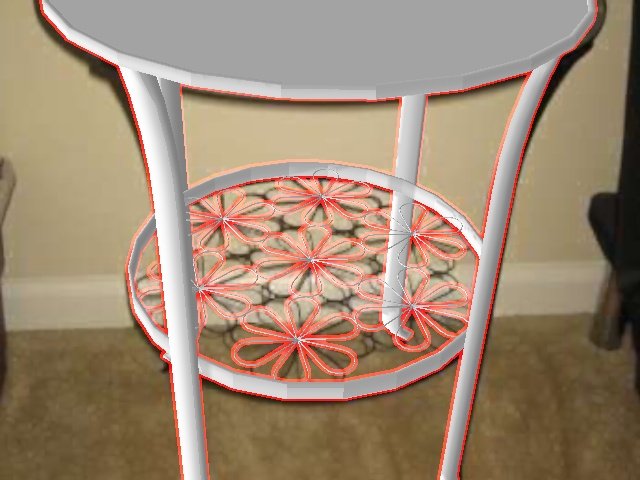}
        \end{minipage}\\[0.5mm]
    \small{\hspace{-3mm}10}
    \begin{minipage}{0.31\columnwidth}
            \centering
            \includegraphics[width=\textwidth]{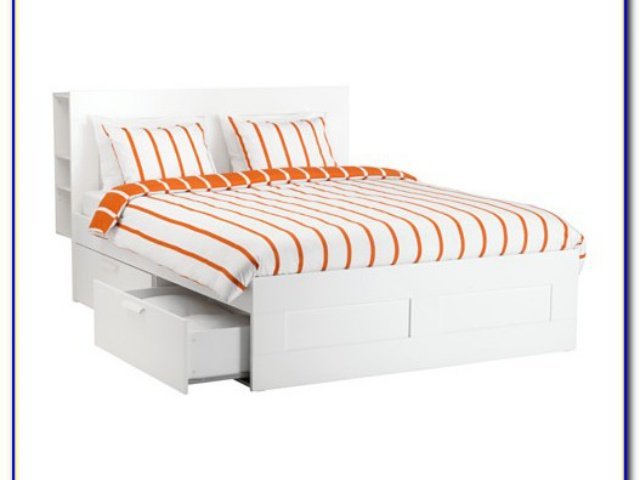}
        \end{minipage}
        \begin{minipage}{0.31\columnwidth}
            \centering
            \includegraphics[width=\textwidth]{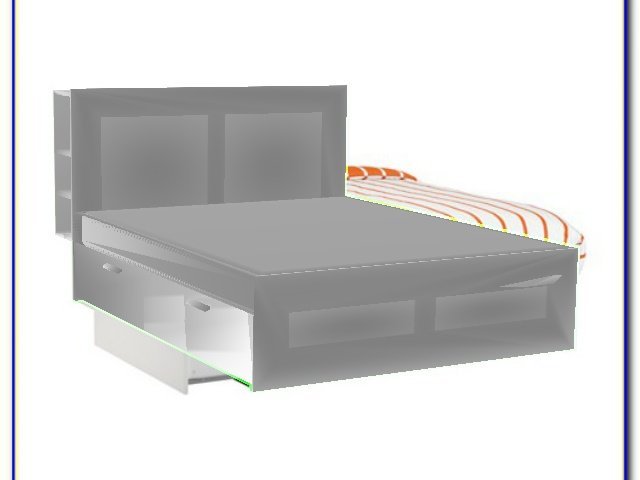}
        \end{minipage}
        \begin{minipage}{0.31\columnwidth}
            \centering
            \includegraphics[width=\textwidth]{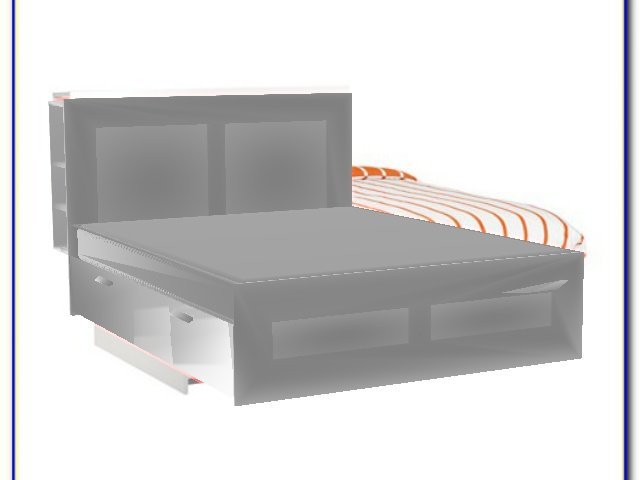}
        \end{minipage}\\[0.5mm]

    \caption{\textbf{Pix3D qualitative results.} 
    For each example (each row), we show the input image (left), ground truth focal length and pose annotation (center) and our prediction (right). We overlay a rendering of the detected 3D model with the jointly estimated 6D pose and focal length. Notice how our method produces precise alignments for truncated objects (rows 1, 8, 9) and handles large perspective effects (rows 3, 5, 6). Notice also that in row 8 our prediction is better than the manually annotated ground truth. 
    }
    \label{fig:qual_examples}
    \vspace*{-5mm}
\end{figure}

\begin{figure}[tbp]
    \centering
    \small{\hspace{1.5mm}1}
        \begin{minipage}{0.31\columnwidth}
            {\small Input image\vspace{1mm}}
            \centering
            \includegraphics[width=\textwidth]{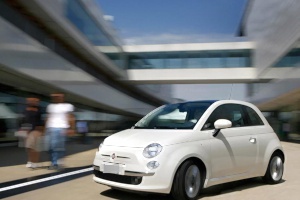}
        \end{minipage}
        \begin{minipage}{0.31\columnwidth}
            {\small Ground truth\vspace{1mm}}
            \centering
            \includegraphics[width=\textwidth]{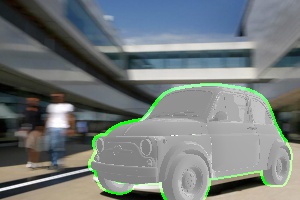}
        \end{minipage}
        \begin{minipage}{0.31\columnwidth}
            {\small Our prediction\vspace{1mm}}
            \centering
            \includegraphics[width=\textwidth]{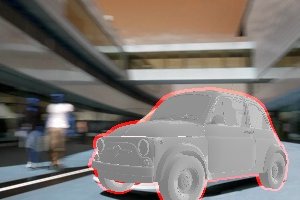}
        \end{minipage}\\[0.5mm]    
    \small{\hspace{1.5mm}2}
        \begin{minipage}{0.31\columnwidth}
            \centering
            \includegraphics[width=\textwidth]{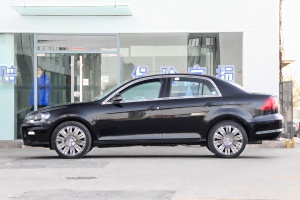}
        \end{minipage}
        \begin{minipage}{0.31\columnwidth}
            \centering
            \includegraphics[width=\textwidth]{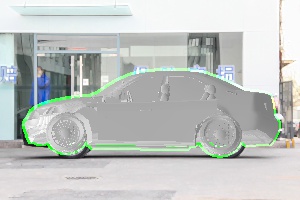}
        \end{minipage}
        \begin{minipage}{0.31\columnwidth}
            \centering
            \includegraphics[width=\textwidth]{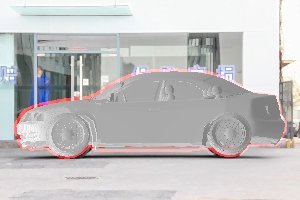}
        \end{minipage}\\[0.5mm]
    \small{\hspace{1.5mm}3}
        \begin{minipage}{0.31\columnwidth}
            \centering
            \includegraphics[width=\textwidth]{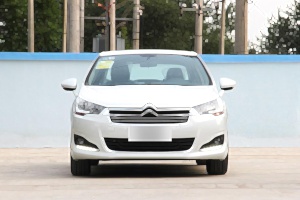}
        \end{minipage}
        \begin{minipage}{0.31\columnwidth}
            \centering
            \includegraphics[width=\textwidth]{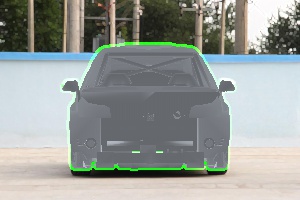}
        \end{minipage}
        \begin{minipage}{0.31\columnwidth}
            \centering
            \includegraphics[width=\textwidth]{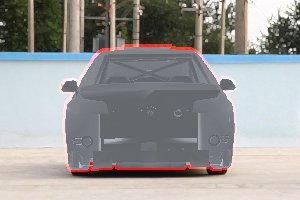}
        \end{minipage}\\[0.5mm]
    \small{\hspace{1.5mm}4}
        \begin{minipage}{0.31\columnwidth}
            \centering
            \includegraphics[width=\textwidth]{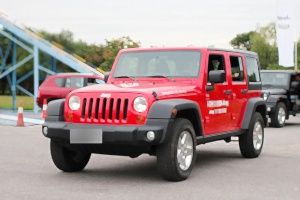}
        \end{minipage}
        \begin{minipage}{0.31\columnwidth}
            \centering
            \includegraphics[width=\textwidth]{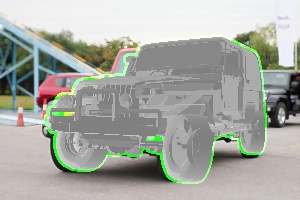}
        \end{minipage}
        \begin{minipage}{0.31\columnwidth}
            \centering
            \includegraphics[width=\textwidth]{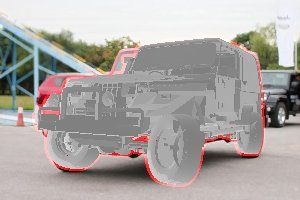}
        \end{minipage}\\[0.5mm]
    \small{\hspace{1.5mm}5}
        \begin{minipage}{0.31\columnwidth}
            \centering
            \includegraphics[width=\textwidth]{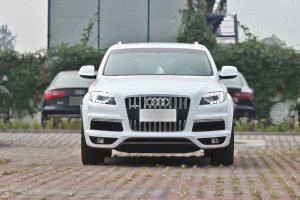}
        \end{minipage}
        \begin{minipage}{0.31\columnwidth}
            \centering
            \includegraphics[width=\textwidth]{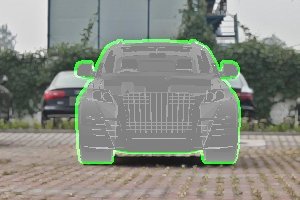}
        \end{minipage}
        \begin{minipage}{0.31\columnwidth}
            \centering
            \includegraphics[width=\textwidth]{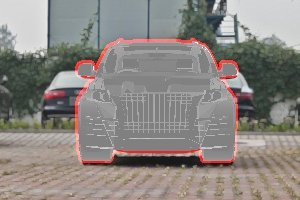}
        \end{minipage}\\[0.5mm]
    \small{\hspace{1.5mm}6}
        \begin{minipage}{0.31\columnwidth}
            \centering
            \includegraphics[width=\textwidth]{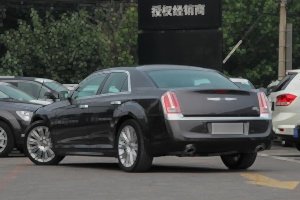}
        \end{minipage}
        \begin{minipage}{0.31\columnwidth}
            \centering
            \includegraphics[width=\textwidth]{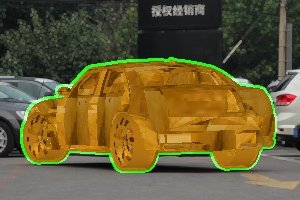}
        \end{minipage}
        \begin{minipage}{0.31\columnwidth}
            \centering
            \includegraphics[width=\textwidth]{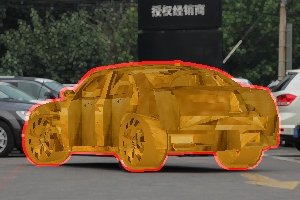}
        \end{minipage}\\[0.5mm]

    \small{\hspace{1.5mm}7}
        \begin{minipage}{0.31\columnwidth}
            \centering
            \includegraphics[width=\textwidth]{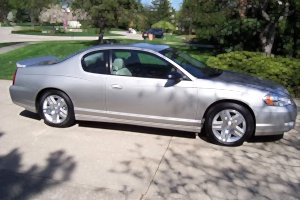}
        \end{minipage}
        \begin{minipage}{0.31\columnwidth}
            \centering
            \includegraphics[width=\textwidth]{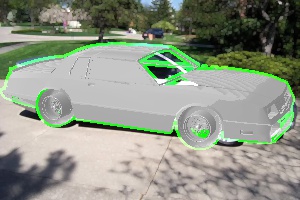}
        \end{minipage}
        \begin{minipage}{0.31\columnwidth}
            \centering
            \includegraphics[width=\textwidth]{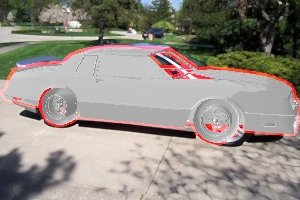}
        \end{minipage}\\[0.5mm]
     \small{\hspace{1.5mm}8}
        \begin{minipage}{0.31\columnwidth}
            \centering
            \includegraphics[width=\textwidth]{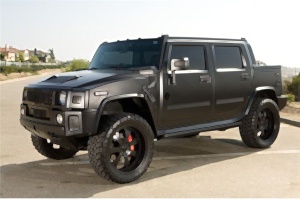}
        \end{minipage}
        \begin{minipage}{0.31\columnwidth}
            \centering
            \includegraphics[width=\textwidth]{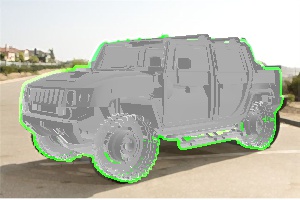}
        \end{minipage}
        \begin{minipage}{0.31\columnwidth}
            \centering
            \includegraphics[width=\textwidth]{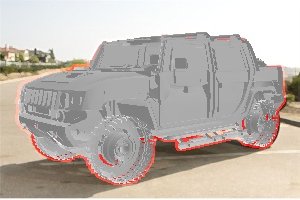}
        \end{minipage}\\[0.5mm]
    
    \small{\hspace{1.5mm}9}
    \begin{minipage}{0.31\columnwidth}
            \centering
            \includegraphics[width=\textwidth]{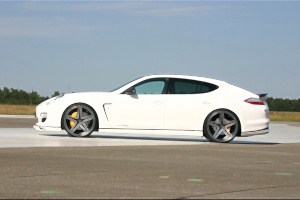}
        \end{minipage}
        \begin{minipage}{0.31\columnwidth}
            \centering
            \includegraphics[width=\textwidth]{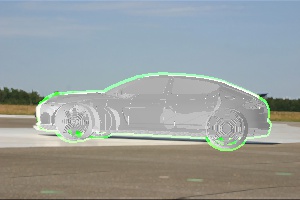}
        \end{minipage}
        \begin{minipage}{0.31\columnwidth}
            \centering
            \includegraphics[width=\textwidth]{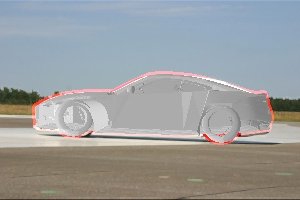}
        \end{minipage}\\[0.5mm]

    \small{10}    
    \begin{minipage}{0.31\columnwidth}
            \centering
            \includegraphics[width=\textwidth]{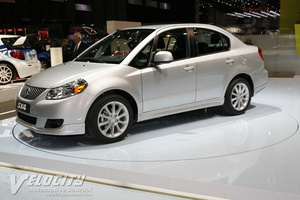}
        \end{minipage}
        \begin{minipage}{0.31\columnwidth}
            \centering
            \includegraphics[width=\textwidth]{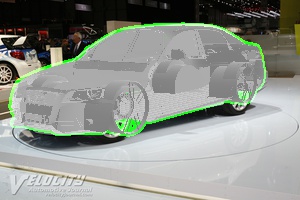}
        \end{minipage}
        \begin{minipage}{0.31\columnwidth}
            \centering
            \includegraphics[width=\textwidth]{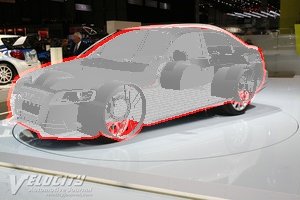}
        \end{minipage}\\[0.5mm]
        
    \small{11}    
    \begin{minipage}{0.31\columnwidth}
            \centering
            \includegraphics[width=\textwidth]{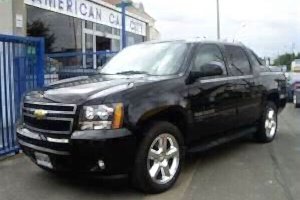}
        \end{minipage}
        \begin{minipage}{0.31\columnwidth}
            \centering
            \includegraphics[width=\textwidth]{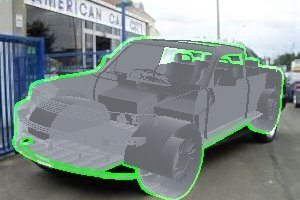}
        \end{minipage}
        \begin{minipage}{0.31\columnwidth}
            \centering
            \includegraphics[width=\textwidth]{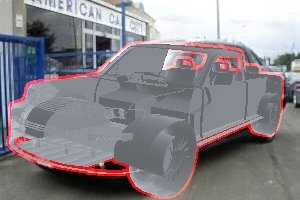}
        \end{minipage}\\[0.5mm]
    \small{12}    
    \begin{minipage}{0.31\columnwidth}
            \centering
            \includegraphics[width=\textwidth]{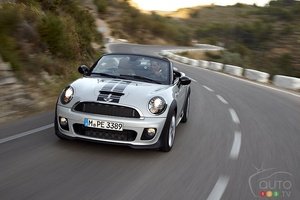}
        \end{minipage}
        \begin{minipage}{0.31\columnwidth}
            \centering
            \includegraphics[width=\textwidth]{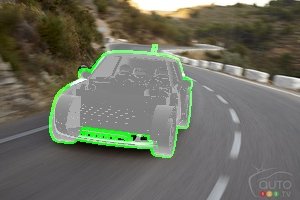}
        \end{minipage}
        \begin{minipage}{0.31\columnwidth}
            \centering
            \includegraphics[width=\textwidth]{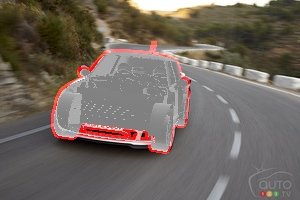}
        \end{minipage}\\[0.5mm]

    \caption{{Example qualitative results on the \textbf{CompCars} (rows 1-6) and \textbf{Stanford cars} (rows 7-12) datasets.} 
    }
    \label{fig:qual_examples_stanford}
    \vspace*{-3mm}
\end{figure}

\paragraph{Qualitative results.} We report examples of qualitative results for our method on the four classes of the Pix3D dataset in Fig.~\ref{fig:qual_examples} and on Stanford cars and CompCars datasets in Fig.~\ref{fig:qual_examples_stanford}. Please note that the renderings of the predictions (taking into account focal length and object 6D pose) show precise alignment with the observed image for in-the-wild photographs. Notably, these qualitative results demonstrate the robustness of our approach to large object truncation and strong perspective effects. 
Please see the \textbf{supplementary material} for additional qualitative results.

\section{Conclusion}

We have demonstrated successful joint estimation of camera-object 6D pose and camera focal length given a single still image. Key to our success was our extension of render and compare that incorporated the estimated focal length in the iterative update rules and a disentangled loss for training. We have shown that our approach produces lower-error focal length and pose estimates compared to prior art.
Our approach can be extended to other camera intrinsic parameters besides focal length, including different forms of camera distortions, provided that they can be reliably rendered.
This work opens up the possibility of downstream applications in augmented reality/computer graphics and reasoning over ``in-the-wild'' articulated and interacted objects in video.

\ifCLASSOPTIONcompsoc
  \section*{Acknowledgments}
\else
  \section*{Acknowledgment}
\fi
{
This work was partly supported by the Ministry of Education, Youth and Sports of the Czech Republic through the e-INFRA CZ (ID:90140), the French government under management of Agence Nationale de la Recherche as part of the ``Investissements d'avenir'' program, reference ANR19-P3IA-0001 (PRAIRIE 3IA Institute), and by the European Union's Horizon Europe projects euROBIN (No. 101070596),  AGIMUS (No. 101070165), ERC DISCOVER (No. 101076028) and ERC FRONTIER (No. 101097822). Views and opinions expressed are however those of the author(s) only and do not necessarily reflect those of the European Union or the European Commission. Neither the European Union nor the European Commission can be held responsible for them.
}

\FloatBarrier
\beginsuppmat
\section{Supplementary Material}
The supplementary material is organized as follows. Sec.~\ref{sec:update_rule_derivation} shows the full derivation of the 6D pose update rule.
In Sec.~\ref{sec:metrics} we give the details of the evaluation metrics. In Sec.~\ref{refiner-iterations} we show detailed results on the benefits of multiple refiner iterations. Sec.~\ref{detailed-results} provides additional quantitative evaluation of the performance of our model. Finally, in Sec.~\ref{sec:additional-qualitative-results} we provide more qualitative results on Pix3D, Stanford Cars, and CompCars datasets.

\subsection{Derivation of the 6D pose update rule}
\label{sec:update_rule_derivation}
In this section, we show the full derivation of the 6D pose update rule used for coarse and refiner networks, as presented in the main paper, and discuss the difference compared to the update rule used in FocalPose~\cite{ponimatkin2022focal}.
Let $[x,y,z]$ be the 3D coordinates of the center of the object and $[a,b]$ its image coordinates after projection. We use a simplified pinhole camera model where the camera principal point is at the origin of the image coordinate system ($c_x=c_y=0$) and the world coordinate system is placed at the object center (see Fig.~2~(b) in the main paper). We get the following camera projection equation in homogeneous coordinates:
\begin{equation}
    \lambda \begin{pmatrix}a\\b\\1 \end{pmatrix} =
    \begin{pmatrix}
        f_x & 0   & c_x\\
        0   & f_y & c_y\\
        0   & 0   & 1  \\
    \end{pmatrix} \begin{pmatrix}x\\y\\z\end{pmatrix} = 
    \begin{pmatrix}
        f & 0 & 0\\
        0 & f & 0\\
        0 & 0 & 1\\
    \end{pmatrix} \begin{pmatrix}x\\y\\z\end{pmatrix} \, , 
\end{equation}
which can be written as separate equations for $x$, $y$ and $z$:
\begin{equation}\begin{aligned}
    \lambda a &= fx \, , \\
    \lambda b &= fy \, , \\
    \lambda   &= z \, ,
\end{aligned}\end{equation}
and further simplified as:
\begin{equation}
\label{eq:ab_proj}
\begin{aligned}
    a &= \frac{fx}{z} \, , \\
    b &= \frac{fy}{z} \, .
\end{aligned}
\end{equation}
Let $[v_x^k, v_y^k, v_y^z]$ be the object translation update predicted by the neural network $F$. We want $v_z^k$ to represent the ratio of camera-to-object depth between the object observed in the real image and the rendered image, thus we define the update rule for $z^{k+1}$ as:
\begin{equation}
    z^{k+1} = v_z^kz^k  \,,  
\end{equation}
where $z^{k}$ and $z^{k+1}$ are the old and new estimations of the $z$-translation respectively, and $v_z^{k}$ is the corresponding network output for the $z$-translation at iteration $k$.
We want $[v_x^k$, $v_y^k]$ to represent the translation of the projected object center measured in pixels, i.e:
\begin{equation}
\label{eq:ab}
\begin{aligned}
    a^{k+1} &= a^k + v_x^k \, , \\
    b^{k+1} &= b^k + v_y^k \, .
\end{aligned}\end{equation}
Next, in eq.~\eqref{eq:ab} we substitute the image space coordinates $a$ and $b$ by the expressions from eq.~\eqref{eq:ab_proj} obtaining:
\begin{equation}\begin{aligned}
    \frac{f^{k+1}x^{k+1}}{z^{k+1}} &= \frac{f^kx^k}{z^k} + v_x^k \, , \\
    \frac{f^{k+1}y^{k+1}}{z^{k+1}} &= \frac{f^ky^k}{z^k} + v_y^k \, . 
\end{aligned}\end{equation}
By rearranging the above equations, we derive the final update rule for $x^{k+1}$ and $y^{k+1}$ as:
\begin{equation}\begin{aligned}
\label{eq:xy_update}
    x^{k+1} &= \left(v_x^k + \frac{f^kx^k}{z^k}\right)\frac{z^{k+1}}{f^{k+1}} \, , \\
    y^{k+1} &= \left(v_y^k + \frac{f^ky^k}{z^k}\right)\frac{z^{k+1}}{f^{k+1}} \, . 
\end{aligned}\end{equation}

\paragraph{Discussion.} To analyze the difference between~\eqref{eq:xy_update} and the update rule used in FocalPose~\cite{ponimatkin2022focal}, we first recall the translation part of the FocalPose~\cite{ponimatkin2022focal} update rule:
\begin{equation}\begin{aligned}
\label{eq:xy_update_old}
x^{k+1} &= \left(\frac{v_x^{k}}{f^{k+1}} + \frac{x^k}{z^k}\right)z^{k+1} \, , \\
y^{k+1} &= \left(\frac{v_y^{k}}{f^{k+1}} + \frac{y^k}{z^k}\right)z^{k+1} \, .
\end{aligned}\end{equation}
Comparing the new update rule~\eqref{eq:xy_update} to the original FocalPose~\cite{ponimatkin2022focal} update rule~\eqref{eq:xy_update_old}, it can be seen that eq.~\eqref{eq:xy_update_old} can be considered as an approximation of eq.~\eqref{eq:xy_update} -- the right-hand sides of the equations are equivalent up to the multiplicative factor $f^{k}/f^{k+1}$ for the terms $x^k/z^k$ and $y^k/z^k$ in parentheses. If the focal length does not change much between the update iterations, this factor will be close to $1$ and could be ignored. In other words, the original FocalPose~\cite{ponimatkin2022focal} update rule assumes the focal length to be constant when updating the translation components $x$ and $y$ of the 6D pose in each iteration and applies only the new focal length estimate $f^{k+1}$. As shown in the ablations in the main paper, the new update rule derived in this work given by~\eqref{eq:xy_update} achieves slightly better results.

\subsection{Evaluation criteria}
\label{sec:metrics}

We now recall the metrics presented in~\cite{wang20183d}, commonly~\cite{wang20183d,grabner2019gp2c,han2020gcvnet} used on these datasets and also used in this work. 

\paragraph{Detection metric.} We report the detection accuracy $Acc_{D_{0.5}}$ which corresponds to the percentage of images for which the intersection over union between the ground truth and predicted 2D bounding box is larger than $0.5$. 

Note that an incorrect object prediction is not penalized by this metric as our method can predict the focal length and object 6D pose even if the model is only approximate as long as it belongs to the correct category for which the 3D models are approximately aligned, similar to~\cite{wang20183d,grabner2019gp2c,han2020gcvnet}.

\paragraph{6D pose metrics.} We report the point matching error $e_{R,t}$ that measures the error between the 3D points of the object model transformed with the ground truth and with the estimated 6D pose with respect to the camera:
\begin{equation}
\label{eq:errorRt}
  e_{R,t} = \frac{d_{\text{bbox}}}{d_{\text{img}}} \underset{p \in \mathcal{M}^{\star}}{\mathrm{avg}} \frac{||(R p + t) -  (\hat{R} p + \hat{t})||_{2}}{||\hat{t}||_{2}},
\end{equation}
where $d_{{\text{bbox}}}$ is the diagonal of the ground truth 2D bounding box, $d_{\text{img}}$ is the diagonal of the image, $\mathcal{M}^{\star}$ is the 3D model of the ground truth object instance, $(R, t)$ is the predicted 6D pose and $(\hat{R}, \hat{t})$ is the ground truth 6D pose. Note that the point error in 3D (the numerator of~\eqref{eq:errorRt})  is normalized by the ground truth object-to-camera distance $||\hat{t}||_{2}$ and multiplied by the relative size of the object in the image $\frac{d_{\text{bbox}}}{d_{\text{img}}}$~\cite{grabner2019gp2c}.

Following~\cite{grabner2019gp2c}, we also use metrics that evaluate separately the quality of the estimated 3D translation and rotation. We use the rotation error computed using the geometric distance between the predicted rotation $R$ and the ground truth rotation $\hat{R}$ $e_{R}=\frac{||\mathrm{log}(\hat{R}^T R)||_{F}}{\sqrt{2}}$, and the normalized translation error
$e_{t}=\frac{||t - \hat{t}||_{2}}{||\hat{t}||_{2}}$, where $t$ is the predicted translation and $\hat{t}$ is the ground truth translation. For all the errors, we report the median value (denoted as $MedErr_{Rt}$, $MedErr_{R}$, $MedErr_{t}$, respectively). Following~\cite{grabner2019gp2c}, for the rotation error we also report the percentage of images with $e_{R} \leq \frac{\pi}{6}$ denoted as $Acc_{R\frac{\pi}{6}}$.

\paragraph{Focal length and reprojection metrics.} Following~\cite{grabner2019gp2c}, we report the relative focal length error $e_{f}=\frac{|\hat{f} - f|}{\hat{f}}$ between the estimated focal length $f$ and the ground truth focal length $\hat{f}$. We also report the reprojection error $e_{P}$ which is similar to the error of 6D pose (eq.~\eqref{eq:errorRt} but reprojects the 3D points into the image, also taking into account the estimated focal length $f$:
\begin{equation}
  e_{P} = \underset{p \in \mathcal{M}^{\star}}{\mathrm{avg}} \frac{||\pi(R, t, f, p) - \pi(\hat{R}, \hat{t}, \hat{f}, p)||_{2}}{d_{\text{bbox}}},
\end{equation}
where $p$ are the 3D points of the object model $\mathcal{M}^{\star}$ of the ground truth object instance, $\pi(K(f), R, t, p)$ is the reprojection of a 3D point $p$ using the estimated parameters, and $\pi(K(\hat{f}), \hat{R}, \hat{t}, p)$ is the reprojection of the same 3D point $p$ using ground truth parameters, and $d_{bbox}$ is the diagonal of the ground truth 2D bounding box. We report the median value of the reprojection error $MedErr_{P}$ and the percentage of images where the reprojection error is below $0.1$ of the image, $Acc_{P_{0.1}}$

\subsection{Multiple refiner iterations}
\label{refiner-iterations}
Finally, in Figure~\ref{fig:iter_errs}, we show how the model performance evolves with an increasing number of refiner iterations at inference time. Two effects can be observed. First, the translation and focal length errors tend to go down with the number of iterations and they empirically reach a fixed error value. On the other hand, we observe that the rotation errors can increase with the number of iterations, which can be seen for the Pix3D table class.

We believe this finding could be attributed to the fact that our refiner model is trained only for one iteration.  These results can be potentially improved by increasing the number of refiner iterations during training at the cost of additional compute.

\begin{figure}[t]
    \centering
        \begin{minipage}{0.31\columnwidth}
            {\small Input image\vspace{1mm}}
            \centering
            \includegraphics[width=\textwidth]{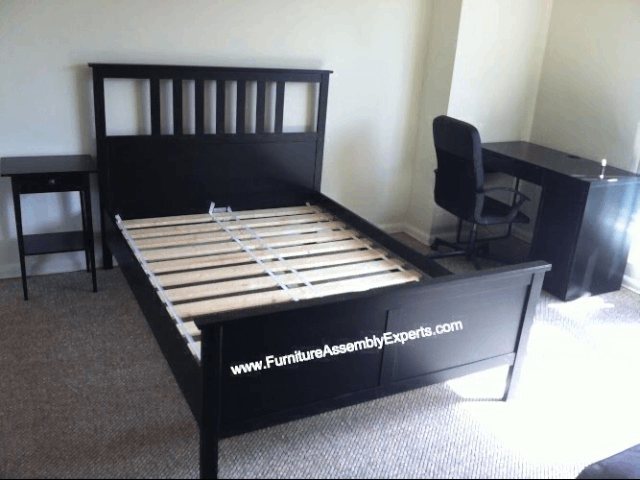}
        \end{minipage}
        \begin{minipage}{0.31\columnwidth}
            {\small Ground truth\vspace{1mm}}
            \centering
            \includegraphics[width=\textwidth]{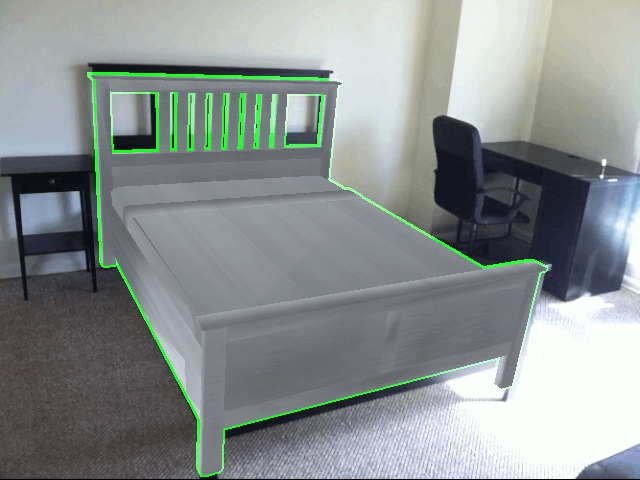}
        \end{minipage}
        \begin{minipage}{0.31\columnwidth}
            {\small Our prediction\vspace{1mm}}
            \centering
            \includegraphics[width=\textwidth]{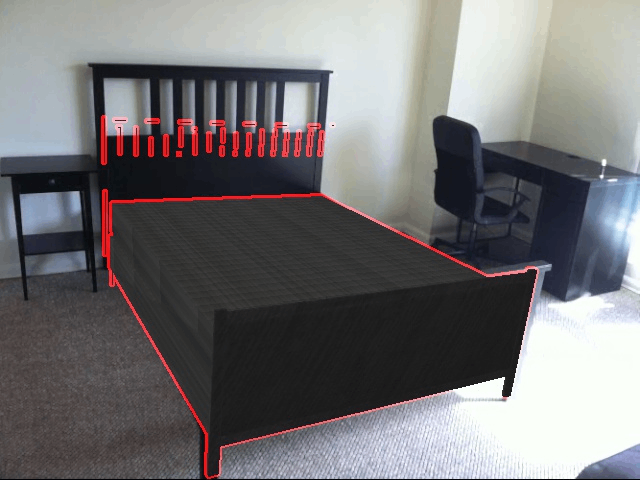}
        \end{minipage}\\[1mm]
    \caption{\textbf{Inaccuracies in ground truth annotations in the Pix3D dataset.} Example of an alignment with an incorrect 3D model predicted by our approach (right) that results in a lower 3D translation and focal length errors compared to the aligned ground truth 3D model (middle). This is caused by a mismatch between the bed depicted in the input image (with no mattress) and the ground truth 3D model. }
    \label{fig:bad_model_good_error}
    \vspace*{-5mm}
\end{figure}

\subsection{Detailed results}
\label{detailed-results}
To show fine-grained information about the errors of our model, we provide a set of histograms and plots that are complementary to the results tables in the main paper. 

In Figure~\ref{fig:err_hist_pix3d} we show the distributions of rotation and reprojection errors for the Pix3D dataset and in Figure~\ref{fig:err_hist_cars} for the CompCars and Stanford Cars datasets. For the Pix3D chair and table classes we observe peaks at $\sim\hspace{-0.75mm}90^\circ$ intervals, which suggests that many errors in those classes come from symmetrical objects that cause problems for our approach. For the car datasets we observe a large peak at $\sim\hspace{-0.75mm}180^\circ$, which also shows that some of the car models are fitted to incorrect orientations due to (almost) symmetrical models.

Figure~\ref{fig:threshold_errs} shows rotation and projection accuracies at different projection and rotation error thresholds. The standard thresholds used in previous work are quite loose and correspond to the right-most endpoints of the reported graphs, \ie, reprojection error of 0.1 (10\% of the object bounding box size) and rotation error of 30 degrees. We observe that the accuracy of our approach drops only slightly over a range of tighter thresholds, up to 0.05 relative reprojection error and up to about 15$^\circ$ rotation error.  For stricter thresholds (below around 0.05 and 15$^\circ$), the accuracy of our model starts dropping  significantly, which shows that there is still space for improvement in future work. 

\begin{figure*}[t]
    \centering
    \small{1}
        \begin{minipage}{0.666\columnwidth}
            \textbf{Focal length error}
            \centering
            \includegraphics[width=\textwidth]{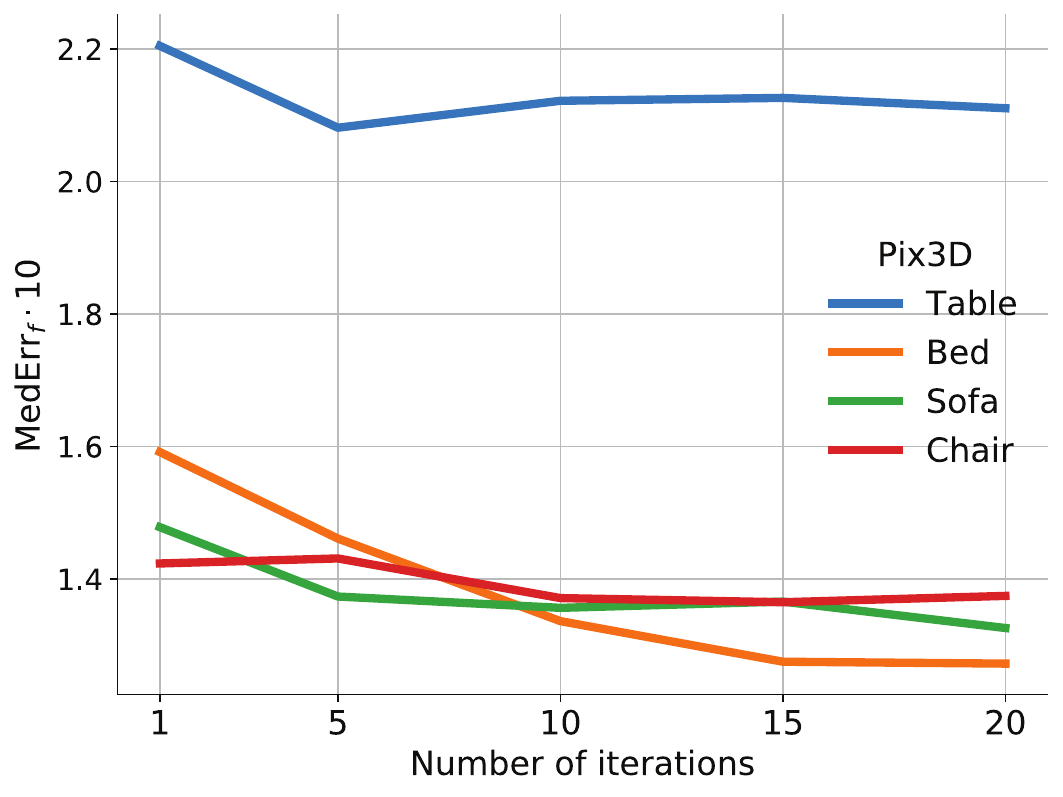}
        \end{minipage}
        \begin{minipage}{0.666\columnwidth}
            \textbf{Translation error}
            \centering
            \includegraphics[width=\textwidth]{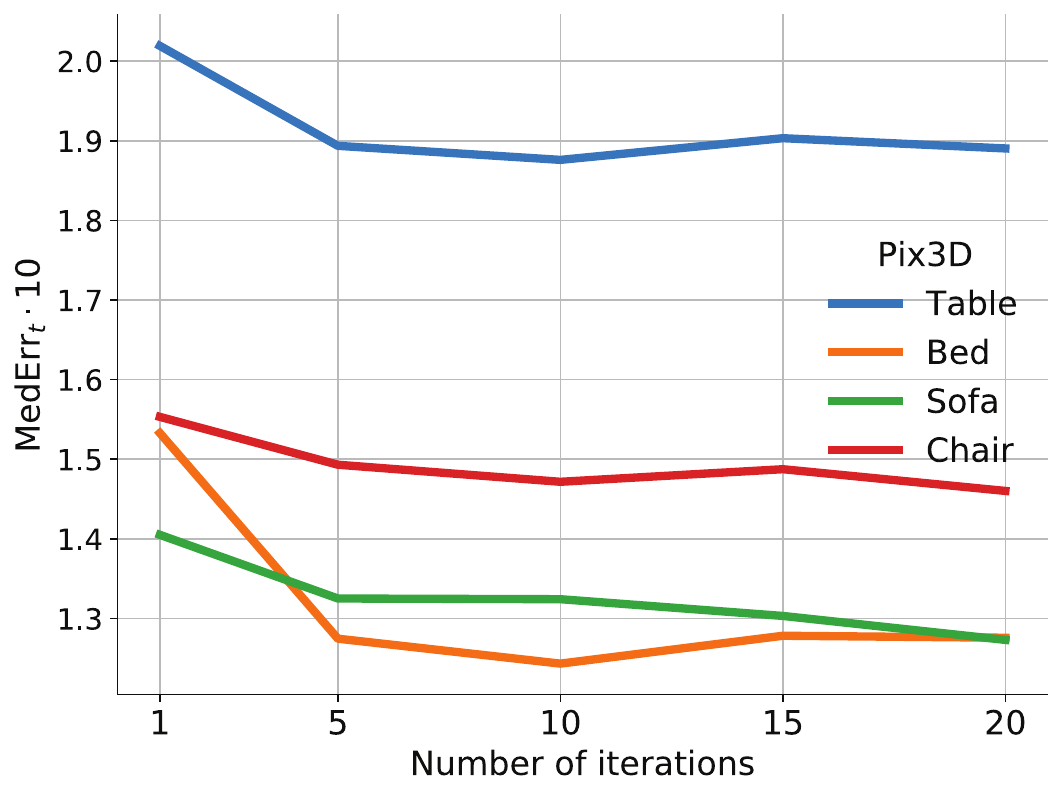}
        \end{minipage}
        \begin{minipage}{0.666\columnwidth}
            \textbf{Rotation error}
            \centering
            \includegraphics[width=\textwidth]{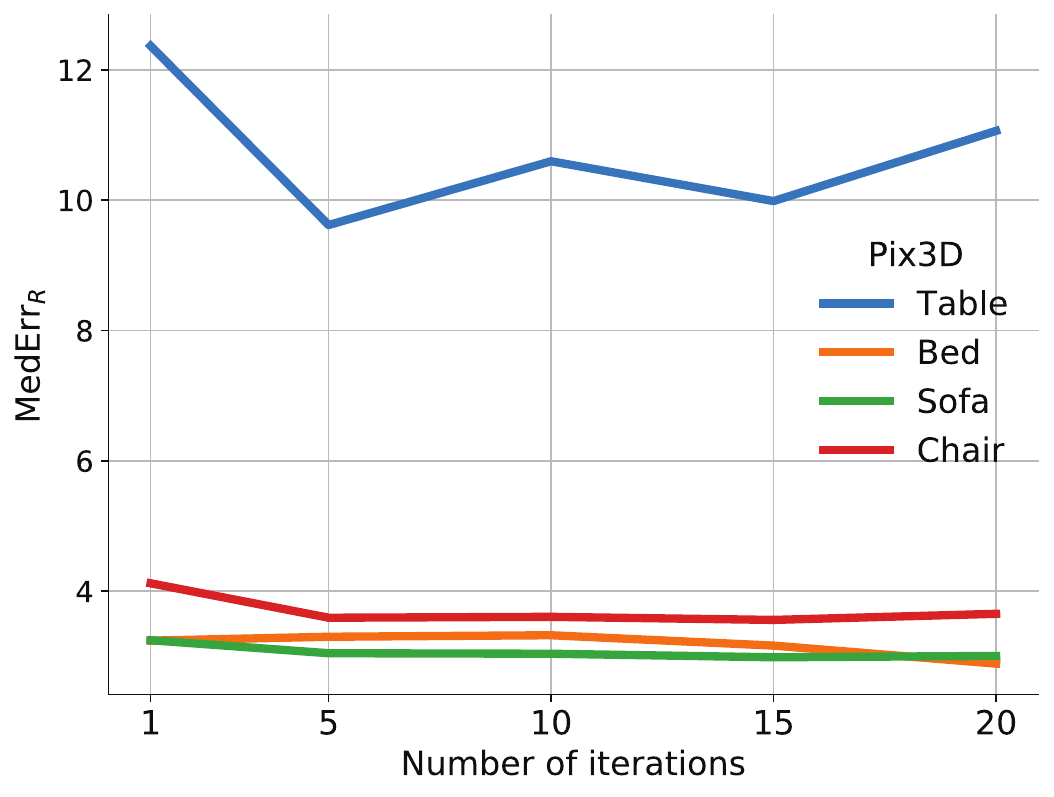}
        \end{minipage}\\[1mm]
    
    \caption{\textbf{Evolution of errors with an increasing number of refiner iterations at inference time for different object classes on the Pix3D dataset}. } 
    \label{fig:iter_errs}
    \vspace*{-5mm}
\end{figure*}

\begin{figure*}[t]
    \centering
    \small{1}
        \begin{minipage}{0.8\columnwidth}
            \textbf{Projection error histograms}
            \centering
            \includegraphics[width=\textwidth]{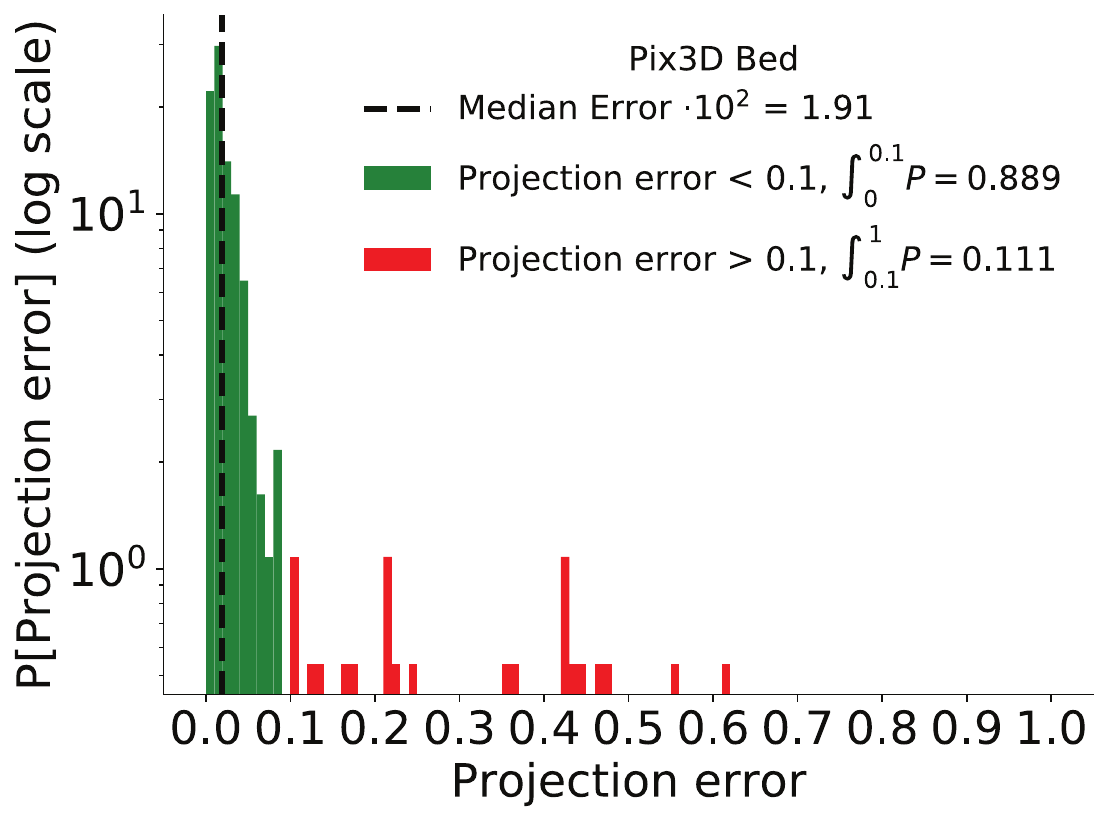}
        \end{minipage}
        \begin{minipage}{0.8\columnwidth}
            \textbf{Rotation error histograms}
            \centering
            \includegraphics[width=\textwidth]{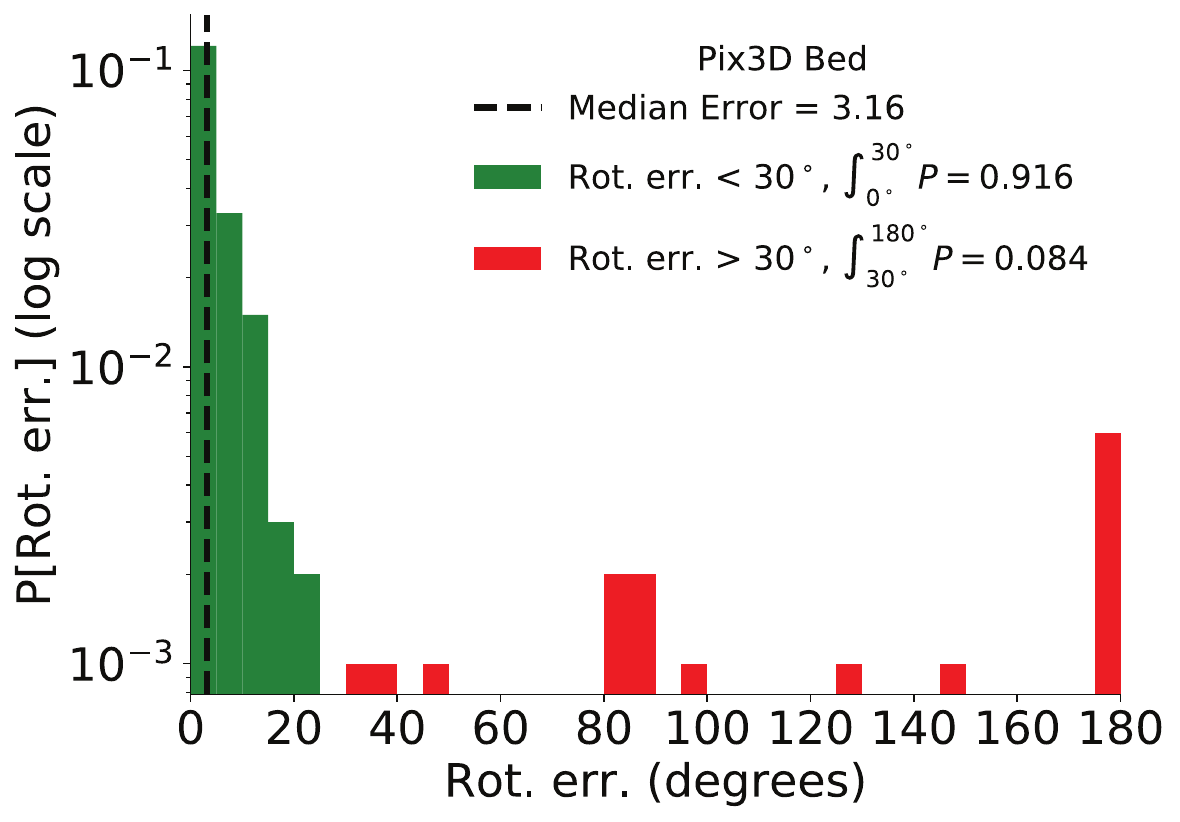}
        \end{minipage}\\[1mm]
    \small{2}
        \begin{minipage}{0.8\columnwidth}
            \centering
            \includegraphics[width=\textwidth]{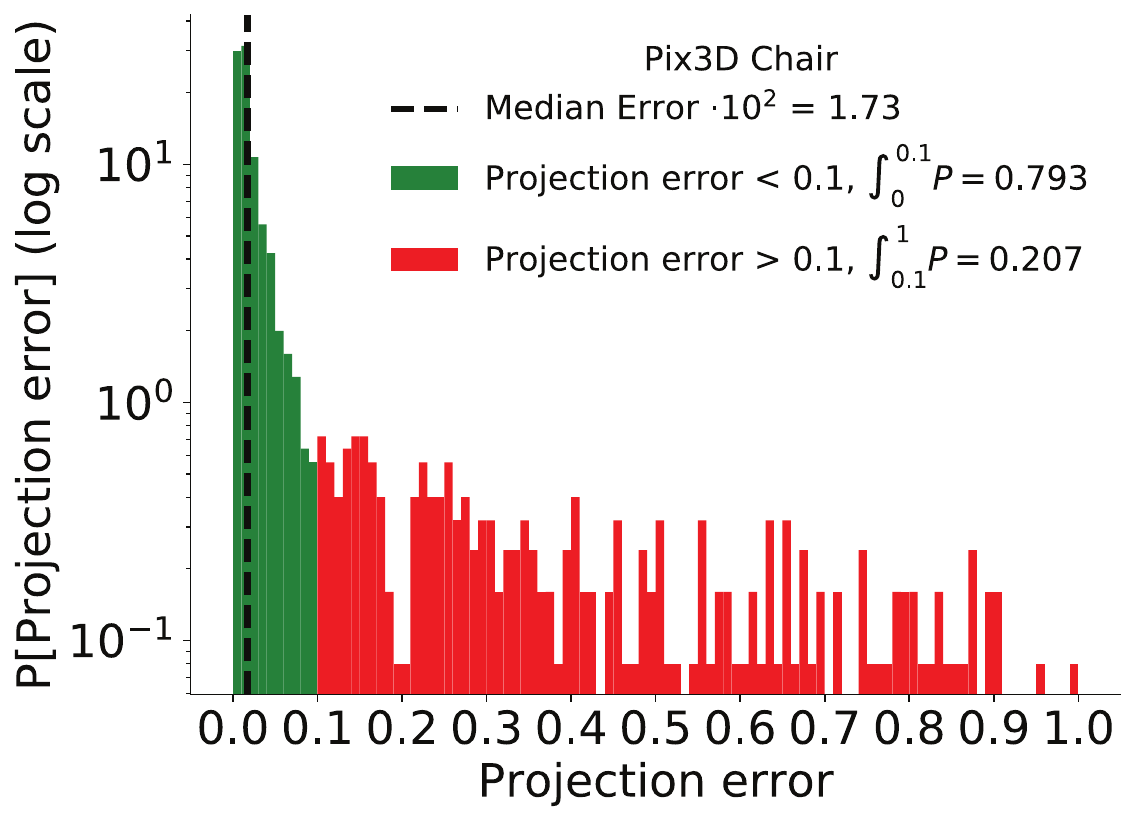}
        \end{minipage}
        \begin{minipage}{0.8\columnwidth}
            \centering
            \includegraphics[width=\textwidth]{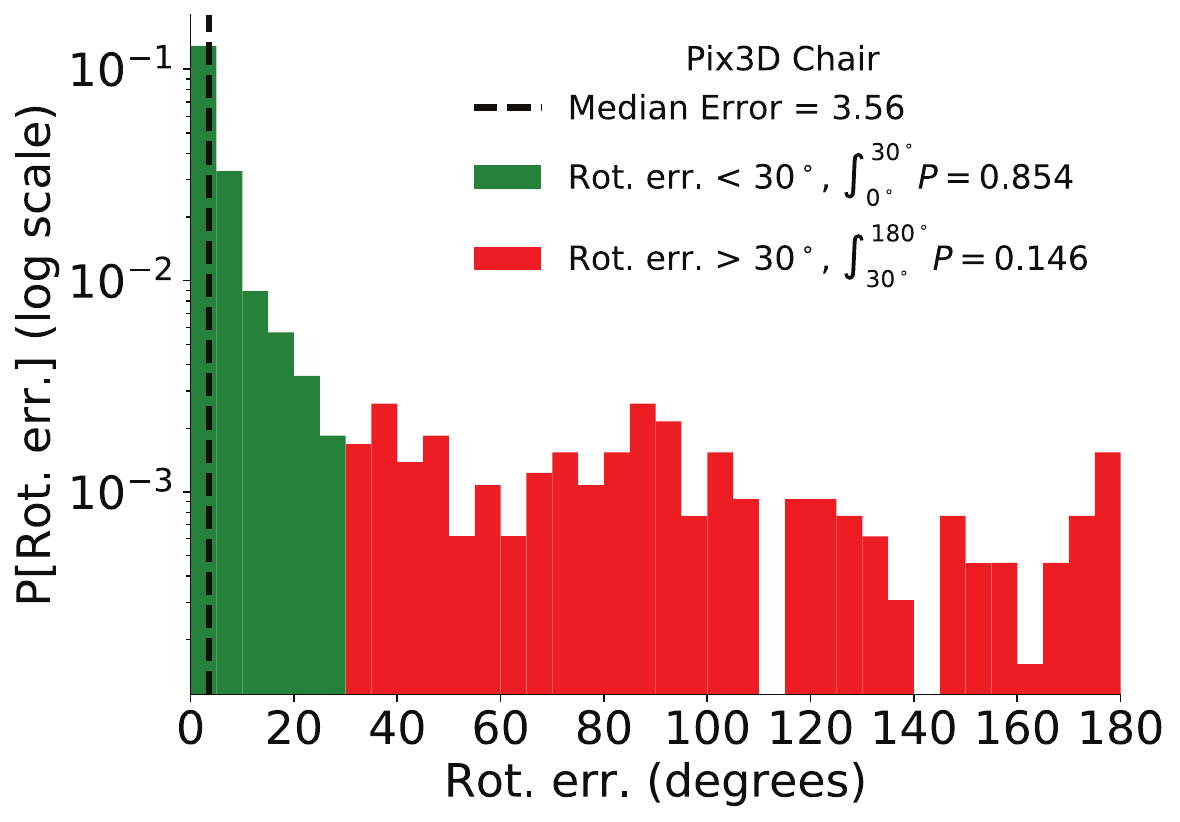}
        \end{minipage}\\[1mm]
     \small{3}
        \begin{minipage}{0.8\columnwidth}
            \centering
            \includegraphics[width=\textwidth]{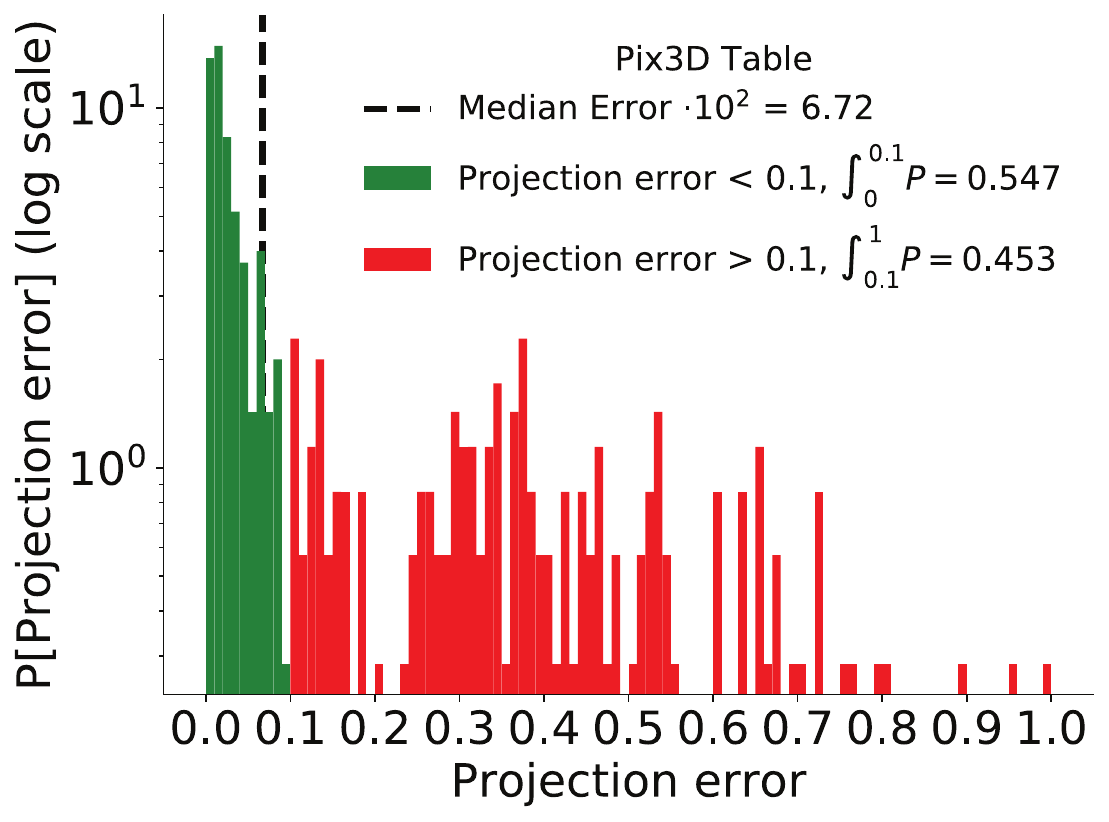}
        \end{minipage}
        \begin{minipage}{0.8\columnwidth}
            \centering
            \includegraphics[width=\textwidth]{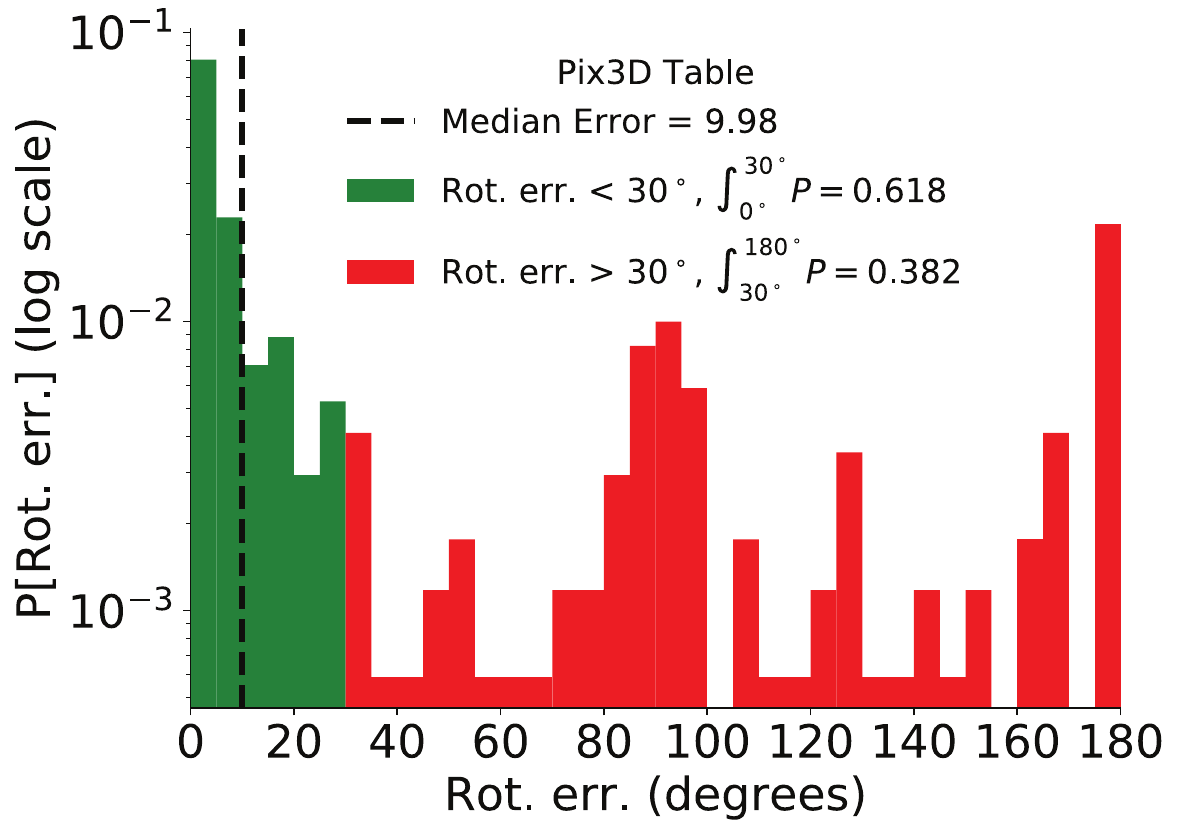}
        \end{minipage}\\[1mm]
     \small{4}
        \begin{minipage}{0.8\columnwidth}
            \centering
            \includegraphics[width=\textwidth]{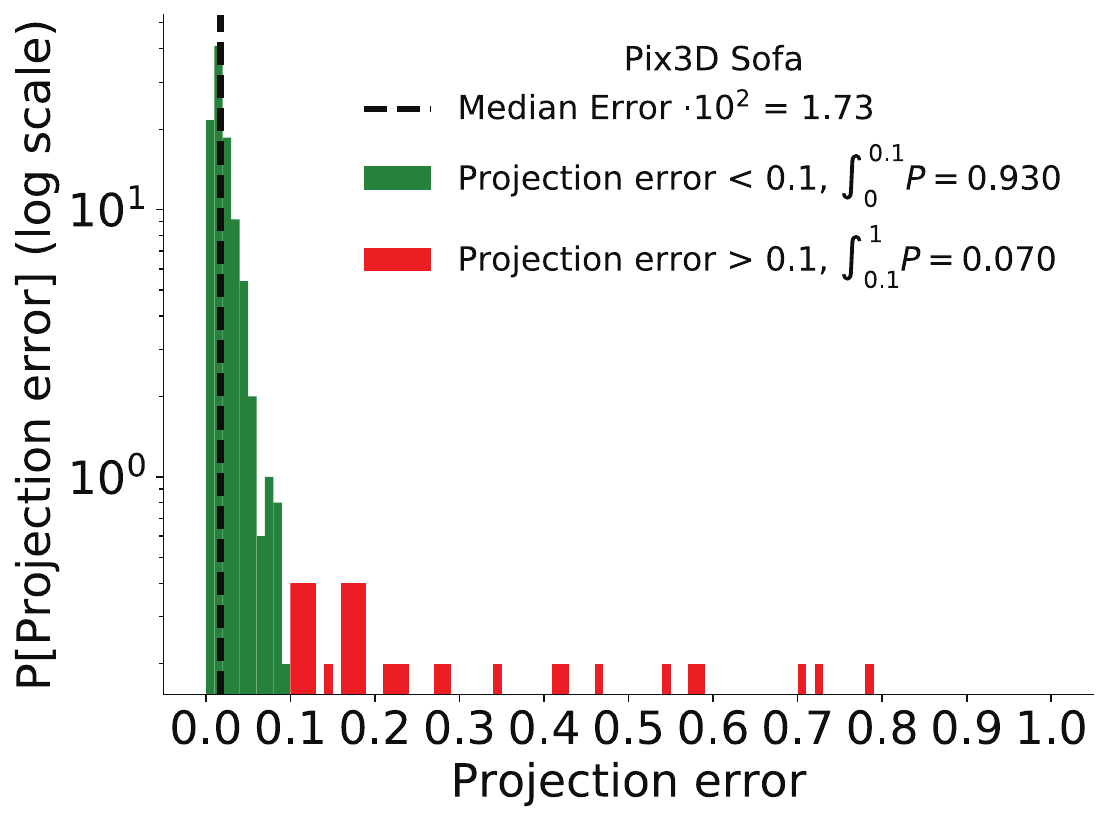}
        \end{minipage}
        \begin{minipage}{0.8\columnwidth}
            \centering
            \includegraphics[width=\textwidth]{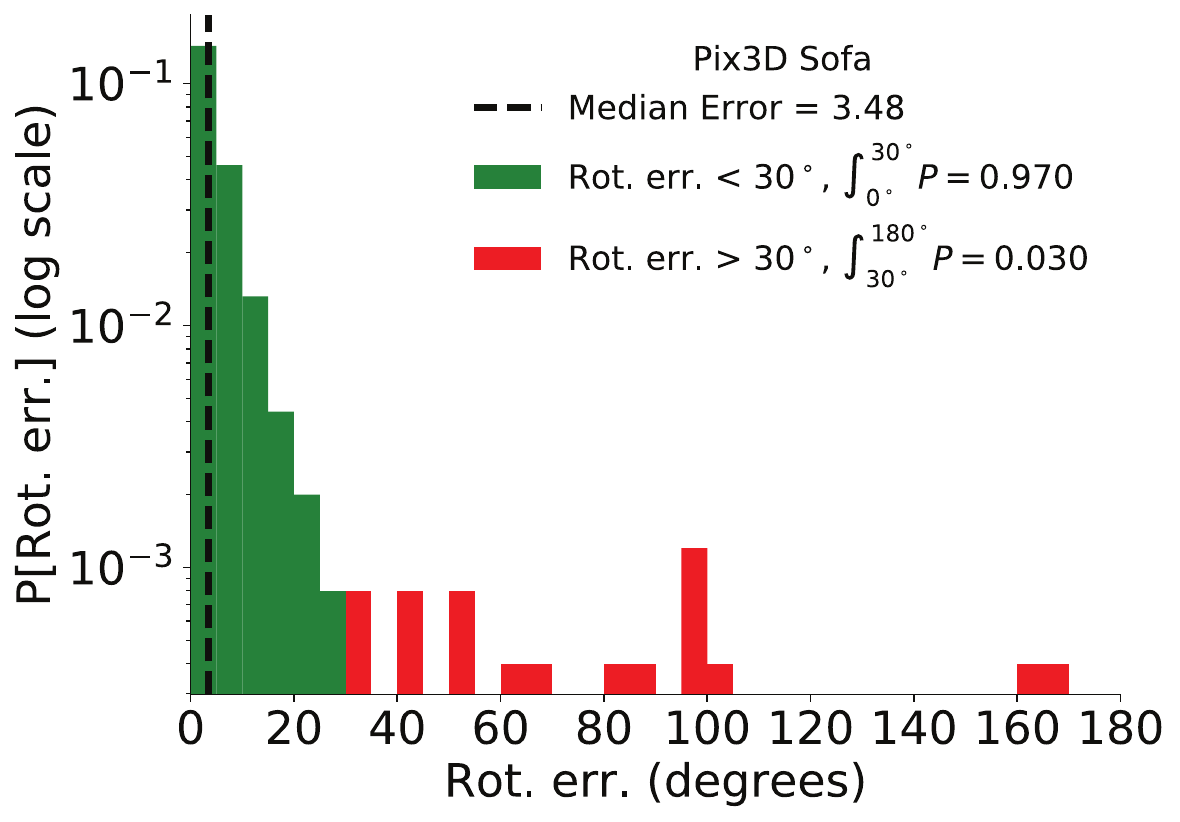}
        \end{minipage}\\[1mm]
            \vspace*{-3mm}
    \caption{\textbf{Projection error histograms (left) and rotation error histograms (right) for the Pix3D object classes.} Please note the logarithmic scale of the y-axis.} 
    \label{fig:err_hist_pix3d}
\end{figure*}

\begin{figure*}[t]
    \centering
    \small{1}
        \begin{minipage}{0.8\columnwidth}
            \textbf{Projection error histograms}
            \centering
            \includegraphics[width=\textwidth]{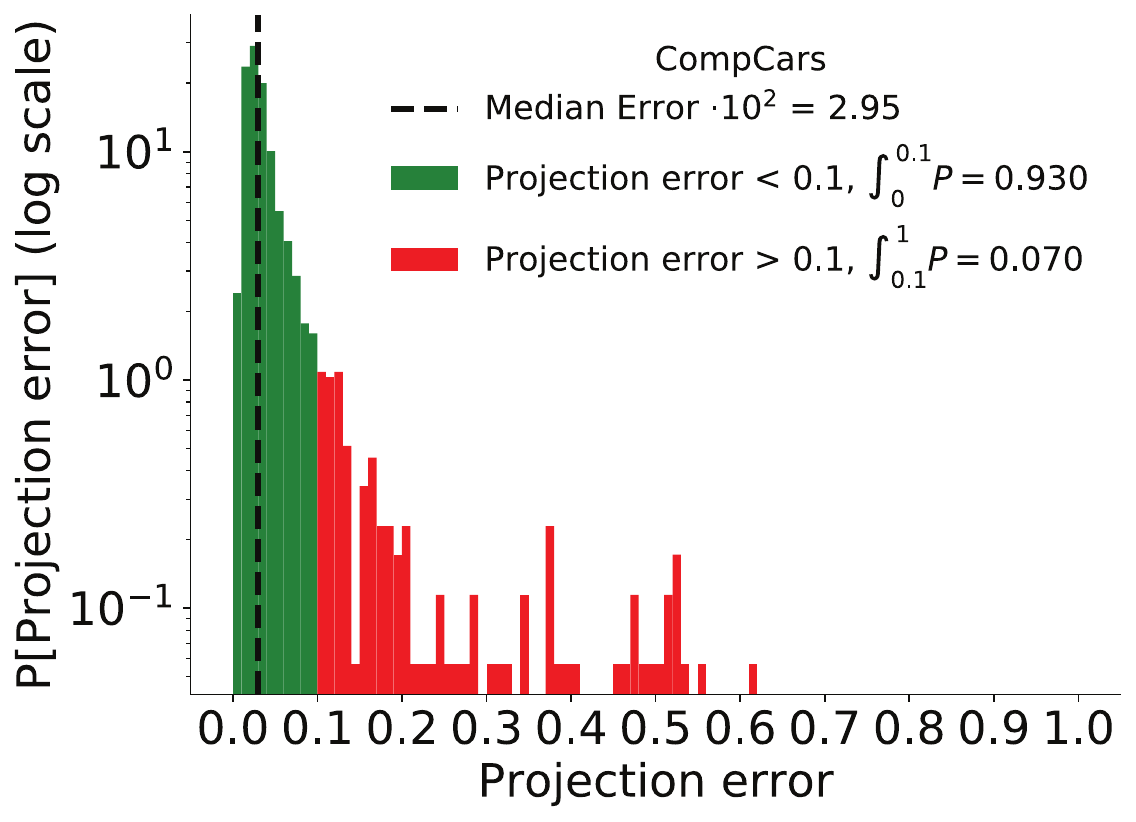}
        \end{minipage}
        \begin{minipage}{0.8\columnwidth}
            \textbf{Rotation error histograms}
            \centering
            \includegraphics[width=\textwidth]{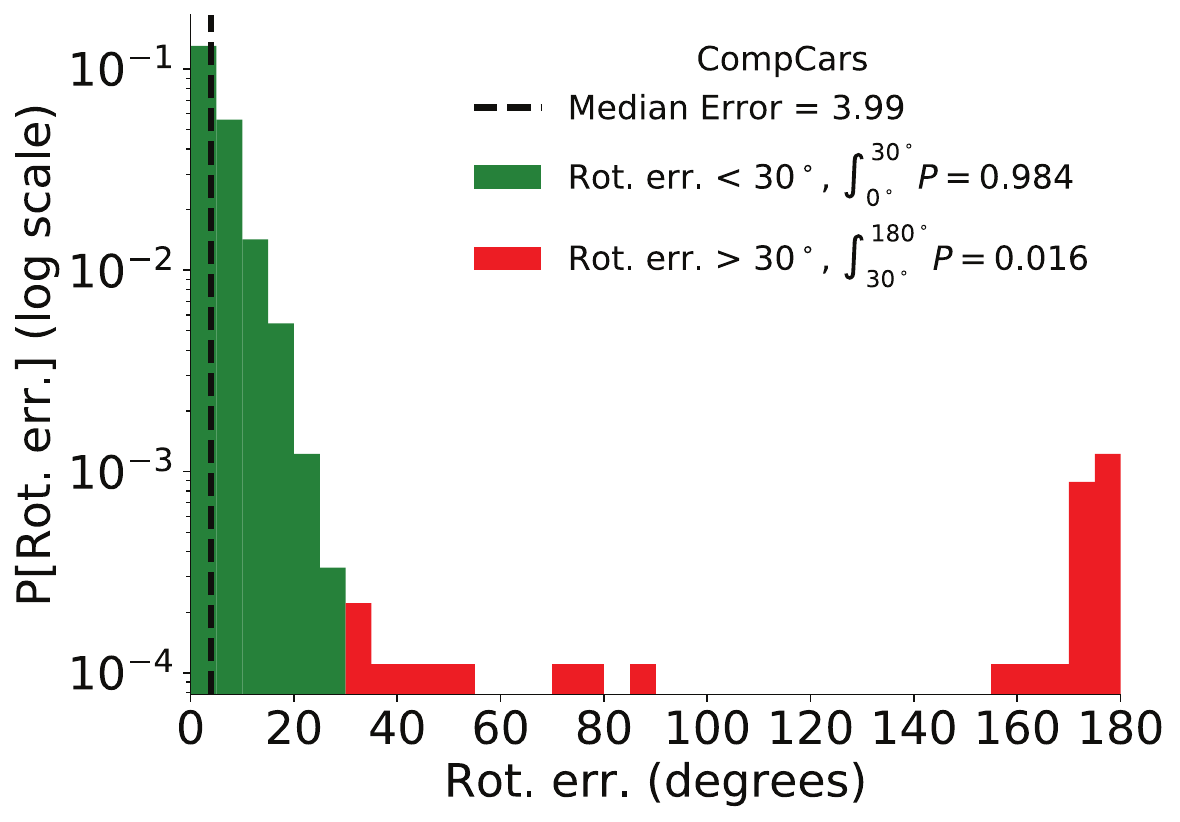}
        \end{minipage}\\[1mm]
    \small{2}
        \begin{minipage}{0.8\columnwidth}
            \centering
            \includegraphics[width=\textwidth]{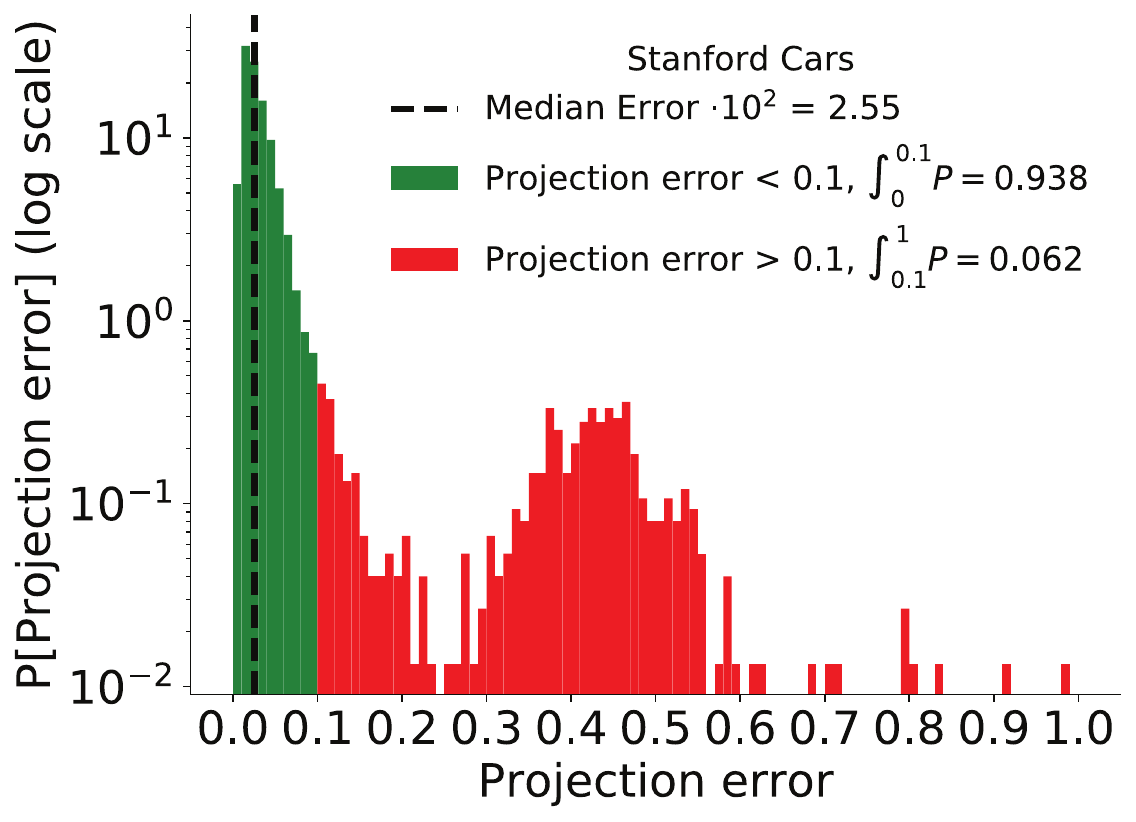}
        \end{minipage}
        \begin{minipage}{0.8\columnwidth}
            \centering
            \includegraphics[width=\textwidth]{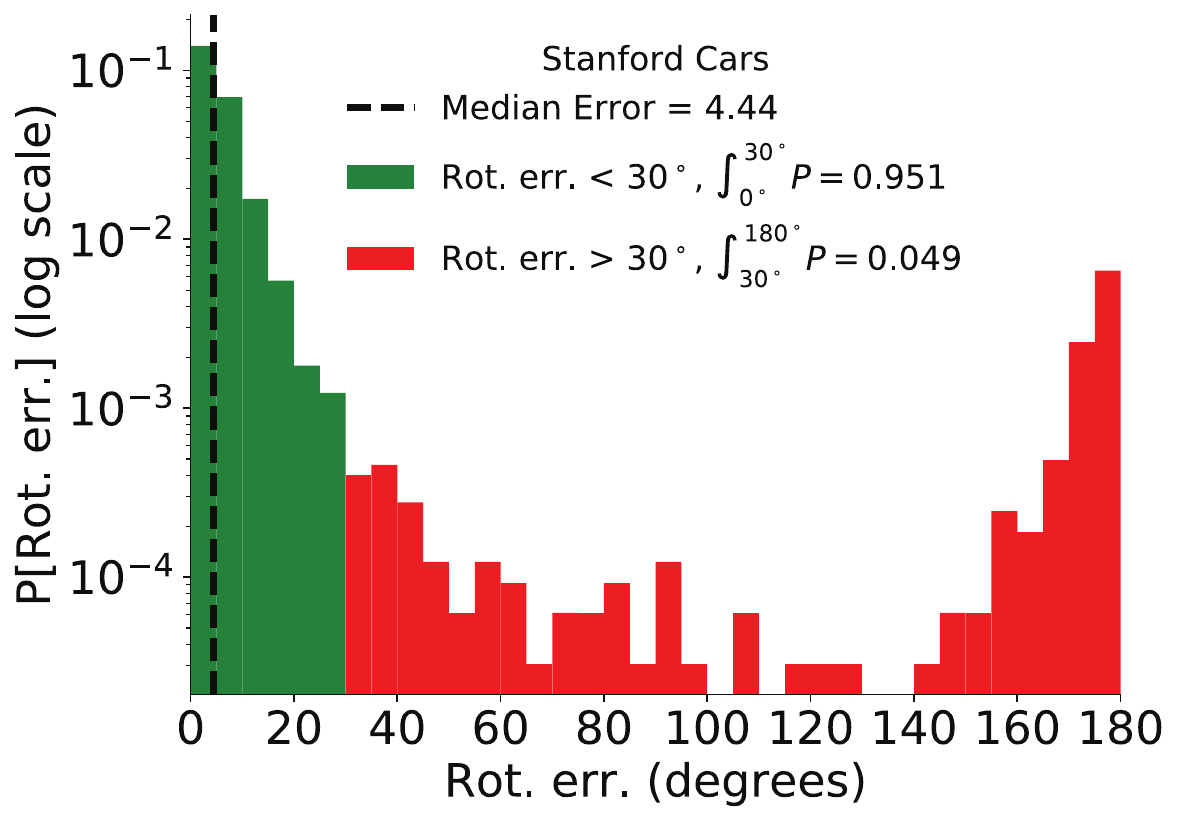}
        \end{minipage}\\[1mm]
    \vspace*{-3mm}
    \caption{\textbf{Projection error histograms (left) and rotation error histograms (right) for the CompCars (first row) and Stanford Cars (second row) datasets.} Please note the logarithmic scale of the y-axis.} 
    \label{fig:err_hist_cars}
\end{figure*}

\begin{figure*}[t]
    \centering
    \small{1}
        \begin{minipage}{0.8\columnwidth}
            \textbf{Projection accuracy at different thresholds}
            \centering
            \includegraphics[width=\textwidth]{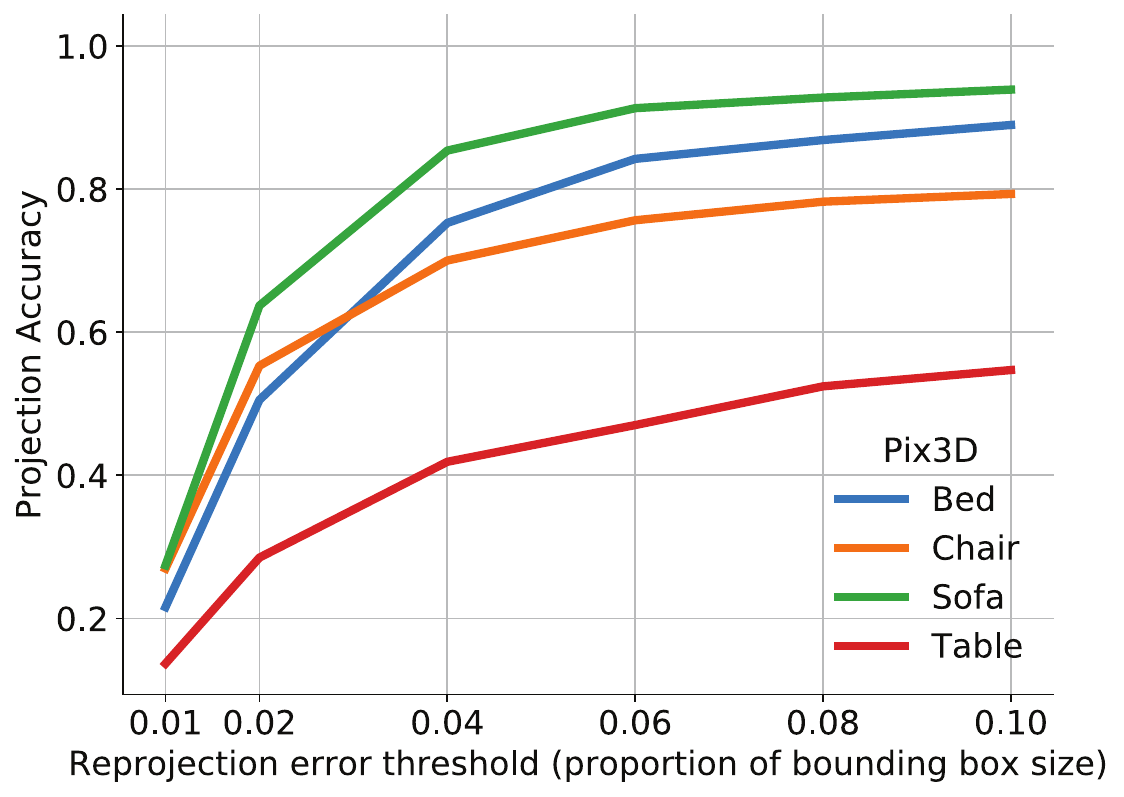}
        \end{minipage}
        \begin{minipage}{0.8\columnwidth}
            \textbf{Rotation accuracy at different thresholds}
            \centering
            \includegraphics[width=\textwidth]{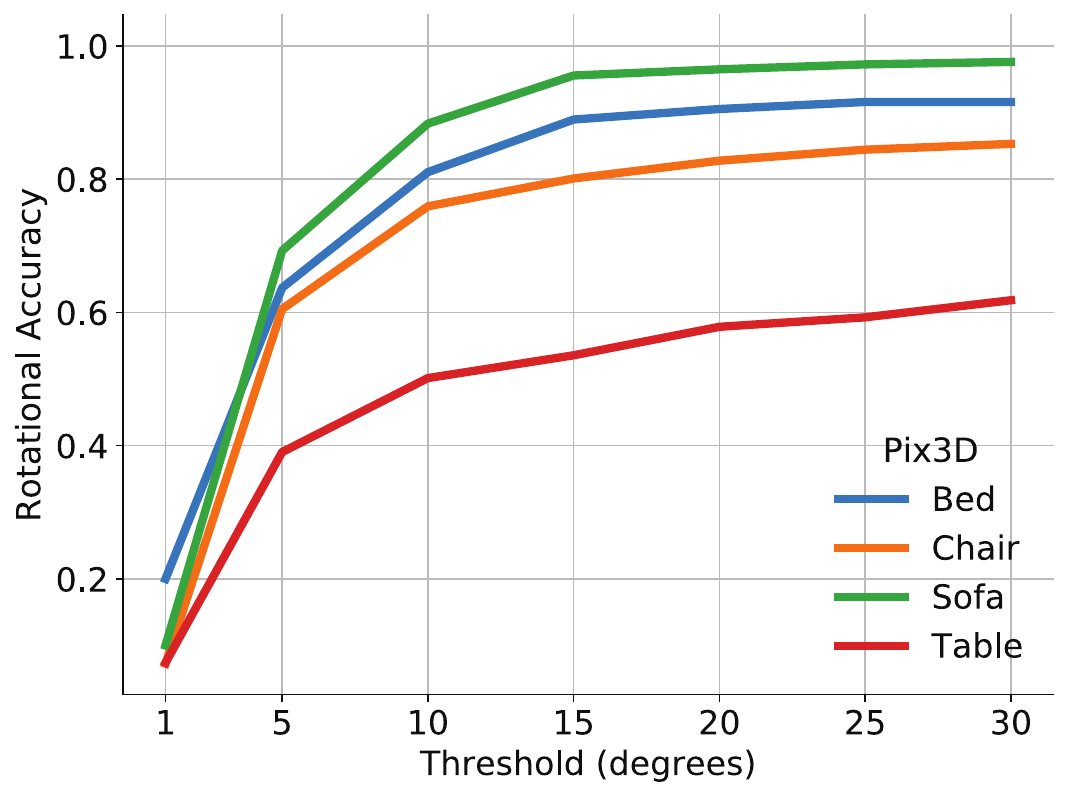}
        \end{minipage}\\[1mm]
    \small{2}
        \begin{minipage}{0.8\columnwidth}
            \centering
            \includegraphics[width=\textwidth]{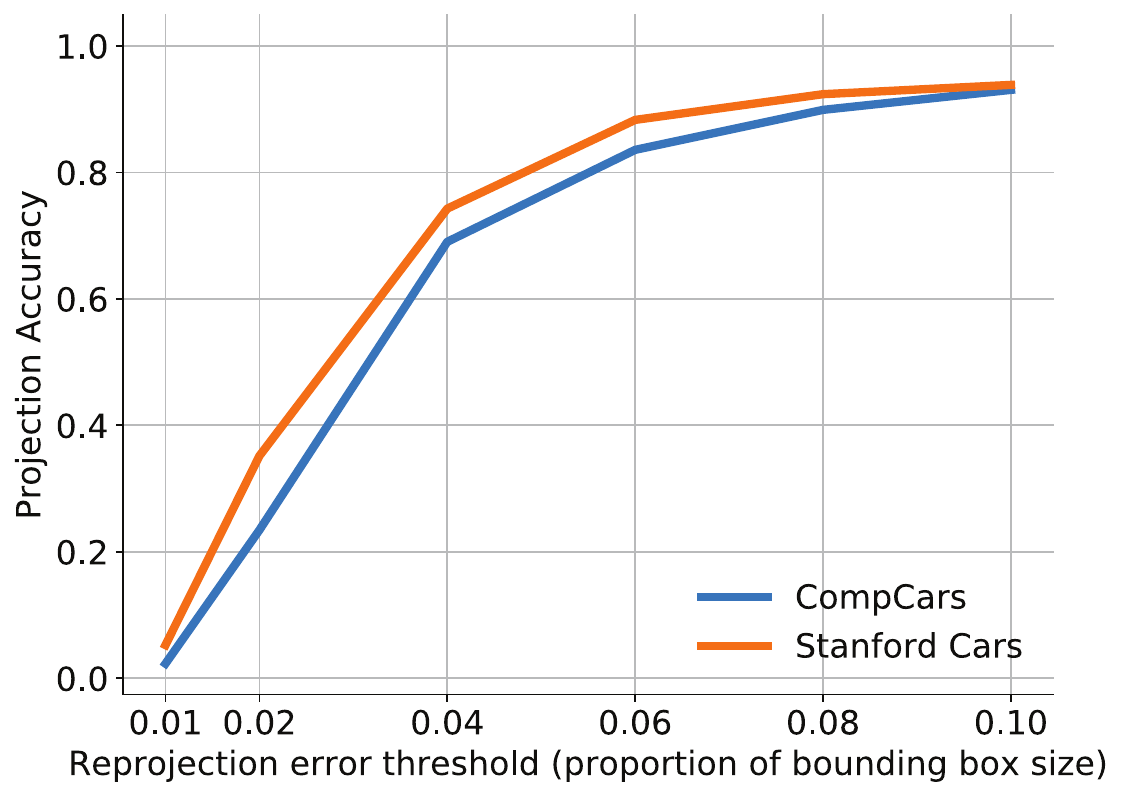}
        \end{minipage}
        \begin{minipage}{0.8\columnwidth}
            \centering
            \includegraphics[width=\textwidth]{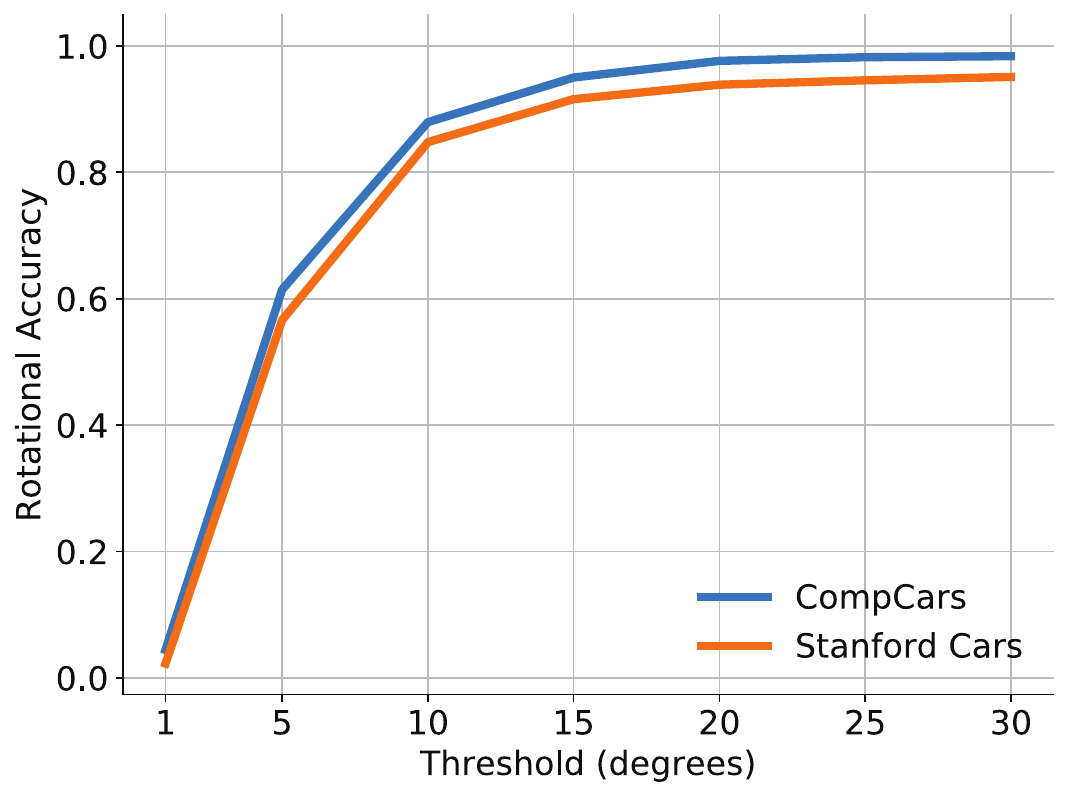}
        \end{minipage}\\[1mm]
        \vspace*{-3mm}
    \caption{\textbf{Projection and rotation accuracies at different error thresholds.}} 
    \label{fig:threshold_errs}
\end{figure*}

\subsection{Additional qualitative results}
\label{sec:additional-qualitative-results}
\begin{figure}[ht]
    \centering
    \small{1}
        \begin{minipage}{0.31\columnwidth}
            {\small Input image\vspace{1mm}}
            \centering
            \includegraphics[width=\textwidth]{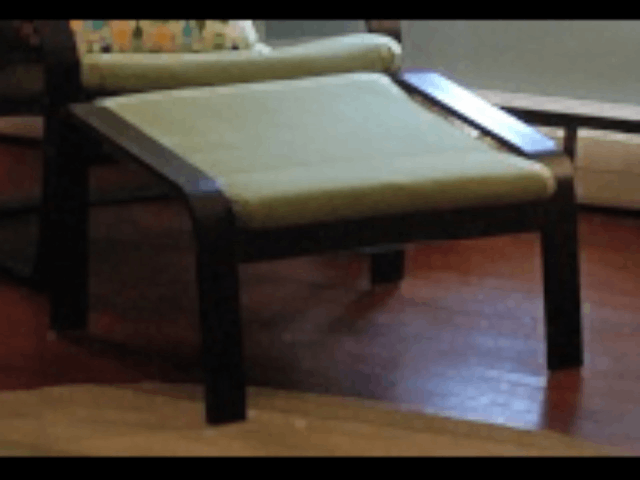}
        \end{minipage}
        \begin{minipage}{0.31\columnwidth}
            {\small Ground truth\vspace{1mm}}
            \centering
            \includegraphics[width=\textwidth]{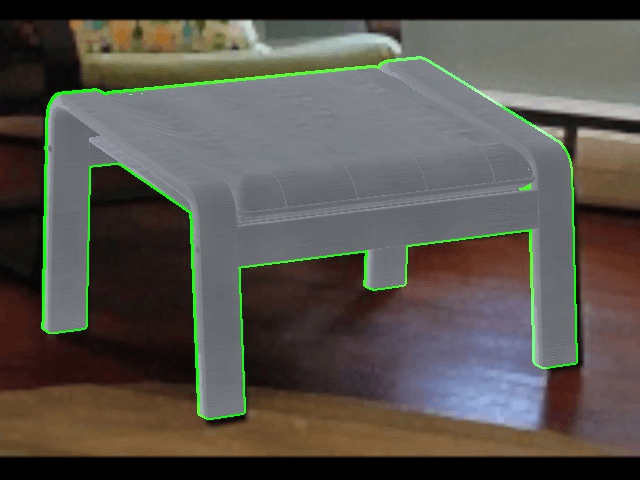}
        \end{minipage}
        \begin{minipage}{0.31\columnwidth}
            {\small Our prediction\vspace{1mm}}
            \centering
            \includegraphics[width=\textwidth]{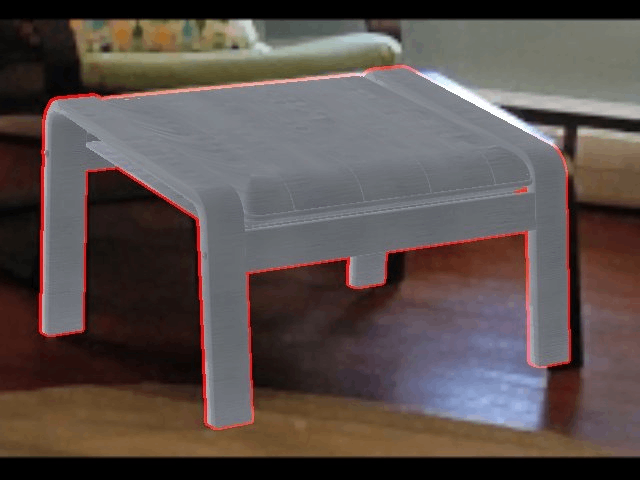}
        \end{minipage}\\[0.5mm]
    \small{2}
        \begin{minipage}{0.31\columnwidth}
            \centering
            \includegraphics[width=\textwidth]{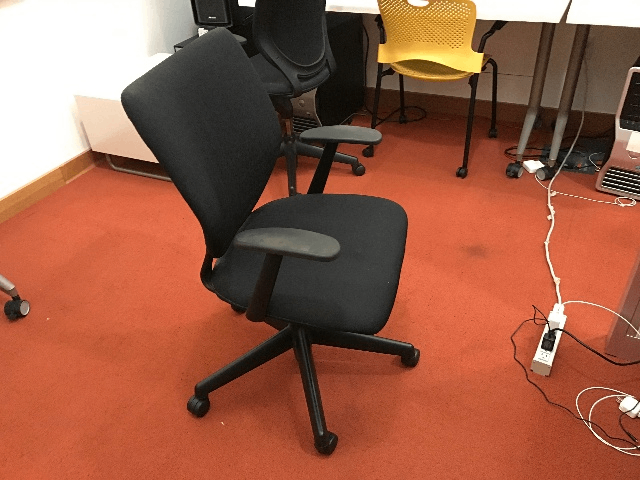}
        \end{minipage}
        \begin{minipage}{0.31\columnwidth}
            \centering
            \includegraphics[width=\textwidth]{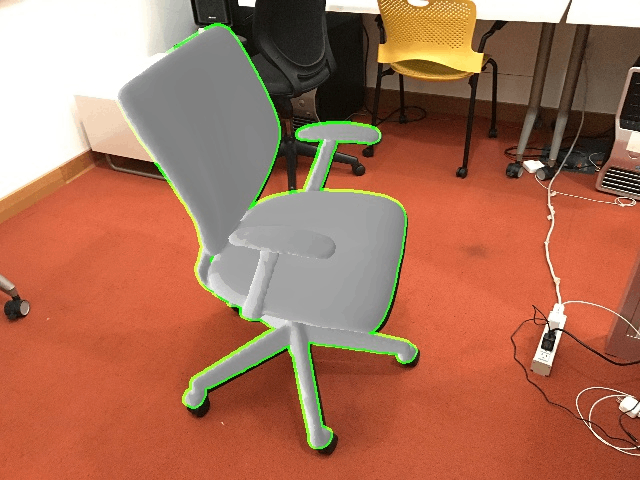}
        \end{minipage}
        \begin{minipage}{0.31\columnwidth}
            \centering
            \includegraphics[width=\textwidth]{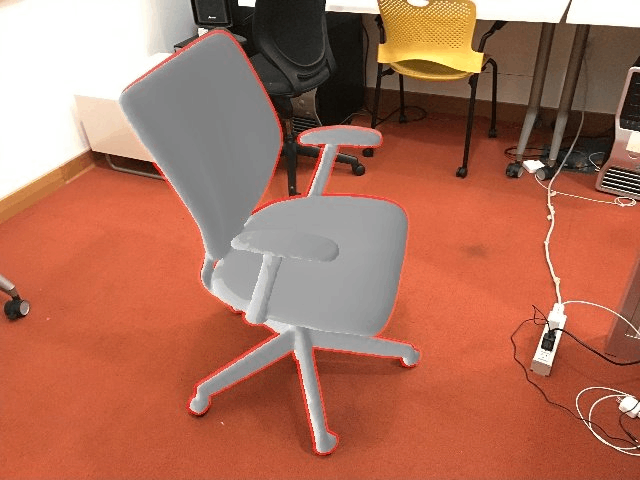}
        \end{minipage}\\[0.5mm]
    \small{3}
        \begin{minipage}{0.31\columnwidth}
            \centering
            \includegraphics[width=\textwidth]{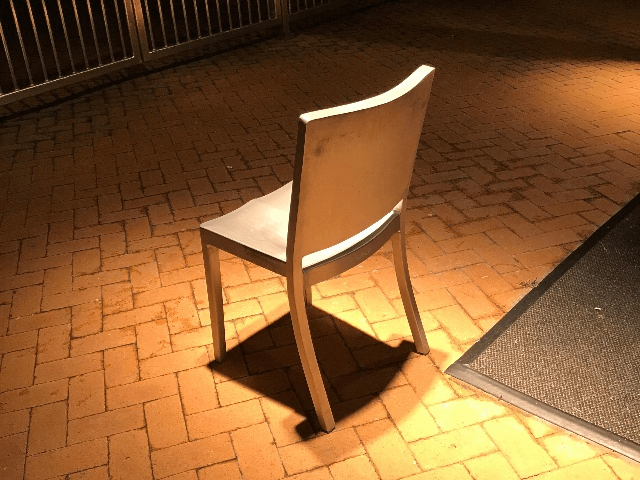}
        \end{minipage}
        \begin{minipage}{0.31\columnwidth}
            \centering
            \includegraphics[width=\textwidth]{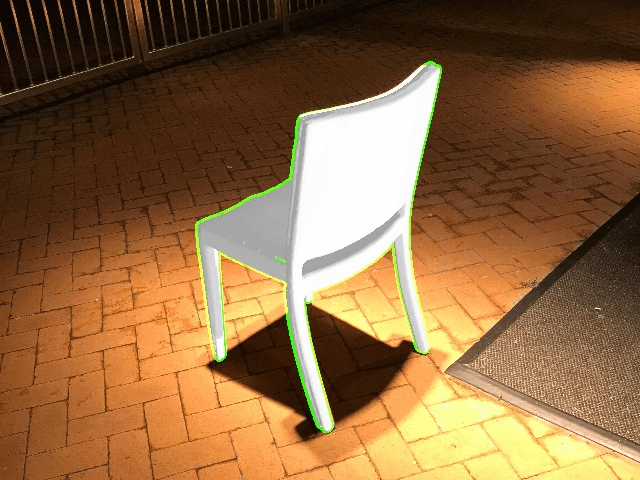}
        \end{minipage}
        \begin{minipage}{0.31\columnwidth}
            \centering
            \includegraphics[width=\textwidth]{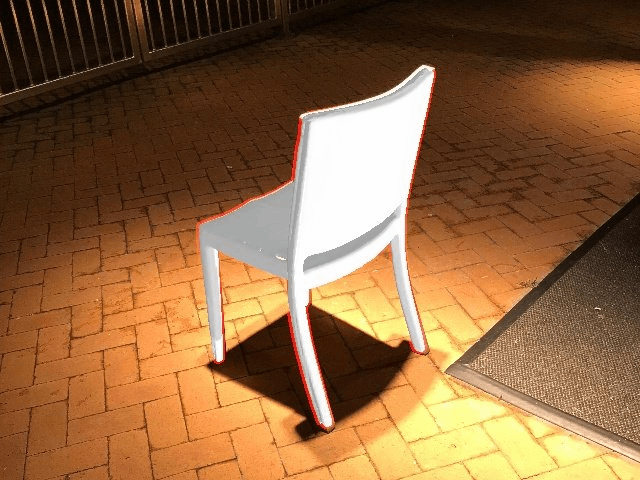}
        \end{minipage}\\[0.5mm]
    \small{4}
        \begin{minipage}{0.31\columnwidth}
            \centering
            \includegraphics[width=\textwidth]{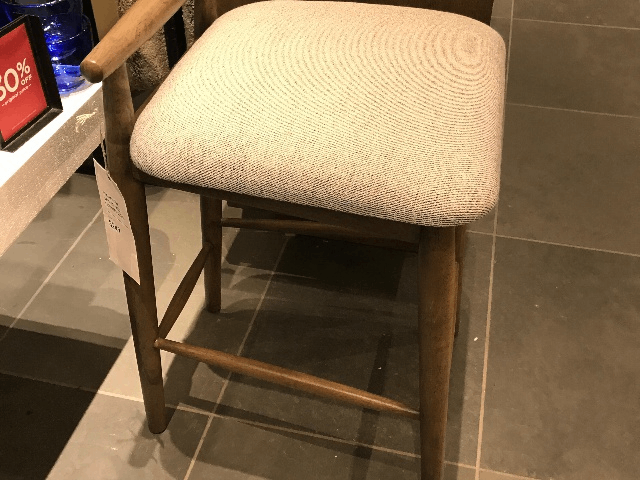}
        \end{minipage}
        \begin{minipage}{0.31\columnwidth}
            \centering
            \includegraphics[width=\textwidth]{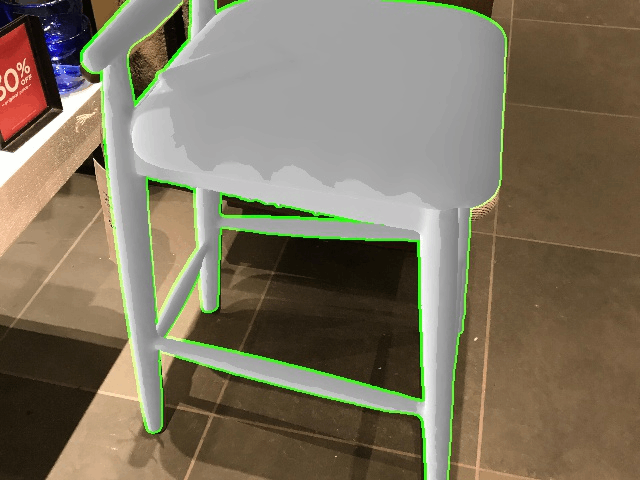}
        \end{minipage}
        \begin{minipage}{0.31\columnwidth}
            \centering
            \includegraphics[width=\textwidth]{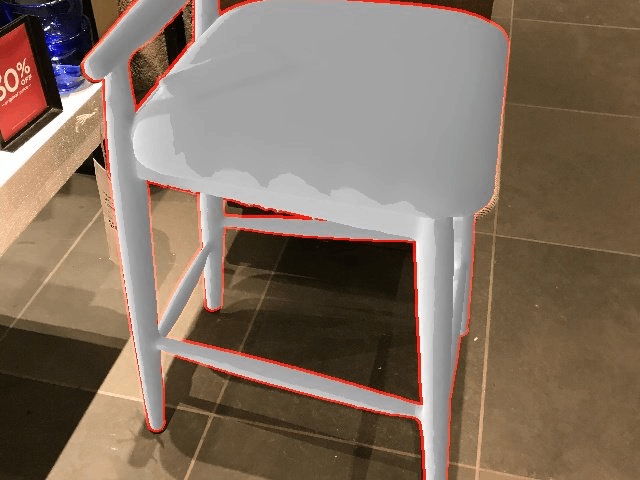}
        \end{minipage}\\[0.5mm]
    \small{5}
        \begin{minipage}{0.31\columnwidth}
            \centering
            \includegraphics[width=\textwidth]{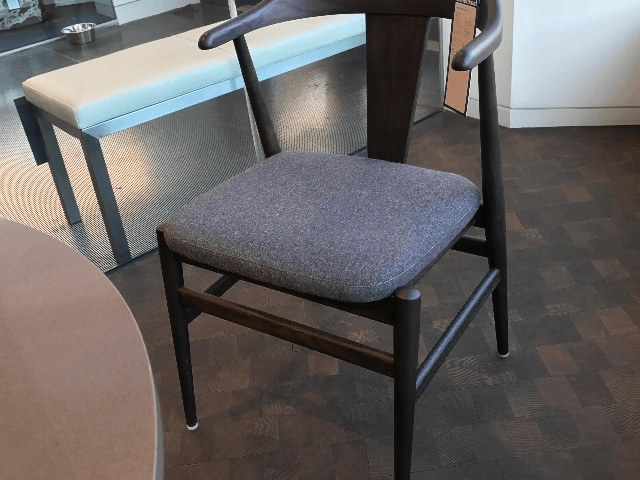}
        \end{minipage}
        \begin{minipage}{0.31\columnwidth}
            \centering
            \includegraphics[width=\textwidth]{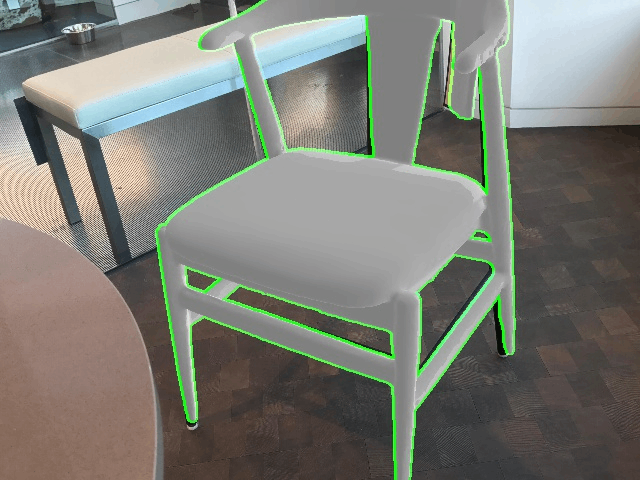}
        \end{minipage}
        \begin{minipage}{0.31\columnwidth}
            \centering
            \includegraphics[width=\textwidth]{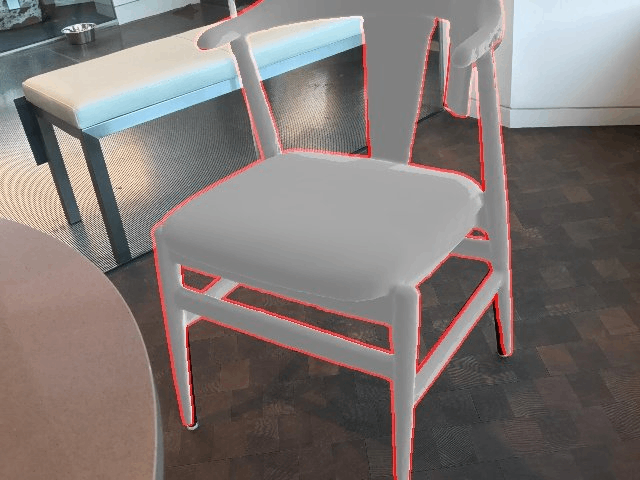}
        \end{minipage}\\[0.5mm]
    \small{6}
        \begin{minipage}{0.31\columnwidth}
            \centering
            \includegraphics[width=\textwidth]{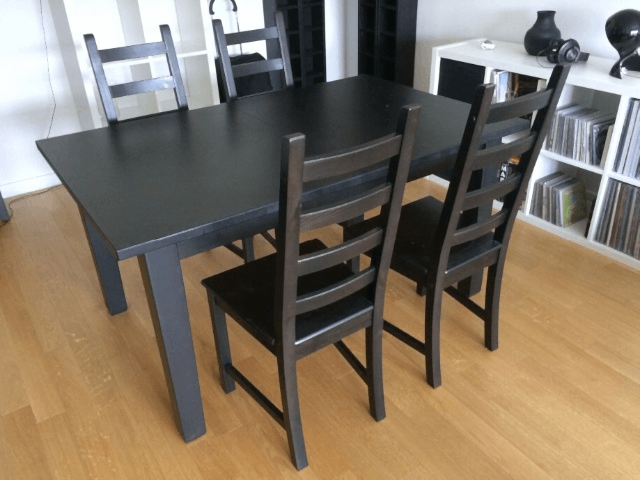}
        \end{minipage}
        \begin{minipage}{0.31\columnwidth}
            \centering
            \includegraphics[width=\textwidth]{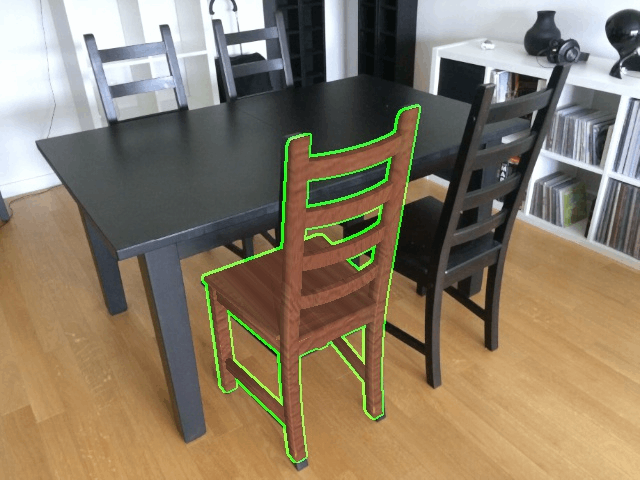}
        \end{minipage}
        \begin{minipage}{0.31\columnwidth}
            \centering
            \includegraphics[width=\textwidth]{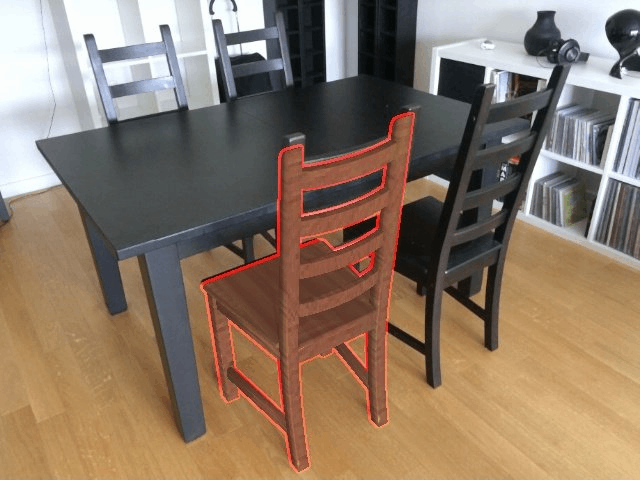}
        \end{minipage}\\[0.5mm]
    \small{7}
        \begin{minipage}{0.31\columnwidth}
            \centering
            \includegraphics[width=\textwidth]{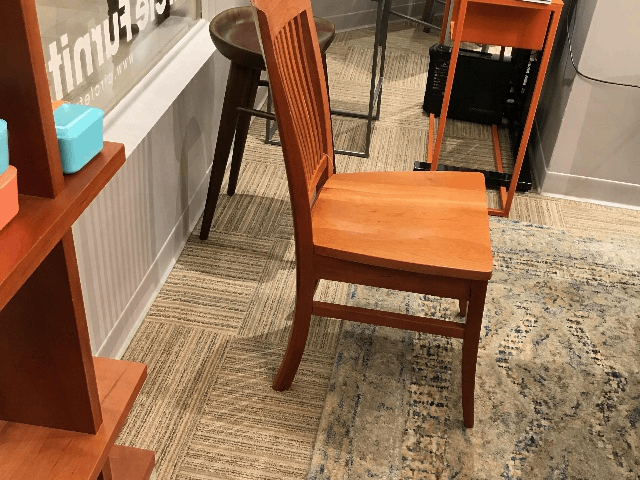}
        \end{minipage}
        \begin{minipage}{0.31\columnwidth}
            \centering
            \includegraphics[width=\textwidth]{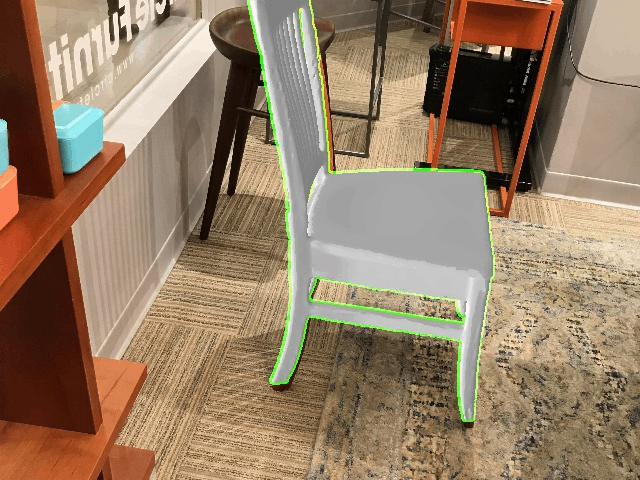}
        \end{minipage}
        \begin{minipage}{0.31\columnwidth}
            \centering
            \includegraphics[width=\textwidth]{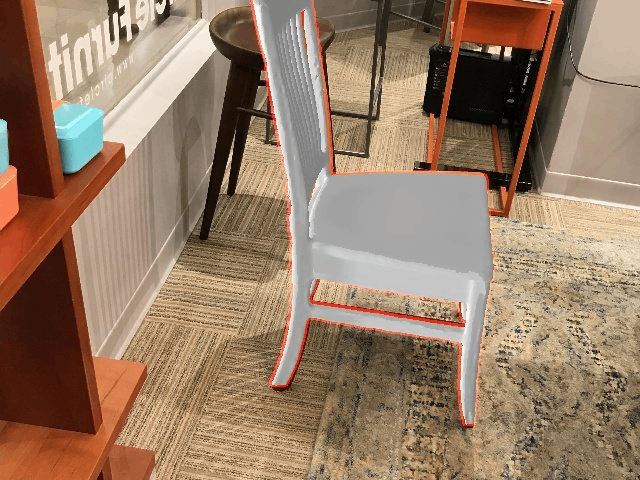}
        \end{minipage}\\[0.5mm]
    \small{8}
        \begin{minipage}{0.31\columnwidth}
            \centering
            \includegraphics[width=\textwidth]{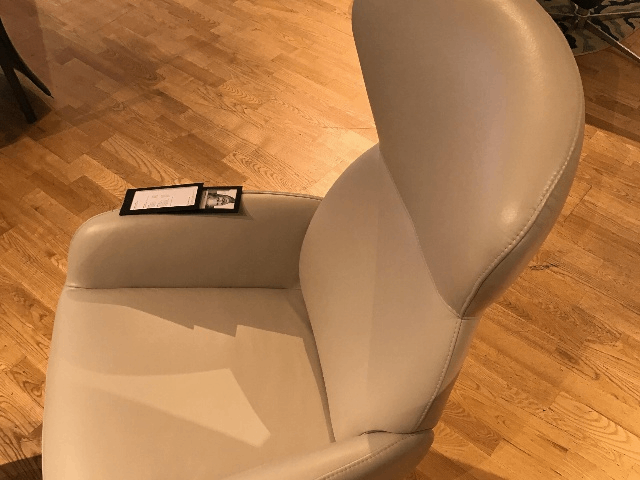}
        \end{minipage}
        \begin{minipage}{0.31\columnwidth}
            \centering
            \includegraphics[width=\textwidth]{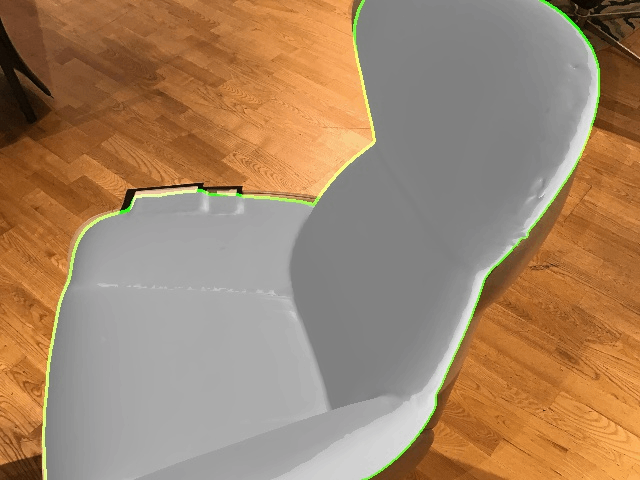}
        \end{minipage}
        \begin{minipage}{0.31\columnwidth}
            \centering
            \includegraphics[width=\textwidth]{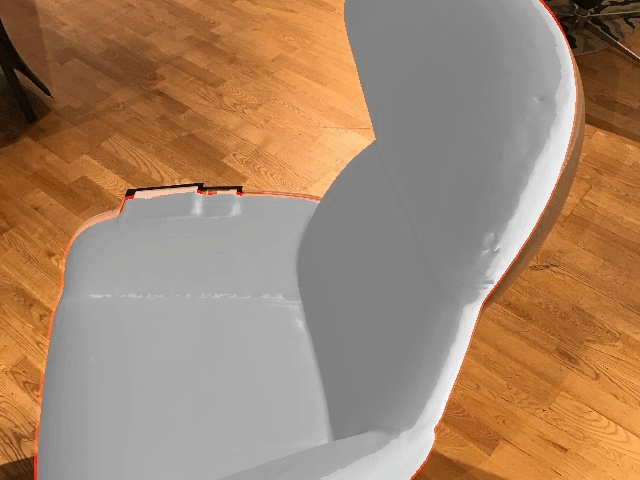}
        \end{minipage}\\[0.5mm]

    \caption{\textbf{Qualitative results for Pix3D chairs.}}
    \label{pix3d-chair-q}
    \vspace*{-5mm}
\end{figure}

\begin{figure}[ht]
    \centering
    \small{1}
        \begin{minipage}{0.31\columnwidth}
            {\small Input image\vspace{1mm}}
            \centering
            \includegraphics[width=\textwidth]{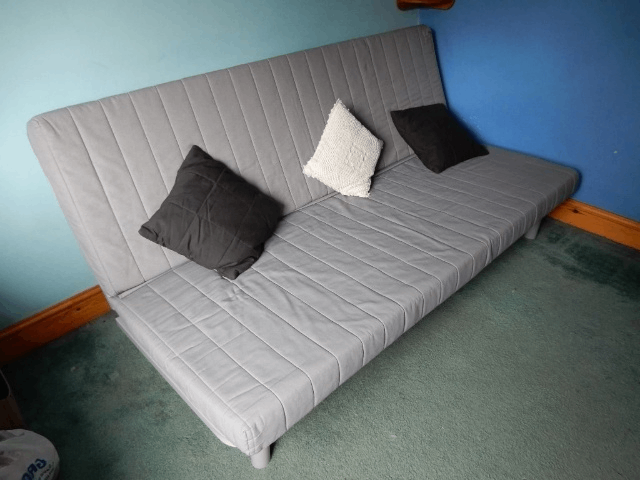}
        \end{minipage}
        \begin{minipage}{0.31\columnwidth}
            {\small Ground truth\vspace{1mm}}
            \centering
            \includegraphics[width=\textwidth]{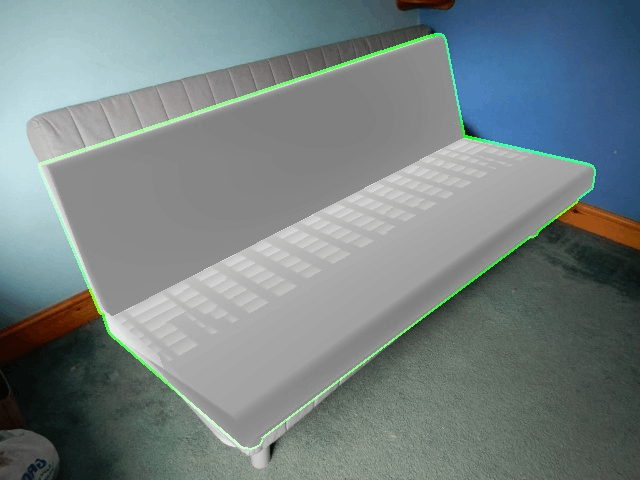}
        \end{minipage}
        \begin{minipage}{0.31\columnwidth}
            {\small Our prediction\vspace{1mm}}
            \centering
            \includegraphics[width=\textwidth]{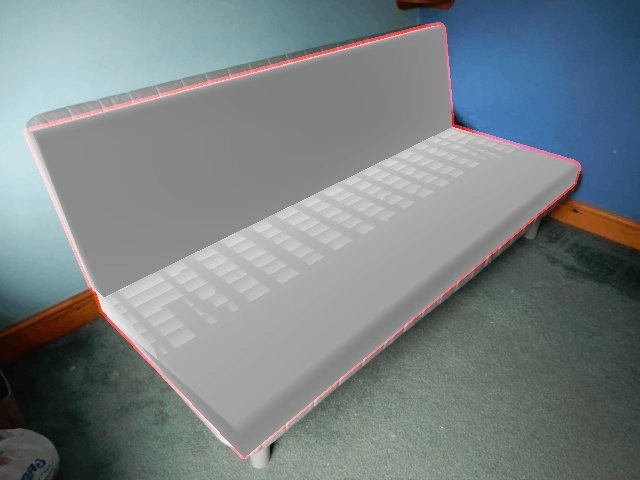}
        \end{minipage}\\[0.5mm]
    \small{2}
        \begin{minipage}{0.31\columnwidth}
            \centering
            \includegraphics[width=\textwidth]{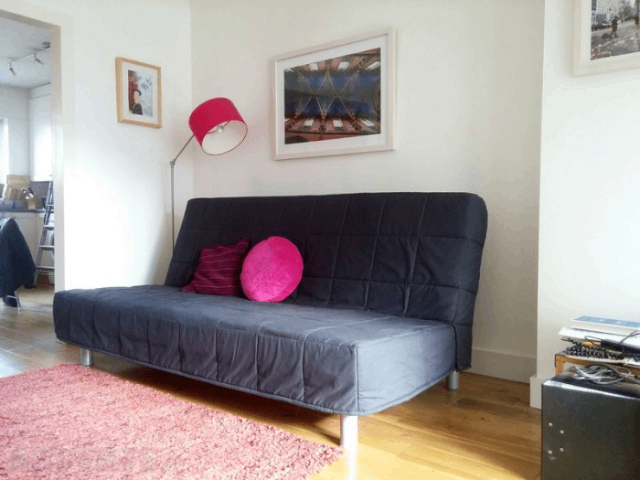}
        \end{minipage}
        \begin{minipage}{0.31\columnwidth}
            \centering
            \includegraphics[width=\textwidth]{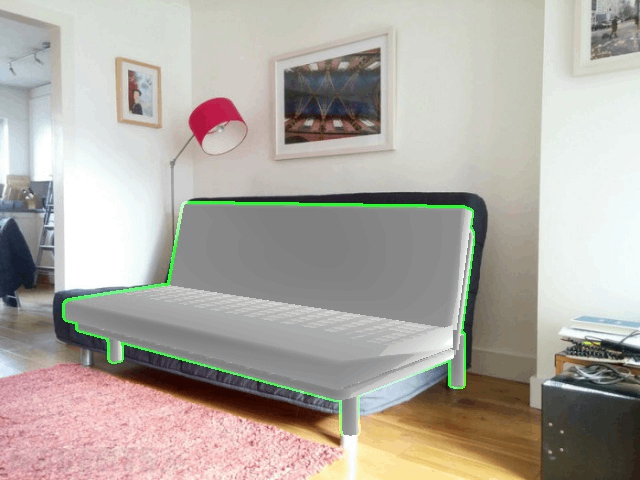}
        \end{minipage}
        \begin{minipage}{0.31\columnwidth}
            \centering
            \includegraphics[width=\textwidth]{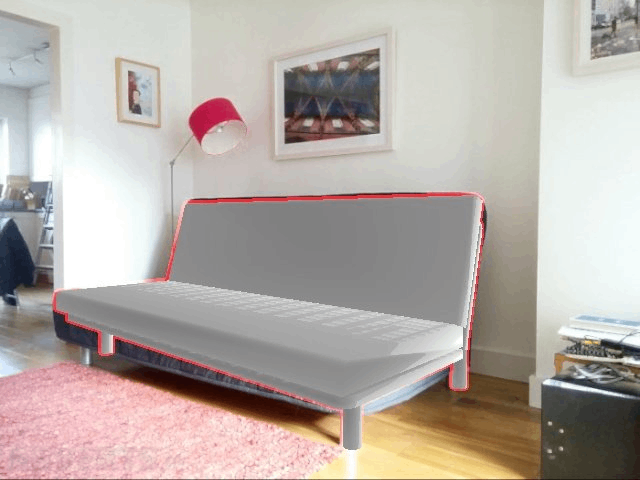}
        \end{minipage}\\[0.5mm]
    \small{3}
        \begin{minipage}{0.31\columnwidth}
            \centering
            \includegraphics[width=\textwidth]{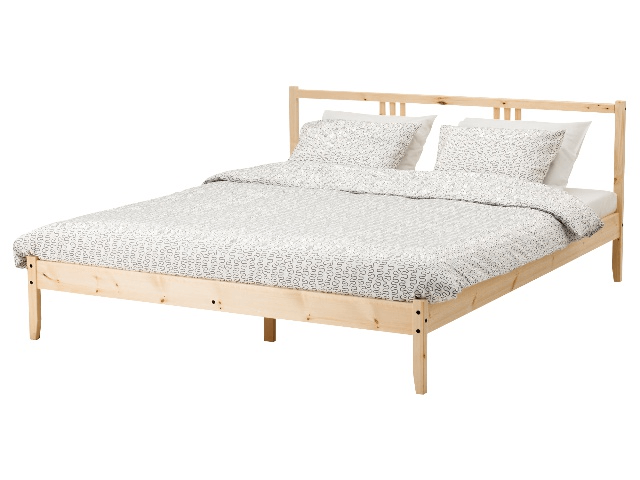}
        \end{minipage}
        \begin{minipage}{0.31\columnwidth}
            \centering
            \includegraphics[width=\textwidth]{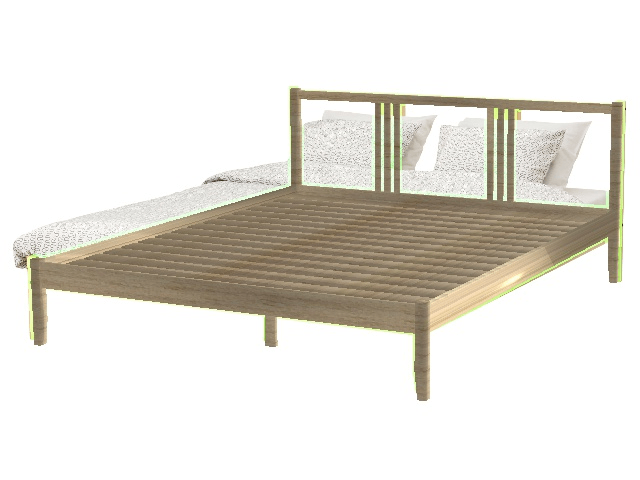}
        \end{minipage}
        \begin{minipage}{0.31\columnwidth}
            \centering
            \includegraphics[width=\textwidth]{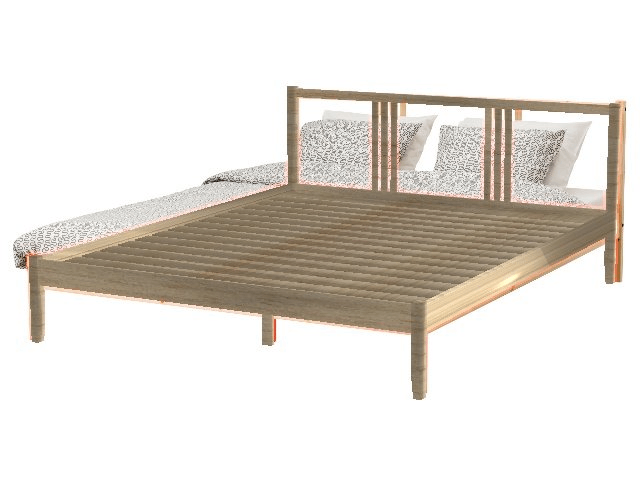}
        \end{minipage}\\[0.5mm]
    \small{4}
        \begin{minipage}{0.31\columnwidth}
            \centering
            \includegraphics[width=\textwidth]{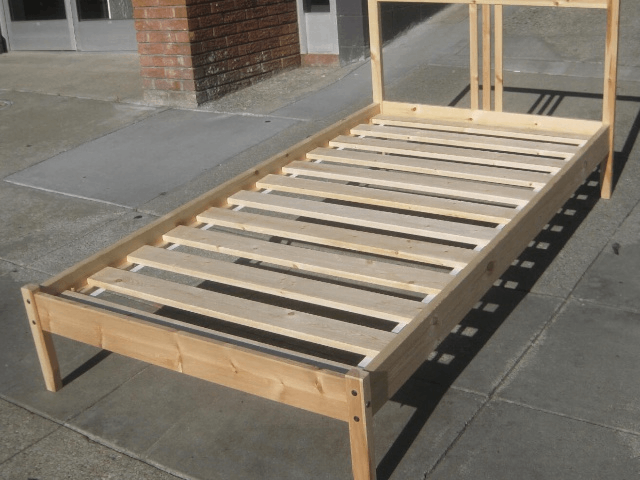}
        \end{minipage}
        \begin{minipage}{0.31\columnwidth}
            \centering
            \includegraphics[width=\textwidth]{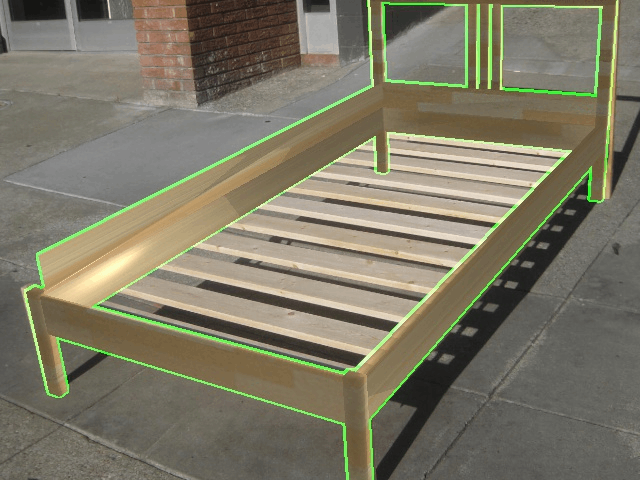}
        \end{minipage}
        \begin{minipage}{0.31\columnwidth}
            \centering
            \includegraphics[width=\textwidth]{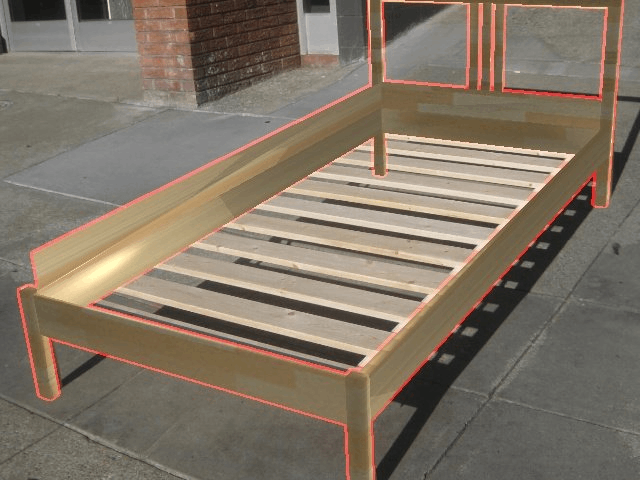}
        \end{minipage}\\[0.5mm]
    \small{5}
        \begin{minipage}{0.31\columnwidth}
            \centering
            \includegraphics[width=\textwidth]{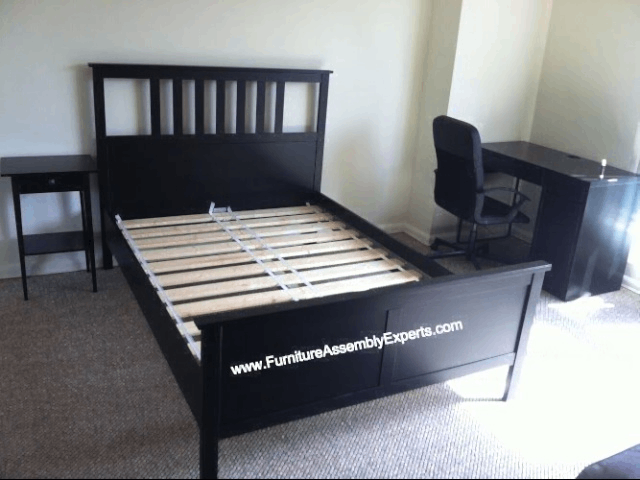}
        \end{minipage}
        \begin{minipage}{0.31\columnwidth}
            \centering
            \includegraphics[width=\textwidth]{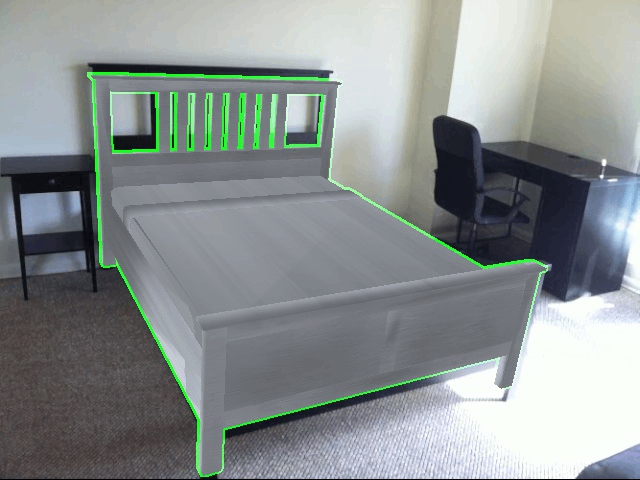}
        \end{minipage}
        \begin{minipage}{0.31\columnwidth}
            \centering
            \includegraphics[width=\textwidth]{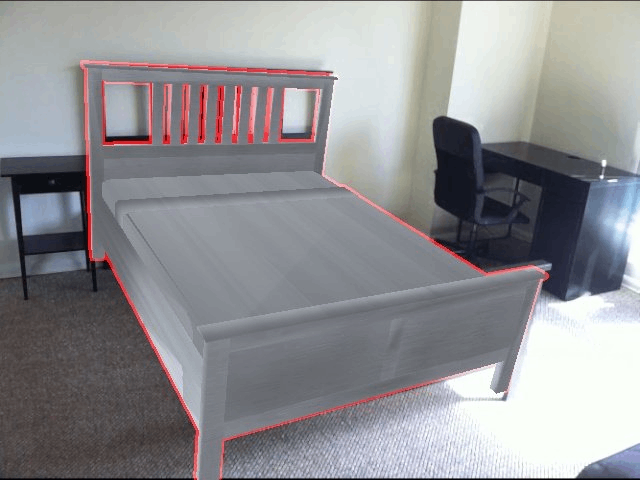}
        \end{minipage}\\[0.5mm]
    \small{6}
        \begin{minipage}{0.31\columnwidth}
            \centering
            \includegraphics[width=\textwidth]{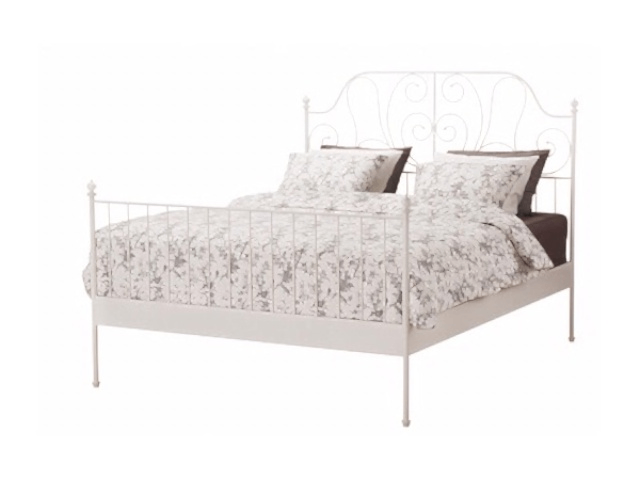}
        \end{minipage}
        \begin{minipage}{0.31\columnwidth}
            \centering
            \includegraphics[width=\textwidth]{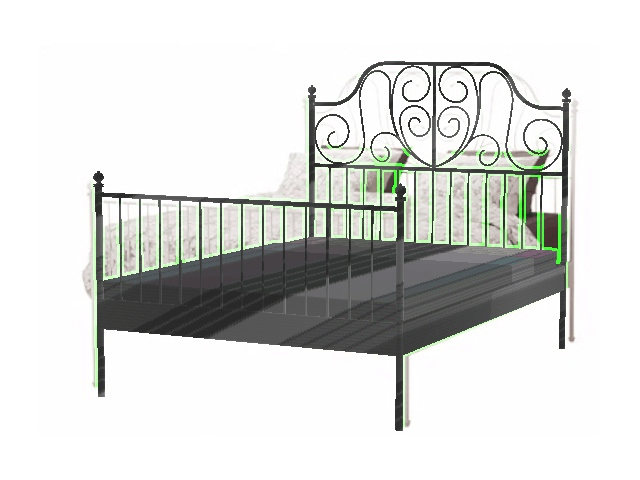}
        \end{minipage}
        \begin{minipage}{0.31\columnwidth}
            \centering
            \includegraphics[width=\textwidth]{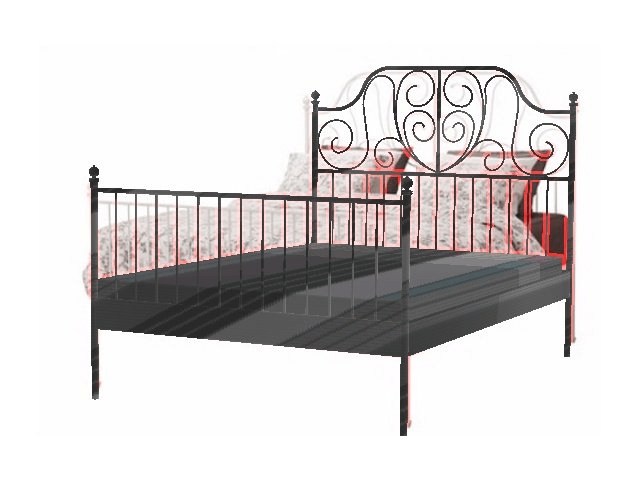}
        \end{minipage}\\[0.5mm]
    \small{7}
        \begin{minipage}{0.31\columnwidth}
            \centering
            \includegraphics[width=\textwidth]{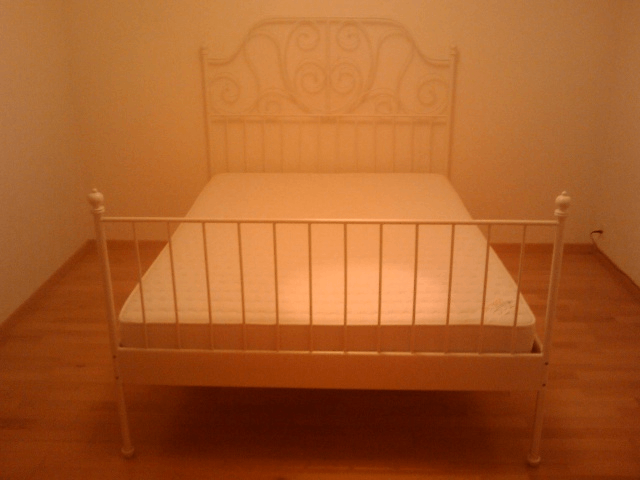}
        \end{minipage}
        \begin{minipage}{0.31\columnwidth}
            \centering
            \includegraphics[width=\textwidth]{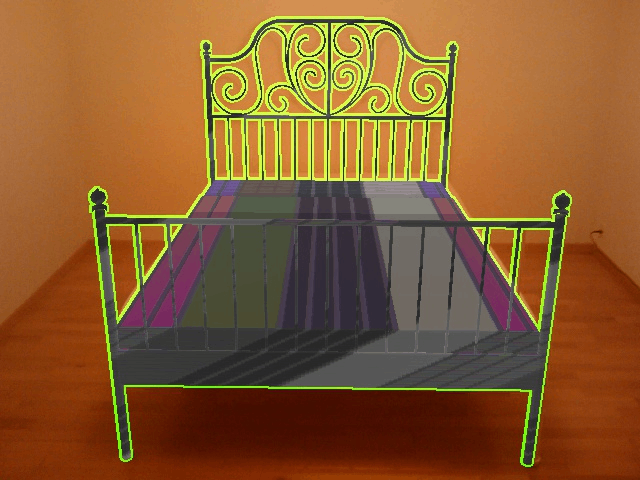}
        \end{minipage}
        \begin{minipage}{0.31\columnwidth}
            \centering
            \includegraphics[width=\textwidth]{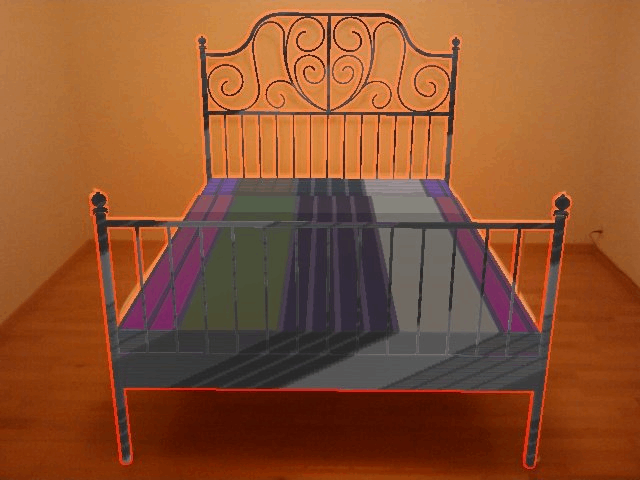}
        \end{minipage}\\[0.5mm]
    \small{8}
        \begin{minipage}{0.31\columnwidth}
            \centering
            \includegraphics[width=\textwidth]{figures/beds/entry_179_input.png}
        \end{minipage}
        \begin{minipage}{0.31\columnwidth}
            \centering
            \includegraphics[width=\textwidth]{figures/beds/entry_179_gt.png}
        \end{minipage}
        \begin{minipage}{0.31\columnwidth}
            \centering
            \includegraphics[width=\textwidth]{figures/beds/entry_179_pred.png}
        \end{minipage}\\[0.5mm]

    \caption{\textbf{Qualitative results for Pix3D beds.}}
    \label{pix3d-beds-q}
    \vspace*{-5mm}
\end{figure}

\begin{figure}[ht]
    \centering
    \small{1}
        \begin{minipage}{0.31\columnwidth}
            {\small Input image\vspace{1mm}}
            \centering
            \includegraphics[width=\textwidth]{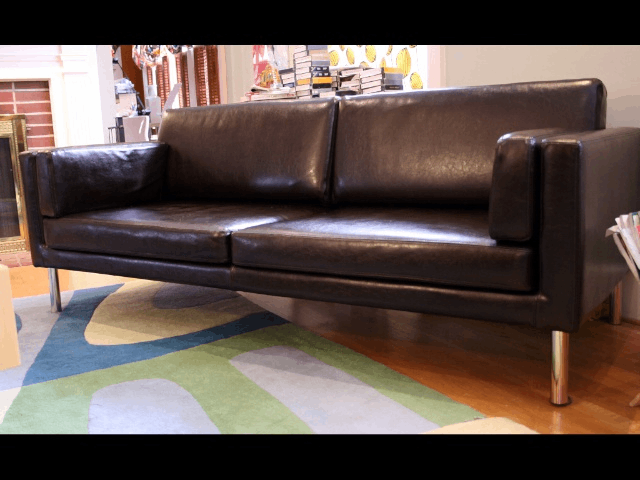}
        \end{minipage}
        \begin{minipage}{0.31\columnwidth}
            {\small Ground truth\vspace{1mm}}
            \centering
            \includegraphics[width=\textwidth]{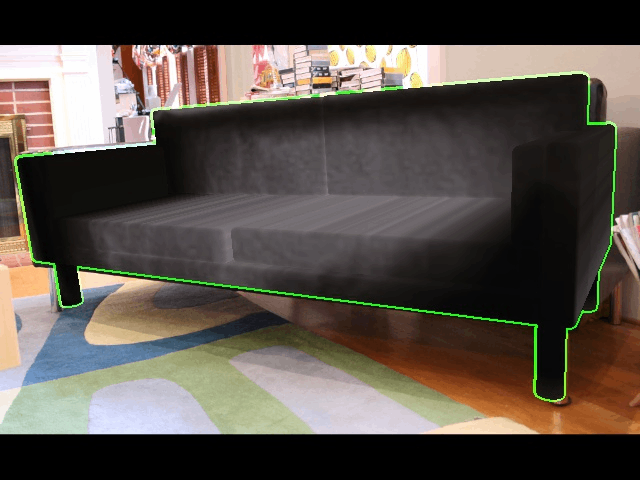}
        \end{minipage}
        \begin{minipage}{0.31\columnwidth}
            {\small Our prediction\vspace{1mm}}
            \centering
            \includegraphics[width=\textwidth]{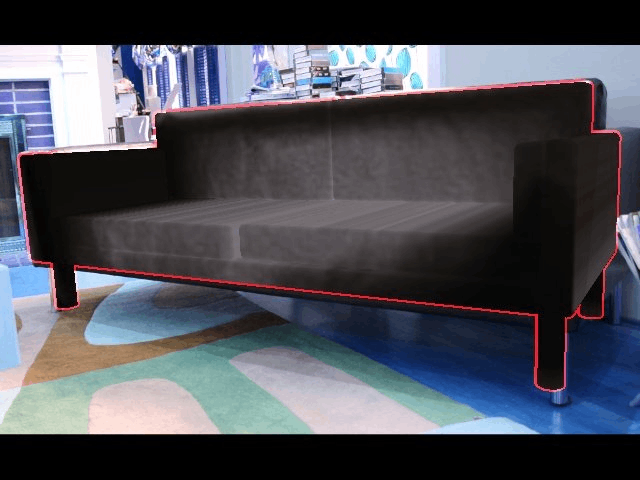}
        \end{minipage}\\[0.5mm]
    \small{2}
        \begin{minipage}{0.31\columnwidth}
            \centering
            \includegraphics[width=\textwidth]{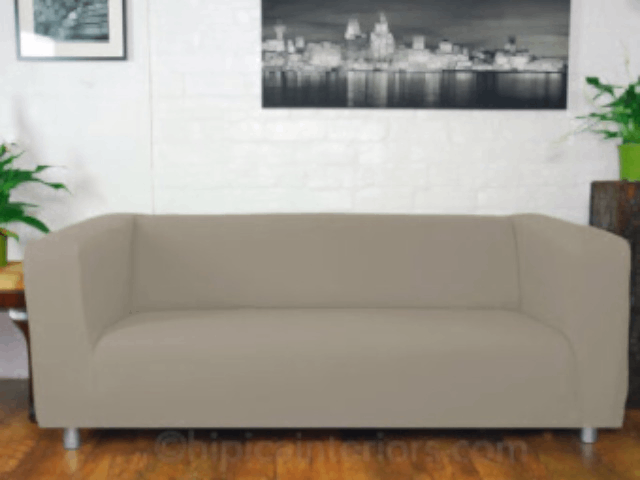}
        \end{minipage}
        \begin{minipage}{0.31\columnwidth}
            \centering
            \includegraphics[width=\textwidth]{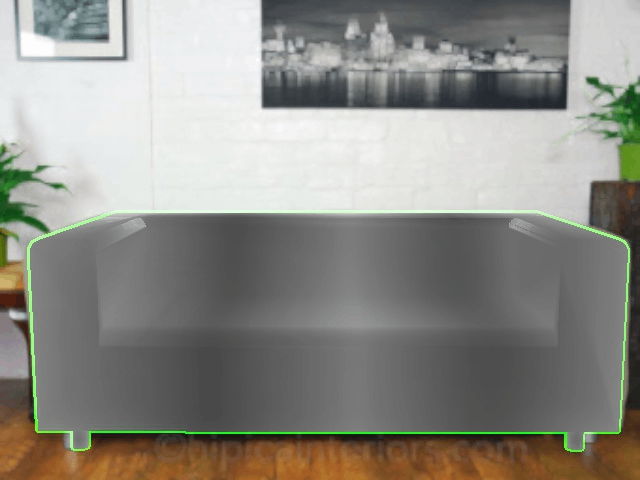}
        \end{minipage}
        \begin{minipage}{0.31\columnwidth}
            \centering
            \includegraphics[width=\textwidth]{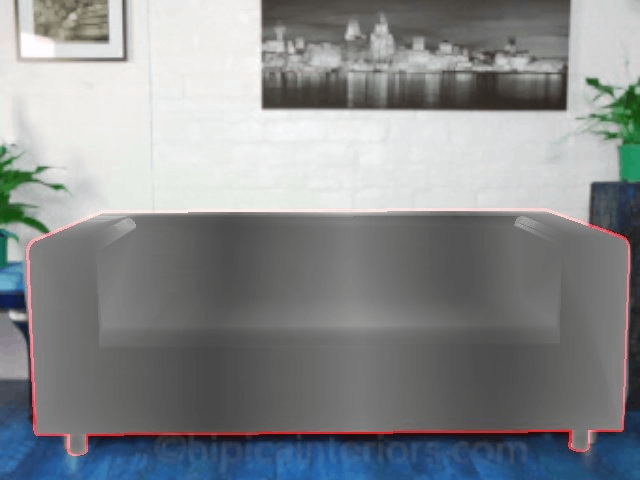}
        \end{minipage}\\[0.5mm]
    \small{3}
        \begin{minipage}{0.31\columnwidth}
            \centering
            \includegraphics[width=\textwidth]{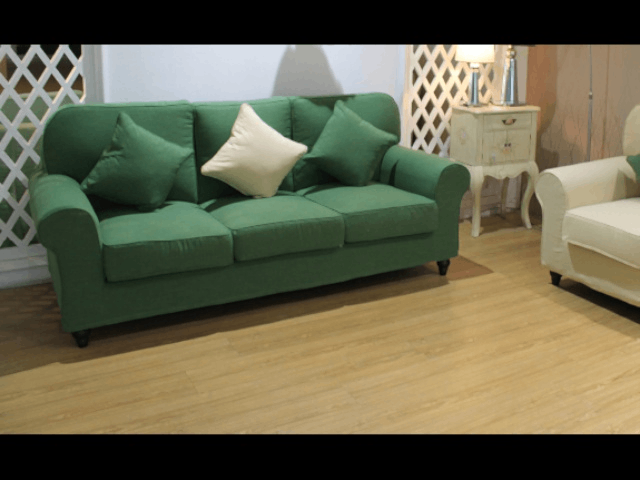}
        \end{minipage}
        \begin{minipage}{0.31\columnwidth}
            \centering
            \includegraphics[width=\textwidth]{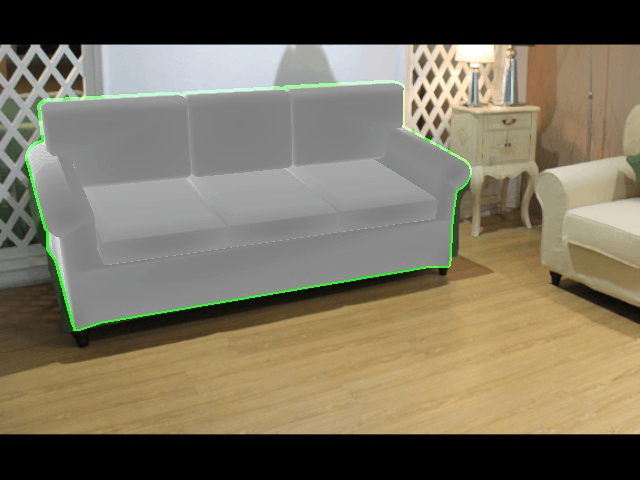}
        \end{minipage}
        \begin{minipage}{0.31\columnwidth}
            \centering
            \includegraphics[width=\textwidth]{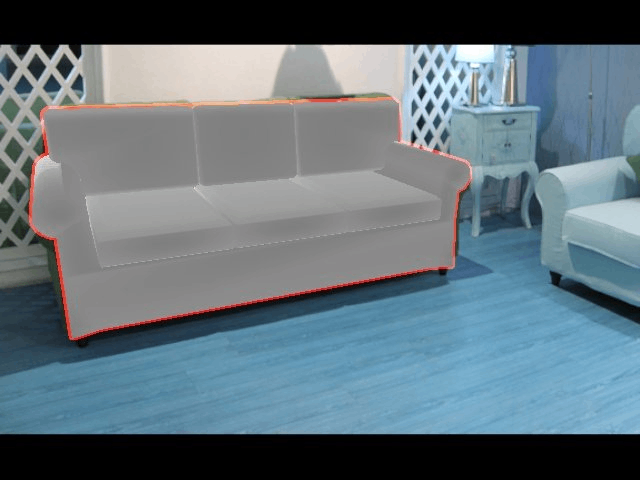}
        \end{minipage}\\[0.5mm]
    \small{4}
        \begin{minipage}{0.31\columnwidth}
            \centering
            \includegraphics[width=\textwidth]{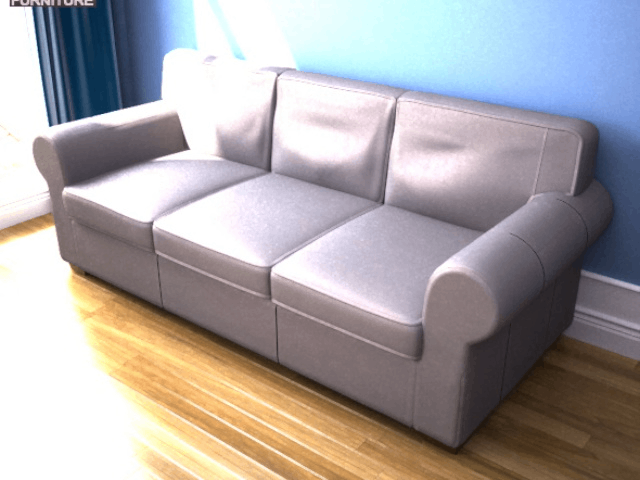}
        \end{minipage}
        \begin{minipage}{0.31\columnwidth}
            \centering
            \includegraphics[width=\textwidth]{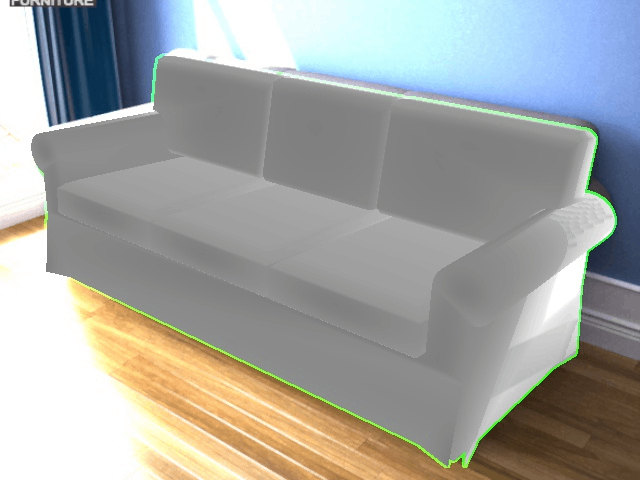}
        \end{minipage}
        \begin{minipage}{0.31\columnwidth}
            \centering
            \includegraphics[width=\textwidth]{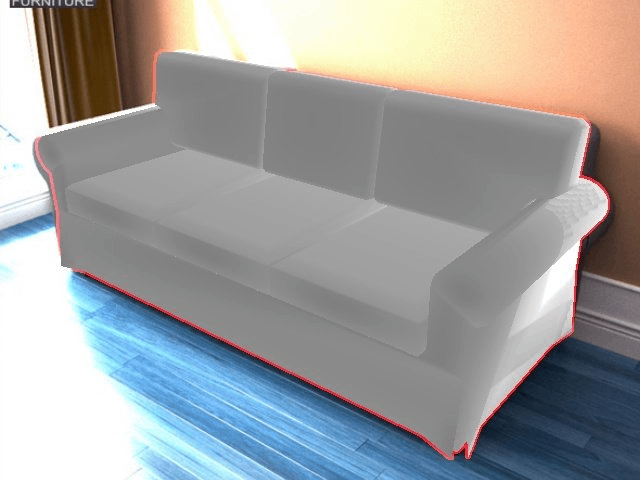}
        \end{minipage}\\[0.5mm]
    \small{5}
        \begin{minipage}{0.31\columnwidth}
            \centering
            \includegraphics[width=\textwidth]{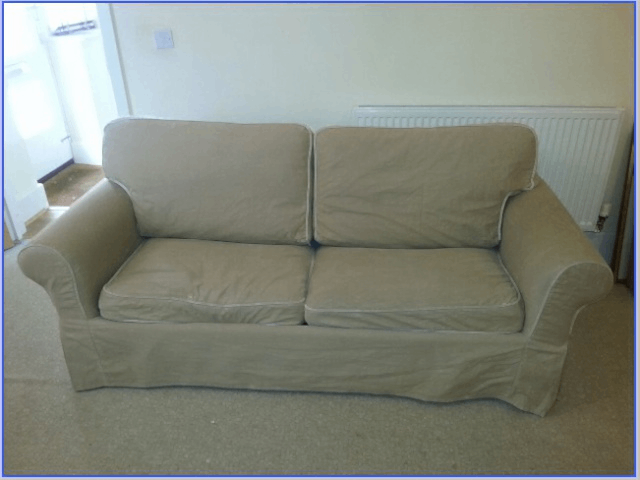}
        \end{minipage}
        \begin{minipage}{0.31\columnwidth}
            \centering
            \includegraphics[width=\textwidth]{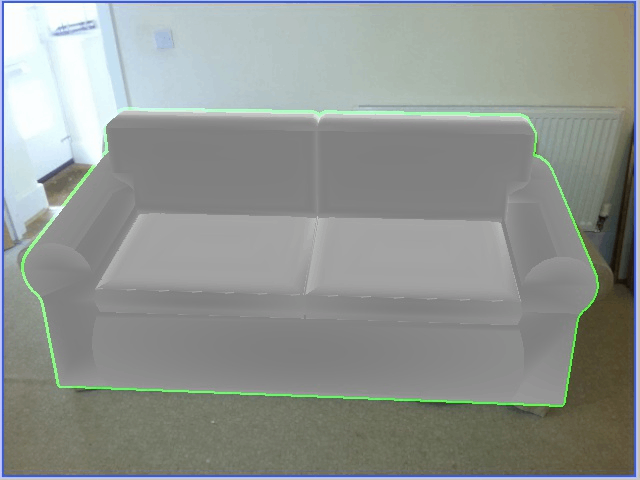}
        \end{minipage}
        \begin{minipage}{0.31\columnwidth}
            \centering
            \includegraphics[width=\textwidth]{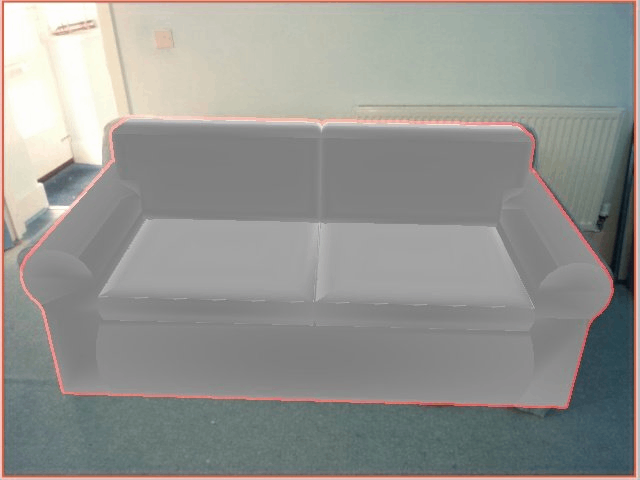}
        \end{minipage}\\[0.5mm]
    \small{6}
        \begin{minipage}{0.31\columnwidth}
            \centering
            \includegraphics[width=\textwidth]{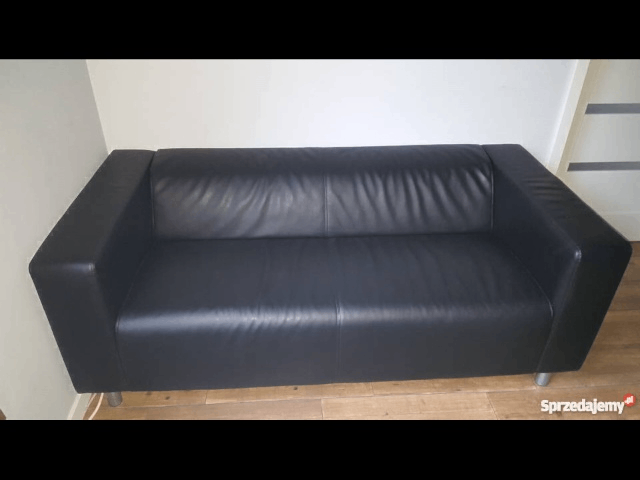}
        \end{minipage}
        \begin{minipage}{0.31\columnwidth}
            \centering
            \includegraphics[width=\textwidth]{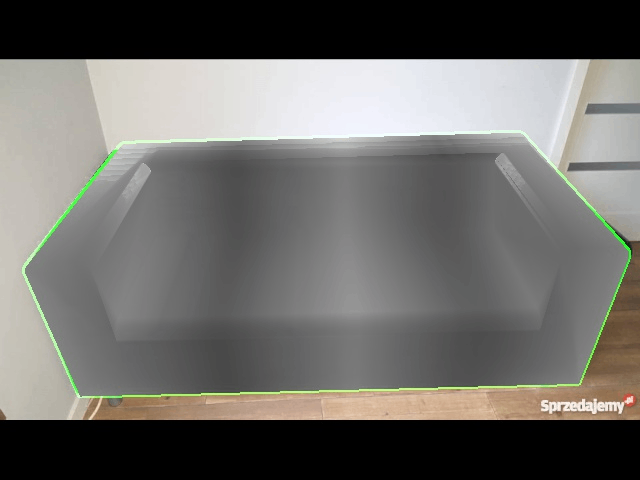}
        \end{minipage}
        \begin{minipage}{0.31\columnwidth}
            \centering
            \includegraphics[width=\textwidth]{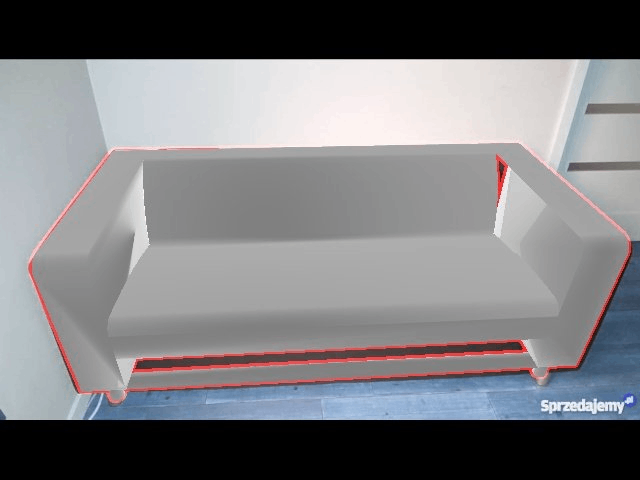}
        \end{minipage}\\[0.5mm]
    \small{7}
        \begin{minipage}{0.31\columnwidth}
            \centering
            \includegraphics[width=\textwidth]{figures/sofas/entry_343_input.png}
        \end{minipage}
        \begin{minipage}{0.31\columnwidth}
            \centering
            \includegraphics[width=\textwidth]{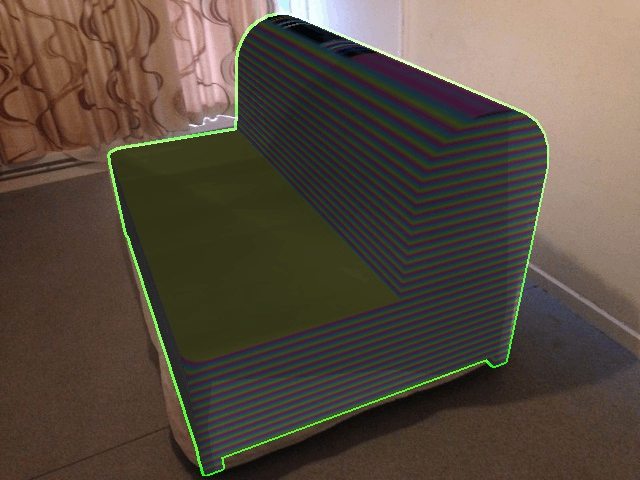}
        \end{minipage}
        \begin{minipage}{0.31\columnwidth}
            \centering
            \includegraphics[width=\textwidth]{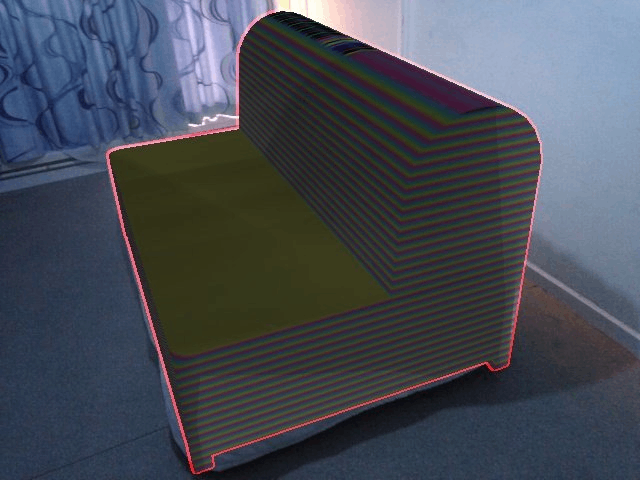}
        \end{minipage}\\[0.5mm]
    \small{8}
        \begin{minipage}{0.31\columnwidth}
            \centering
            \includegraphics[width=\textwidth]{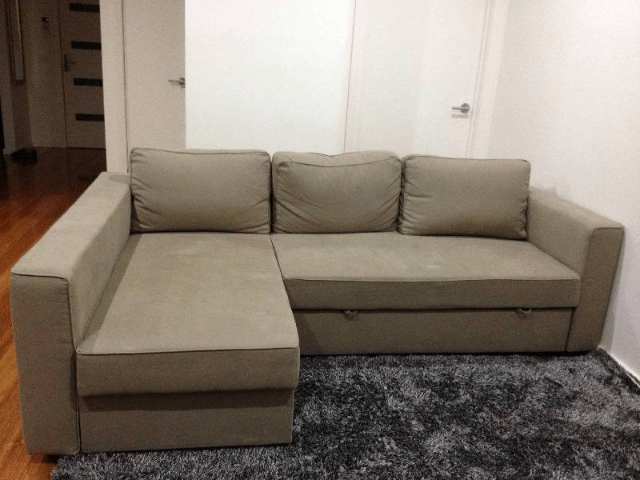}
        \end{minipage}
        \begin{minipage}{0.31\columnwidth}
            \centering
            \includegraphics[width=\textwidth]{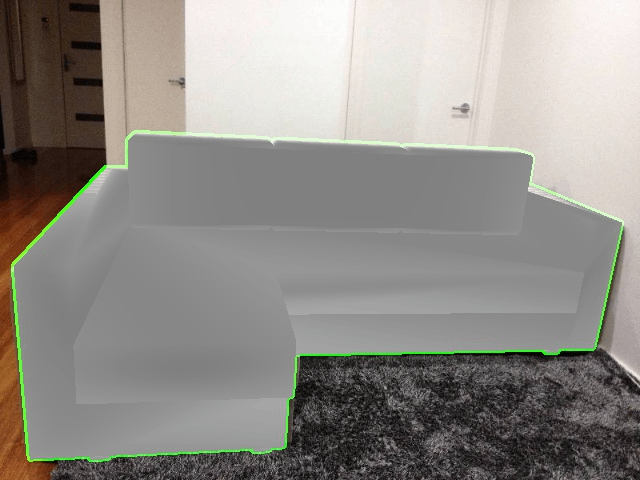}
        \end{minipage}
        \begin{minipage}{0.31\columnwidth}
            \centering
            \includegraphics[width=\textwidth]{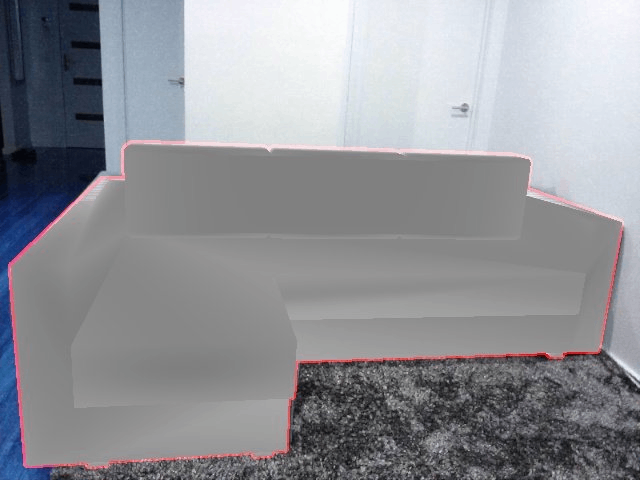}
        \end{minipage}\\[0.5mm]

    \caption{\textbf{Qualitative results for Pix3D sofas.}}
    \label{pix3d-sofas-q}
    \vspace*{-5mm}
\end{figure}

\begin{figure}[ht]
    \centering
    \small{1}
        \begin{minipage}{0.31\columnwidth}
            {\small Input image\vspace{1mm}}
            \centering
            \includegraphics[width=\textwidth]{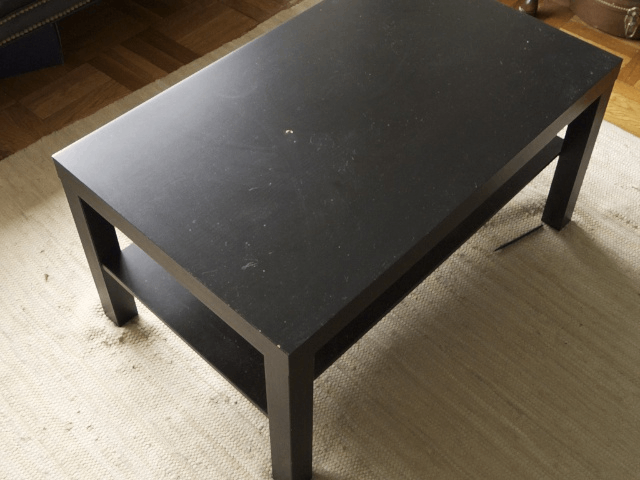}
        \end{minipage}
        \begin{minipage}{0.31\columnwidth}
            {\small Ground truth\vspace{1mm}}
            \centering
            \includegraphics[width=\textwidth]{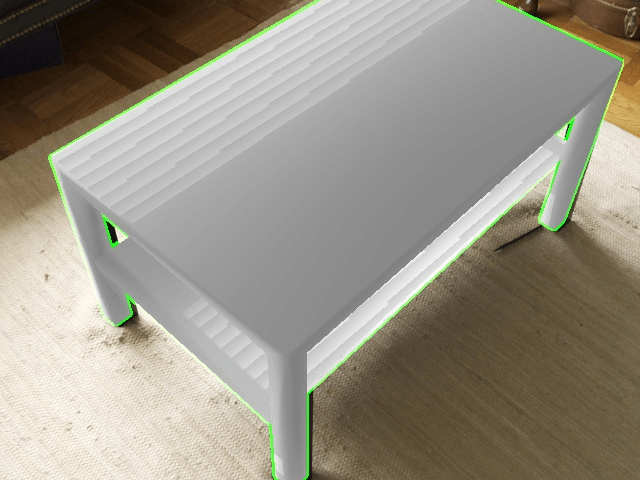}
        \end{minipage}
        \begin{minipage}{0.31\columnwidth}
            {\small Our prediction\vspace{1mm}}
            \centering
            \includegraphics[width=\textwidth]{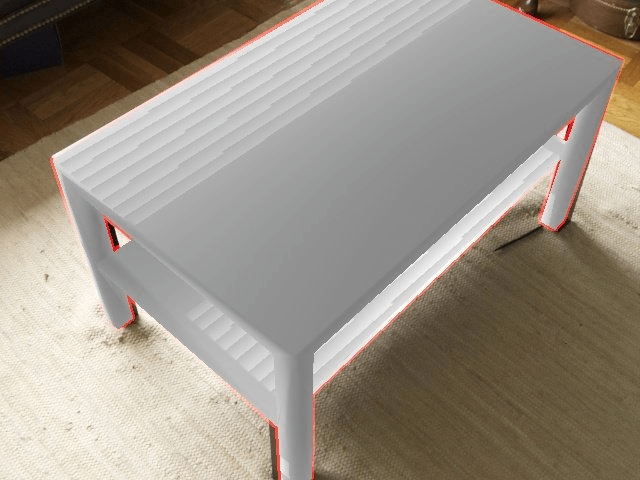}
        \end{minipage}\\[0.5mm]
    \small{2}
        \begin{minipage}{0.31\columnwidth}
            \centering
            \includegraphics[width=\textwidth]{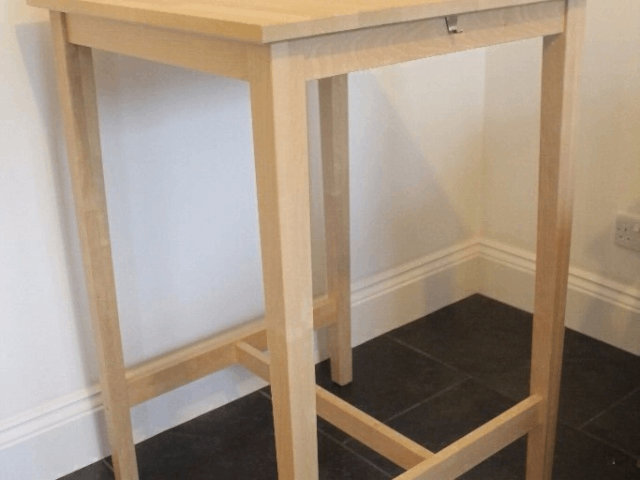}
        \end{minipage}
        \begin{minipage}{0.31\columnwidth}
            \centering
            \includegraphics[width=\textwidth]{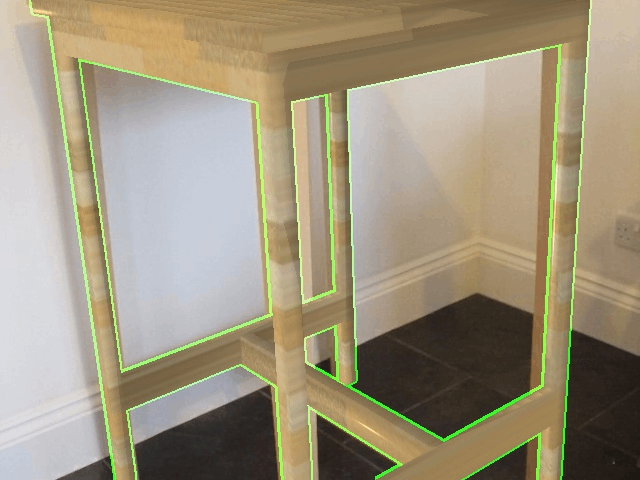}
        \end{minipage}
        \begin{minipage}{0.31\columnwidth}
            \centering
            \includegraphics[width=\textwidth]{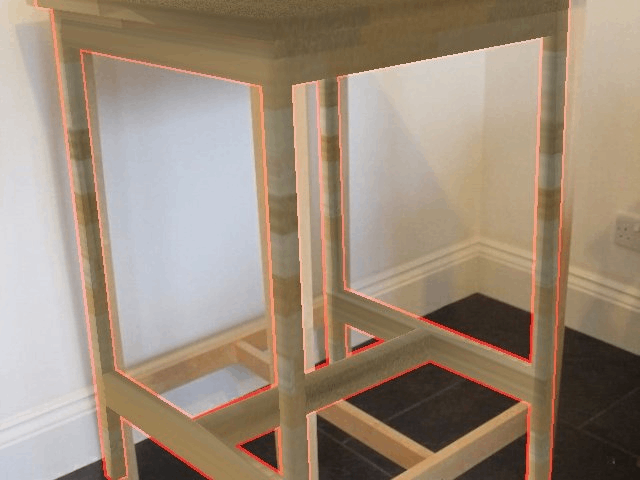}
        \end{minipage}\\[0.5mm]
    \small{3}
        \begin{minipage}{0.31\columnwidth}
            \centering
            \includegraphics[width=\textwidth]{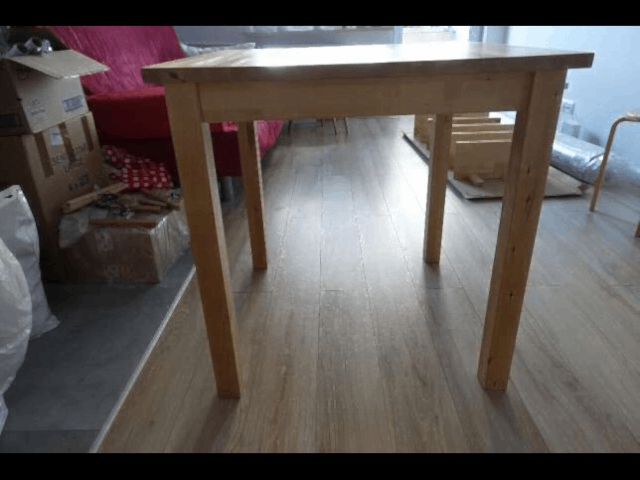}
        \end{minipage}
        \begin{minipage}{0.31\columnwidth}
            \centering
            \includegraphics[width=\textwidth]{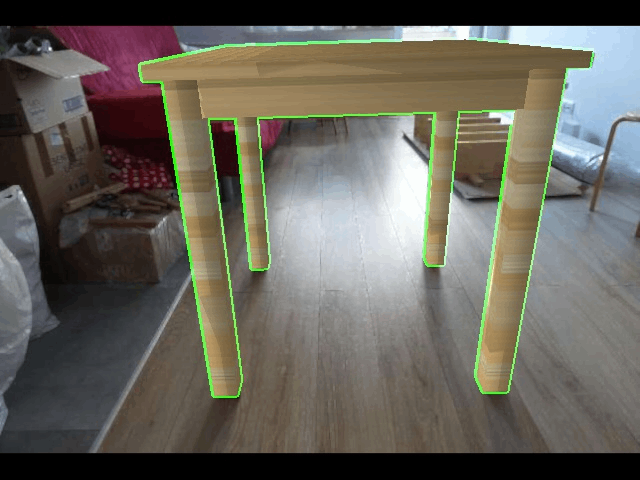}
        \end{minipage}
        \begin{minipage}{0.31\columnwidth}
            \centering
            \includegraphics[width=\textwidth]{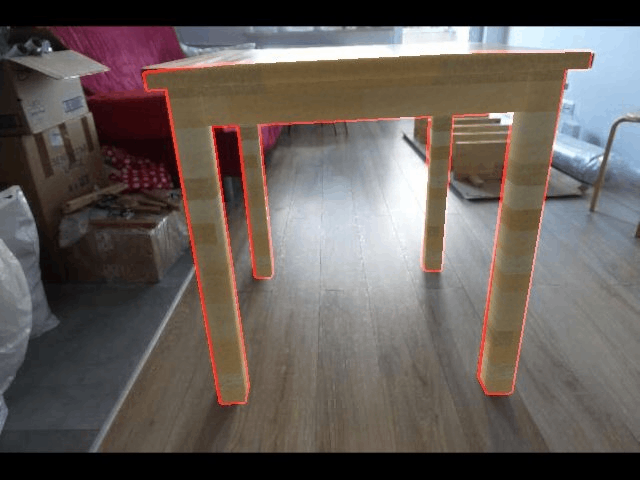}
        \end{minipage}\\[0.5mm]
    \small{4}
        \begin{minipage}{0.31\columnwidth}
            \centering
            \includegraphics[width=\textwidth]{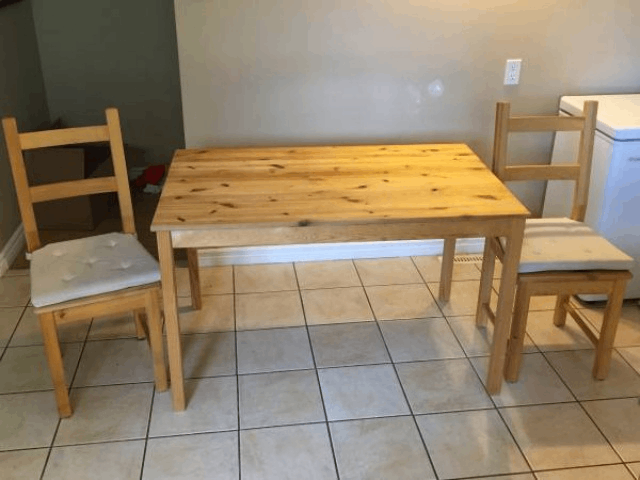}
        \end{minipage}
        \begin{minipage}{0.31\columnwidth}
            \centering
            \includegraphics[width=\textwidth]{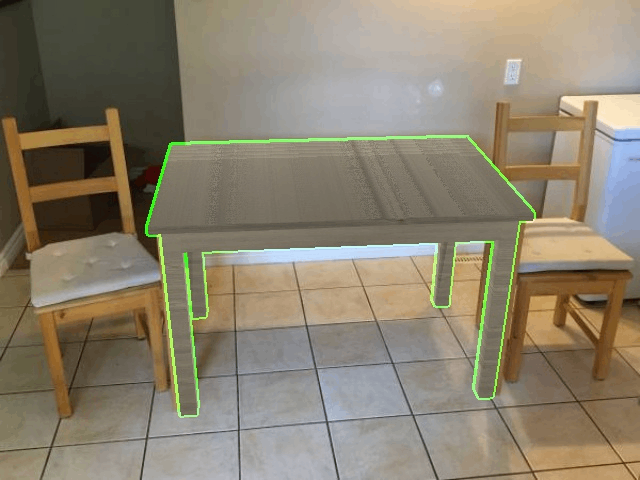}
        \end{minipage}
        \begin{minipage}{0.31\columnwidth}
            \centering
            \includegraphics[width=\textwidth]{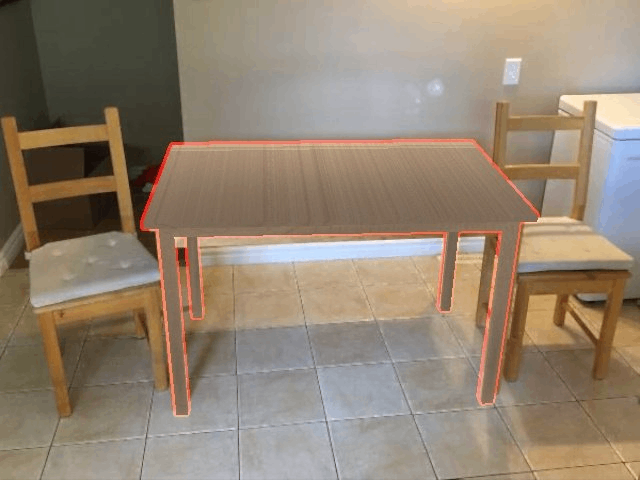}
        \end{minipage}\\[0.5mm]
    \small{5}
        \begin{minipage}{0.31\columnwidth}
            \centering
            \includegraphics[width=\textwidth]{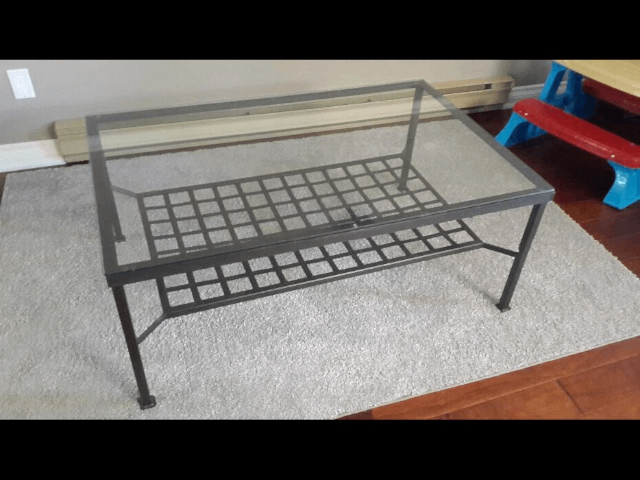}
        \end{minipage}
        \begin{minipage}{0.31\columnwidth}
            \centering
            \includegraphics[width=\textwidth]{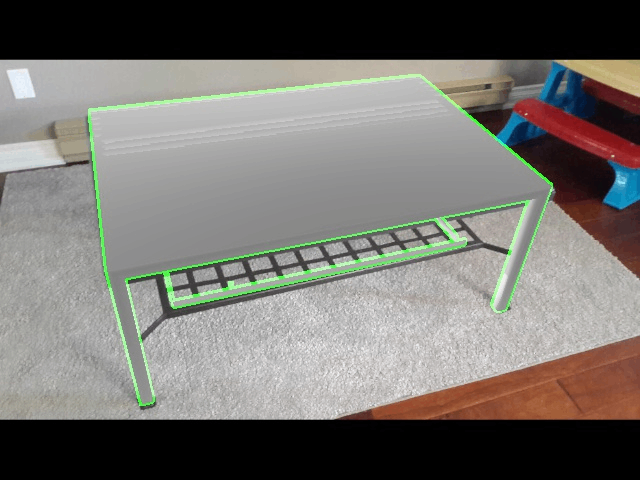}
        \end{minipage}
        \begin{minipage}{0.31\columnwidth}
            \centering
            \includegraphics[width=\textwidth]{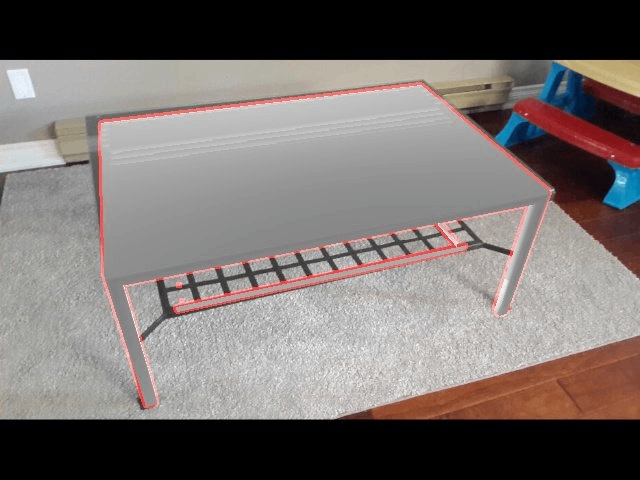}
        \end{minipage}\\[0.5mm]
    \small{6}
        \begin{minipage}{0.31\columnwidth}
            \centering
            \includegraphics[width=\textwidth]{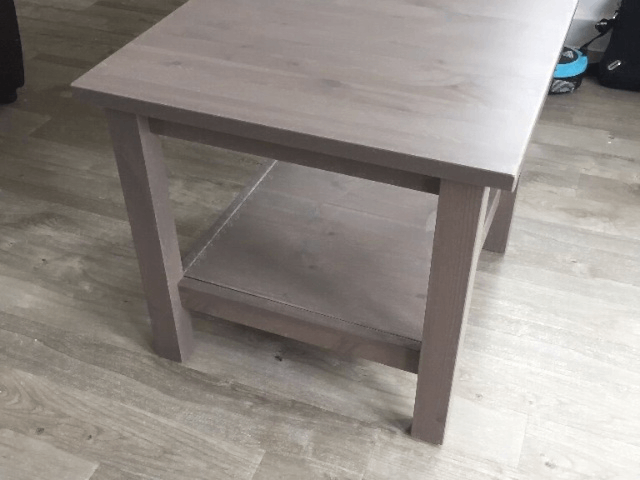}
        \end{minipage}
        \begin{minipage}{0.31\columnwidth}
            \centering
            \includegraphics[width=\textwidth]{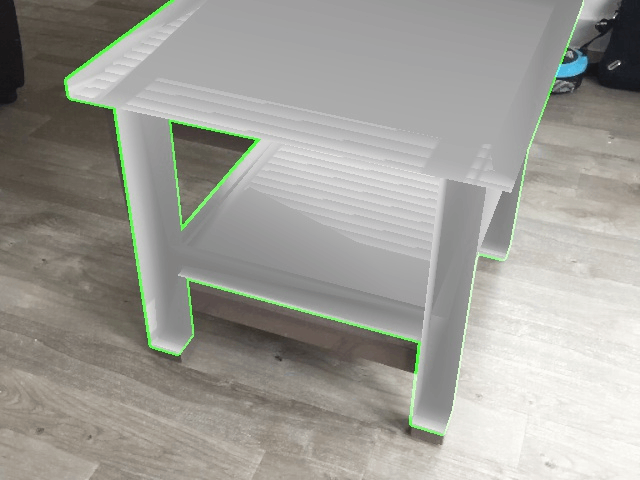}
        \end{minipage}
        \begin{minipage}{0.31\columnwidth}
            \centering
            \includegraphics[width=\textwidth]{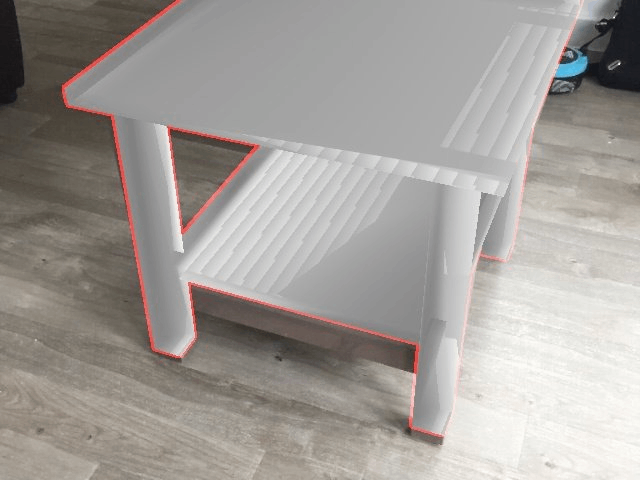}
        \end{minipage}\\[0.5mm]
    \small{7}
        \begin{minipage}{0.31\columnwidth}
            \centering
            \includegraphics[width=\textwidth]{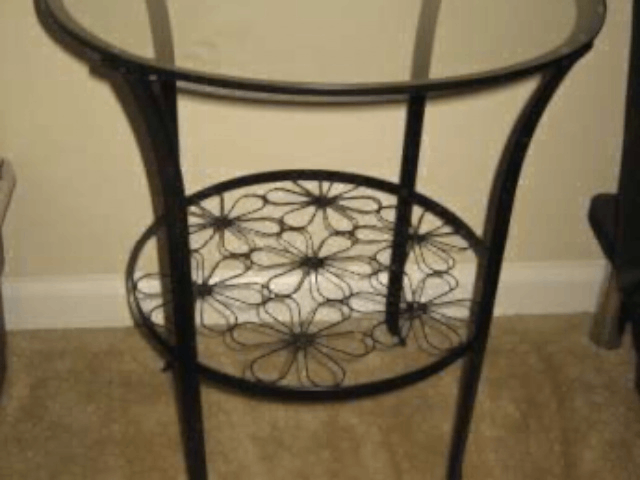}
        \end{minipage}
        \begin{minipage}{0.31\columnwidth}
            \centering
            \includegraphics[width=\textwidth]{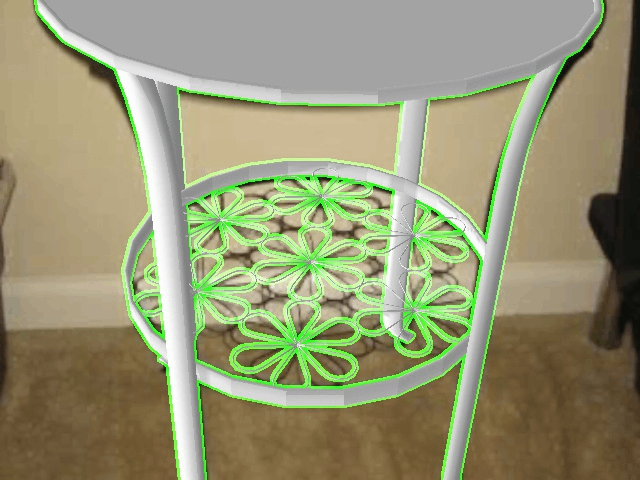}
        \end{minipage}
        \begin{minipage}{0.31\columnwidth}
            \centering
            \includegraphics[width=\textwidth]{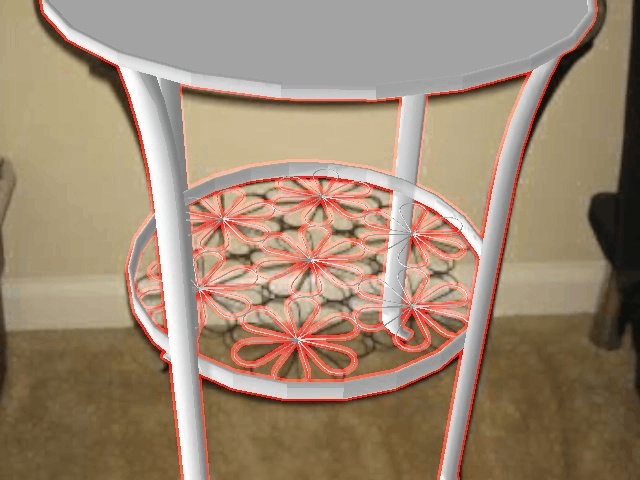}
        \end{minipage}\\[0.5mm]
    \small{8}
        \begin{minipage}{0.31\columnwidth}
            \centering
            \includegraphics[width=\textwidth]{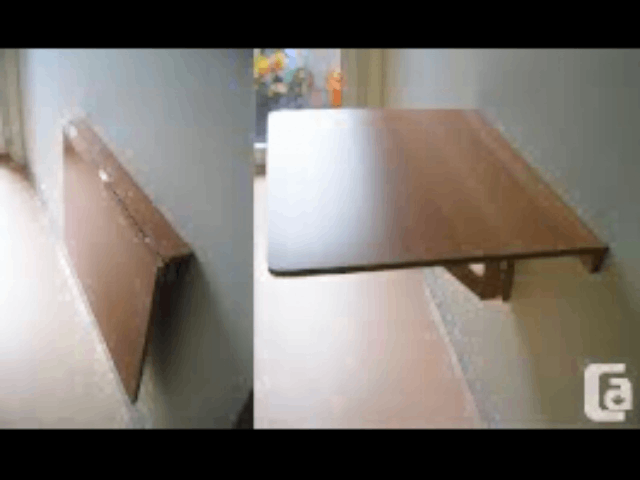}
        \end{minipage}
        \begin{minipage}{0.31\columnwidth}
            \centering
            \includegraphics[width=\textwidth]{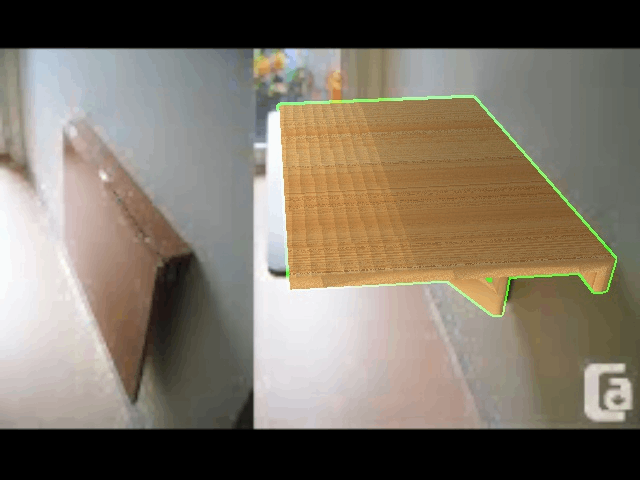}
        \end{minipage}
        \begin{minipage}{0.31\columnwidth}
            \centering
            \includegraphics[width=\textwidth]{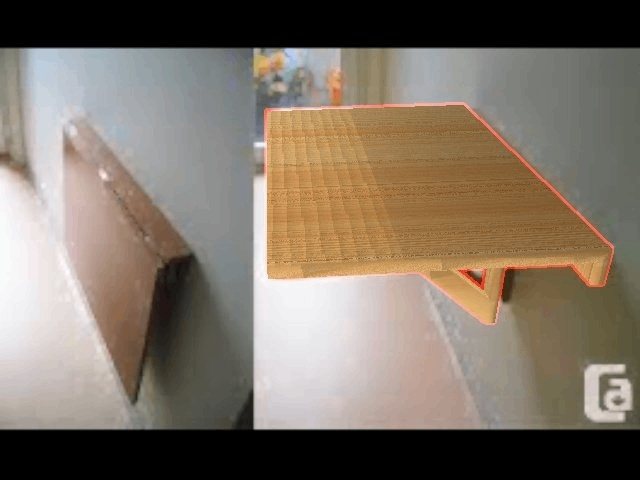}
        \end{minipage}\\[0.5mm]

    \caption{\textbf{Qualitative results for Pix3D tables.}}
    \label{pix3d-tables-q}
    \vspace*{-5mm}
\end{figure}

\begin{figure}[ht]
    \centering
    \small{1}
        \begin{minipage}{0.31\columnwidth}
            {\small Input image\vspace{1mm}}
            \centering
            \includegraphics[width=\textwidth]{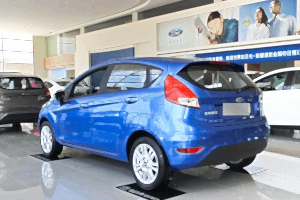}
        \end{minipage}
        \begin{minipage}{0.31\columnwidth}
            {\small Ground truth\vspace{1mm}}
            \centering
            \includegraphics[width=\textwidth]{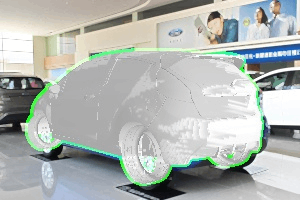}
        \end{minipage}
        \begin{minipage}{0.31\columnwidth}
            {\small Our prediction\vspace{1mm}}
            \centering
            \includegraphics[width=\textwidth]{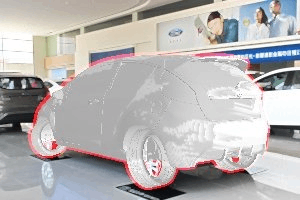}
        \end{minipage}\\[0.5mm]
    \small{2}
        \begin{minipage}{0.31\columnwidth}
            \centering
            \includegraphics[width=\textwidth]{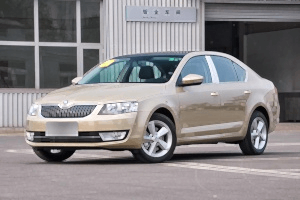}
        \end{minipage}
        \begin{minipage}{0.31\columnwidth}
            \centering
            \includegraphics[width=\textwidth]{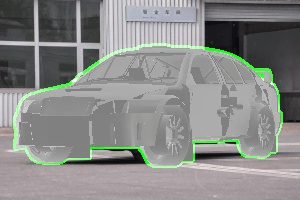}
        \end{minipage}
        \begin{minipage}{0.31\columnwidth}
            \centering
            \includegraphics[width=\textwidth]{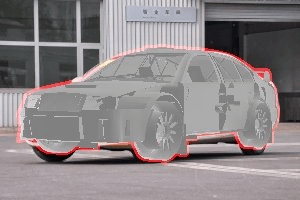}
        \end{minipage}\\[0.5mm]
    \small{3}
        \begin{minipage}{0.31\columnwidth}
            \centering
            \includegraphics[width=\textwidth]{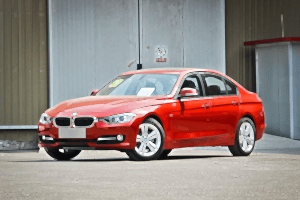}
        \end{minipage}
        \begin{minipage}{0.31\columnwidth}
            \centering
            \includegraphics[width=\textwidth]{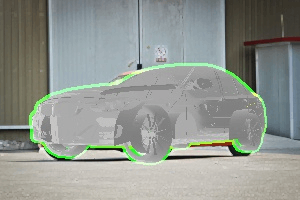}
        \end{minipage}
        \begin{minipage}{0.31\columnwidth}
            \centering
            \includegraphics[width=\textwidth]{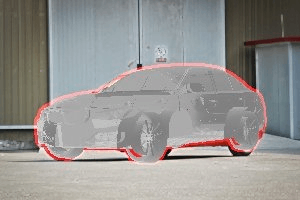}
        \end{minipage}\\[0.5mm]
    \small{4}
        \begin{minipage}{0.31\columnwidth}
            \centering
            \includegraphics[width=\textwidth]{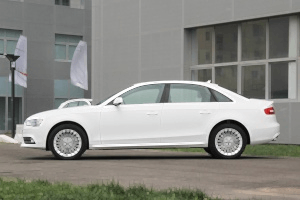}
        \end{minipage}
        \begin{minipage}{0.31\columnwidth}
            \centering
            \includegraphics[width=\textwidth]{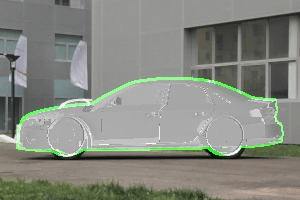}
        \end{minipage}
        \begin{minipage}{0.31\columnwidth}
            \centering
            \includegraphics[width=\textwidth]{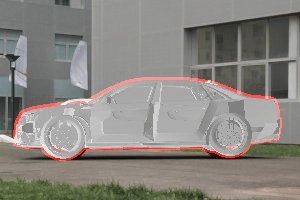}
        \end{minipage}\\[0.5mm]
    \small{5}
        \begin{minipage}{0.31\columnwidth}
            \centering
            \includegraphics[width=\textwidth]{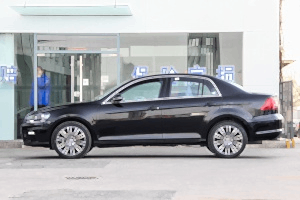}
        \end{minipage}
        \begin{minipage}{0.31\columnwidth}
            \centering
            \includegraphics[width=\textwidth]{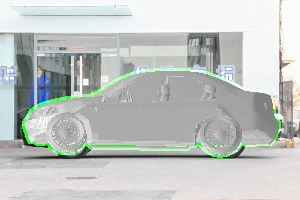}
        \end{minipage}
        \begin{minipage}{0.31\columnwidth}
            \centering
            \includegraphics[width=\textwidth]{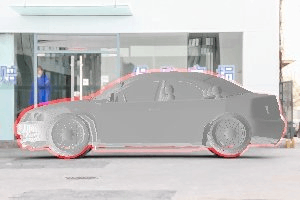}
        \end{minipage}\\[0.5mm]
    \small{6}
        \begin{minipage}{0.31\columnwidth}
            \centering
            \includegraphics[width=\textwidth]{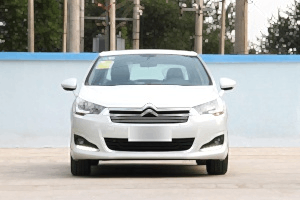}
        \end{minipage}
        \begin{minipage}{0.31\columnwidth}
            \centering
            \includegraphics[width=\textwidth]{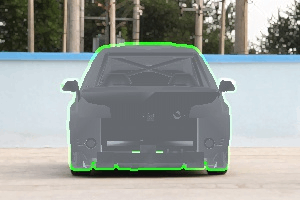}
        \end{minipage}
        \begin{minipage}{0.31\columnwidth}
            \centering
            \includegraphics[width=\textwidth]{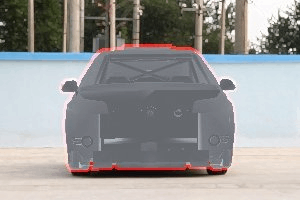}
        \end{minipage}\\[0.5mm]
    \small{7}
        \begin{minipage}{0.31\columnwidth}
            \centering
            \includegraphics[width=\textwidth]{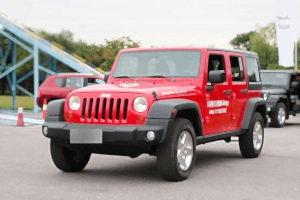}
        \end{minipage}
        \begin{minipage}{0.31\columnwidth}
            \centering
            \includegraphics[width=\textwidth]{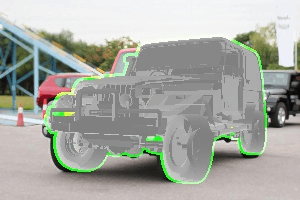}
        \end{minipage}
        \begin{minipage}{0.31\columnwidth}
            \centering
            \includegraphics[width=\textwidth]{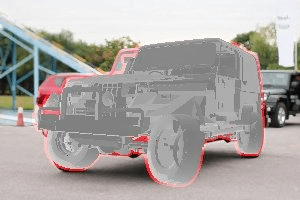}
        \end{minipage}\\[0.5mm]
    \small{8}
        \begin{minipage}{0.31\columnwidth}
            \centering
            \includegraphics[width=\textwidth]{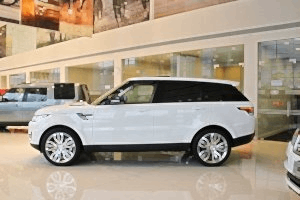}
        \end{minipage}
        \begin{minipage}{0.31\columnwidth}
            \centering
            \includegraphics[width=\textwidth]{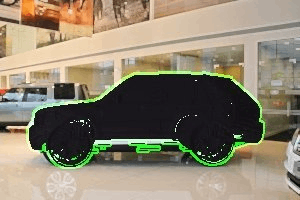}
        \end{minipage}
        \begin{minipage}{0.31\columnwidth}
            \centering
            \includegraphics[width=\textwidth]{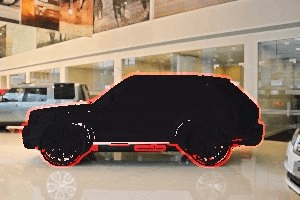}
        \end{minipage}\\[0.5mm]

    \caption{\textbf{Qualitative results for CompCars.}}
    \label{compcars-q}
    \vspace*{-5mm}
\end{figure}

\begin{figure}[ht]
    \centering
    \small{1}
        \begin{minipage}{0.31\columnwidth}
            {\small Input image\vspace{1mm}}
            \centering
            \includegraphics[width=\textwidth]{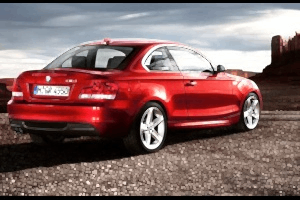}
        \end{minipage}
        \begin{minipage}{0.31\columnwidth}
            {\small Ground truth\vspace{1mm}}
            \centering
            \includegraphics[width=\textwidth]{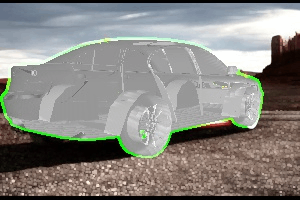}
        \end{minipage}
        \begin{minipage}{0.31\columnwidth}
            {\small Our prediction\vspace{1mm}}
            \centering
            \includegraphics[width=\textwidth]{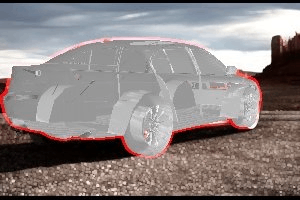}
        \end{minipage}\\[0.5mm]
    \small{2}
        \begin{minipage}{0.31\columnwidth}
            \centering
            \includegraphics[width=\textwidth]{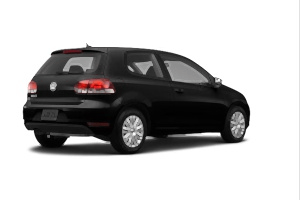}
        \end{minipage}
        \begin{minipage}{0.31\columnwidth}
            \centering
            \includegraphics[width=\textwidth]{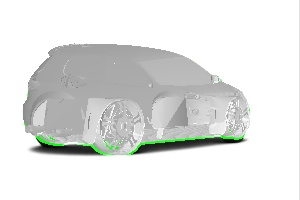}
        \end{minipage}
        \begin{minipage}{0.31\columnwidth}
            \centering
            \includegraphics[width=\textwidth]{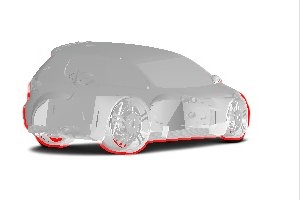}
        \end{minipage}\\[0.5mm]
    \small{3}
        \begin{minipage}{0.31\columnwidth}
            \centering
            \includegraphics[width=\textwidth]{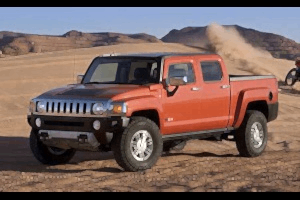}
        \end{minipage}
        \begin{minipage}{0.31\columnwidth}
            \centering
            \includegraphics[width=\textwidth]{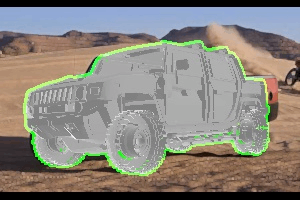}
        \end{minipage}
        \begin{minipage}{0.31\columnwidth}
            \centering
            \includegraphics[width=\textwidth]{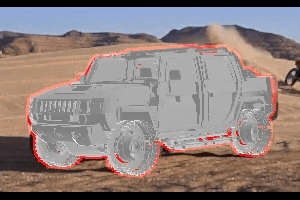}
        \end{minipage}\\[0.5mm]
    \small{4}
        \begin{minipage}{0.31\columnwidth}
            \centering
            \includegraphics[width=\textwidth]{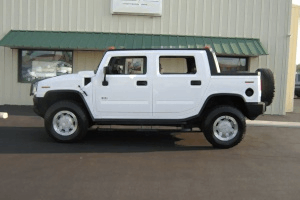}
        \end{minipage}
        \begin{minipage}{0.31\columnwidth}
            \centering
            \includegraphics[width=\textwidth]{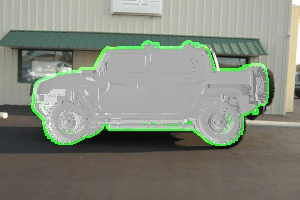}
        \end{minipage}
        \begin{minipage}{0.31\columnwidth}
            \centering
            \includegraphics[width=\textwidth]{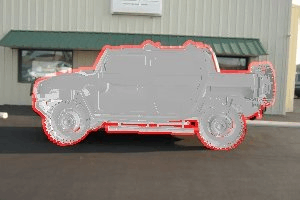}
        \end{minipage}\\[0.5mm]
    \small{5}
        \begin{minipage}{0.31\columnwidth}
            \centering
            \includegraphics[width=\textwidth]{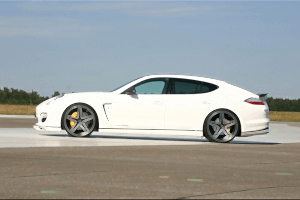}
        \end{minipage}
        \begin{minipage}{0.31\columnwidth}
            \centering
            \includegraphics[width=\textwidth]{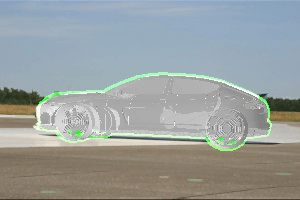}
        \end{minipage}
        \begin{minipage}{0.31\columnwidth}
            \centering
            \includegraphics[width=\textwidth]{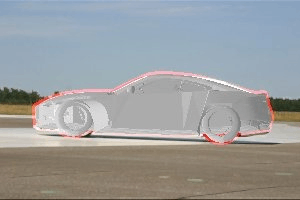}
        \end{minipage}\\[0.5mm]
    \small{6}
        \begin{minipage}{0.31\columnwidth}
            \centering
            \includegraphics[width=\textwidth]{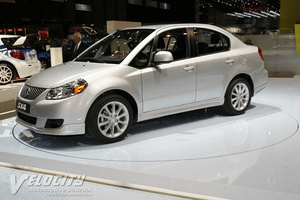}
        \end{minipage}
        \begin{minipage}{0.31\columnwidth}
            \centering
            \includegraphics[width=\textwidth]{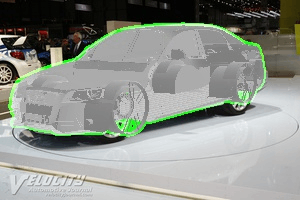}
        \end{minipage}
        \begin{minipage}{0.31\columnwidth}
            \centering
            \includegraphics[width=\textwidth]{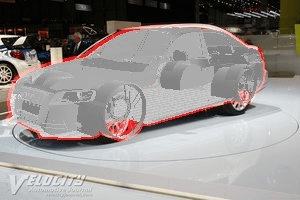}
        \end{minipage}\\[0.5mm]
    \small{7}
        \begin{minipage}{0.31\columnwidth}
            \centering
            \includegraphics[width=\textwidth]{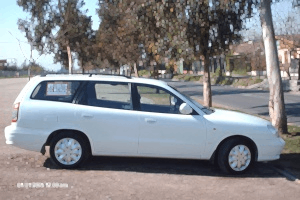}
        \end{minipage}
        \begin{minipage}{0.31\columnwidth}
            \centering
            \includegraphics[width=\textwidth]{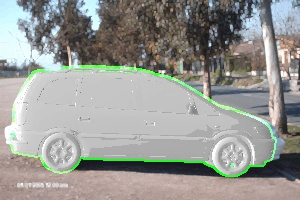}
        \end{minipage}
        \begin{minipage}{0.31\columnwidth}
            \centering
            \includegraphics[width=\textwidth]{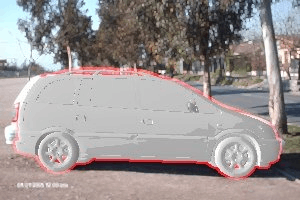}
        \end{minipage}\\[0.5mm]
    \small{8}
        \begin{minipage}{0.31\columnwidth}
            \centering
            \includegraphics[width=\textwidth]{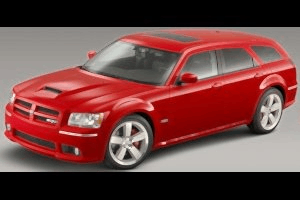}
        \end{minipage}
        \begin{minipage}{0.31\columnwidth}
            \centering
            \includegraphics[width=\textwidth]{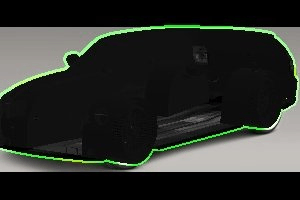}
        \end{minipage}
        \begin{minipage}{0.31\columnwidth}
            \centering
            \includegraphics[width=\textwidth]{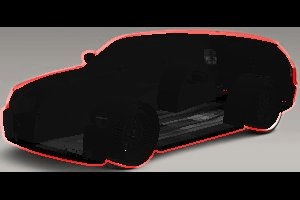}
        \end{minipage}\\[0.5mm]

    \caption{\textbf{Qualitative results for StanfordCars.}}
    \label{stanfordcars-q}
    \vspace*{-5mm}
\end{figure}

\begin{figure*}[t]
    \centering
    \small{1}
    \begin{minipage}{0.225\textwidth}
        {\small Input image\vspace{1mm}}
            \centering
            \includegraphics[width=\textwidth]{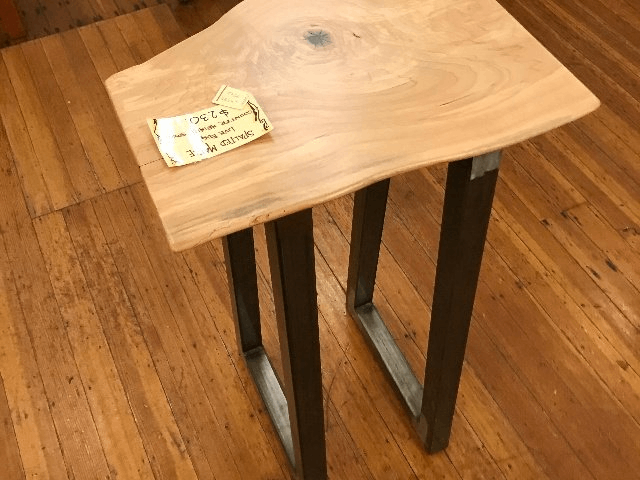}
        \end{minipage}
        \begin{minipage}{0.225\textwidth}
        {\small Ground truth\vspace{1mm}}
            \centering
            \includegraphics[width=\textwidth]{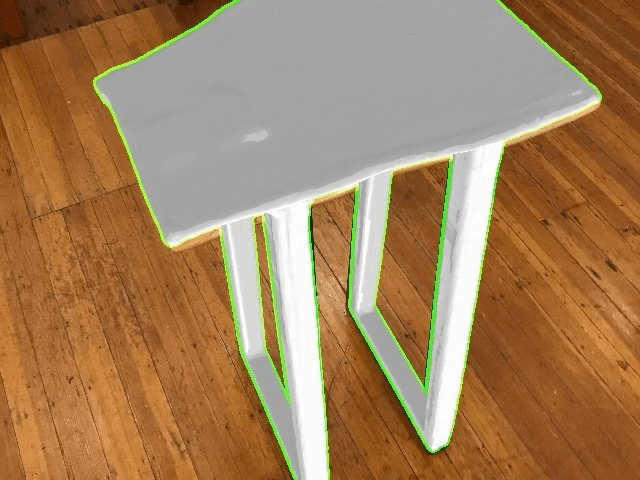}
        \end{minipage}
        \begin{minipage}{0.225\textwidth}
        {\small Our prediction\vspace{1mm}}
            \centering
            \includegraphics[width=\textwidth]{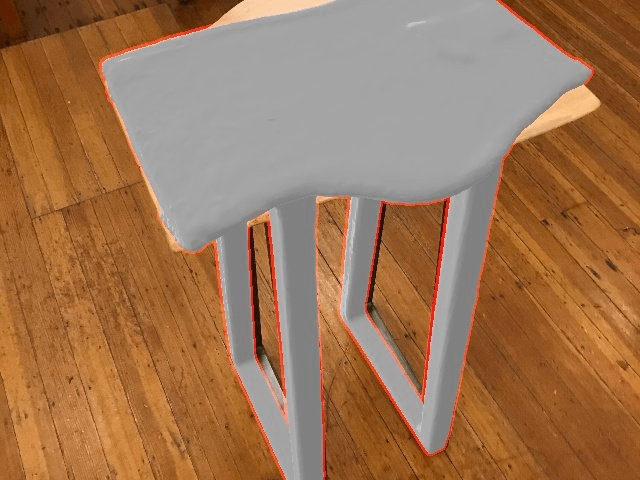}
        \end{minipage}\\[1mm]
        
        \small{2}
        \begin{minipage}{0.225\textwidth}
            \centering
            \includegraphics[width=\textwidth]{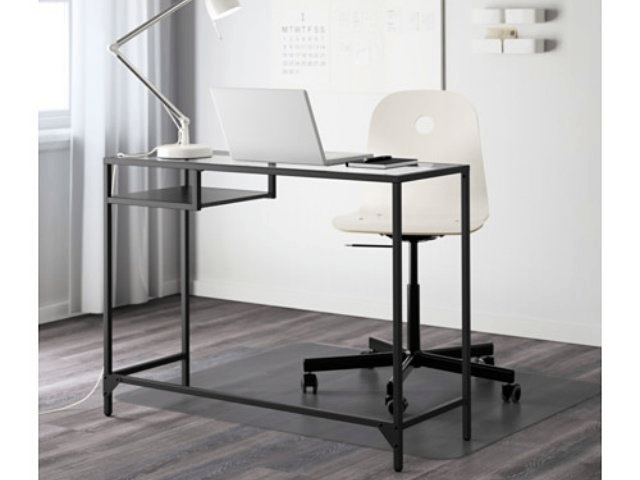}
        \end{minipage}
        \begin{minipage}{0.225\textwidth}
            \centering
            \includegraphics[width=\textwidth]{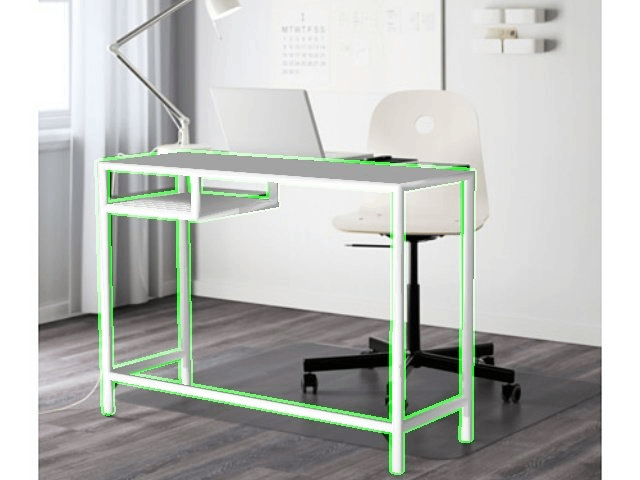}
        \end{minipage}
        \begin{minipage}{0.225\textwidth}
            \centering
            \includegraphics[width=\textwidth]{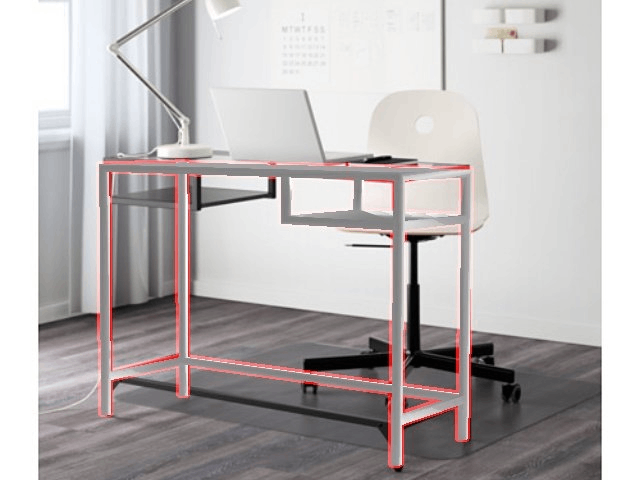}
        \end{minipage}\\[1mm]
        \small{3}
        \begin{minipage}{0.225\textwidth}
        \centering
            \includegraphics[width=\textwidth]{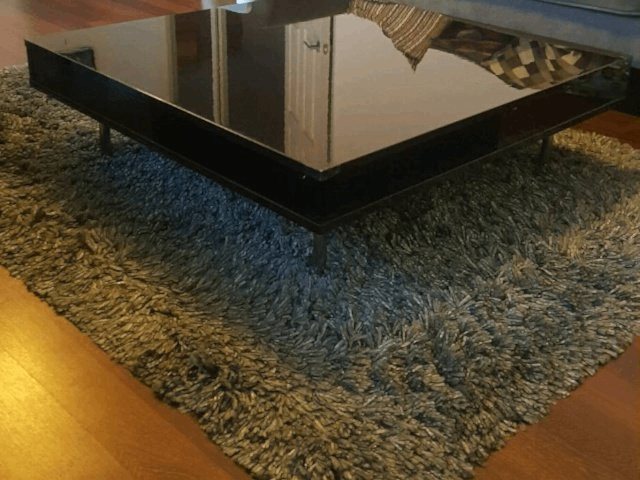}
        \end{minipage}
        \begin{minipage}{0.225\textwidth}
            \centering
            \includegraphics[width=\textwidth]{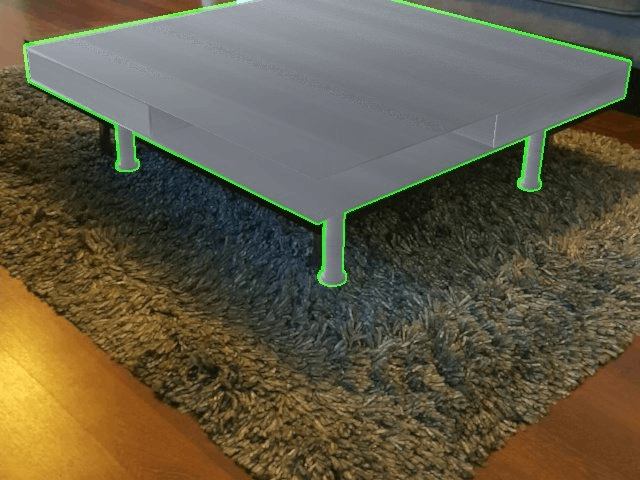}
        \end{minipage}
        \begin{minipage}{0.225\textwidth}
            \centering
            \includegraphics[width=\textwidth]{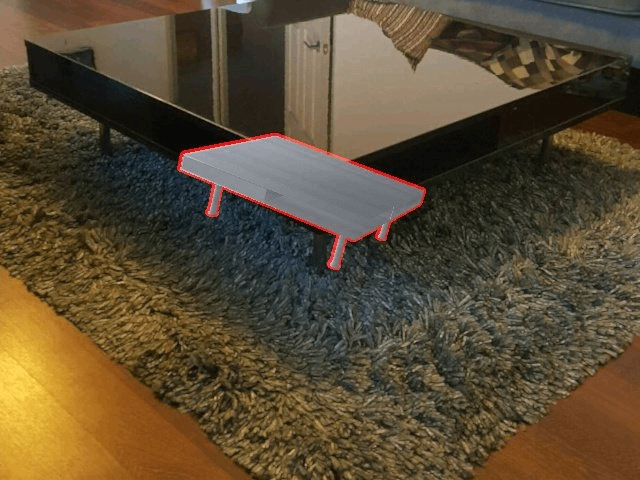}
        \end{minipage}\\[1mm]
        \small{4}
        \begin{minipage}{0.225\textwidth}
        \centering
            \includegraphics[width=\textwidth]{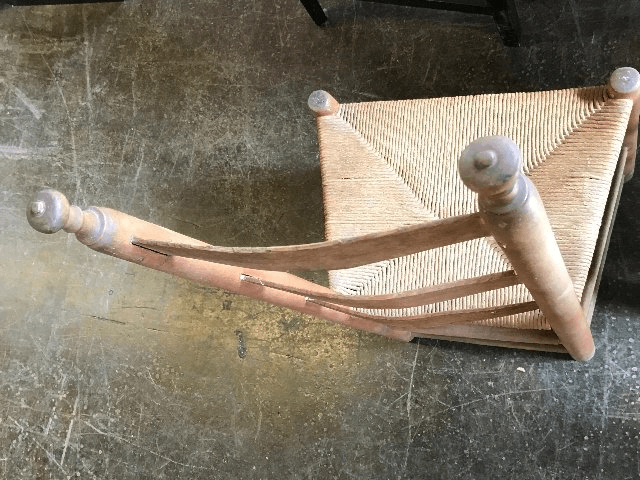}
        \end{minipage}
        \begin{minipage}{0.225\textwidth}
            \centering
            \includegraphics[width=\textwidth]{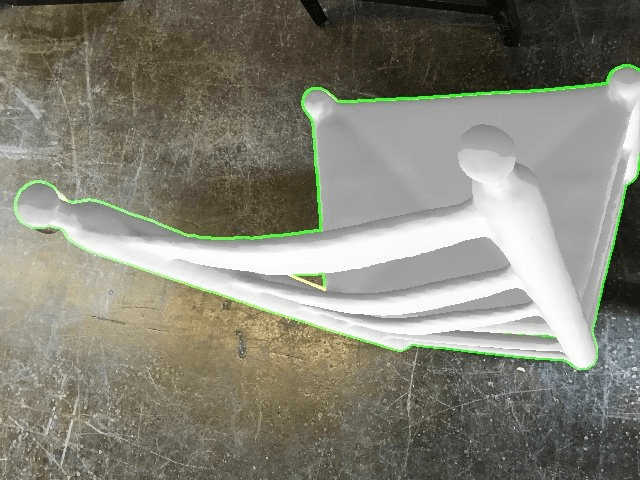}
        \end{minipage}
        \begin{minipage}{0.225\textwidth}
            \centering
            \includegraphics[width=\textwidth]{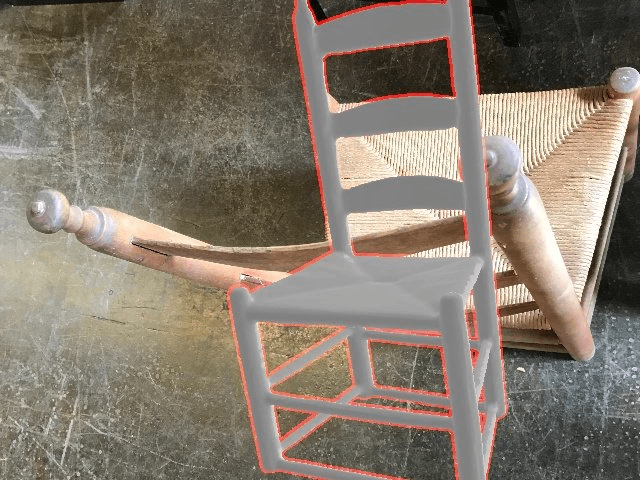}
        \end{minipage}\\[1mm]
        \small{5}
        \begin{minipage}{0.225\textwidth}
            \centering
            \includegraphics[width=\textwidth]{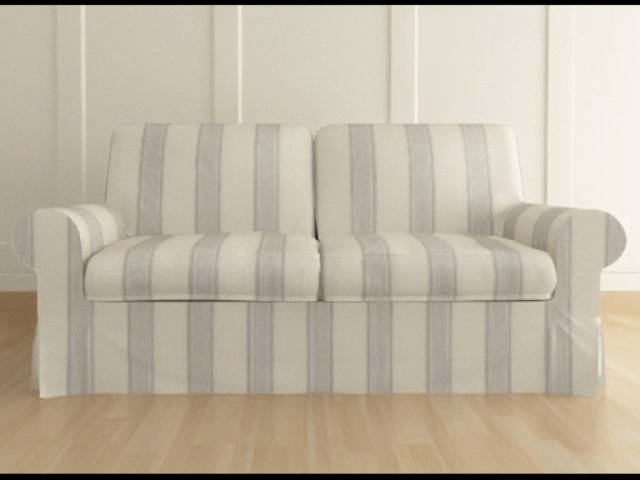}
        \end{minipage}
        \begin{minipage}{0.225\textwidth}
            \centering
            \includegraphics[width=\textwidth]{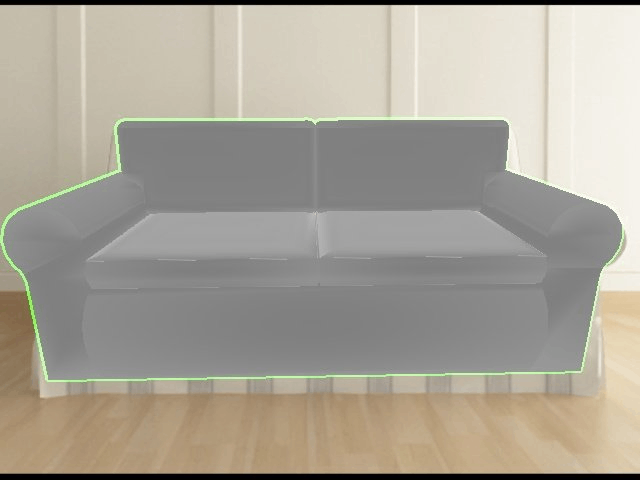}
        \end{minipage}
        \begin{minipage}{0.225\textwidth}
            \centering
            \includegraphics[width=\textwidth]{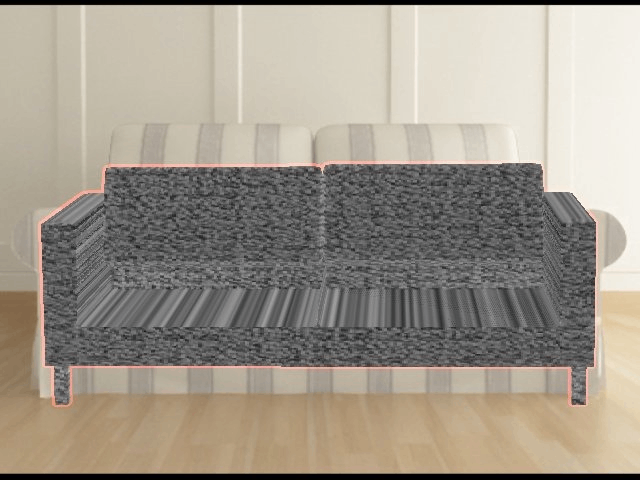}
        \end{minipage}\\[1mm]
        \small{6}
         \begin{minipage}{0.225\textwidth}
            \centering
            \includegraphics[width=\textwidth]{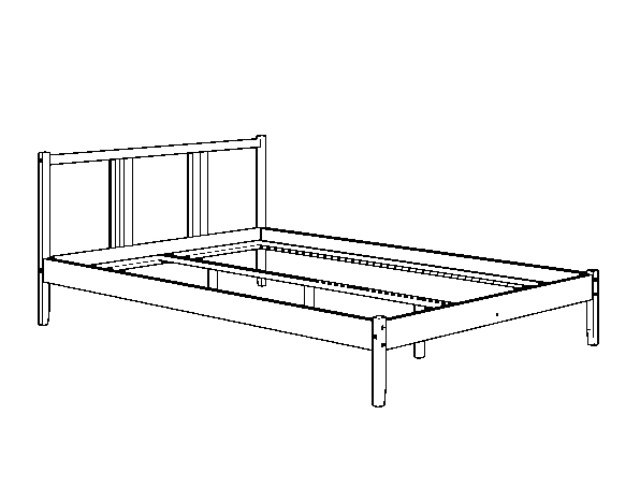}
        \end{minipage}
        \begin{minipage}{0.225\textwidth}
            \centering
            \includegraphics[width=\textwidth]{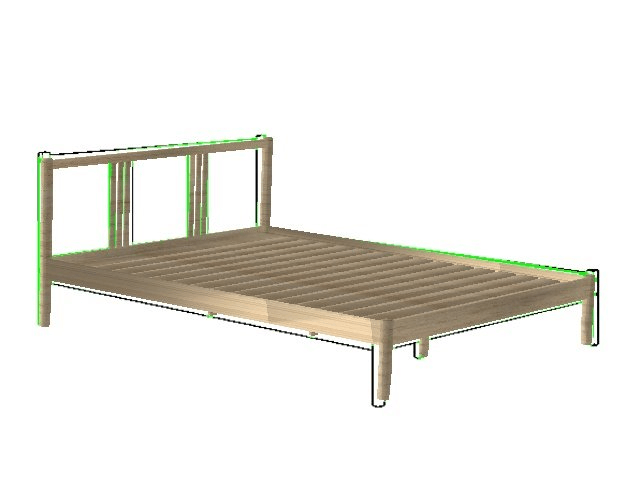}
        \end{minipage}
        \begin{minipage}{0.225\textwidth}
            \centering
            \includegraphics[width=\textwidth]{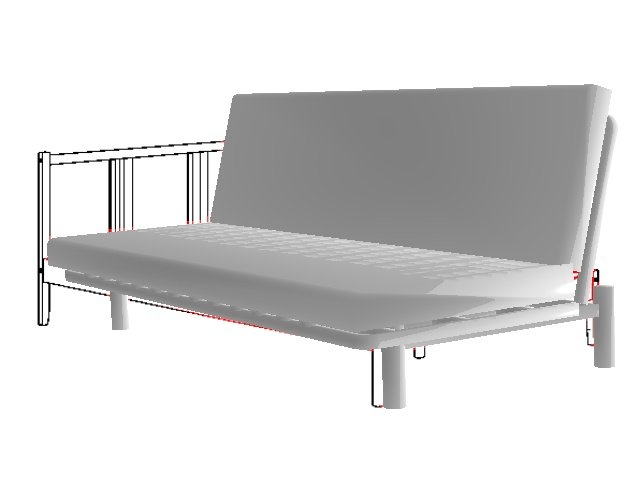}
        \end{minipage}\\[1mm]
        \small{7}
         \begin{minipage}{0.225\textwidth}
            \centering
            \includegraphics[width=\textwidth]{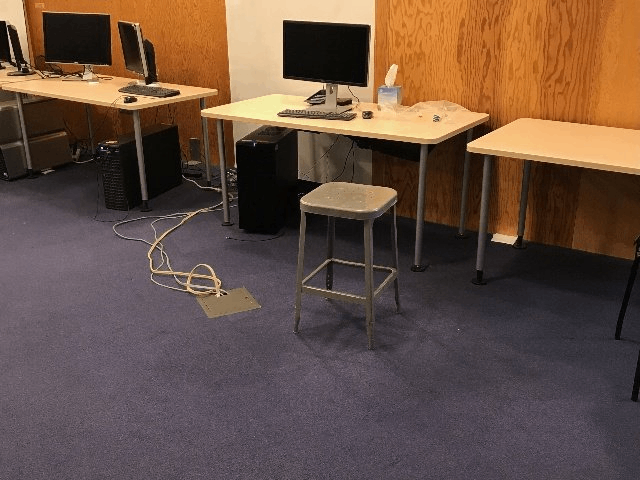}
        \end{minipage}
        \begin{minipage}{0.225\textwidth}
            \centering
            \includegraphics[width=\textwidth]{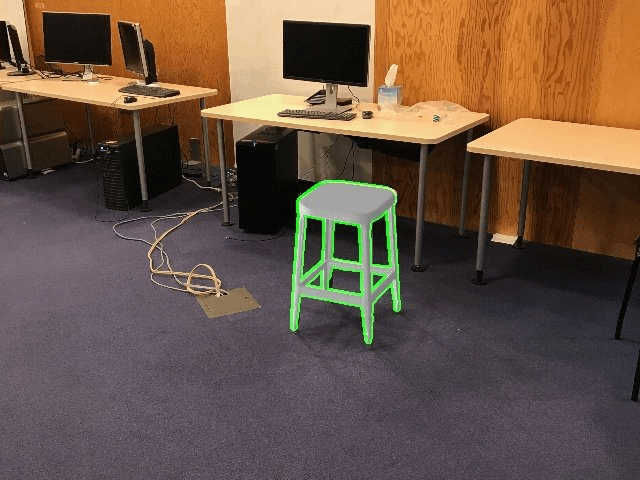}
        \end{minipage}
        \begin{minipage}{0.225\textwidth}
            \centering
            \includegraphics[width=\textwidth]{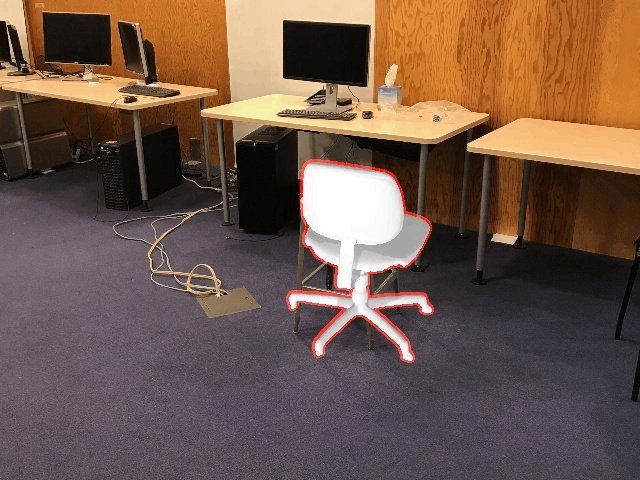}
        \end{minipage}\\[1mm]
        
        \caption{\textbf{Examples of failures in the Pix3D dataset.} Typical failures include symmetric objects (rows 1-2), local minima (rows 3-4) and misalignment due to the incorrect model (row 5-7). For more details please see the section about limitations in the main paper.}
        \label{pix3d-q-fail}
\end{figure*}

In this section, we provide more qualitative results of our approach. 
Figures~\ref{pix3d-chair-q}--\ref{pix3d-tables-q} show additional results for the chair, bed, sofa, and table classes in the Pix3D dataset. Figures~\ref{compcars-q} and~\ref{stanfordcars-q} show additional results for the Stanford cars and CompCars datasets, respectively. The qualitative results demonstrate the high accuracy of the alignments obtained by our approach despite variation in focal length, variability of the 3D models that have often very little texture, occlusions, and cluttered backgrounds.  Finally, Figure~\ref{pix3d-q-fail} shows additional examples of failure modes on the Pix3D dataset. 

For Pix3D, we provide good results for the chair class in Fig.~\ref{pix3d-chair-q}, for the bed class in Fig.~\ref{pix3d-beds-q}, for the sofa class in Fig.~\ref{pix3d-sofas-q} and for the table class in Fig.~\ref{pix3d-tables-q}. We also provide qualitative results for Stanford cars in Fig.~\ref{compcars-q} and for CompCars in Fig.~\ref{stanfordcars-q}. Please notice the quality of alignment that our approach can achieve.
We provide the failure cases for the Pix3D dataset in Fig.~\ref{pix3d-q-fail}.

\bibliographystyle{IEEEtran}
\bibliography{main.bib}

\ifCLASSOPTIONcaptionsoff
  \newpage
\fi

\begin{IEEEbiography}[{\includegraphics[width=1in,height=1.25in,clip,keepaspectratio]{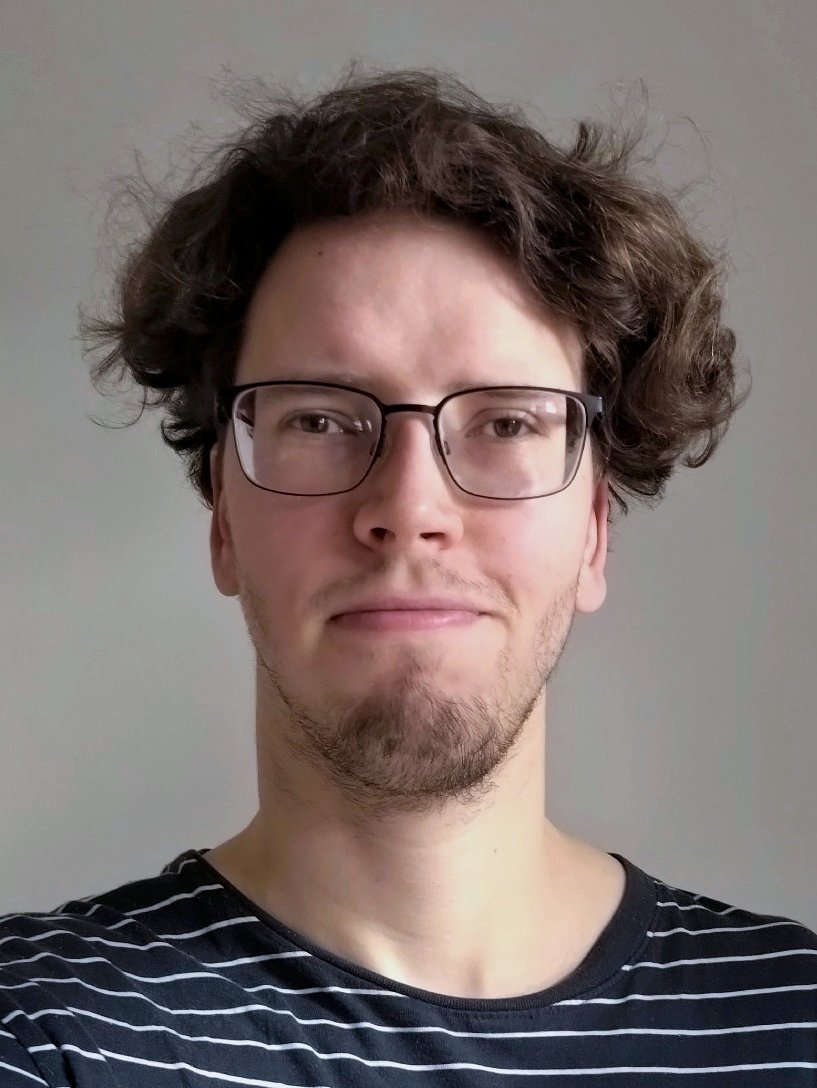}}]{Martin Cífka}
is a Ph.D. student at Faculty of Electrical Engineering, Czech Technical University in Prague, and currently works in the Intelligent Machine Perception group with the Czech Institute of Informatics, Robotics and Cybernetics (CTU Prague) under the supervision of Josef Sivic. He obtained his M.Sc. degree in Visual Computing at Faculty of Mathematics and Physics, Charles University in Prague in 2024. His research interests include 6D pose estimation and object detection.
\end{IEEEbiography}

\begin{IEEEbiography}[{\includegraphics[width=1in,height=1.25in,clip,keepaspectratio]{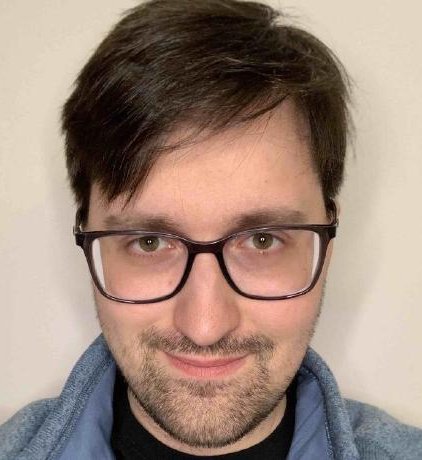}}]{Georgy Ponimatkin}
is a Ph.D. student at Faculty of Electrical Engineering, Czech Technical University in Prague, and currently works in the Intelligent Machine Perception group with the Czech Institute of Informatics, Robotics and Cybernetics (CTU Prague) under the supervision of Josef Sivic. He obtained his M.Sc. degree in high-energy physics at Faculty of Nuclear Sciences and Physical Engineering of Czech Technical University in Prague in 2021. His research interests include 6D pose estimation in the wild and in applications to robotic manipulation.
\end{IEEEbiography}

\begin{IEEEbiography}[{\includegraphics[width=1in,height=1.25in,clip,keepaspectratio]{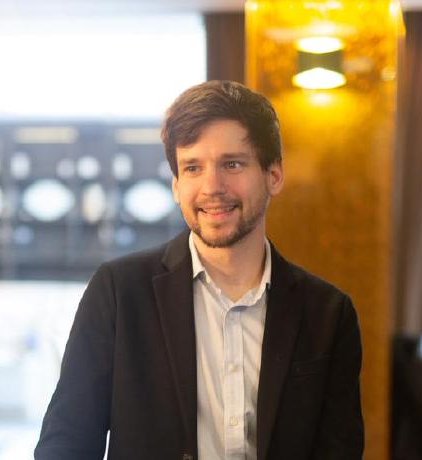}}]{Yann Labbé}
graduated from Ecole Normale Supérieure de Cachan where he received a dual Masters degree in Mathematics, Machine Learning, Computer Vision (MVA) and in Electrical Engineering. He obtained his PhD from Inria Paris and Ecole Normale Supérieure where he was supervised by Josef Sivic. He is now a research scientist at Meta Reality Labs. His current research focuses on various flavors of pose estimation.
\end{IEEEbiography}

\begin{IEEEbiography}[{\includegraphics[width=1in,height=1.25in,clip,keepaspectratio]{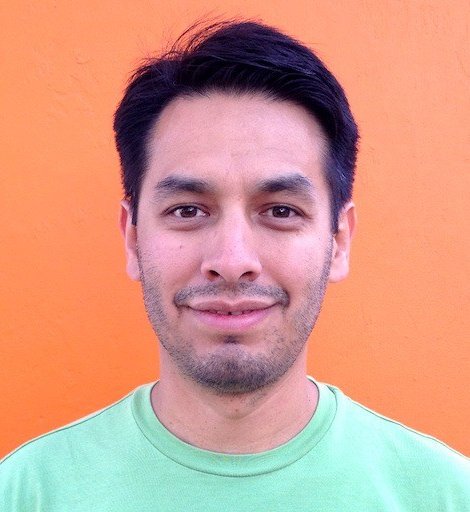}}]{Bryan Russell}
is a Senior Research Scientist at Adobe Systems. His research interests are primarily in computer vision. Bryan received his Ph.D. from MIT in the Computer Science and Artificial Intelligence Laboratory and was a post-doctoral fellow in the INRIA Willow team at the Département d’Informatique of Ecole Normale Supérieure in Paris. He was a Research Scientist with Intel Labs as part of the Intel Science and Technology Center for Visual Computing (ISTC-VC) and has been affiliated with the University of Washington.
\end{IEEEbiography}

\begin{IEEEbiography}[{\includegraphics[width=1in,height=1.25in,clip,keepaspectratio]{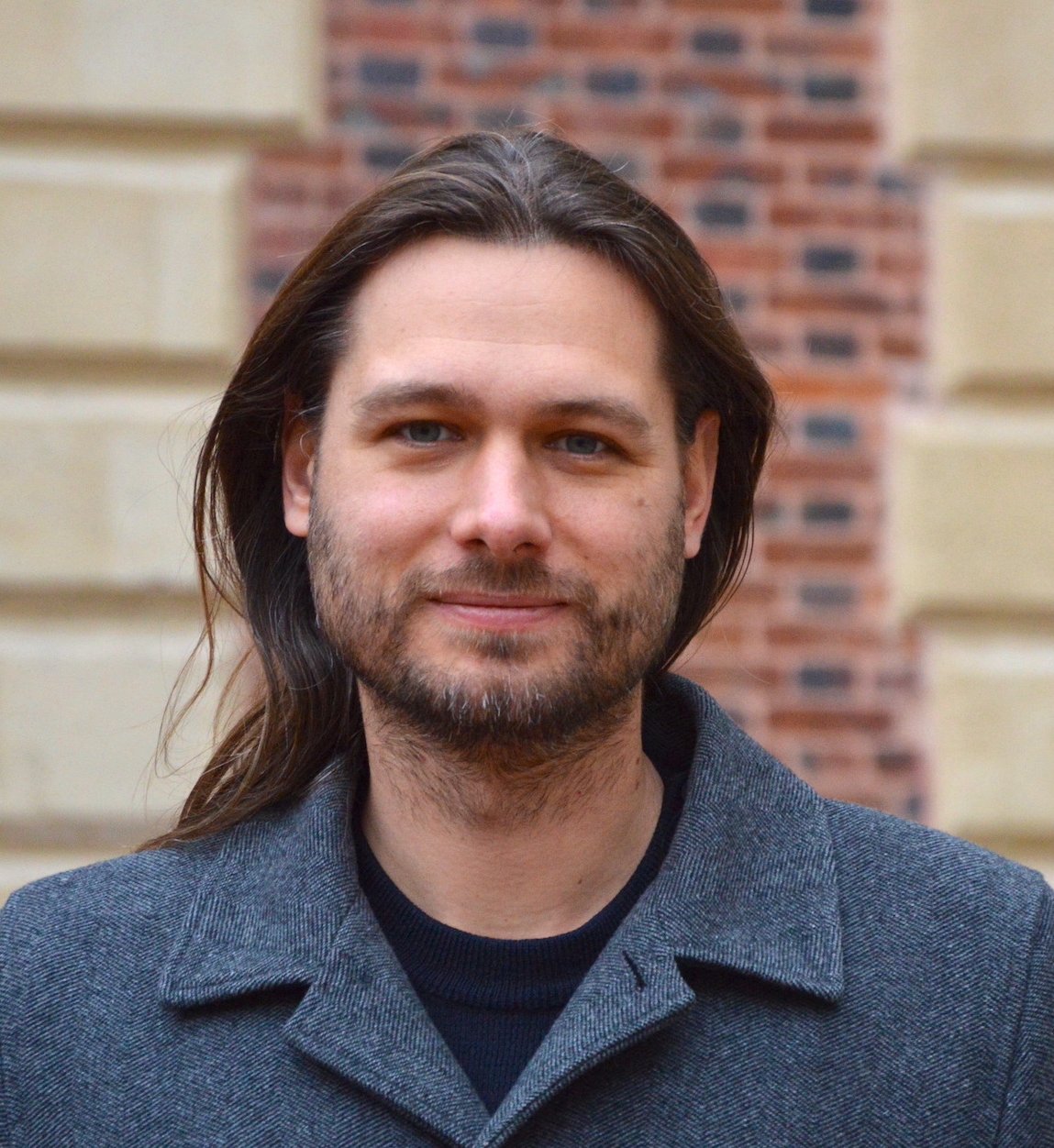}}]{Mathieu Aubry}
Mathieu Aubry is a tenured researcher in Computer Vision at École des Ponts ParisTech in the LIGM lab (UMR8049).  He obtained his PhD at ENS in 2015, co-advised by Josef Sivic (INRIA) and Daniel Cremers (TUM). In 2015, he spent a year working as a postdoc with Alexei Efros in UC Berkeley. He has a leading role in the ANR EnHerit, VHS and EIDA projects and the ERC DISCOVER project on interpretable visual structures discovery.
\end{IEEEbiography}

\begin{IEEEbiography}[{\includegraphics[width=1in,height=1.25in,clip,keepaspectratio]{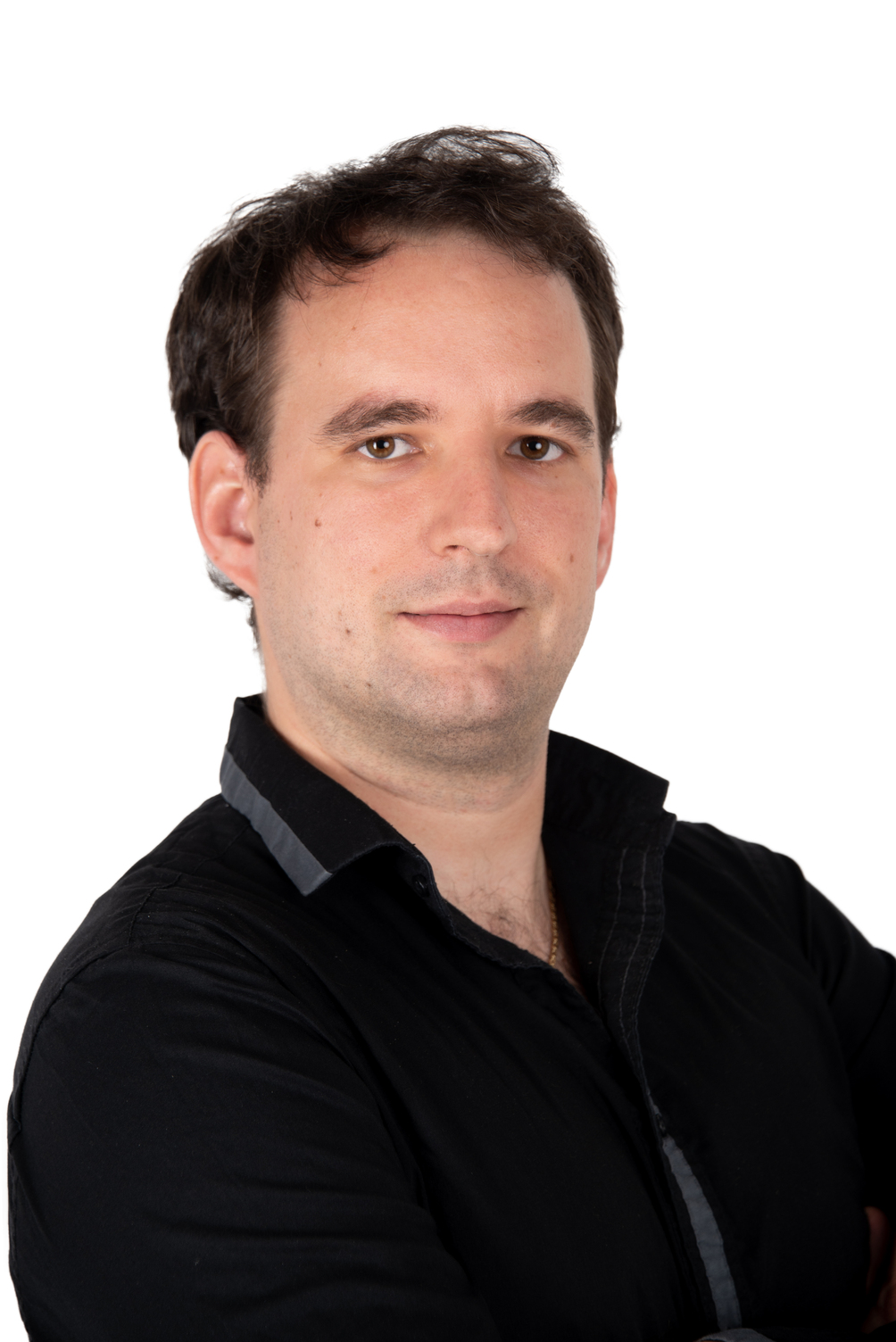}}]{Vladimir Petrik}
received the M.Sc. degree in cybernetics and robotics and the Ph.D. degree in artificial intelligence and biocybernetics from the Czech Technical University in Prague, Prague, Czech Republic, in 2014 and 2019, respectively. From 2018 to 2019, he was a Visiting Researcher with Aalto University, Espoo, Finland, after which he spent four months at Inria Research Institute, Paris, France. Since 2019, he has been a Junior Researcher with the Czech Institute of Informatics, Robotics and Cybernetics, Czech Technical University in Prague. His research interests include soft material manipulation and machine learning for robotics.
\end{IEEEbiography}

\begin{IEEEbiography}[{\includegraphics[width=1in,height=1.25in,clip,keepaspectratio]{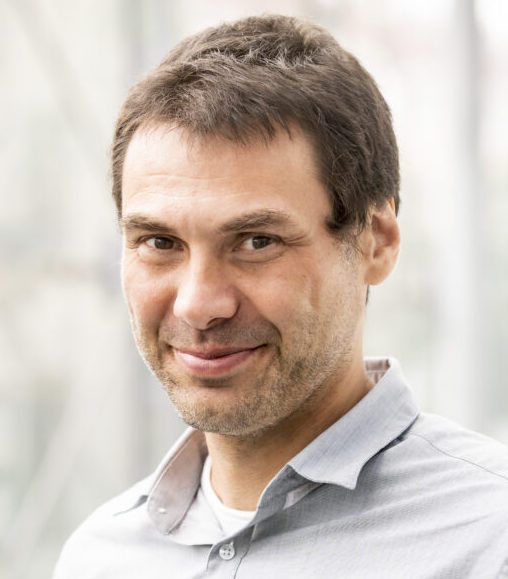}}]{Josef Sivic}
holds a distinguished researcher position at the Czech Institute of Robotics, Informatics and Cybernetics at the Czech Technical University in Prague where he heads the Intelligent Machine Perception team and the ELLIS Unit Prague. He received the habilitation degree from Ecole Normale Superieure in Paris in 2014 and PhD from the University of Oxford in 2006. After Phd he was a post-doctoral associate at the Computer Science and Artificial Intelligence Laboratory at the Massachusetts Institute of Technology. He received the British Machine Vision Association Sullivan Thesis Prize, three test-of-time awards at major computer vision conferences, an ERC Starting Grant and an ERC Advanced Grant.
\end{IEEEbiography}

\end{document}